%% file: main.tex
\documentclass[10pt,twocolumn,letterpaper]{article}
\usepackage[pagenumbers]{cvpr}

\usepackage{times,graphicx,amsmath,amssymb,booktabs,array,multirow,colortbl,tabularx,pifont}
\definecolor{BAgroup}{RGB}{236,238,241}\definecolor{BAours}{RGB}{230,240,250}
\definecolor{BAgain}{RGB}{29,122,51}\definecolor{BAloss}{RGB}{180,40,40}\definecolor{BAdash}{gray}{0.6}\definecolor{BAdesc}{gray}{0.45}

\newcommand{\BApm}[1]{{\scriptsize\textcolor{BAdesc}{$\pm$#1}}}
\usepackage{algorithm,algpseudocode}
\usepackage{iftex}
\ifXeTeX
\usepackage{fontspec}
\IfFontExistsTF{Times New Roman}{\setmainfont{Times New Roman}\def\BAtnrbold{"Times New Roman/B"}}%
  {\setmainfont{times.ttf}[Path=./,BoldFont=timesbd.ttf,ItalicFont=timesi.ttf,BoldItalicFont=timesbi.ttf]\def\BAtnrbold{"[timesbd.ttf]"}}
=\BAtnrbold\space at 11pt
=\BAtnrbold\space at 10pt
\fi
\ifPDFTeX\usepackage[accsupp]{axessibility}\pdfminorversion=7\fi
\definecolor{cvprblue}{rgb}{0.21,0.49,0.74}
\usepackage[pagebackref,breaklinks,colorlinks,allcolors=cvprblue]{hyperref}
\def\confName{CVPR}\def\confYear{2027}

\newcommand{\sid}{S_{\mathrm{ID}}}
\input{numbers.tex}\input{numbers_v38.tex}\input{numbers_v41.tex}\input{numbers_update_20261117.tex}\input{numbers_v44.tex}
\input{unsure_flags.tex}
\usepackage[normalem]{ulem}
\newcommand{\BAunsure}[1]{{\setlength{\fboxsep}{0.5pt}\colorbox{yellow}{#1}}}
\let\BAvalraw\BAval
\renewcommand{\BAval}[1]{\ifcsname BAflag@#1\endcsname\BAunsure{\BAvalraw{#1}}\else\BAvalraw{#1}\fi}
\DeclareRobustCommand{\BAunsuretext}[1]{\bgroup\markoverwith{\textcolor{yellow}{\rule[-0.55ex]{2pt}{2.6ex}}}\ULon{#1}}
\makeatletter\let\BAtable\@@input
\renewcommand{\paragraph}{\@startsection{paragraph}{4}{\z@}{0.5pt plus 1pt}{-0.5em}{\normalfont\normalsize\bfseries}}\makeatother
\title{\raisebox{-0.15em}{\includegraphics[height=1.30em]{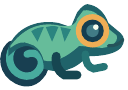}}\hspace{0.24em}BindAdapt: Joint Camera--Computation Scheduling\\for Persistent Target Identity in PTZ Tracking}
\author{Haoyang Wu\textsuperscript{1,*}\hspace{0.75em}Ziqian Wang\textsuperscript{2,*}\hspace{0.75em}Shoudong Han\textsuperscript{1,\textdagger}\hspace{0.75em}Chaoyue Li\textsuperscript{1}\hspace{0.75em}Taoran Fu\textsuperscript{3}\hspace{0.75em}Zhenhao Zhang\textsuperscript{4}\\
Jiachen Luo\textsuperscript{5}\hspace{0.75em}Heng Li\textsuperscript{1}\hspace{0.75em}Hao Geng\textsuperscript{6}\hspace{0.75em}Zhenyang Xie\textsuperscript{7}\hspace{0.75em}Sihan Wang\textsuperscript{8}\\
{\small \textsuperscript{1}Huazhong University of Science and Technology\quad \textsuperscript{2}Beijing Institute of Technology}\\
{\small \textsuperscript{3}Xi'an Jiaotong-Liverpool University\quad \textsuperscript{4}Tsinghua University}\\
{\small \textsuperscript{5}Independent Researcher\quad \textsuperscript{6}Shanghai Maritime University}\\
{\small \textsuperscript{7}Jiangxi University of Water Resources and Electric Power\quad \textsuperscript{8}Zhongnan University of Economics and Law}\\[-1pt]{\small \textsuperscript{*}Equal contribution.\qquad \textsuperscript{\textdagger}Corresponding author.}}
\hypersetup{pdfauthor={Haoyang Wu, Ziqian Wang, Shoudong Han, Chaoyue Li, Taoran Fu, Zhenhao Zhang, Jiachen Luo, Heng Li, Hao Geng, Zhenyang Xie, Sihan Wang},pdftitle={BindAdapt: Joint Camera-Computation Scheduling for Persistent Target Identity in PTZ Tracking}}
\begin{document}
\maketitle
\input{sec/0_abstract}
\input{sec/1_introduction}
\input{sec/2_related}
\input{sec/3_methodology}

\input{sec/4_experiments}
\input{sec/5_conclusion}
\section*{LLM Usage Statement}
The authors conceived the research question and hypotheses, designed every experiment and analysis, made all research decisions, and checked every derivation and number against the result files. Under their direction, LLM-based assistants (Claude, Codex, Gemini and ChatGPT) served as tools to help draft and polish text and to implement and run code. The authors take full responsibility for the content.

{\small\bibliographystyle{ieeenat_fullname}\bibliography{references}}
\end{document}


\maketitle
\input{supp/s01_implementation}
\input{supp/s02_data_adapters}
\input{supp/s03_measures}
\input{supp/s04_robustness}
\input{supp/s05_mechanism}
\input{supp/s06_transfer}
\input{supp/s07_oracle}
{\small\bibliographystyle{ieeenat_fullname}\bibliography{supp_references}}

%% file: numbers.tex
\providecommand{\BAval}[1]{\ifcsname BAval@#1\endcsname\csname BAval@#1\endcsname\else\PackageError{BindAdapt}{Unknown numeric ledger key #1}{}\fi}
\expandafter\def\csname BAval@t01.r01.sid\endcsname{44.0}
\expandafter\def\csname BAval@t01.r01.wrong\endcsname{19.8}
\expandafter\def\csname BAval@t01.r01.recovery\endcsname{42.0}
\expandafter\def\csname BAval@t01.r01.t5\endcsname{3.8}
\expandafter\def\csname BAval@t01.r02.sid\endcsname{58.2}
\expandafter\def\csname BAval@t01.r02.wrong\endcsname{15.3}
\expandafter\def\csname BAval@t01.r02.recovery\endcsname{56.0}
\expandafter\def\csname BAval@t01.r02.t5\endcsname{3.2}
\expandafter\def\csname BAval@t01.r03.sid\endcsname{55.6}
\expandafter\def\csname BAval@t01.r03.wrong\endcsname{17.0}
\expandafter\def\csname BAval@t01.r03.recovery\endcsname{54.0}
\expandafter\def\csname BAval@t01.r03.t5\endcsname{3.3}
\expandafter\def\csname BAval@t01.r04.sid\endcsname{64.5}
\expandafter\def\csname BAval@t01.r04.wrong\endcsname{12.2}
\expandafter\def\csname BAval@t01.r04.recovery\endcsname{64.0}
\expandafter\def\csname BAval@t01.r04.t5\endcsname{2.8}
\expandafter\def\csname BAval@t01.r05.sid\endcsname{65.8}
\expandafter\def\csname BAval@t01.r05.wrong\endcsname{11.8}
\expandafter\def\csname BAval@t01.r05.recovery\endcsname{65.0}
\expandafter\def\csname BAval@t01.r05.t5\endcsname{2.7}
\expandafter\def\csname BAval@t01.r06.sid\endcsname{68.7}
\expandafter\def\csname BAval@t01.r06.wrong\endcsname{10.1}
\expandafter\def\csname BAval@t01.r06.recovery\endcsname{70.0}
\expandafter\def\csname BAval@t01.r06.t5\endcsname{2.5}
\expandafter\def\csname BAval@t01.r07.sid\endcsname{70.3}
\expandafter\def\csname BAval@t01.r07.wrong\endcsname{8.8}
\expandafter\def\csname BAval@t01.r07.recovery\endcsname{72.0}
\expandafter\def\csname BAval@t01.r07.t5\endcsname{2.4}
\expandafter\def\csname BAval@t01.r08.sid\endcsname{82.3}
\expandafter\def\csname BAval@t01.r08.wrong\endcsname{4.4}
\expandafter\def\csname BAval@t01.r08.recovery\endcsname{86.2}
\expandafter\def\csname BAval@t01.r08.t5\endcsname{1.4}
\expandafter\def\csname BAval@t01.r09.sid\endcsname{87.9}
\expandafter\def\csname BAval@t01.r09.wrong\endcsname{1.6}
\expandafter\def\csname BAval@t01.r09.recovery\endcsname{92.0}
\expandafter\def\csname BAval@t01.r09.t5\endcsname{1.2}
\expandafter\def\csname BAval@t02.r01.short\endcsname{78.0}
\expandafter\def\csname BAval@t02.r01.similar\endcsname{67.0}
\expandafter\def\csname BAval@t02.r01.long\endcsname{53.0}
\expandafter\def\csname BAval@t02.r02.short\endcsname{79.0}
\expandafter\def\csname BAval@t02.r02.similar\endcsname{68.0}
\expandafter\def\csname BAval@t02.r02.long\endcsname{54.8}
\expandafter\def\csname BAval@t02.r03.short\endcsname{80.0}
\expandafter\def\csname BAval@t02.r03.similar\endcsname{72.0}
\expandafter\def\csname BAval@t02.r03.long\endcsname{57.6}
\expandafter\def\csname BAval@t02.r04.short\endcsname{82.0}
\expandafter\def\csname BAval@t02.r04.similar\endcsname{74.0}
\expandafter\def\csname BAval@t02.r04.long\endcsname{58.3}
\expandafter\def\csname BAval@t02.r05.short\endcsname{90.2}
\expandafter\def\csname BAval@t02.r05.similar\endcsname{85.4}
\expandafter\def\csname BAval@t02.r05.long\endcsname{73.9}
\expandafter\def\csname BAval@t02.r06.short\endcsname{94.0}
\expandafter\def\csname BAval@t02.r06.similar\endcsname{90.0}
\expandafter\def\csname BAval@t02.r06.long\endcsname{82.0}
\expandafter\def\csname BAval@t03.r01.sid\endcsname{65.9}
\expandafter\def\csname BAval@t03.r01.wrong\endcsname{11.9}
\expandafter\def\csname BAval@t03.r02.sid\endcsname{70.3}
\expandafter\def\csname BAval@t03.r02.wrong\endcsname{8.8}
\expandafter\def\csname BAval@t03.r03.sid\endcsname{77.6}
\expandafter\def\csname BAval@t03.r03.wrong\endcsname{6.3}
\expandafter\def\csname BAval@t03.r04.sid\endcsname{82.3}
\expandafter\def\csname BAval@t03.r04.wrong\endcsname{4.4}
\expandafter\def\csname BAval@t04.r01.sid\endcsname{82.3}
\expandafter\def\csname BAval@t04.r01.wrong\endcsname{4.4}
\expandafter\def\csname BAval@t04.r01.t5\endcsname{1.4}
\expandafter\def\csname BAval@t04.r02.sid\endcsname{77.9}
\expandafter\def\csname BAval@t04.r02.wrong\endcsname{7.1}
\expandafter\def\csname BAval@t04.r02.t5\endcsname{1.9}
\expandafter\def\csname BAval@t04.r03.sid\endcsname{76.0}
\expandafter\def\csname BAval@t04.r03.wrong\endcsname{9.0}
\expandafter\def\csname BAval@t04.r03.t5\endcsname{2.0}
\expandafter\def\csname BAval@t04.r04.sid\endcsname{78.0}
\expandafter\def\csname BAval@t04.r04.wrong\endcsname{7.7}
\expandafter\def\csname BAval@t04.r04.t5\endcsname{1.8}
\expandafter\def\csname BAval@t04.r05.sid\endcsname{79.3}
\expandafter\def\csname BAval@t04.r05.wrong\endcsname{6.8}
\expandafter\def\csname BAval@t04.r05.t5\endcsname{1.6}
\expandafter\def\csname BAval@t04.r06.sid\endcsname{73.7}
\expandafter\def\csname BAval@t04.r06.wrong\endcsname{8.8}
\expandafter\def\csname BAval@t04.r06.t5\endcsname{2.2}
\expandafter\def\csname BAval@t04.r07.sid\endcsname{77.6}
\expandafter\def\csname BAval@t04.r07.wrong\endcsname{6.3}
\expandafter\def\csname BAval@t04.r07.t5\endcsname{1.7}
\expandafter\def\csname BAval@t04.r08.sid\endcsname{80.5}
\expandafter\def\csname BAval@t04.r08.wrong\endcsname{5.8}
\expandafter\def\csname BAval@t04.r08.t5\endcsname{1.5}
\expandafter\def\csname BAval@t05.r01.separable\endcsname{78.0}
\expandafter\def\csname BAval@t05.r01.temporal.q\endcsname{80.1}
\expandafter\def\csname BAval@t05.r01.bindadapt\endcsname{86.1}
\expandafter\def\csname BAval@t05.r01.oracle\endcsname{91.7}
\expandafter\def\csname BAval@t05.r02.separable\endcsname{75.6}
\expandafter\def\csname BAval@t05.r02.temporal.q\endcsname{77.2}
\expandafter\def\csname BAval@t05.r02.bindadapt\endcsname{85.3}
\expandafter\def\csname BAval@t05.r02.oracle\endcsname{90.9}
\expandafter\def\csname BAval@t05.r03.separable\endcsname{71.1}
\expandafter\def\csname BAval@t05.r03.temporal.q\endcsname{73.2}
\expandafter\def\csname BAval@t05.r03.bindadapt\endcsname{83.8}
\expandafter\def\csname BAval@t05.r03.oracle\endcsname{90.1}
\expandafter\def\csname BAval@t05.r04.separable\endcsname{65.9}
\expandafter\def\csname BAval@t05.r04.temporal.q\endcsname{68.7}
\expandafter\def\csname BAval@t05.r04.bindadapt\endcsname{82.3}
\expandafter\def\csname BAval@t05.r04.oracle\endcsname{87.9}
\expandafter\def\csname BAval@t05.r05.separable\endcsname{53.9}
\expandafter\def\csname BAval@t05.r05.temporal.q\endcsname{57.7}
\expandafter\def\csname BAval@t05.r05.bindadapt\endcsname{75.6}
\expandafter\def\csname BAval@t05.r05.oracle\endcsname{82.9}
\expandafter\def\csname BAval@t05.r06.separable\endcsname{37.9}
\expandafter\def\csname BAval@t05.r06.temporal.q\endcsname{42.1}
\expandafter\def\csname BAval@t05.r06.bindadapt\endcsname{64.0}
\expandafter\def\csname BAval@t05.r06.oracle\endcsname{72.9}
\expandafter\def\csname BAval@t06.r01.temporal.q\endcsname{54.1}
\expandafter\def\csname BAval@t06.r01.joint.belief\endcsname{56.1}
\expandafter\def\csname BAval@t06.r01.bindadapt\endcsname{70.2}
\expandafter\def\csname BAval@t06.r01.oracle\endcsname{77.0}
\expandafter\def\csname BAval@t06.r02.temporal.q\endcsname{62.0}
\expandafter\def\csname BAval@t06.r02.joint.belief\endcsname{64.0}
\expandafter\def\csname BAval@t06.r02.bindadapt\endcsname{77.1}
\expandafter\def\csname BAval@t06.r02.oracle\endcsname{84.0}
\expandafter\def\csname BAval@t06.r03.temporal.q\endcsname{68.7}
\expandafter\def\csname BAval@t06.r03.joint.belief\endcsname{70.3}
\expandafter\def\csname BAval@t06.r03.bindadapt\endcsname{82.3}
\expandafter\def\csname BAval@t06.r03.oracle\endcsname{87.9}
\expandafter\def\csname BAval@t06.r04.temporal.q\endcsname{72.3}
\expandafter\def\csname BAval@t06.r04.joint.belief\endcsname{74.3}
\expandafter\def\csname BAval@t06.r04.bindadapt\endcsname{84.5}
\expandafter\def\csname BAval@t06.r04.oracle\endcsname{90.2}
\expandafter\def\csname BAval@t06.r05.temporal.q\endcsname{74.7}
\expandafter\def\csname BAval@t06.r05.joint.belief\endcsname{76.1}
\expandafter\def\csname BAval@t06.r05.bindadapt\endcsname{85.6}
\expandafter\def\csname BAval@t06.r05.oracle\endcsname{91.0}
\expandafter\def\csname BAval@t06.r06.temporal.q\endcsname{75.9}
\expandafter\def\csname BAval@t06.r06.joint.belief\endcsname{77.6}
\expandafter\def\csname BAval@t06.r06.bindadapt\endcsname{86.2}
\expandafter\def\csname BAval@t06.r06.oracle\endcsname{91.4}
\expandafter\def\csname BAval@t07.r01.pair.rank.pct\endcsname{67.0}
\expandafter\def\csname BAval@t07.r01.risk.mae\endcsname{0.140}
\expandafter\def\csname BAval@t07.r01.calibration.error\endcsname{0.098}
\expandafter\def\csname BAval@t07.r01.action.regret\endcsname{0.125}
\expandafter\def\csname BAval@t07.r02.pair.rank.pct\endcsname{72.5}
\expandafter\def\csname BAval@t07.r02.risk.mae\endcsname{0.118}
\expandafter\def\csname BAval@t07.r02.calibration.error\endcsname{0.073}
\expandafter\def\csname BAval@t07.r02.action.regret\endcsname{0.091}
\expandafter\def\csname BAval@t07.r03.pair.rank.pct\endcsname{78.0}
\expandafter\def\csname BAval@t07.r03.risk.mae\endcsname{0.096}
\expandafter\def\csname BAval@t07.r03.calibration.error\endcsname{0.056}
\expandafter\def\csname BAval@t07.r03.action.regret\endcsname{0.068}
\expandafter\def\csname BAval@t07.r04.pair.rank.pct\endcsname{88.7}
\expandafter\def\csname BAval@t07.r04.risk.mae\endcsname{0.053}
\expandafter\def\csname BAval@t07.r04.calibration.error\endcsname{0.026}
\expandafter\def\csname BAval@t07.r04.action.regret\endcsname{0.032}
\expandafter\def\csname BAval@t08.r01.sid\endcsname{82.3}
\expandafter\def\csname BAval@t08.r02.sid\endcsname{82.8}
\expandafter\def\csname BAval@t08.r03.sid\endcsname{86.1}
\expandafter\def\csname BAval@t08.r04.sid\endcsname{87.9}
\expandafter\def\csname BAval@t08.r05.sid\endcsname{94.1}
\expandafter\def\csname BAval@t08.r06.sid\endcsname{98.0}
\expandafter\def\csname BAval@t09.r01.groups\endcsname{60}
\expandafter\def\csname BAval@t09.r01.episodes\endcsname{1200}
\expandafter\def\csname BAval@t09.r02.groups\endcsname{20}
\expandafter\def\csname BAval@t09.r02.episodes\endcsname{200}
\expandafter\def\csname BAval@t09.r03.groups\endcsname{40}
\expandafter\def\csname BAval@t09.r03.episodes\endcsname{480}
\expandafter\def\csname BAval@t10.r01.separable\endcsname{68.6}
\expandafter\def\csname BAval@t10.r01.temporal.q\endcsname{70.1}
\expandafter\def\csname BAval@t10.r01.bindadapt\endcsname{82.3}
\expandafter\def\csname BAval@t10.r02.separable\endcsname{65.3}
\expandafter\def\csname BAval@t10.r02.temporal.q\endcsname{67.0}
\expandafter\def\csname BAval@t10.r02.bindadapt\endcsname{80.1}
\expandafter\def\csname BAval@t10.r03.separable\endcsname{61.8}
\expandafter\def\csname BAval@t10.r03.temporal.q\endcsname{65.2}
\expandafter\def\csname BAval@t10.r03.bindadapt\endcsname{78.1}
\expandafter\def\csname BAval@t10.r04.separable\endcsname{58.9}
\expandafter\def\csname BAval@t10.r04.temporal.q\endcsname{62.9}
\expandafter\def\csname BAval@t10.r04.bindadapt\endcsname{76.1}
\expandafter\def\csname BAval@t10.r05.separable\endcsname{62.4}
\expandafter\def\csname BAval@t10.r05.temporal.q\endcsname{64.9}
\expandafter\def\csname BAval@t10.r05.bindadapt\endcsname{77.1}
\expandafter\def\csname BAval@t11.r01.reference\endcsname{91.2}
\expandafter\def\csname BAval@t11.r01.bindadapt\endcsname{92.0}
\expandafter\def\csname BAval@t11.r01.gain.pp\endcsname{0.8}
\expandafter\def\csname BAval@t11.r02.reference\endcsname{93.9}
\expandafter\def\csname BAval@t11.r02.bindadapt\endcsname{94.1}
\expandafter\def\csname BAval@t11.r02.gain.pp\endcsname{0.2}
\expandafter\def\csname BAval@t11.r03.reference\endcsname{65.9}
\expandafter\def\csname BAval@t11.r03.bindadapt\endcsname{82.3}
\expandafter\def\csname BAval@t11.r03.gain.pp\endcsname{16.4}
\expandafter\def\csname BAval@t11.r04.reference\endcsname{40.9}
\expandafter\def\csname BAval@t11.r04.bindadapt\endcsname{41.3}
\expandafter\def\csname BAval@t11.r04.gain.pp\endcsname{0.4}
\expandafter\def\csname BAval@t11.r05.reference\endcsname{20.5}
\expandafter\def\csname BAval@t11.r05.bindadapt\endcsname{20.7}
\expandafter\def\csname BAval@t11.r05.gain.pp\endcsname{0.2}
\expandafter\def\csname BAval@t12.r01.sid\endcsname{60.9}
\expandafter\def\csname BAval@t12.r01.wrong\endcsname{14.6}
\expandafter\def\csname BAval@t12.r01.recovery\endcsname{59.0}
\expandafter\def\csname BAval@t12.r01.t5\endcsname{3.0}
\expandafter\def\csname BAval@t12.r02.sid\endcsname{62.5}
\expandafter\def\csname BAval@t12.r02.wrong\endcsname{13.8}
\expandafter\def\csname BAval@t12.r02.recovery\endcsname{61.1}
\expandafter\def\csname BAval@t12.r02.t5\endcsname{2.9}
\expandafter\def\csname BAval@t12.r03.sid\endcsname{65.3}
\expandafter\def\csname BAval@t12.r03.wrong\endcsname{12.2}
\expandafter\def\csname BAval@t12.r03.recovery\endcsname{64.9}
\expandafter\def\csname BAval@t12.r03.t5\endcsname{2.7}
\expandafter\def\csname BAval@t12.r04.sid\endcsname{66.4}
\expandafter\def\csname BAval@t12.r04.wrong\endcsname{11.8}
\expandafter\def\csname BAval@t12.r04.recovery\endcsname{67.1}
\expandafter\def\csname BAval@t12.r04.t5\endcsname{2.6}
\expandafter\def\csname BAval@t12.r05.sid\endcsname{71.0}
\expandafter\def\csname BAval@t12.r05.wrong\endcsname{8.8}
\expandafter\def\csname BAval@t12.r05.recovery\endcsname{73.0}
\expandafter\def\csname BAval@t12.r05.t5\endcsname{2.3}
\expandafter\def\csname BAval@t12.r06.sid\endcsname{62.0}
\expandafter\def\csname BAval@t12.r06.wrong\endcsname{14.3}
\expandafter\def\csname BAval@t12.r06.recovery\endcsname{64.0}
\expandafter\def\csname BAval@t12.r06.t5\endcsname{2.8}
\expandafter\def\csname BAval@t12.r07.sid\endcsname{65.6}
\expandafter\def\csname BAval@t12.r07.wrong\endcsname{10.4}
\expandafter\def\csname BAval@t12.r07.recovery\endcsname{68.0}
\expandafter\def\csname BAval@t12.r07.t5\endcsname{2.5}
\expandafter\def\csname BAval@t12.r08.sid\endcsname{70.3}
\expandafter\def\csname BAval@t12.r08.wrong\endcsname{8.1}
\expandafter\def\csname BAval@t12.r08.recovery\endcsname{74.0}
\expandafter\def\csname BAval@t12.r08.t5\endcsname{2.2}
\expandafter\def\csname BAval@t12.r09.sid\endcsname{66.5}
\expandafter\def\csname BAval@t12.r09.wrong\endcsname{10.9}
\expandafter\def\csname BAval@t12.r09.recovery\endcsname{68.0}
\expandafter\def\csname BAval@t12.r09.t5\endcsname{2.6}
\expandafter\def\csname BAval@t12.r10.sid\endcsname{72.8}
\expandafter\def\csname BAval@t12.r10.wrong\endcsname{7.8}
\expandafter\def\csname BAval@t12.r10.recovery\endcsname{76.1}
\expandafter\def\csname BAval@t12.r10.t5\endcsname{2.1}
\expandafter\def\csname BAval@t12.r11.sid\endcsname{68.0}
\expandafter\def\csname BAval@t12.r11.wrong\endcsname{10.1}
\expandafter\def\csname BAval@t12.r11.recovery\endcsname{69.0}
\expandafter\def\csname BAval@t12.r11.t5\endcsname{2.5}
\expandafter\def\csname BAval@t12.r12.sid\endcsname{71.6}
\expandafter\def\csname BAval@t12.r12.wrong\endcsname{8.6}
\expandafter\def\csname BAval@t12.r12.recovery\endcsname{73.9}
\expandafter\def\csname BAval@t12.r12.t5\endcsname{2.2}
\expandafter\def\csname BAval@t12.r13.sid\endcsname{74.1}
\expandafter\def\csname BAval@t12.r13.wrong\endcsname{7.4}
\expandafter\def\csname BAval@t12.r13.recovery\endcsname{77.1}
\expandafter\def\csname BAval@t12.r13.t5\endcsname{2.1}
\expandafter\def\csname BAval@t12.r14.sid\endcsname{74.6}
\expandafter\def\csname BAval@t12.r14.wrong\endcsname{6.9}
\expandafter\def\csname BAval@t12.r14.recovery\endcsname{77.9}
\expandafter\def\csname BAval@t12.r14.t5\endcsname{2.0}
\expandafter\def\csname BAval@t12.r15.sid\endcsname{75.2}
\expandafter\def\csname BAval@t12.r15.wrong\endcsname{7.3}
\expandafter\def\csname BAval@t12.r15.recovery\endcsname{79.0}
\expandafter\def\csname BAval@t12.r15.t5\endcsname{2.0}
\expandafter\def\csname BAval@t12.r16.sid\endcsname{82.3}
\expandafter\def\csname BAval@t12.r16.wrong\endcsname{4.4}
\expandafter\def\csname BAval@t12.r16.recovery\endcsname{86.2}
\expandafter\def\csname BAval@t12.r16.t5\endcsname{1.4}
\expandafter\def\csname BAval@t13.r01.ao\endcsname{63.9}
\expandafter\def\csname BAval@t13.r01.p20.pct\endcsname{76.6}
\expandafter\def\csname BAval@t13.r01.pn20\endcsname{78.1}
\expandafter\def\csname BAval@t13.r01.sr50.pct\endcsname{74.2}
\expandafter\def\csname BAval@t13.r01.sr75.pct\endcsname{49.6}
\expandafter\def\csname BAval@t13.r01.sao.pct\endcsname{58.3}
\expandafter\def\csname BAval@t13.r01.ssr50.pct\endcsname{66.7}
\expandafter\def\csname BAval@t13.r01.ssr75.pct\endcsname{41.0}
\expandafter\def\csname BAval@t13.r02.ao\endcsname{65.0}
\expandafter\def\csname BAval@t13.r02.p20.pct\endcsname{77.9}
\expandafter\def\csname BAval@t13.r02.pn20\endcsname{79.2}
\expandafter\def\csname BAval@t13.r02.sr50.pct\endcsname{75.5}
\expandafter\def\csname BAval@t13.r02.sr75.pct\endcsname{51.4}
\expandafter\def\csname BAval@t13.r02.sao.pct\endcsname{59.2}
\expandafter\def\csname BAval@t13.r02.ssr50.pct\endcsname{67.9}
\expandafter\def\csname BAval@t13.r02.ssr75.pct\endcsname{42.4}
\expandafter\def\csname BAval@t13.r03.ao\endcsname{67.7}
\expandafter\def\csname BAval@t13.r03.p20.pct\endcsname{79.7}
\expandafter\def\csname BAval@t13.r03.pn20\endcsname{81.5}
\expandafter\def\csname BAval@t13.r03.sr50.pct\endcsname{78.5}
\expandafter\def\csname BAval@t13.r03.sr75.pct\endcsname{55.8}
\expandafter\def\csname BAval@t13.r03.sao.pct\endcsname{61.9}
\expandafter\def\csname BAval@t13.r03.ssr50.pct\endcsname{70.6}
\expandafter\def\csname BAval@t13.r03.ssr75.pct\endcsname{45.7}
\expandafter\def\csname BAval@t13.r04.ao\endcsname{68.6}
\expandafter\def\csname BAval@t13.r04.p20.pct\endcsname{80.2}
\expandafter\def\csname BAval@t13.r04.pn20\endcsname{82.0}
\expandafter\def\csname BAval@t13.r04.sr50.pct\endcsname{79.5}
\expandafter\def\csname BAval@t13.r04.sr75.pct\endcsname{57.1}
\expandafter\def\csname BAval@t13.r04.sao.pct\endcsname{62.5}
\expandafter\def\csname BAval@t13.r04.ssr50.pct\endcsname{71.4}
\expandafter\def\csname BAval@t13.r04.ssr75.pct\endcsname{47.2}
\expandafter\def\csname BAval@t13.r05.ao\endcsname{72.7}
\expandafter\def\csname BAval@t13.r05.p20.pct\endcsname{84.9}
\expandafter\def\csname BAval@t13.r05.pn20\endcsname{86.0}
\expandafter\def\csname BAval@t13.r05.sr50.pct\endcsname{83.6}
\expandafter\def\csname BAval@t13.r05.sr75.pct\endcsname{63.2}
\expandafter\def\csname BAval@t13.r05.sao.pct\endcsname{67.0}
\expandafter\def\csname BAval@t13.r05.ssr50.pct\endcsname{76.1}
\expandafter\def\csname BAval@t13.r05.ssr75.pct\endcsname{53.6}
\expandafter\def\csname BAval@t13.r06.ao\endcsname{63.7}
\expandafter\def\csname BAval@t13.r06.p20.pct\endcsname{75.3}
\expandafter\def\csname BAval@t13.r06.pn20\endcsname{77.1}
\expandafter\def\csname BAval@t13.r06.sr50.pct\endcsname{74.1}
\expandafter\def\csname BAval@t13.r06.sr75.pct\endcsname{48.3}
\expandafter\def\csname BAval@t13.r06.sao.pct\endcsname{57.8}
\expandafter\def\csname BAval@t13.r06.ssr50.pct\endcsname{66.1}
\expandafter\def\csname BAval@t13.r06.ssr75.pct\endcsname{40.6}
\expandafter\def\csname BAval@t13.r07.ao\endcsname{67.0}
\expandafter\def\csname BAval@t13.r07.p20.pct\endcsname{78.9}
\expandafter\def\csname BAval@t13.r07.pn20\endcsname{80.5}
\expandafter\def\csname BAval@t13.r07.sr50.pct\endcsname{77.9}
\expandafter\def\csname BAval@t13.r07.sr75.pct\endcsname{54.1}
\expandafter\def\csname BAval@t13.r07.sao.pct\endcsname{61.9}
\expandafter\def\csname BAval@t13.r07.ssr50.pct\endcsname{69.9}
\expandafter\def\csname BAval@t13.r07.ssr75.pct\endcsname{46.3}
\expandafter\def\csname BAval@t13.r08.ao\endcsname{73.3}
\expandafter\def\csname BAval@t13.r08.p20.pct\endcsname{84.5}
\expandafter\def\csname BAval@t13.r08.pn20\endcsname{85.9}
\expandafter\def\csname BAval@t13.r08.sr50.pct\endcsname{84.3}
\expandafter\def\csname BAval@t13.r08.sr75.pct\endcsname{63.6}
\expandafter\def\csname BAval@t13.r08.sao.pct\endcsname{66.7}
\expandafter\def\csname BAval@t13.r08.ssr50.pct\endcsname{75.8}
\expandafter\def\csname BAval@t13.r08.ssr75.pct\endcsname{53.0}
\expandafter\def\csname BAval@t13.r09.ao\endcsname{67.3}
\expandafter\def\csname BAval@t13.r09.p20.pct\endcsname{79.2}
\expandafter\def\csname BAval@t13.r09.pn20\endcsname{80.6}
\expandafter\def\csname BAval@t13.r09.sr50.pct\endcsname{78.4}
\expandafter\def\csname BAval@t13.r09.sr75.pct\endcsname{54.7}
\expandafter\def\csname BAval@t13.r09.sao.pct\endcsname{64.6}
\expandafter\def\csname BAval@t13.r09.ssr50.pct\endcsname{73.6}
\expandafter\def\csname BAval@t13.r09.ssr75.pct\endcsname{49.9}
\expandafter\def\csname BAval@t13.r10.ao\endcsname{72.0}
\expandafter\def\csname BAval@t13.r10.p20.pct\endcsname{83.2}
\expandafter\def\csname BAval@t13.r10.pn20\endcsname{84.7}
\expandafter\def\csname BAval@t13.r10.sr50.pct\endcsname{82.7}
\expandafter\def\csname BAval@t13.r10.sr75.pct\endcsname{61.8}
\expandafter\def\csname BAval@t13.r10.sao.pct\endcsname{68.3}
\expandafter\def\csname BAval@t13.r10.ssr50.pct\endcsname{77.0}
\expandafter\def\csname BAval@t13.r10.ssr75.pct\endcsname{55.4}
\expandafter\def\csname BAval@t13.r11.ao\endcsname{79.1}
\expandafter\def\csname BAval@t13.r11.p20.pct\endcsname{88.0}
\expandafter\def\csname BAval@t13.r11.pn20\endcsname{89.3}
\expandafter\def\csname BAval@t13.r11.sr50.pct\endcsname{88.0}
\expandafter\def\csname BAval@t13.r11.sr75.pct\endcsname{71.5}
\expandafter\def\csname BAval@t13.r11.sao.pct\endcsname{76.9}
\expandafter\def\csname BAval@t13.r11.ssr50.pct\endcsname{83.9}
\expandafter\def\csname BAval@t13.r11.ssr75.pct\endcsname{67.2}
\expandafter\def\csname BAval@t14.r01.short\endcsname{84.0}
\expandafter\def\csname BAval@t14.r01.similar\endcsname{75.9}
\expandafter\def\csname BAval@t14.r01.long\endcsname{55.5}
\expandafter\def\csname BAval@t14.r01.weighted\endcsname{70.3}
\expandafter\def\csname BAval@t14.r02.short\endcsname{85.5}
\expandafter\def\csname BAval@t14.r02.similar\endcsname{77.2}
\expandafter\def\csname BAval@t14.r02.long\endcsname{59.9}
\expandafter\def\csname BAval@t14.r02.weighted\endcsname{72.8}
\expandafter\def\csname BAval@t14.r03.short\endcsname{84.8}
\expandafter\def\csname BAval@t14.r03.similar\endcsname{76.5}
\expandafter\def\csname BAval@t14.r03.long\endcsname{58.0}
\expandafter\def\csname BAval@t14.r03.weighted\endcsname{71.6}
\expandafter\def\csname BAval@t14.r04.short\endcsname{85.4}
\expandafter\def\csname BAval@t14.r04.similar\endcsname{78.8}
\expandafter\def\csname BAval@t14.r04.long\endcsname{61.9}
\expandafter\def\csname BAval@t14.r04.weighted\endcsname{74.1}
\expandafter\def\csname BAval@t14.r05.short\endcsname{86.2}
\expandafter\def\csname BAval@t14.r05.similar\endcsname{79.2}
\expandafter\def\csname BAval@t14.r05.long\endcsname{63.9}
\expandafter\def\csname BAval@t14.r05.weighted\endcsname{75.2}
\expandafter\def\csname BAval@t14.r06.short\endcsname{90.2}
\expandafter\def\csname BAval@t14.r06.similar\endcsname{85.4}
\expandafter\def\csname BAval@t14.r06.long\endcsname{73.9}
\expandafter\def\csname BAval@t14.r06.weighted\endcsname{82.3}
\expandafter\def\csname BAval@t14.r07.short\endcsname{74.9}
\expandafter\def\csname BAval@t14.r07.similar\endcsname{65.7}
\expandafter\def\csname BAval@t14.r07.long\endcsname{46.8}
\expandafter\def\csname BAval@t14.r07.weighted\endcsname{60.9}
\expandafter\def\csname BAval@t14.r08.short\endcsname{76.5}
\expandafter\def\csname BAval@t14.r08.similar\endcsname{67.3}
\expandafter\def\csname BAval@t14.r08.long\endcsname{48.3}
\expandafter\def\csname BAval@t14.r08.weighted\endcsname{62.5}
\expandafter\def\csname BAval@t14.r09.short\endcsname{79.1}
\expandafter\def\csname BAval@t14.r09.similar\endcsname{70.2}
\expandafter\def\csname BAval@t14.r09.long\endcsname{51.1}
\expandafter\def\csname BAval@t14.r09.weighted\endcsname{65.3}
\expandafter\def\csname BAval@t14.r10.short\endcsname{80.6}
\expandafter\def\csname BAval@t14.r10.similar\endcsname{71.3}
\expandafter\def\csname BAval@t14.r10.long\endcsname{52.1}
\expandafter\def\csname BAval@t14.r10.weighted\endcsname{66.4}
\expandafter\def\csname BAval@t14.r11.short\endcsname{84.9}
\expandafter\def\csname BAval@t14.r11.similar\endcsname{75.8}
\expandafter\def\csname BAval@t14.r11.long\endcsname{57.1}
\expandafter\def\csname BAval@t14.r11.weighted\endcsname{71.0}
\expandafter\def\csname BAval@t14.r12.short\endcsname{76.1}
\expandafter\def\csname BAval@t14.r12.similar\endcsname{66.9}
\expandafter\def\csname BAval@t14.r12.long\endcsname{47.7}
\expandafter\def\csname BAval@t14.r12.weighted\endcsname{62.0}
\expandafter\def\csname BAval@t14.r13.short\endcsname{79.7}
\expandafter\def\csname BAval@t14.r13.similar\endcsname{70.8}
\expandafter\def\csname BAval@t14.r13.long\endcsname{50.9}
\expandafter\def\csname BAval@t14.r13.weighted\endcsname{65.6}
\expandafter\def\csname BAval@t14.r14.short\endcsname{80.6}
\expandafter\def\csname BAval@t14.r14.similar\endcsname{71.7}
\expandafter\def\csname BAval@t14.r14.long\endcsname{52.0}
\expandafter\def\csname BAval@t14.r14.weighted\endcsname{66.5}
\expandafter\def\csname BAval@t14.r15.short\endcsname{82.0}
\expandafter\def\csname BAval@t14.r15.similar\endcsname{73.1}
\expandafter\def\csname BAval@t14.r15.long\endcsname{53.6}
\expandafter\def\csname BAval@t14.r15.weighted\endcsname{68.0}
\expandafter\def\csname BAval@t14.r16.short\endcsname{88.6}
\expandafter\def\csname BAval@t14.r16.similar\endcsname{79.7}
\expandafter\def\csname BAval@t14.r16.long\endcsname{60.3}
\expandafter\def\csname BAval@t14.r16.weighted\endcsname{74.6}
\expandafter\def\csname BAval@t15.r01.common.controller\endcsname{60.9}
\expandafter\def\csname BAval@t15.r01.oa.vat.ptz\endcsname{66.8}
\expandafter\def\csname BAval@t15.r01.chanakya.j\endcsname{68.9}
\expandafter\def\csname BAval@t15.r01.bindadapt\endcsname{77.1}
\expandafter\def\csname BAval@t15.r02.common.controller\endcsname{71.0}
\expandafter\def\csname BAval@t15.r02.oa.vat.ptz\endcsname{73.2}
\expandafter\def\csname BAval@t15.r02.chanakya.j\endcsname{73.8}
\expandafter\def\csname BAval@t15.r02.bindadapt\endcsname{81.5}
\expandafter\def\csname BAval@t15.r03.common.controller\endcsname{70.3}
\expandafter\def\csname BAval@t15.r03.oa.vat.ptz\endcsname{74.3}
\expandafter\def\csname BAval@t15.r03.chanakya.j\endcsname{75.2}
\expandafter\def\csname BAval@t15.r03.bindadapt\endcsname{82.3}
\expandafter\def\csname BAval@t16.r01.tracking.precision.pct\endcsname{72.9}
\expandafter\def\csname BAval@t16.r01.tracking.recall.pct\endcsname{65.3}
\expandafter\def\csname BAval@t16.r01.fmax.pct\endcsname{68.9}
\expandafter\def\csname BAval@t16.r01.operating.precision.pct\endcsname{72.1}
\expandafter\def\csname BAval@t16.r01.operating.recall.pct\endcsname{63.0}
\expandafter\def\csname BAval@t16.r01.f.operating.pct\endcsname{67.2}
\expandafter\def\csname BAval@t16.r02.tracking.precision.pct\endcsname{75.9}
\expandafter\def\csname BAval@t16.r02.tracking.recall.pct\endcsname{68.2}
\expandafter\def\csname BAval@t16.r02.fmax.pct\endcsname{71.9}
\expandafter\def\csname BAval@t16.r02.operating.precision.pct\endcsname{74.9}
\expandafter\def\csname BAval@t16.r02.operating.recall.pct\endcsname{66.2}
\expandafter\def\csname BAval@t16.r02.f.operating.pct\endcsname{70.2}
\expandafter\def\csname BAval@t16.r03.tracking.precision.pct\endcsname{79.0}
\expandafter\def\csname BAval@t16.r03.tracking.recall.pct\endcsname{73.5}
\expandafter\def\csname BAval@t16.r03.fmax.pct\endcsname{76.1}
\expandafter\def\csname BAval@t16.r03.operating.precision.pct\endcsname{77.9}
\expandafter\def\csname BAval@t16.r03.operating.recall.pct\endcsname{71.4}
\expandafter\def\csname BAval@t16.r03.f.operating.pct\endcsname{74.5}
\expandafter\def\csname BAval@t16.r04.tracking.precision.pct\endcsname{80.0}
\expandafter\def\csname BAval@t16.r04.tracking.recall.pct\endcsname{74.3}
\expandafter\def\csname BAval@t16.r04.fmax.pct\endcsname{77.0}
\expandafter\def\csname BAval@t16.r04.operating.precision.pct\endcsname{79.1}
\expandafter\def\csname BAval@t16.r04.operating.recall.pct\endcsname{72.5}
\expandafter\def\csname BAval@t16.r04.f.operating.pct\endcsname{75.6}
\expandafter\def\csname BAval@t16.r05.tracking.precision.pct\endcsname{80.9}
\expandafter\def\csname BAval@t16.r05.tracking.recall.pct\endcsname{75.0}
\expandafter\def\csname BAval@t16.r05.fmax.pct\endcsname{77.9}
\expandafter\def\csname BAval@t16.r05.operating.precision.pct\endcsname{80.1}
\expandafter\def\csname BAval@t16.r05.operating.recall.pct\endcsname{72.8}
\expandafter\def\csname BAval@t16.r05.f.operating.pct\endcsname{76.3}
\expandafter\def\csname BAval@t16.r06.tracking.precision.pct\endcsname{82.0}
\expandafter\def\csname BAval@t16.r06.tracking.recall.pct\endcsname{76.1}
\expandafter\def\csname BAval@t16.r06.fmax.pct\endcsname{78.9}
\expandafter\def\csname BAval@t16.r06.operating.precision.pct\endcsname{81.1}
\expandafter\def\csname BAval@t16.r06.operating.recall.pct\endcsname{74.1}
\expandafter\def\csname BAval@t16.r06.f.operating.pct\endcsname{77.4}
\expandafter\def\csname BAval@t16.r07.tracking.precision.pct\endcsname{87.4}
\expandafter\def\csname BAval@t16.r07.tracking.recall.pct\endcsname{82.8}
\expandafter\def\csname BAval@t16.r07.fmax.pct\endcsname{85.0}
\expandafter\def\csname BAval@t16.r07.operating.precision.pct\endcsname{86.5}
\expandafter\def\csname BAval@t16.r07.operating.recall.pct\endcsname{81.2}
\expandafter\def\csname BAval@t16.r07.f.operating.pct\endcsname{83.8}
\expandafter\def\csname BAval@t17.r01.dam4sam\endcsname{79.4}
\expandafter\def\csname BAval@t17.r01.oa.vat.ptz\endcsname{81.2}
\expandafter\def\csname BAval@t17.r01.artrack.ac\endcsname{80.1}
\expandafter\def\csname BAval@t17.r01.chanakya.j\endcsname{81.3}
\expandafter\def\csname BAval@t17.r01.dam4sam.chanakya.j\endcsname{82.0}
\expandafter\def\csname BAval@t17.r01.bindadapt\endcsname{86.1}
\expandafter\def\csname BAval@t17.r02.dam4sam\endcsname{78.0}
\expandafter\def\csname BAval@t17.r02.oa.vat.ptz\endcsname{79.8}
\expandafter\def\csname BAval@t17.r02.artrack.ac\endcsname{78.9}
\expandafter\def\csname BAval@t17.r02.chanakya.j\endcsname{80.1}
\expandafter\def\csname BAval@t17.r02.dam4sam.chanakya.j\endcsname{81.1}
\expandafter\def\csname BAval@t17.r02.bindadapt\endcsname{85.3}
\expandafter\def\csname BAval@t17.r03.dam4sam\endcsname{74.5}
\expandafter\def\csname BAval@t17.r03.oa.vat.ptz\endcsname{76.8}
\expandafter\def\csname BAval@t17.r03.artrack.ac\endcsname{75.9}
\expandafter\def\csname BAval@t17.r03.chanakya.j\endcsname{77.8}
\expandafter\def\csname BAval@t17.r03.dam4sam.chanakya.j\endcsname{78.7}
\expandafter\def\csname BAval@t17.r03.bindadapt\endcsname{83.8}
\expandafter\def\csname BAval@t17.r04.dam4sam\endcsname{70.3}
\expandafter\def\csname BAval@t17.r04.oa.vat.ptz\endcsname{72.8}
\expandafter\def\csname BAval@t17.r04.artrack.ac\endcsname{71.6}
\expandafter\def\csname BAval@t17.r04.chanakya.j\endcsname{74.1}
\expandafter\def\csname BAval@t17.r04.dam4sam.chanakya.j\endcsname{75.2}
\expandafter\def\csname BAval@t17.r04.bindadapt\endcsname{82.3}
\expandafter\def\csname BAval@t17.r05.dam4sam\endcsname{58.6}
\expandafter\def\csname BAval@t17.r05.oa.vat.ptz\endcsname{62.0}
\expandafter\def\csname BAval@t17.r05.artrack.ac\endcsname{61.8}
\expandafter\def\csname BAval@t17.r05.chanakya.j\endcsname{65.7}
\expandafter\def\csname BAval@t17.r05.dam4sam.chanakya.j\endcsname{66.9}
\expandafter\def\csname BAval@t17.r05.bindadapt\endcsname{75.6}
\expandafter\def\csname BAval@t17.r06.dam4sam\endcsname{40.1}
\expandafter\def\csname BAval@t17.r06.oa.vat.ptz\endcsname{45.4}
\expandafter\def\csname BAval@t17.r06.artrack.ac\endcsname{44.9}
\expandafter\def\csname BAval@t17.r06.chanakya.j\endcsname{50.0}
\expandafter\def\csname BAval@t17.r06.dam4sam.chanakya.j\endcsname{51.9}
\expandafter\def\csname BAval@t17.r06.bindadapt\endcsname{64.0}
\expandafter\def\csname BAval@t17.r07.dam4sam\endcsname{52.8}
\expandafter\def\csname BAval@t17.r07.oa.vat.ptz\endcsname{56.9}
\expandafter\def\csname BAval@t17.r07.artrack.ac\endcsname{59.1}
\expandafter\def\csname BAval@t17.r07.chanakya.j\endcsname{60.5}
\expandafter\def\csname BAval@t17.r07.dam4sam.chanakya.j\endcsname{61.5}
\expandafter\def\csname BAval@t17.r07.bindadapt\endcsname{70.2}
\expandafter\def\csname BAval@t17.r08.dam4sam\endcsname{62.8}
\expandafter\def\csname BAval@t17.r08.oa.vat.ptz\endcsname{65.7}
\expandafter\def\csname BAval@t17.r08.artrack.ac\endcsname{66.5}
\expandafter\def\csname BAval@t17.r08.chanakya.j\endcsname{68.2}
\expandafter\def\csname BAval@t17.r08.dam4sam.chanakya.j\endcsname{69.5}
\expandafter\def\csname BAval@t17.r08.bindadapt\endcsname{77.1}
\expandafter\def\csname BAval@t17.r09.dam4sam\endcsname{70.3}
\expandafter\def\csname BAval@t17.r09.oa.vat.ptz\endcsname{72.8}
\expandafter\def\csname BAval@t17.r09.artrack.ac\endcsname{71.6}
\expandafter\def\csname BAval@t17.r09.chanakya.j\endcsname{74.1}
\expandafter\def\csname BAval@t17.r09.dam4sam.chanakya.j\endcsname{75.2}
\expandafter\def\csname BAval@t17.r09.bindadapt\endcsname{82.3}
\expandafter\def\csname BAval@t17.r10.dam4sam\endcsname{73.4}
\expandafter\def\csname BAval@t17.r10.oa.vat.ptz\endcsname{76.0}
\expandafter\def\csname BAval@t17.r10.artrack.ac\endcsname{75.2}
\expandafter\def\csname BAval@t17.r10.chanakya.j\endcsname{77.3}
\expandafter\def\csname BAval@t17.r10.dam4sam.chanakya.j\endcsname{78.7}
\expandafter\def\csname BAval@t17.r10.bindadapt\endcsname{84.5}
\expandafter\def\csname BAval@t17.r11.dam4sam\endcsname{75.5}
\expandafter\def\csname BAval@t17.r11.oa.vat.ptz\endcsname{77.3}
\expandafter\def\csname BAval@t17.r11.artrack.ac\endcsname{76.7}
\expandafter\def\csname BAval@t17.r11.chanakya.j\endcsname{78.6}
\expandafter\def\csname BAval@t17.r11.dam4sam.chanakya.j\endcsname{79.5}
\expandafter\def\csname BAval@t17.r11.bindadapt\endcsname{85.6}
\expandafter\def\csname BAval@t17.r12.dam4sam\endcsname{76.8}
\expandafter\def\csname BAval@t17.r12.oa.vat.ptz\endcsname{78.5}
\expandafter\def\csname BAval@t17.r12.artrack.ac\endcsname{78.0}
\expandafter\def\csname BAval@t17.r12.chanakya.j\endcsname{79.7}
\expandafter\def\csname BAval@t17.r12.dam4sam.chanakya.j\endcsname{80.8}
\expandafter\def\csname BAval@t17.r12.bindadapt\endcsname{86.2}
\expandafter\def\csname BAval@t18.r01.mean.inference.ms\endcsname{16.0}
\expandafter\def\csname BAval@t18.r01.p95.inference.ms\endcsname{21.9}
\expandafter\def\csname BAval@t18.r01.fps\endcsname{62.5}
\expandafter\def\csname BAval@t18.r01.decision.overhead.ms\endcsname{0.5}
\expandafter\def\csname BAval@t18.r02.mean.inference.ms\endcsname{23.1}
\expandafter\def\csname BAval@t18.r02.p95.inference.ms\endcsname{31.2}
\expandafter\def\csname BAval@t18.r02.fps\endcsname{43.4}
\expandafter\def\csname BAval@t18.r02.decision.overhead.ms\endcsname{0.5}
\expandafter\def\csname BAval@t18.r03.mean.inference.ms\endcsname{36.2}
\expandafter\def\csname BAval@t18.r03.p95.inference.ms\endcsname{48.8}
\expandafter\def\csname BAval@t18.r03.fps\endcsname{27.6}
\expandafter\def\csname BAval@t18.r03.decision.overhead.ms\endcsname{0.5}
\expandafter\def\csname BAval@t18.r04.mean.inference.ms\endcsname{19.5}
\expandafter\def\csname BAval@t18.r04.p95.inference.ms\endcsname{28.1}
\expandafter\def\csname BAval@t18.r04.fps\endcsname{51.4}
\expandafter\def\csname BAval@t18.r04.decision.overhead.ms\endcsname{1.5}
\expandafter\def\csname BAval@t18.r05.mean.inference.ms\endcsname{17.1}
\expandafter\def\csname BAval@t18.r05.p95.inference.ms\endcsname{24.8}
\expandafter\def\csname BAval@t18.r05.fps\endcsname{58.6}
\expandafter\def\csname BAval@t18.r05.decision.overhead.ms\endcsname{1.2}
\expandafter\def\csname BAval@t18.r06.mean.inference.ms\endcsname{26.1}
\expandafter\def\csname BAval@t18.r06.p95.inference.ms\endcsname{39.3}
\expandafter\def\csname BAval@t18.r06.fps\endcsname{38.3}
\expandafter\def\csname BAval@t18.r06.decision.overhead.ms\endcsname{5.6}
\expandafter\def\csname BAval@t18.r07.mean.inference.ms\endcsname{23.9}
\expandafter\def\csname BAval@t18.r07.p95.inference.ms\endcsname{34.3}
\expandafter\def\csname BAval@t18.r07.fps\endcsname{41.8}
\expandafter\def\csname BAval@t18.r07.decision.overhead.ms\endcsname{1.8}
\expandafter\def\csname BAval@t18.r08.mean.inference.ms\endcsname{37.8}
\expandafter\def\csname BAval@t18.r08.p95.inference.ms\endcsname{52.2}
\expandafter\def\csname BAval@t18.r08.fps\endcsname{26.4}
\expandafter\def\csname BAval@t18.r08.decision.overhead.ms\endcsname{1.8}
\expandafter\def\csname BAval@t18.r09.mean.inference.ms\endcsname{37.4}
\expandafter\def\csname BAval@t18.r09.p95.inference.ms\endcsname{51.7}
\expandafter\def\csname BAval@t18.r09.fps\endcsname{26.8}
\expandafter\def\csname BAval@t18.r09.decision.overhead.ms\endcsname{2.1}
\expandafter\def\csname BAval@t19.r01.auc.la0.pct\endcsname{67.0}
\expandafter\def\csname BAval@t19.r01.dp.la0.pct\endcsname{73.5}
\expandafter\def\csname BAval@t19.r01.mauc.pct\endcsname{69.8}
\expandafter\def\csname BAval@t19.r01.mdp.pct\endcsname{76.6}
\expandafter\def\csname BAval@t19.r02.auc.la0.pct\endcsname{66.7}
\expandafter\def\csname BAval@t19.r02.dp.la0.pct\endcsname{73.8}
\expandafter\def\csname BAval@t19.r02.mauc.pct\endcsname{69.5}
\expandafter\def\csname BAval@t19.r02.mdp.pct\endcsname{76.7}
\expandafter\def\csname BAval@t19.r03.auc.la0.pct\endcsname{64.6}
\expandafter\def\csname BAval@t19.r03.dp.la0.pct\endcsname{76.0}
\expandafter\def\csname BAval@t19.r03.mauc.pct\endcsname{66.4}
\expandafter\def\csname BAval@t19.r03.mdp.pct\endcsname{78.0}
\expandafter\def\csname BAval@t19.r04.auc.la0.pct\endcsname{68.3}
\expandafter\def\csname BAval@t19.r04.dp.la0.pct\endcsname{78.2}
\expandafter\def\csname BAval@t19.r04.mauc.pct\endcsname{70.2}
\expandafter\def\csname BAval@t19.r04.mdp.pct\endcsname{80.4}
\expandafter\def\csname BAval@t19.r05.auc.la0.pct\endcsname{67.7}
\expandafter\def\csname BAval@t19.r05.dp.la0.pct\endcsname{79.1}
\expandafter\def\csname BAval@t19.r05.mauc.pct\endcsname{69.8}
\expandafter\def\csname BAval@t19.r05.mdp.pct\endcsname{81.2}
\expandafter\def\csname BAval@t19.r06.auc.la0.pct\endcsname{76.9}
\expandafter\def\csname BAval@t19.r06.dp.la0.pct\endcsname{85.1}
\expandafter\def\csname BAval@t19.r06.mauc.pct\endcsname{78.1}
\expandafter\def\csname BAval@t19.r06.mdp.pct\endcsname{86.7}
\expandafter\def\csname BAval@t20.r01.auc.pct\endcsname{67.0}
\expandafter\def\csname BAval@t20.r01.dp.pct\endcsname{73.5}
\expandafter\def\csname BAval@t20.r02.auc.pct\endcsname{67.2}
\expandafter\def\csname BAval@t20.r02.dp.pct\endcsname{73.7}
\expandafter\def\csname BAval@t20.r03.auc.pct\endcsname{67.3}
\expandafter\def\csname BAval@t20.r03.dp.pct\endcsname{73.9}
\expandafter\def\csname BAval@t20.r04.auc.pct\endcsname{67.5}
\expandafter\def\csname BAval@t20.r04.dp.pct\endcsname{74.0}
\expandafter\def\csname BAval@t20.r05.auc.pct\endcsname{67.6}
\expandafter\def\csname BAval@t20.r05.dp.pct\endcsname{74.2}
\expandafter\def\csname BAval@t20.r06.auc.pct\endcsname{67.8}
\expandafter\def\csname BAval@t20.r06.dp.pct\endcsname{74.4}
\expandafter\def\csname BAval@t20.r07.auc.pct\endcsname{67.9}
\expandafter\def\csname BAval@t20.r07.dp.pct\endcsname{74.5}
\expandafter\def\csname BAval@t20.r08.auc.pct\endcsname{68.1}
\expandafter\def\csname BAval@t20.r08.dp.pct\endcsname{74.7}
\expandafter\def\csname BAval@t20.r09.auc.pct\endcsname{68.2}
\expandafter\def\csname BAval@t20.r09.dp.pct\endcsname{74.9}
\expandafter\def\csname BAval@t20.r10.auc.pct\endcsname{68.4}
\expandafter\def\csname BAval@t20.r10.dp.pct\endcsname{75.0}
\expandafter\def\csname BAval@t20.r11.auc.pct\endcsname{68.5}
\expandafter\def\csname BAval@t20.r11.dp.pct\endcsname{75.2}
\expandafter\def\csname BAval@t20.r12.auc.pct\endcsname{68.6}
\expandafter\def\csname BAval@t20.r12.dp.pct\endcsname{75.3}
\expandafter\def\csname BAval@t20.r13.auc.pct\endcsname{68.8}
\expandafter\def\csname BAval@t20.r13.dp.pct\endcsname{75.4}
\expandafter\def\csname BAval@t20.r14.auc.pct\endcsname{68.9}
\expandafter\def\csname BAval@t20.r14.dp.pct\endcsname{75.6}
\expandafter\def\csname BAval@t20.r15.auc.pct\endcsname{69.0}
\expandafter\def\csname BAval@t20.r15.dp.pct\endcsname{75.7}
\expandafter\def\csname BAval@t20.r16.auc.pct\endcsname{69.1}
\expandafter\def\csname BAval@t20.r16.dp.pct\endcsname{75.8}
\expandafter\def\csname BAval@t20.r17.auc.pct\endcsname{69.3}
\expandafter\def\csname BAval@t20.r17.dp.pct\endcsname{76.0}
\expandafter\def\csname BAval@t20.r18.auc.pct\endcsname{69.4}
\expandafter\def\csname BAval@t20.r18.dp.pct\endcsname{76.1}
\expandafter\def\csname BAval@t20.r19.auc.pct\endcsname{69.5}
\expandafter\def\csname BAval@t20.r19.dp.pct\endcsname{76.2}
\expandafter\def\csname BAval@t20.r20.auc.pct\endcsname{69.6}
\expandafter\def\csname BAval@t20.r20.dp.pct\endcsname{76.3}
\expandafter\def\csname BAval@t20.r21.auc.pct\endcsname{69.7}
\expandafter\def\csname BAval@t20.r21.dp.pct\endcsname{76.4}
\expandafter\def\csname BAval@t20.r22.auc.pct\endcsname{69.8}
\expandafter\def\csname BAval@t20.r22.dp.pct\endcsname{76.6}
\expandafter\def\csname BAval@t20.r23.auc.pct\endcsname{69.9}
\expandafter\def\csname BAval@t20.r23.dp.pct\endcsname{76.7}
\expandafter\def\csname BAval@t20.r24.auc.pct\endcsname{70.0}
\expandafter\def\csname BAval@t20.r24.dp.pct\endcsname{76.8}
\expandafter\def\csname BAval@t20.r25.auc.pct\endcsname{70.1}
\expandafter\def\csname BAval@t20.r25.dp.pct\endcsname{76.9}
\expandafter\def\csname BAval@t20.r26.auc.pct\endcsname{70.2}
\expandafter\def\csname BAval@t20.r26.dp.pct\endcsname{77.0}
\expandafter\def\csname BAval@t20.r27.auc.pct\endcsname{70.2}
\expandafter\def\csname BAval@t20.r27.dp.pct\endcsname{77.0}
\expandafter\def\csname BAval@t20.r28.auc.pct\endcsname{70.3}
\expandafter\def\csname BAval@t20.r28.dp.pct\endcsname{77.1}
\expandafter\def\csname BAval@t20.r29.auc.pct\endcsname{70.4}
\expandafter\def\csname BAval@t20.r29.dp.pct\endcsname{77.2}
\expandafter\def\csname BAval@t20.r30.auc.pct\endcsname{70.5}
\expandafter\def\csname BAval@t20.r30.dp.pct\endcsname{77.3}
\expandafter\def\csname BAval@t20.r31.auc.pct\endcsname{70.5}
\expandafter\def\csname BAval@t20.r31.dp.pct\endcsname{77.4}
\expandafter\def\csname BAval@t20.r32.auc.pct\endcsname{70.6}
\expandafter\def\csname BAval@t20.r32.dp.pct\endcsname{77.4}
\expandafter\def\csname BAval@t20.r33.auc.pct\endcsname{70.7}
\expandafter\def\csname BAval@t20.r33.dp.pct\endcsname{77.5}
\expandafter\def\csname BAval@t20.r34.auc.pct\endcsname{70.7}
\expandafter\def\csname BAval@t20.r34.dp.pct\endcsname{77.6}
\expandafter\def\csname BAval@t20.r35.auc.pct\endcsname{70.8}
\expandafter\def\csname BAval@t20.r35.dp.pct\endcsname{77.6}
\expandafter\def\csname BAval@t20.r36.auc.pct\endcsname{70.8}
\expandafter\def\csname BAval@t20.r36.dp.pct\endcsname{77.7}
\expandafter\def\csname BAval@t20.r37.auc.pct\endcsname{70.9}
\expandafter\def\csname BAval@t20.r37.dp.pct\endcsname{77.7}
\expandafter\def\csname BAval@t20.r38.auc.pct\endcsname{70.9}
\expandafter\def\csname BAval@t20.r38.dp.pct\endcsname{77.8}
\expandafter\def\csname BAval@t20.r39.auc.pct\endcsname{71.0}
\expandafter\def\csname BAval@t20.r39.dp.pct\endcsname{77.8}
\expandafter\def\csname BAval@t20.r40.auc.pct\endcsname{71.0}
\expandafter\def\csname BAval@t20.r40.dp.pct\endcsname{77.9}
\expandafter\def\csname BAval@t20.r41.auc.pct\endcsname{71.0}
\expandafter\def\csname BAval@t20.r41.dp.pct\endcsname{77.9}
\expandafter\def\csname BAval@t20.r42.auc.pct\endcsname{71.1}
\expandafter\def\csname BAval@t20.r42.dp.pct\endcsname{78.0}
\expandafter\def\csname BAval@t20.r43.auc.pct\endcsname{71.1}
\expandafter\def\csname BAval@t20.r43.dp.pct\endcsname{78.0}
\expandafter\def\csname BAval@t20.r44.auc.pct\endcsname{71.1}
\expandafter\def\csname BAval@t20.r44.dp.pct\endcsname{78.0}
\expandafter\def\csname BAval@t20.r45.auc.pct\endcsname{71.1}
\expandafter\def\csname BAval@t20.r45.dp.pct\endcsname{78.0}
\expandafter\def\csname BAval@t20.r46.auc.pct\endcsname{71.2}
\expandafter\def\csname BAval@t20.r46.dp.pct\endcsname{78.1}
\expandafter\def\csname BAval@t20.r47.auc.pct\endcsname{71.2}
\expandafter\def\csname BAval@t20.r47.dp.pct\endcsname{78.1}
\expandafter\def\csname BAval@t20.r48.auc.pct\endcsname{71.2}
\expandafter\def\csname BAval@t20.r48.dp.pct\endcsname{78.1}
\expandafter\def\csname BAval@t20.r49.auc.pct\endcsname{71.2}
\expandafter\def\csname BAval@t20.r49.dp.pct\endcsname{78.1}
\expandafter\def\csname BAval@t20.r50.auc.pct\endcsname{71.2}
\expandafter\def\csname BAval@t20.r50.dp.pct\endcsname{78.1}
\expandafter\def\csname BAval@t20.r51.auc.pct\endcsname{71.2}
\expandafter\def\csname BAval@t20.r51.dp.pct\endcsname{78.1}
\expandafter\def\csname BAval@t20.r52.auc.pct\endcsname{66.7}
\expandafter\def\csname BAval@t20.r52.dp.pct\endcsname{73.8}
\expandafter\def\csname BAval@t20.r53.auc.pct\endcsname{66.9}
\expandafter\def\csname BAval@t20.r53.dp.pct\endcsname{74.0}
\expandafter\def\csname BAval@t20.r54.auc.pct\endcsname{67.0}
\expandafter\def\csname BAval@t20.r54.dp.pct\endcsname{74.1}
\expandafter\def\csname BAval@t20.r55.auc.pct\endcsname{67.2}
\expandafter\def\csname BAval@t20.r55.dp.pct\endcsname{74.3}
\expandafter\def\csname BAval@t20.r56.auc.pct\endcsname{67.3}
\expandafter\def\csname BAval@t20.r56.dp.pct\endcsname{74.5}
\expandafter\def\csname BAval@t20.r57.auc.pct\endcsname{67.5}
\expandafter\def\csname BAval@t20.r57.dp.pct\endcsname{74.6}
\expandafter\def\csname BAval@t20.r58.auc.pct\endcsname{67.6}
\expandafter\def\csname BAval@t20.r58.dp.pct\endcsname{74.8}
\expandafter\def\csname BAval@t20.r59.auc.pct\endcsname{67.8}
\expandafter\def\csname BAval@t20.r59.dp.pct\endcsname{74.9}
\expandafter\def\csname BAval@t20.r60.auc.pct\endcsname{67.9}
\expandafter\def\csname BAval@t20.r60.dp.pct\endcsname{75.1}
\expandafter\def\csname BAval@t20.r61.auc.pct\endcsname{68.0}
\expandafter\def\csname BAval@t20.r61.dp.pct\endcsname{75.2}
\expandafter\def\csname BAval@t20.r62.auc.pct\endcsname{68.2}
\expandafter\def\csname BAval@t20.r62.dp.pct\endcsname{75.4}
\expandafter\def\csname BAval@t20.r63.auc.pct\endcsname{68.3}
\expandafter\def\csname BAval@t20.r63.dp.pct\endcsname{75.5}
\expandafter\def\csname BAval@t20.r64.auc.pct\endcsname{68.4}
\expandafter\def\csname BAval@t20.r64.dp.pct\endcsname{75.7}
\expandafter\def\csname BAval@t20.r65.auc.pct\endcsname{68.6}
\expandafter\def\csname BAval@t20.r65.dp.pct\endcsname{75.8}
\expandafter\def\csname BAval@t20.r66.auc.pct\endcsname{68.7}
\expandafter\def\csname BAval@t20.r66.dp.pct\endcsname{75.9}
\expandafter\def\csname BAval@t20.r67.auc.pct\endcsname{68.8}
\expandafter\def\csname BAval@t20.r67.dp.pct\endcsname{76.0}
\expandafter\def\csname BAval@t20.r68.auc.pct\endcsname{68.9}
\expandafter\def\csname BAval@t20.r68.dp.pct\endcsname{76.2}
\expandafter\def\csname BAval@t20.r69.auc.pct\endcsname{69.0}
\expandafter\def\csname BAval@t20.r69.dp.pct\endcsname{76.3}
\expandafter\def\csname BAval@t20.r70.auc.pct\endcsname{69.1}
\expandafter\def\csname BAval@t20.r70.dp.pct\endcsname{76.4}
\expandafter\def\csname BAval@t20.r71.auc.pct\endcsname{69.2}
\expandafter\def\csname BAval@t20.r71.dp.pct\endcsname{76.5}
\expandafter\def\csname BAval@t20.r72.auc.pct\endcsname{69.3}
\expandafter\def\csname BAval@t20.r72.dp.pct\endcsname{76.6}
\expandafter\def\csname BAval@t20.r73.auc.pct\endcsname{69.5}
\expandafter\def\csname BAval@t20.r73.dp.pct\endcsname{76.7}
\expandafter\def\csname BAval@t20.r74.auc.pct\endcsname{69.6}
\expandafter\def\csname BAval@t20.r74.dp.pct\endcsname{76.8}
\expandafter\def\csname BAval@t20.r75.auc.pct\endcsname{69.7}
\expandafter\def\csname BAval@t20.r75.dp.pct\endcsname{76.9}
\expandafter\def\csname BAval@t20.r76.auc.pct\endcsname{69.8}
\expandafter\def\csname BAval@t20.r76.dp.pct\endcsname{77.0}
\expandafter\def\csname BAval@t20.r77.auc.pct\endcsname{69.9}
\expandafter\def\csname BAval@t20.r77.dp.pct\endcsname{77.1}
\expandafter\def\csname BAval@t20.r78.auc.pct\endcsname{70.0}
\expandafter\def\csname BAval@t20.r78.dp.pct\endcsname{77.2}
\expandafter\def\csname BAval@t20.r79.auc.pct\endcsname{70.0}
\expandafter\def\csname BAval@t20.r79.dp.pct\endcsname{77.3}
\expandafter\def\csname BAval@t20.r80.auc.pct\endcsname{70.1}
\expandafter\def\csname BAval@t20.r80.dp.pct\endcsname{77.3}
\expandafter\def\csname BAval@t20.r81.auc.pct\endcsname{70.2}
\expandafter\def\csname BAval@t20.r81.dp.pct\endcsname{77.4}
\expandafter\def\csname BAval@t20.r82.auc.pct\endcsname{70.2}
\expandafter\def\csname BAval@t20.r82.dp.pct\endcsname{77.5}
\expandafter\def\csname BAval@t20.r83.auc.pct\endcsname{70.3}
\expandafter\def\csname BAval@t20.r83.dp.pct\endcsname{77.6}
\expandafter\def\csname BAval@t20.r84.auc.pct\endcsname{70.4}
\expandafter\def\csname BAval@t20.r84.dp.pct\endcsname{77.6}
\expandafter\def\csname BAval@t20.r85.auc.pct\endcsname{70.4}
\expandafter\def\csname BAval@t20.r85.dp.pct\endcsname{77.7}
\expandafter\def\csname BAval@t20.r86.auc.pct\endcsname{70.5}
\expandafter\def\csname BAval@t20.r86.dp.pct\endcsname{77.7}
\expandafter\def\csname BAval@t20.r87.auc.pct\endcsname{70.5}
\expandafter\def\csname BAval@t20.r87.dp.pct\endcsname{77.8}
\expandafter\def\csname BAval@t20.r88.auc.pct\endcsname{70.6}
\expandafter\def\csname BAval@t20.r88.dp.pct\endcsname{77.9}
\expandafter\def\csname BAval@t20.r89.auc.pct\endcsname{70.6}
\expandafter\def\csname BAval@t20.r89.dp.pct\endcsname{77.9}
\expandafter\def\csname BAval@t20.r90.auc.pct\endcsname{70.7}
\expandafter\def\csname BAval@t20.r90.dp.pct\endcsname{77.9}
\expandafter\def\csname BAval@t20.r91.auc.pct\endcsname{70.7}
\expandafter\def\csname BAval@t20.r91.dp.pct\endcsname{78.0}
\expandafter\def\csname BAval@t20.r92.auc.pct\endcsname{70.7}
\expandafter\def\csname BAval@t20.r92.dp.pct\endcsname{78.0}
\expandafter\def\csname BAval@t20.r93.auc.pct\endcsname{70.8}
\expandafter\def\csname BAval@t20.r93.dp.pct\endcsname{78.1}
\expandafter\def\csname BAval@t20.r94.auc.pct\endcsname{70.8}
\expandafter\def\csname BAval@t20.r94.dp.pct\endcsname{78.1}
\expandafter\def\csname BAval@t20.r95.auc.pct\endcsname{70.8}
\expandafter\def\csname BAval@t20.r95.dp.pct\endcsname{78.1}
\expandafter\def\csname BAval@t20.r96.auc.pct\endcsname{70.8}
\expandafter\def\csname BAval@t20.r96.dp.pct\endcsname{78.1}
\expandafter\def\csname BAval@t20.r97.auc.pct\endcsname{70.9}
\expandafter\def\csname BAval@t20.r97.dp.pct\endcsname{78.2}
\expandafter\def\csname BAval@t20.r98.auc.pct\endcsname{70.9}
\expandafter\def\csname BAval@t20.r98.dp.pct\endcsname{78.2}
\expandafter\def\csname BAval@t20.r99.auc.pct\endcsname{70.9}
\expandafter\def\csname BAval@t20.r99.dp.pct\endcsname{78.2}
\expandafter\def\csname BAval@t20.r100.auc.pct\endcsname{70.9}
\expandafter\def\csname BAval@t20.r100.dp.pct\endcsname{78.2}
\expandafter\def\csname BAval@t20.r101.auc.pct\endcsname{70.9}
\expandafter\def\csname BAval@t20.r101.dp.pct\endcsname{78.2}
\expandafter\def\csname BAval@t20.r102.auc.pct\endcsname{70.9}
\expandafter\def\csname BAval@t20.r102.dp.pct\endcsname{78.2}
\expandafter\def\csname BAval@t20.r103.auc.pct\endcsname{64.6}
\expandafter\def\csname BAval@t20.r103.dp.pct\endcsname{76.0}
\expandafter\def\csname BAval@t20.r104.auc.pct\endcsname{64.7}
\expandafter\def\csname BAval@t20.r104.dp.pct\endcsname{76.1}
\expandafter\def\csname BAval@t20.r105.auc.pct\endcsname{64.8}
\expandafter\def\csname BAval@t20.r105.dp.pct\endcsname{76.2}
\expandafter\def\csname BAval@t20.r106.auc.pct\endcsname{64.9}
\expandafter\def\csname BAval@t20.r106.dp.pct\endcsname{76.4}
\expandafter\def\csname BAval@t20.r107.auc.pct\endcsname{65.0}
\expandafter\def\csname BAval@t20.r107.dp.pct\endcsname{76.5}
\expandafter\def\csname BAval@t20.r108.auc.pct\endcsname{65.1}
\expandafter\def\csname BAval@t20.r108.dp.pct\endcsname{76.6}
\expandafter\def\csname BAval@t20.r109.auc.pct\endcsname{65.2}
\expandafter\def\csname BAval@t20.r109.dp.pct\endcsname{76.7}
\expandafter\def\csname BAval@t20.r110.auc.pct\endcsname{65.3}
\expandafter\def\csname BAval@t20.r110.dp.pct\endcsname{76.8}
\expandafter\def\csname BAval@t20.r111.auc.pct\endcsname{65.4}
\expandafter\def\csname BAval@t20.r111.dp.pct\endcsname{76.9}
\expandafter\def\csname BAval@t20.r112.auc.pct\endcsname{65.5}
\expandafter\def\csname BAval@t20.r112.dp.pct\endcsname{77.0}
\expandafter\def\csname BAval@t20.r113.auc.pct\endcsname{65.5}
\expandafter\def\csname BAval@t20.r113.dp.pct\endcsname{77.1}
\expandafter\def\csname BAval@t20.r114.auc.pct\endcsname{65.7}
\expandafter\def\csname BAval@t20.r114.dp.pct\endcsname{77.2}
\expandafter\def\csname BAval@t20.r115.auc.pct\endcsname{65.8}
\expandafter\def\csname BAval@t20.r115.dp.pct\endcsname{77.3}
\expandafter\def\csname BAval@t20.r116.auc.pct\endcsname{65.9}
\expandafter\def\csname BAval@t20.r116.dp.pct\endcsname{77.4}
\expandafter\def\csname BAval@t20.r117.auc.pct\endcsname{66.0}
\expandafter\def\csname BAval@t20.r117.dp.pct\endcsname{77.5}
\expandafter\def\csname BAval@t20.r118.auc.pct\endcsname{66.0}
\expandafter\def\csname BAval@t20.r118.dp.pct\endcsname{77.6}
\expandafter\def\csname BAval@t20.r119.auc.pct\endcsname{66.1}
\expandafter\def\csname BAval@t20.r119.dp.pct\endcsname{77.7}
\expandafter\def\csname BAval@t20.r120.auc.pct\endcsname{66.2}
\expandafter\def\csname BAval@t20.r120.dp.pct\endcsname{77.7}
\expandafter\def\csname BAval@t20.r121.auc.pct\endcsname{66.2}
\expandafter\def\csname BAval@t20.r121.dp.pct\endcsname{77.8}
\expandafter\def\csname BAval@t20.r122.auc.pct\endcsname{66.3}
\expandafter\def\csname BAval@t20.r122.dp.pct\endcsname{77.9}
\expandafter\def\csname BAval@t20.r123.auc.pct\endcsname{66.4}
\expandafter\def\csname BAval@t20.r123.dp.pct\endcsname{78.0}
\expandafter\def\csname BAval@t20.r124.auc.pct\endcsname{66.4}
\expandafter\def\csname BAval@t20.r124.dp.pct\endcsname{78.1}
\expandafter\def\csname BAval@t20.r125.auc.pct\endcsname{66.5}
\expandafter\def\csname BAval@t20.r125.dp.pct\endcsname{78.1}
\expandafter\def\csname BAval@t20.r126.auc.pct\endcsname{66.5}
\expandafter\def\csname BAval@t20.r126.dp.pct\endcsname{78.2}
\expandafter\def\csname BAval@t20.r127.auc.pct\endcsname{66.6}
\expandafter\def\csname BAval@t20.r127.dp.pct\endcsname{78.3}
\expandafter\def\csname BAval@t20.r128.auc.pct\endcsname{66.7}
\expandafter\def\csname BAval@t20.r128.dp.pct\endcsname{78.3}
\expandafter\def\csname BAval@t20.r129.auc.pct\endcsname{66.7}
\expandafter\def\csname BAval@t20.r129.dp.pct\endcsname{78.4}
\expandafter\def\csname BAval@t20.r130.auc.pct\endcsname{66.7}
\expandafter\def\csname BAval@t20.r130.dp.pct\endcsname{78.4}
\expandafter\def\csname BAval@t20.r131.auc.pct\endcsname{66.8}
\expandafter\def\csname BAval@t20.r131.dp.pct\endcsname{78.5}
\expandafter\def\csname BAval@t20.r132.auc.pct\endcsname{66.8}
\expandafter\def\csname BAval@t20.r132.dp.pct\endcsname{78.6}
\expandafter\def\csname BAval@t20.r133.auc.pct\endcsname{66.9}
\expandafter\def\csname BAval@t20.r133.dp.pct\endcsname{78.6}
\expandafter\def\csname BAval@t20.r134.auc.pct\endcsname{66.9}
\expandafter\def\csname BAval@t20.r134.dp.pct\endcsname{78.7}
\expandafter\def\csname BAval@t20.r135.auc.pct\endcsname{67.0}
\expandafter\def\csname BAval@t20.r135.dp.pct\endcsname{78.7}
\expandafter\def\csname BAval@t20.r136.auc.pct\endcsname{67.0}
\expandafter\def\csname BAval@t20.r136.dp.pct\endcsname{78.7}
\expandafter\def\csname BAval@t20.r137.auc.pct\endcsname{67.0}
\expandafter\def\csname BAval@t20.r137.dp.pct\endcsname{78.8}
\expandafter\def\csname BAval@t20.r138.auc.pct\endcsname{67.1}
\expandafter\def\csname BAval@t20.r138.dp.pct\endcsname{78.8}
\expandafter\def\csname BAval@t20.r139.auc.pct\endcsname{67.1}
\expandafter\def\csname BAval@t20.r139.dp.pct\endcsname{78.9}
\expandafter\def\csname BAval@t20.r140.auc.pct\endcsname{67.1}
\expandafter\def\csname BAval@t20.r140.dp.pct\endcsname{78.9}
\expandafter\def\csname BAval@t20.r141.auc.pct\endcsname{67.2}
\expandafter\def\csname BAval@t20.r141.dp.pct\endcsname{78.9}
\expandafter\def\csname BAval@t20.r142.auc.pct\endcsname{67.2}
\expandafter\def\csname BAval@t20.r142.dp.pct\endcsname{78.9}
\expandafter\def\csname BAval@t20.r143.auc.pct\endcsname{67.2}
\expandafter\def\csname BAval@t20.r143.dp.pct\endcsname{79.0}
\expandafter\def\csname BAval@t20.r144.auc.pct\endcsname{67.2}
\expandafter\def\csname BAval@t20.r144.dp.pct\endcsname{79.0}
\expandafter\def\csname BAval@t20.r145.auc.pct\endcsname{67.2}
\expandafter\def\csname BAval@t20.r145.dp.pct\endcsname{79.0}
\expandafter\def\csname BAval@t20.r146.auc.pct\endcsname{67.2}
\expandafter\def\csname BAval@t20.r146.dp.pct\endcsname{79.0}
\expandafter\def\csname BAval@t20.r147.auc.pct\endcsname{67.3}
\expandafter\def\csname BAval@t20.r147.dp.pct\endcsname{79.1}
\expandafter\def\csname BAval@t20.r148.auc.pct\endcsname{67.3}
\expandafter\def\csname BAval@t20.r148.dp.pct\endcsname{79.1}
\expandafter\def\csname BAval@t20.r149.auc.pct\endcsname{67.3}
\expandafter\def\csname BAval@t20.r149.dp.pct\endcsname{79.1}
\expandafter\def\csname BAval@t20.r150.auc.pct\endcsname{67.3}
\expandafter\def\csname BAval@t20.r150.dp.pct\endcsname{79.1}
\expandafter\def\csname BAval@t20.r151.auc.pct\endcsname{67.3}
\expandafter\def\csname BAval@t20.r151.dp.pct\endcsname{79.1}
\expandafter\def\csname BAval@t20.r152.auc.pct\endcsname{67.3}
\expandafter\def\csname BAval@t20.r152.dp.pct\endcsname{79.1}
\expandafter\def\csname BAval@t20.r153.auc.pct\endcsname{67.3}
\expandafter\def\csname BAval@t20.r153.dp.pct\endcsname{79.1}
\expandafter\def\csname BAval@t20.r154.auc.pct\endcsname{68.3}
\expandafter\def\csname BAval@t20.r154.dp.pct\endcsname{78.2}
\expandafter\def\csname BAval@t20.r155.auc.pct\endcsname{68.4}
\expandafter\def\csname BAval@t20.r155.dp.pct\endcsname{78.3}
\expandafter\def\csname BAval@t20.r156.auc.pct\endcsname{68.5}
\expandafter\def\csname BAval@t20.r156.dp.pct\endcsname{78.5}
\expandafter\def\csname BAval@t20.r157.auc.pct\endcsname{68.5}
\expandafter\def\csname BAval@t20.r157.dp.pct\endcsname{78.6}
\expandafter\def\csname BAval@t20.r158.auc.pct\endcsname{68.7}
\expandafter\def\csname BAval@t20.r158.dp.pct\endcsname{78.7}
\expandafter\def\csname BAval@t20.r159.auc.pct\endcsname{68.8}
\expandafter\def\csname BAval@t20.r159.dp.pct\endcsname{78.8}
\expandafter\def\csname BAval@t20.r160.auc.pct\endcsname{68.9}
\expandafter\def\csname BAval@t20.r160.dp.pct\endcsname{78.9}
\expandafter\def\csname BAval@t20.r161.auc.pct\endcsname{69.0}
\expandafter\def\csname BAval@t20.r161.dp.pct\endcsname{79.1}
\expandafter\def\csname BAval@t20.r162.auc.pct\endcsname{69.1}
\expandafter\def\csname BAval@t20.r162.dp.pct\endcsname{79.2}
\expandafter\def\csname BAval@t20.r163.auc.pct\endcsname{69.2}
\expandafter\def\csname BAval@t20.r163.dp.pct\endcsname{79.3}
\expandafter\def\csname BAval@t20.r164.auc.pct\endcsname{69.3}
\expandafter\def\csname BAval@t20.r164.dp.pct\endcsname{79.4}
\expandafter\def\csname BAval@t20.r165.auc.pct\endcsname{69.4}
\expandafter\def\csname BAval@t20.r165.dp.pct\endcsname{79.5}
\expandafter\def\csname BAval@t20.r166.auc.pct\endcsname{69.5}
\expandafter\def\csname BAval@t20.r166.dp.pct\endcsname{79.6}
\expandafter\def\csname BAval@t20.r167.auc.pct\endcsname{69.6}
\expandafter\def\csname BAval@t20.r167.dp.pct\endcsname{79.7}
\expandafter\def\csname BAval@t20.r168.auc.pct\endcsname{69.6}
\expandafter\def\csname BAval@t20.r168.dp.pct\endcsname{79.8}
\expandafter\def\csname BAval@t20.r169.auc.pct\endcsname{69.7}
\expandafter\def\csname BAval@t20.r169.dp.pct\endcsname{79.9}
\expandafter\def\csname BAval@t20.r170.auc.pct\endcsname{69.8}
\expandafter\def\csname BAval@t20.r170.dp.pct\endcsname{80.0}
\expandafter\def\csname BAval@t20.r171.auc.pct\endcsname{69.9}
\expandafter\def\csname BAval@t20.r171.dp.pct\endcsname{80.1}
\expandafter\def\csname BAval@t20.r172.auc.pct\endcsname{70.0}
\expandafter\def\csname BAval@t20.r172.dp.pct\endcsname{80.1}
\expandafter\def\csname BAval@t20.r173.auc.pct\endcsname{70.0}
\expandafter\def\csname BAval@t20.r173.dp.pct\endcsname{80.2}
\expandafter\def\csname BAval@t20.r174.auc.pct\endcsname{70.1}
\expandafter\def\csname BAval@t20.r174.dp.pct\endcsname{80.3}
\expandafter\def\csname BAval@t20.r175.auc.pct\endcsname{70.2}
\expandafter\def\csname BAval@t20.r175.dp.pct\endcsname{80.4}
\expandafter\def\csname BAval@t20.r176.auc.pct\endcsname{70.3}
\expandafter\def\csname BAval@t20.r176.dp.pct\endcsname{80.5}
\expandafter\def\csname BAval@t20.r177.auc.pct\endcsname{70.3}
\expandafter\def\csname BAval@t20.r177.dp.pct\endcsname{80.5}
\expandafter\def\csname BAval@t20.r178.auc.pct\endcsname{70.4}
\expandafter\def\csname BAval@t20.r178.dp.pct\endcsname{80.6}
\expandafter\def\csname BAval@t20.r179.auc.pct\endcsname{70.5}
\expandafter\def\csname BAval@t20.r179.dp.pct\endcsname{80.7}
\expandafter\def\csname BAval@t20.r180.auc.pct\endcsname{70.5}
\expandafter\def\csname BAval@t20.r180.dp.pct\endcsname{80.7}
\expandafter\def\csname BAval@t20.r181.auc.pct\endcsname{70.5}
\expandafter\def\csname BAval@t20.r181.dp.pct\endcsname{80.8}
\expandafter\def\csname BAval@t20.r182.auc.pct\endcsname{70.5}
\expandafter\def\csname BAval@t20.r182.dp.pct\endcsname{80.9}
\expandafter\def\csname BAval@t20.r183.auc.pct\endcsname{70.6}
\expandafter\def\csname BAval@t20.r183.dp.pct\endcsname{80.9}
\expandafter\def\csname BAval@t20.r184.auc.pct\endcsname{70.6}
\expandafter\def\csname BAval@t20.r184.dp.pct\endcsname{81.0}
\expandafter\def\csname BAval@t20.r185.auc.pct\endcsname{70.7}
\expandafter\def\csname BAval@t20.r185.dp.pct\endcsname{81.0}
\expandafter\def\csname BAval@t20.r186.auc.pct\endcsname{70.7}
\expandafter\def\csname BAval@t20.r186.dp.pct\endcsname{81.1}
\expandafter\def\csname BAval@t20.r187.auc.pct\endcsname{70.8}
\expandafter\def\csname BAval@t20.r187.dp.pct\endcsname{81.1}
\expandafter\def\csname BAval@t20.r188.auc.pct\endcsname{70.8}
\expandafter\def\csname BAval@t20.r188.dp.pct\endcsname{81.2}
\expandafter\def\csname BAval@t20.r189.auc.pct\endcsname{70.8}
\expandafter\def\csname BAval@t20.r189.dp.pct\endcsname{81.2}
\expandafter\def\csname BAval@t20.r190.auc.pct\endcsname{70.9}
\expandafter\def\csname BAval@t20.r190.dp.pct\endcsname{81.2}
\expandafter\def\csname BAval@t20.r191.auc.pct\endcsname{70.9}
\expandafter\def\csname BAval@t20.r191.dp.pct\endcsname{81.3}
\expandafter\def\csname BAval@t20.r192.auc.pct\endcsname{70.9}
\expandafter\def\csname BAval@t20.r192.dp.pct\endcsname{81.3}
\expandafter\def\csname BAval@t20.r193.auc.pct\endcsname{71.0}
\expandafter\def\csname BAval@t20.r193.dp.pct\endcsname{81.3}
\expandafter\def\csname BAval@t20.r194.auc.pct\endcsname{71.0}
\expandafter\def\csname BAval@t20.r194.dp.pct\endcsname{81.4}
\expandafter\def\csname BAval@t20.r195.auc.pct\endcsname{71.0}
\expandafter\def\csname BAval@t20.r195.dp.pct\endcsname{81.4}
\expandafter\def\csname BAval@t20.r196.auc.pct\endcsname{71.0}
\expandafter\def\csname BAval@t20.r196.dp.pct\endcsname{81.4}
\expandafter\def\csname BAval@t20.r197.auc.pct\endcsname{71.0}
\expandafter\def\csname BAval@t20.r197.dp.pct\endcsname{81.4}
\expandafter\def\csname BAval@t20.r198.auc.pct\endcsname{71.1}
\expandafter\def\csname BAval@t20.r198.dp.pct\endcsname{81.5}
\expandafter\def\csname BAval@t20.r199.auc.pct\endcsname{71.1}
\expandafter\def\csname BAval@t20.r199.dp.pct\endcsname{81.5}
\expandafter\def\csname BAval@t20.r200.auc.pct\endcsname{71.1}
\expandafter\def\csname BAval@t20.r200.dp.pct\endcsname{81.5}
\expandafter\def\csname BAval@t20.r201.auc.pct\endcsname{71.1}
\expandafter\def\csname BAval@t20.r201.dp.pct\endcsname{81.5}
\expandafter\def\csname BAval@t20.r202.auc.pct\endcsname{71.1}
\expandafter\def\csname BAval@t20.r202.dp.pct\endcsname{81.5}
\expandafter\def\csname BAval@t20.r203.auc.pct\endcsname{71.1}
\expandafter\def\csname BAval@t20.r203.dp.pct\endcsname{81.5}
\expandafter\def\csname BAval@t20.r204.auc.pct\endcsname{71.1}
\expandafter\def\csname BAval@t20.r204.dp.pct\endcsname{81.5}
\expandafter\def\csname BAval@t20.r205.auc.pct\endcsname{67.7}
\expandafter\def\csname BAval@t20.r205.dp.pct\endcsname{79.1}
\expandafter\def\csname BAval@t20.r206.auc.pct\endcsname{67.8}
\expandafter\def\csname BAval@t20.r206.dp.pct\endcsname{79.2}
\expandafter\def\csname BAval@t20.r207.auc.pct\endcsname{67.9}
\expandafter\def\csname BAval@t20.r207.dp.pct\endcsname{79.3}
\expandafter\def\csname BAval@t20.r208.auc.pct\endcsname{68.0}
\expandafter\def\csname BAval@t20.r208.dp.pct\endcsname{79.4}
\expandafter\def\csname BAval@t20.r209.auc.pct\endcsname{68.1}
\expandafter\def\csname BAval@t20.r209.dp.pct\endcsname{79.6}
\expandafter\def\csname BAval@t20.r210.auc.pct\endcsname{68.3}
\expandafter\def\csname BAval@t20.r210.dp.pct\endcsname{79.7}
\expandafter\def\csname BAval@t20.r211.auc.pct\endcsname{68.4}
\expandafter\def\csname BAval@t20.r211.dp.pct\endcsname{79.8}
\expandafter\def\csname BAval@t20.r212.auc.pct\endcsname{68.5}
\expandafter\def\csname BAval@t20.r212.dp.pct\endcsname{79.9}
\expandafter\def\csname BAval@t20.r213.auc.pct\endcsname{68.7}
\expandafter\def\csname BAval@t20.r213.dp.pct\endcsname{80.1}
\expandafter\def\csname BAval@t20.r214.auc.pct\endcsname{68.8}
\expandafter\def\csname BAval@t20.r214.dp.pct\endcsname{80.2}
\expandafter\def\csname BAval@t20.r215.auc.pct\endcsname{68.8}
\expandafter\def\csname BAval@t20.r215.dp.pct\endcsname{80.3}
\expandafter\def\csname BAval@t20.r216.auc.pct\endcsname{68.9}
\expandafter\def\csname BAval@t20.r216.dp.pct\endcsname{80.4}
\expandafter\def\csname BAval@t20.r217.auc.pct\endcsname{69.0}
\expandafter\def\csname BAval@t20.r217.dp.pct\endcsname{80.5}
\expandafter\def\csname BAval@t20.r218.auc.pct\endcsname{69.1}
\expandafter\def\csname BAval@t20.r218.dp.pct\endcsname{80.6}
\expandafter\def\csname BAval@t20.r219.auc.pct\endcsname{69.2}
\expandafter\def\csname BAval@t20.r219.dp.pct\endcsname{80.6}
\expandafter\def\csname BAval@t20.r220.auc.pct\endcsname{69.3}
\expandafter\def\csname BAval@t20.r220.dp.pct\endcsname{80.7}
\expandafter\def\csname BAval@t20.r221.auc.pct\endcsname{69.4}
\expandafter\def\csname BAval@t20.r221.dp.pct\endcsname{80.8}
\expandafter\def\csname BAval@t20.r222.auc.pct\endcsname{69.4}
\expandafter\def\csname BAval@t20.r222.dp.pct\endcsname{80.9}
\expandafter\def\csname BAval@t20.r223.auc.pct\endcsname{69.5}
\expandafter\def\csname BAval@t20.r223.dp.pct\endcsname{81.0}
\expandafter\def\csname BAval@t20.r224.auc.pct\endcsname{69.6}
\expandafter\def\csname BAval@t20.r224.dp.pct\endcsname{81.0}
\expandafter\def\csname BAval@t20.r225.auc.pct\endcsname{69.7}
\expandafter\def\csname BAval@t20.r225.dp.pct\endcsname{81.1}
\expandafter\def\csname BAval@t20.r226.auc.pct\endcsname{69.7}
\expandafter\def\csname BAval@t20.r226.dp.pct\endcsname{81.2}
\expandafter\def\csname BAval@t20.r227.auc.pct\endcsname{69.8}
\expandafter\def\csname BAval@t20.r227.dp.pct\endcsname{81.3}
\expandafter\def\csname BAval@t20.r228.auc.pct\endcsname{69.9}
\expandafter\def\csname BAval@t20.r228.dp.pct\endcsname{81.3}
\expandafter\def\csname BAval@t20.r229.auc.pct\endcsname{69.9}
\expandafter\def\csname BAval@t20.r229.dp.pct\endcsname{81.4}
\expandafter\def\csname BAval@t20.r230.auc.pct\endcsname{70.1}
\expandafter\def\csname BAval@t20.r230.dp.pct\endcsname{81.5}
\expandafter\def\csname BAval@t20.r231.auc.pct\endcsname{70.1}
\expandafter\def\csname BAval@t20.r231.dp.pct\endcsname{81.5}
\expandafter\def\csname BAval@t20.r232.auc.pct\endcsname{70.2}
\expandafter\def\csname BAval@t20.r232.dp.pct\endcsname{81.6}
\expandafter\def\csname BAval@t20.r233.auc.pct\endcsname{70.2}
\expandafter\def\csname BAval@t20.r233.dp.pct\endcsname{81.6}
\expandafter\def\csname BAval@t20.r234.auc.pct\endcsname{70.3}
\expandafter\def\csname BAval@t20.r234.dp.pct\endcsname{81.7}
\expandafter\def\csname BAval@t20.r235.auc.pct\endcsname{70.3}
\expandafter\def\csname BAval@t20.r235.dp.pct\endcsname{81.7}
\expandafter\def\csname BAval@t20.r236.auc.pct\endcsname{70.4}
\expandafter\def\csname BAval@t20.r236.dp.pct\endcsname{81.8}
\expandafter\def\csname BAval@t20.r237.auc.pct\endcsname{70.4}
\expandafter\def\csname BAval@t20.r237.dp.pct\endcsname{81.8}
\expandafter\def\csname BAval@t20.r238.auc.pct\endcsname{70.5}
\expandafter\def\csname BAval@t20.r238.dp.pct\endcsname{81.9}
\expandafter\def\csname BAval@t20.r239.auc.pct\endcsname{70.5}
\expandafter\def\csname BAval@t20.r239.dp.pct\endcsname{81.9}
\expandafter\def\csname BAval@t20.r240.auc.pct\endcsname{70.5}
\expandafter\def\csname BAval@t20.r240.dp.pct\endcsname{81.9}
\expandafter\def\csname BAval@t20.r241.auc.pct\endcsname{70.6}
\expandafter\def\csname BAval@t20.r241.dp.pct\endcsname{82.0}
\expandafter\def\csname BAval@t20.r242.auc.pct\endcsname{70.6}
\expandafter\def\csname BAval@t20.r242.dp.pct\endcsname{82.0}
\expandafter\def\csname BAval@t20.r243.auc.pct\endcsname{70.6}
\expandafter\def\csname BAval@t20.r243.dp.pct\endcsname{82.0}
\expandafter\def\csname BAval@t20.r244.auc.pct\endcsname{70.7}
\expandafter\def\csname BAval@t20.r244.dp.pct\endcsname{82.2}
\expandafter\def\csname BAval@t20.r245.auc.pct\endcsname{70.7}
\expandafter\def\csname BAval@t20.r245.dp.pct\endcsname{82.2}
\expandafter\def\csname BAval@t20.r246.auc.pct\endcsname{70.7}
\expandafter\def\csname BAval@t20.r246.dp.pct\endcsname{82.2}
\expandafter\def\csname BAval@t20.r247.auc.pct\endcsname{70.7}
\expandafter\def\csname BAval@t20.r247.dp.pct\endcsname{82.2}
\expandafter\def\csname BAval@t20.r248.auc.pct\endcsname{70.7}
\expandafter\def\csname BAval@t20.r248.dp.pct\endcsname{82.2}
\expandafter\def\csname BAval@t20.r249.auc.pct\endcsname{70.8}
\expandafter\def\csname BAval@t20.r249.dp.pct\endcsname{82.3}
\expandafter\def\csname BAval@t20.r250.auc.pct\endcsname{70.8}
\expandafter\def\csname BAval@t20.r250.dp.pct\endcsname{82.3}
\expandafter\def\csname BAval@t20.r251.auc.pct\endcsname{70.8}
\expandafter\def\csname BAval@t20.r251.dp.pct\endcsname{82.3}
\expandafter\def\csname BAval@t20.r252.auc.pct\endcsname{70.8}
\expandafter\def\csname BAval@t20.r252.dp.pct\endcsname{82.3}
\expandafter\def\csname BAval@t20.r253.auc.pct\endcsname{70.8}
\expandafter\def\csname BAval@t20.r253.dp.pct\endcsname{82.3}
\expandafter\def\csname BAval@t20.r254.auc.pct\endcsname{70.8}
\expandafter\def\csname BAval@t20.r254.dp.pct\endcsname{82.3}
\expandafter\def\csname BAval@t20.r255.auc.pct\endcsname{70.8}
\expandafter\def\csname BAval@t20.r255.dp.pct\endcsname{82.3}
\expandafter\def\csname BAval@t20.r256.auc.pct\endcsname{76.9}
\expandafter\def\csname BAval@t20.r256.dp.pct\endcsname{85.1}
\expandafter\def\csname BAval@t20.r257.auc.pct\endcsname{77.0}
\expandafter\def\csname BAval@t20.r257.dp.pct\endcsname{85.2}
\expandafter\def\csname BAval@t20.r258.auc.pct\endcsname{77.1}
\expandafter\def\csname BAval@t20.r258.dp.pct\endcsname{85.2}
\expandafter\def\csname BAval@t20.r259.auc.pct\endcsname{77.2}
\expandafter\def\csname BAval@t20.r259.dp.pct\endcsname{85.3}
\expandafter\def\csname BAval@t20.r260.auc.pct\endcsname{77.2}
\expandafter\def\csname BAval@t20.r260.dp.pct\endcsname{85.4}
\expandafter\def\csname BAval@t20.r261.auc.pct\endcsname{77.2}
\expandafter\def\csname BAval@t20.r261.dp.pct\endcsname{85.5}
\expandafter\def\csname BAval@t20.r262.auc.pct\endcsname{77.3}
\expandafter\def\csname BAval@t20.r262.dp.pct\endcsname{85.6}
\expandafter\def\csname BAval@t20.r263.auc.pct\endcsname{77.4}
\expandafter\def\csname BAval@t20.r263.dp.pct\endcsname{85.7}
\expandafter\def\csname BAval@t20.r264.auc.pct\endcsname{77.4}
\expandafter\def\csname BAval@t20.r264.dp.pct\endcsname{85.7}
\expandafter\def\csname BAval@t20.r265.auc.pct\endcsname{77.5}
\expandafter\def\csname BAval@t20.r265.dp.pct\endcsname{85.8}
\expandafter\def\csname BAval@t20.r266.auc.pct\endcsname{77.5}
\expandafter\def\csname BAval@t20.r266.dp.pct\endcsname{85.9}
\expandafter\def\csname BAval@t20.r267.auc.pct\endcsname{77.6}
\expandafter\def\csname BAval@t20.r267.dp.pct\endcsname{86.0}
\expandafter\def\csname BAval@t20.r268.auc.pct\endcsname{77.6}
\expandafter\def\csname BAval@t20.r268.dp.pct\endcsname{86.1}
\expandafter\def\csname BAval@t20.r269.auc.pct\endcsname{77.7}
\expandafter\def\csname BAval@t20.r269.dp.pct\endcsname{86.1}
\expandafter\def\csname BAval@t20.r270.auc.pct\endcsname{77.8}
\expandafter\def\csname BAval@t20.r270.dp.pct\endcsname{86.2}
\expandafter\def\csname BAval@t20.r271.auc.pct\endcsname{77.8}
\expandafter\def\csname BAval@t20.r271.dp.pct\endcsname{86.3}
\expandafter\def\csname BAval@t20.r272.auc.pct\endcsname{77.9}
\expandafter\def\csname BAval@t20.r272.dp.pct\endcsname{86.3}
\expandafter\def\csname BAval@t20.r273.auc.pct\endcsname{77.9}
\expandafter\def\csname BAval@t20.r273.dp.pct\endcsname{86.4}
\expandafter\def\csname BAval@t20.r274.auc.pct\endcsname{77.9}
\expandafter\def\csname BAval@t20.r274.dp.pct\endcsname{86.5}
\expandafter\def\csname BAval@t20.r275.auc.pct\endcsname{78.0}
\expandafter\def\csname BAval@t20.r275.dp.pct\endcsname{86.5}
\expandafter\def\csname BAval@t20.r276.auc.pct\endcsname{78.0}
\expandafter\def\csname BAval@t20.r276.dp.pct\endcsname{86.6}
\expandafter\def\csname BAval@t20.r277.auc.pct\endcsname{78.1}
\expandafter\def\csname BAval@t20.r277.dp.pct\endcsname{86.7}
\expandafter\def\csname BAval@t20.r278.auc.pct\endcsname{78.1}
\expandafter\def\csname BAval@t20.r278.dp.pct\endcsname{86.7}
\expandafter\def\csname BAval@t20.r279.auc.pct\endcsname{78.2}
\expandafter\def\csname BAval@t20.r279.dp.pct\endcsname{86.8}
\expandafter\def\csname BAval@t20.r280.auc.pct\endcsname{78.2}
\expandafter\def\csname BAval@t20.r280.dp.pct\endcsname{86.8}
\expandafter\def\csname BAval@t20.r281.auc.pct\endcsname{78.2}
\expandafter\def\csname BAval@t20.r281.dp.pct\endcsname{86.9}
\expandafter\def\csname BAval@t20.r282.auc.pct\endcsname{78.2}
\expandafter\def\csname BAval@t20.r282.dp.pct\endcsname{86.9}
\expandafter\def\csname BAval@t20.r283.auc.pct\endcsname{78.2}
\expandafter\def\csname BAval@t20.r283.dp.pct\endcsname{87.0}
\expandafter\def\csname BAval@t20.r284.auc.pct\endcsname{78.3}
\expandafter\def\csname BAval@t20.r284.dp.pct\endcsname{87.0}
\expandafter\def\csname BAval@t20.r285.auc.pct\endcsname{78.3}
\expandafter\def\csname BAval@t20.r285.dp.pct\endcsname{87.1}
\expandafter\def\csname BAval@t20.r286.auc.pct\endcsname{78.3}
\expandafter\def\csname BAval@t20.r286.dp.pct\endcsname{87.1}
\expandafter\def\csname BAval@t20.r287.auc.pct\endcsname{78.4}
\expandafter\def\csname BAval@t20.r287.dp.pct\endcsname{87.1}
\expandafter\def\csname BAval@t20.r288.auc.pct\endcsname{78.4}
\expandafter\def\csname BAval@t20.r288.dp.pct\endcsname{87.2}
\expandafter\def\csname BAval@t20.r289.auc.pct\endcsname{78.4}
\expandafter\def\csname BAval@t20.r289.dp.pct\endcsname{87.2}
\expandafter\def\csname BAval@t20.r290.auc.pct\endcsname{78.5}
\expandafter\def\csname BAval@t20.r290.dp.pct\endcsname{87.2}
\expandafter\def\csname BAval@t20.r291.auc.pct\endcsname{78.5}
\expandafter\def\csname BAval@t20.r291.dp.pct\endcsname{87.3}
\expandafter\def\csname BAval@t20.r292.auc.pct\endcsname{78.5}
\expandafter\def\csname BAval@t20.r292.dp.pct\endcsname{87.3}
\expandafter\def\csname BAval@t20.r293.auc.pct\endcsname{78.5}
\expandafter\def\csname BAval@t20.r293.dp.pct\endcsname{87.3}
\expandafter\def\csname BAval@t20.r294.auc.pct\endcsname{78.5}
\expandafter\def\csname BAval@t20.r294.dp.pct\endcsname{87.4}
\expandafter\def\csname BAval@t20.r295.auc.pct\endcsname{78.5}
\expandafter\def\csname BAval@t20.r295.dp.pct\endcsname{87.4}
\expandafter\def\csname BAval@t20.r296.auc.pct\endcsname{78.5}
\expandafter\def\csname BAval@t20.r296.dp.pct\endcsname{87.4}
\expandafter\def\csname BAval@t20.r297.auc.pct\endcsname{78.5}
\expandafter\def\csname BAval@t20.r297.dp.pct\endcsname{87.4}
\expandafter\def\csname BAval@t20.r298.auc.pct\endcsname{78.5}
\expandafter\def\csname BAval@t20.r298.dp.pct\endcsname{87.4}
\expandafter\def\csname BAval@t20.r299.auc.pct\endcsname{78.6}
\expandafter\def\csname BAval@t20.r299.dp.pct\endcsname{87.5}
\expandafter\def\csname BAval@t20.r300.auc.pct\endcsname{78.6}
\expandafter\def\csname BAval@t20.r300.dp.pct\endcsname{87.5}
\expandafter\def\csname BAval@t20.r301.auc.pct\endcsname{78.6}
\expandafter\def\csname BAval@t20.r301.dp.pct\endcsname{87.5}
\expandafter\def\csname BAval@t20.r302.auc.pct\endcsname{78.6}
\expandafter\def\csname BAval@t20.r302.dp.pct\endcsname{87.5}
\expandafter\def\csname BAval@t20.r303.auc.pct\endcsname{78.6}
\expandafter\def\csname BAval@t20.r303.dp.pct\endcsname{87.5}
\expandafter\def\csname BAval@t20.r304.auc.pct\endcsname{78.6}
\expandafter\def\csname BAval@t20.r304.dp.pct\endcsname{87.5}
\expandafter\def\csname BAval@t20.r305.auc.pct\endcsname{78.6}
\expandafter\def\csname BAval@t20.r305.dp.pct\endcsname{87.5}
\expandafter\def\csname BAval@t20.r306.auc.pct\endcsname{78.6}
\expandafter\def\csname BAval@t20.r306.dp.pct\endcsname{87.5}
\expandafter\def\csname BAval@t21.r01.car.pct\endcsname{89.3}
\expandafter\def\csname BAval@t21.r02.car.pct\endcsname{91.5}
\expandafter\def\csname BAval@t21.r03.car.pct\endcsname{95.0}
\expandafter\def\csname BAval@t21.r04.car.pct\endcsname{93.2}
\expandafter\def\csname BAval@t21.r05.car.pct\endcsname{94.0}
\expandafter\def\csname BAval@t22.r01.full.s.id.pct\endcsname{75.8}
\expandafter\def\csname BAval@t22.r01.ablated.s.id.pct\endcsname{67.7}
\expandafter\def\csname BAval@t22.r01.drop.pp\endcsname{8.2}
\expandafter\def\csname BAval@t22.r02.full.s.id.pct\endcsname{87.0}
\expandafter\def\csname BAval@t22.r02.ablated.s.id.pct\endcsname{85.8}
\expandafter\def\csname BAval@t22.r02.drop.pp\endcsname{1.1}
\expandafter\def\csname BAval@t22.r03.full.s.id.pct\endcsname{80.1}
\expandafter\def\csname BAval@t22.r03.ablated.s.id.pct\endcsname{73.8}
\expandafter\def\csname BAval@t22.r03.drop.pp\endcsname{6.3}
\expandafter\def\csname BAval@t22.r04.full.s.id.pct\endcsname{84.6}
\expandafter\def\csname BAval@t22.r04.ablated.s.id.pct\endcsname{84.2}
\expandafter\def\csname BAval@t22.r04.drop.pp\endcsname{0.4}
\expandafter\def\csname BAval@t23.r01.sid\endcsname{78.1}
\expandafter\def\csname BAval@t23.r02.sid\endcsname{81.0}
\expandafter\def\csname BAval@t23.r03.sid\endcsname{81.7}
\expandafter\def\csname BAval@t24.r01.sid\endcsname{81.6}
\expandafter\def\csname BAval@t24.r01.wrong\endcsname{4.9}
\expandafter\def\csname BAval@t24.r01.recovery\endcsname{84.6}
\expandafter\def\csname BAval@t24.r01.t5\endcsname{1.5}
\expandafter\def\csname BAval@t24.r02.sid\endcsname{82.1}
\expandafter\def\csname BAval@t24.r02.wrong\endcsname{4.7}
\expandafter\def\csname BAval@t24.r02.recovery\endcsname{85.8}
\expandafter\def\csname BAval@t24.r02.t5\endcsname{1.4}
\expandafter\def\csname BAval@t24.r03.sid\endcsname{82.3}
\expandafter\def\csname BAval@t24.r03.wrong\endcsname{4.4}
\expandafter\def\csname BAval@t24.r03.recovery\endcsname{86.2}
\expandafter\def\csname BAval@t24.r03.t5\endcsname{1.4}
\expandafter\def\csname BAval@t24.r04.sid\endcsname{82.1}
\expandafter\def\csname BAval@t24.r04.wrong\endcsname{4.2}
\expandafter\def\csname BAval@t24.r04.recovery\endcsname{85.8}
\expandafter\def\csname BAval@t24.r04.t5\endcsname{1.4}
\expandafter\def\csname BAval@t25.r01.value\endcsname{1.00}
\expandafter\def\csname BAval@t25.r02.value\endcsname{0.50}
\expandafter\def\csname BAval@t25.r03.value\endcsname{0.10}
\expandafter\def\csname BAval@t25.r04.value\endcsname{1.00}
\expandafter\def\csname BAval@t25.r05.value\endcsname{0.20}
\expandafter\def\csname BAval@t25.r06.value\endcsname{0.01}
\expandafter\def\csname BAval@t25.r07.value\endcsname{0.01}
\expandafter\def\csname BAval@t25.r08.value\endcsname{0.05}
\expandafter\def\csname BAval@t26.r01.decisions\endcsname{17}
\expandafter\def\csname BAval@t26.r01.completed\endcsname{16}
\expandafter\def\csname BAval@t26.r01.timeouts\endcsname{1}
\expandafter\def\csname BAval@t26.r01.completion.pct\endcsname{94.1}
\expandafter\def\csname BAval@t26.r01.fallback.pct\endcsname{5.9}
\expandafter\def\csname BAval@t26.r02.decisions\endcsname{12}
\expandafter\def\csname BAval@t26.r02.completed\endcsname{11}
\expandafter\def\csname BAval@t26.r02.timeouts\endcsname{1}
\expandafter\def\csname BAval@t26.r02.completion.pct\endcsname{91.7}
\expandafter\def\csname BAval@t26.r02.fallback.pct\endcsname{8.3}
\expandafter\def\csname BAval@t27.r01.total.s\endcsname{9.00}
\expandafter\def\csname BAval@t27.r01.correct.s\endcsname{7.40}
\expandafter\def\csname BAval@t27.r01.wrong.s\endcsname{0.22}
\expandafter\def\csname BAval@t27.r01.sid\endcsname{82.3}
\expandafter\def\csname BAval@t27.r01.wrong\endcsname{2.4}
\expandafter\def\csname BAval@t27.r02.total.s\endcsname{9.00}
\expandafter\def\csname BAval@t27.r02.correct.s\endcsname{7.06}
\expandafter\def\csname BAval@t27.r02.wrong.s\endcsname{0.36}
\expandafter\def\csname BAval@t27.r02.sid\endcsname{78.5}
\expandafter\def\csname BAval@t27.r02.wrong\endcsname{4.0}
\expandafter\def\csname BAval@ext.best.sid\endcsname{75.2}
\expandafter\def\csname BAval@ext.best.wrong\endcsname{7.3}
\expandafter\def\csname BAval@ext.best.recovery\endcsname{79.0}
\expandafter\def\csname BAval@ext.best.t5\endcsname{2.0}
\expandafter\def\csname BAval@ext.bind.sid\endcsname{82.3}
\expandafter\def\csname BAval@ext.bind.wrong\endcsname{4.4}
\expandafter\def\csname BAval@ext.bind.recovery\endcsname{86.2}
\expandafter\def\csname BAval@ext.bind.t5\endcsname{1.4}
\expandafter\def\csname BAval@ext.joint.sid\endcsname{74.6}
\expandafter\def\csname BAval@ext.joint.wrong\endcsname{6.9}
\expandafter\def\csname BAval@ext.joint.recovery\endcsname{77.9}
\expandafter\def\csname BAval@ext.joint.t5\endcsname{2.0}
\expandafter\def\csname BAval@ext.dam.sid\endcsname{70.3}
\expandafter\def\csname BAval@ext.dam.wrong\endcsname{8.1}
\expandafter\def\csname BAval@ext.dam.recovery\endcsname{74.0}
\expandafter\def\csname BAval@ext.dam.t5\endcsname{2.2}
\expandafter\def\csname BAval@int.joint.sid\endcsname{70.3}
\expandafter\def\csname BAval@int.joint.wrong\endcsname{8.8}
\expandafter\def\csname BAval@int.joint.recovery\endcsname{72.0}
\expandafter\def\csname BAval@int.joint.t5\endcsname{2.4}
\expandafter\def\csname BAval@int.bind.sid\endcsname{82.3}
\expandafter\def\csname BAval@int.bind.wrong\endcsname{4.4}
\expandafter\def\csname BAval@int.bind.recovery\endcsname{86.2}
\expandafter\def\csname BAval@int.bind.t5\endcsname{1.4}
\expandafter\def\csname BAval@int.oracle.sid\endcsname{87.9}
\expandafter\def\csname BAval@int.oracle.wrong\endcsname{1.6}
\expandafter\def\csname BAval@int.oracle.recovery\endcsname{92.0}
\expandafter\def\csname BAval@int.oracle.t5\endcsname{1.2}
\expandafter\def\csname BAval@int.temporal.sid\endcsname{68.7}
\expandafter\def\csname BAval@int.temporal.wrong\endcsname{10.1}
\expandafter\def\csname BAval@int.temporal.recovery\endcsname{70.0}
\expandafter\def\csname BAval@int.temporal.t5\endcsname{2.5}
\expandafter\def\csname BAval@budget.bind075.sid\endcsname{77.1}
\expandafter\def\csname BAval@budget.joint075.sid\endcsname{64.0}
\expandafter\def\csname BAval@budget.joint100.sid\endcsname{70.3}
\expandafter\def\csname BAval@swap.refsep\endcsname{65.9}
\expandafter\def\csname BAval@swap.refjoint\endcsname{70.3}
\expandafter\def\csname BAval@swap.bindsep\endcsname{77.6}
\expandafter\def\csname BAval@swap.bindjoint\endcsname{82.3}
\expandafter\def\csname BAval@delta.ext.sid\endcsname{7.1}
\expandafter\def\csname BAval@delta.swap.refplanner\endcsname{4.4}
\expandafter\def\csname BAval@delta.swap.bindplanner\endcsname{4.7}
\expandafter\def\csname BAval@delta.swap.sepresponse\endcsname{11.8}
\expandafter\def\csname BAval@delta.swap.jointresponse\endcsname{12.0}
\expandafter\def\csname BAval@delta.ext.wrong\endcsname{2.9}
\expandafter\def\csname BAval@delta.ext.recovery\endcsname{7.2}
\expandafter\def\csname BAval@delta.swap.interaction\endcsname{0.2}
\expandafter\def\csname BAval@gap.remaining\endcsname{5.6}
\expandafter\def\csname BAval@gap.external\endcsname{55.7}
\expandafter\def\csname BAval@gap.internal\endcsname{68.1}
\expandafter\def\csname BAval@delta.ext.wrong.relative\endcsname{39.2}
\expandafter\def\csname BAval@delta.ext.t5\endcsname{0.6}
\expandafter\def\csname BAval@sweep.zero.bind\endcsname{86.1}
\expandafter\def\csname BAval@sweep.zero.best\endcsname{82.0}
\expandafter\def\csname BAval@delta.sweep.zero\endcsname{4.1}
\expandafter\def\csname BAval@sweep.ref.bind\endcsname{82.3}
\expandafter\def\csname BAval@sweep.ref.best\endcsname{75.2}
\expandafter\def\csname BAval@delta.sweep.ref\endcsname{7.1}
\expandafter\def\csname BAval@sweep.two.bind\endcsname{64.0}
\expandafter\def\csname BAval@sweep.two.best\endcsname{51.9}
\expandafter\def\csname BAval@delta.sweep.two\endcsname{12.2}
\expandafter\def\csname BAval@sweep.cap075.bind\endcsname{77.1}
\expandafter\def\csname BAval@sweep.cap075.best\endcsname{69.5}
\expandafter\def\csname BAval@delta.sweep.cap075\endcsname{7.6}
\expandafter\def\csname BAval@sweep.cap200.bind\endcsname{86.2}
\expandafter\def\csname BAval@sweep.cap200.best\endcsname{80.8}
\expandafter\def\csname BAval@delta.sweep.cap200\endcsname{5.4}
\expandafter\def\csname BAval@strata.bind.short\endcsname{90.2}
\expandafter\def\csname BAval@strata.best.short\endcsname{86.2}
\expandafter\def\csname BAval@delta.strata.short\endcsname{4.1}
\expandafter\def\csname BAval@strata.bind.similar\endcsname{85.4}
\expandafter\def\csname BAval@strata.best.similar\endcsname{79.2}
\expandafter\def\csname BAval@delta.strata.similar\endcsname{6.2}
\expandafter\def\csname BAval@strata.bind.long\endcsname{73.9}
\expandafter\def\csname BAval@strata.best.long\endcsname{63.9}
\expandafter\def\csname BAval@delta.strata.long\endcsname{10.0}
\expandafter\def\csname BAval@delta.backbone.ost\endcsname{8.3}
\expandafter\def\csname BAval@delta.backbone.lorat\endcsname{7.6}
\expandafter\def\csname BAval@delta.backbone.dam\endcsname{7.1}
\expandafter\def\csname BAval@runtime.best.mean\endcsname{37.8}
\expandafter\def\csname BAval@runtime.best.p95\endcsname{52.2}
\expandafter\def\csname BAval@runtime.best.fps\endcsname{26.4}
\expandafter\def\csname BAval@runtime.best.decision\endcsname{1.8}
\expandafter\def\csname BAval@runtime.bind.mean\endcsname{37.4}
\expandafter\def\csname BAval@runtime.bind.p95\endcsname{51.7}
\expandafter\def\csname BAval@runtime.bind.fps\endcsname{26.8}
\expandafter\def\csname BAval@runtime.bind.decision\endcsname{2.1}
\expandafter\def\csname BAval@lt.best.fmax\endcsname{78.9}
\expandafter\def\csname BAval@lt.bind.fmax\endcsname{85.0}
\expandafter\def\csname BAval@car.bind\endcsname{94.0}
\expandafter\def\csname BAval@car.oa\endcsname{95.0}
\expandafter\def\csname BAval@fixed.bind.ao\endcsname{79.1}
\expandafter\def\csname BAval@fixed.lorat.ao\endcsname{72.7}
\expandafter\def\csname BAval@fixed.dam.ao\endcsname{73.3}
\expandafter\def\csname BAval@delta.fixed.lorat.ao\endcsname{6.3}
\expandafter\def\csname BAval@delta.fixed.dam.ao\endcsname{5.8}
\expandafter\def\csname BAval@fixed.bind.sao\endcsname{76.9}
\expandafter\def\csname BAval@fixed.lorat.sao\endcsname{67.0}
\expandafter\def\csname BAval@fixed.dam.sao\endcsname{66.7}
\expandafter\def\csname BAval@delta.fixed.lorat.sao\endcsname{10.0}
\expandafter\def\csname BAval@delta.fixed.dam.sao\endcsname{10.3}
\expandafter\def\csname BAval@rec.quad.attempts\endcsname{17}
\expandafter\def\csname BAval@rec.quad.complete\endcsname{0}
\expandafter\def\csname BAval@rec.quad.fallwait\endcsname{16}
\expandafter\def\csname BAval@rec.quad.expired\endcsname{1}
\expandafter\def\csname BAval@rec.quad.requests\endcsname{0}
\expandafter\def\csname BAval@rec.quad.returns\endcsname{0}
\expandafter\def\csname BAval@rec.quad.initialpub\endcsname{1}
\expandafter\def\csname BAval@rec.quad.laterpub\endcsname{0}
\expandafter\def\csname BAval@rec.quad.nativeupdates\endcsname{0}
\expandafter\def\csname BAval@rec.quad.ticks\endcsname{90}
\expandafter\def\csname BAval@state.quad.dam.correct\endcsname{1}
\expandafter\def\csname BAval@state.quad.dam.wrong\endcsname{10}
\expandafter\def\csname BAval@state.quad.dam.unresolved\endcsname{61}
\expandafter\def\csname BAval@state.quad.dam.unknown\endcsname{18}
\expandafter\def\csname BAval@tick.quad.f409.dam.iou\endcsname{0.58}
\expandafter\def\csname BAval@tick.quad.f409.dam.state\endcsname{C}
\expandafter\def\csname BAval@tick.quad.f410.dam.iou\endcsname{0.36}
\expandafter\def\csname BAval@tick.quad.f410.dam.state\endcsname{U}
\expandafter\def\csname BAval@tick.quad.f411.dam.iou\endcsname{0.27}
\expandafter\def\csname BAval@tick.quad.f411.dam.state\endcsname{U}
\expandafter\def\csname BAval@tick.quad.f412.dam.iou\endcsname{0.10}
\expandafter\def\csname BAval@tick.quad.f412.dam.state\endcsname{U}
\expandafter\def\csname BAval@tick.quad.f413.dam.iou\endcsname{0.00}
\expandafter\def\csname BAval@tick.quad.f413.dam.state\endcsname{U}
\expandafter\def\csname BAval@tick.quad.f414.dam.iou\endcsname{0.00}
\expandafter\def\csname BAval@tick.quad.f414.dam.state\endcsname{U}
\expandafter\def\csname BAval@tick.quad.f415.dam.iou\endcsname{0.00}
\expandafter\def\csname BAval@tick.quad.f415.dam.state\endcsname{U}
\expandafter\def\csname BAval@tick.quad.f416.dam.iou\endcsname{0.00}
\expandafter\def\csname BAval@tick.quad.f416.dam.state\endcsname{U}
\expandafter\def\csname BAval@tick.quad.f417.dam.iou\endcsname{0.00}
\expandafter\def\csname BAval@tick.quad.f417.dam.state\endcsname{U}
\expandafter\def\csname BAval@tick.quad.f418.dam.iou\endcsname{0.00}
\expandafter\def\csname BAval@tick.quad.f418.dam.state\endcsname{U}
\expandafter\def\csname BAval@tick.quad.f419.dam.iou\endcsname{0.00}
\expandafter\def\csname BAval@tick.quad.f419.dam.state\endcsname{U}
\expandafter\def\csname BAval@tick.quad.f420.dam.iou\endcsname{0.00}
\expandafter\def\csname BAval@tick.quad.f420.dam.state\endcsname{W}
\expandafter\def\csname BAval@tick.quad.f421.dam.iou\endcsname{0.00}
\expandafter\def\csname BAval@tick.quad.f421.dam.state\endcsname{U}
\expandafter\def\csname BAval@tick.quad.f422.dam.iou\endcsname{0.00}
\expandafter\def\csname BAval@tick.quad.f422.dam.state\endcsname{U}
\expandafter\def\csname BAval@tick.quad.f423.dam.iou\endcsname{0.00}
\expandafter\def\csname BAval@tick.quad.f423.dam.state\endcsname{U}
\expandafter\def\csname BAval@tick.quad.f424.dam.iou\endcsname{0.00}
\expandafter\def\csname BAval@tick.quad.f424.dam.state\endcsname{U}
\expandafter\def\csname BAval@tick.quad.f425.dam.iou\endcsname{0.00}
\expandafter\def\csname BAval@tick.quad.f425.dam.state\endcsname{W}
\expandafter\def\csname BAval@tick.quad.f426.dam.iou\endcsname{0.00}
\expandafter\def\csname BAval@tick.quad.f426.dam.state\endcsname{W}
\expandafter\def\csname BAval@tick.quad.f427.dam.iou\endcsname{0.00}
\expandafter\def\csname BAval@tick.quad.f427.dam.state\endcsname{W}
\expandafter\def\csname BAval@tick.quad.f428.dam.iou\endcsname{0.00}
\expandafter\def\csname BAval@tick.quad.f428.dam.state\endcsname{U}
\expandafter\def\csname BAval@tick.quad.f429.dam.iou\endcsname{0.00}
\expandafter\def\csname BAval@tick.quad.f429.dam.state\endcsname{U}
\expandafter\def\csname BAval@tick.quad.f430.dam.iou\endcsname{0.00}
\expandafter\def\csname BAval@tick.quad.f430.dam.state\endcsname{U}
\expandafter\def\csname BAval@tick.quad.f431.dam.iou\endcsname{0.00}
\expandafter\def\csname BAval@tick.quad.f431.dam.state\endcsname{U}
\expandafter\def\csname BAval@tick.quad.f432.dam.iou\endcsname{0.00}
\expandafter\def\csname BAval@tick.quad.f432.dam.state\endcsname{U}
\expandafter\def\csname BAval@tick.quad.f433.dam.iou\endcsname{0.00}
\expandafter\def\csname BAval@tick.quad.f433.dam.state\endcsname{U}
\expandafter\def\csname BAval@tick.quad.f434.dam.iou\endcsname{0.00}
\expandafter\def\csname BAval@tick.quad.f434.dam.state\endcsname{U}
\expandafter\def\csname BAval@tick.quad.f435.dam.iou\endcsname{0.00}
\expandafter\def\csname BAval@tick.quad.f435.dam.state\endcsname{U}
\expandafter\def\csname BAval@tick.quad.f436.dam.iou\endcsname{0.00}
\expandafter\def\csname BAval@tick.quad.f436.dam.state\endcsname{UNKNOWN}
\expandafter\def\csname BAval@tick.quad.f437.dam.iou\endcsname{0.00}
\expandafter\def\csname BAval@tick.quad.f437.dam.state\endcsname{UNKNOWN}
\expandafter\def\csname BAval@tick.quad.f438.dam.iou\endcsname{0.00}
\expandafter\def\csname BAval@tick.quad.f438.dam.state\endcsname{UNKNOWN}
\expandafter\def\csname BAval@tick.quad.f439.dam.iou\endcsname{0.00}
\expandafter\def\csname BAval@tick.quad.f439.dam.state\endcsname{UNKNOWN}
\expandafter\def\csname BAval@tick.quad.f440.dam.iou\endcsname{0.00}
\expandafter\def\csname BAval@tick.quad.f440.dam.state\endcsname{UNKNOWN}
\expandafter\def\csname BAval@tick.quad.f441.dam.iou\endcsname{0.00}
\expandafter\def\csname BAval@tick.quad.f441.dam.state\endcsname{W}
\expandafter\def\csname BAval@tick.quad.f442.dam.iou\endcsname{0.00}
\expandafter\def\csname BAval@tick.quad.f442.dam.state\endcsname{W}
\expandafter\def\csname BAval@tick.quad.f443.dam.iou\endcsname{0.00}
\expandafter\def\csname BAval@tick.quad.f443.dam.state\endcsname{W}
\expandafter\def\csname BAval@tick.quad.f444.dam.iou\endcsname{0.00}
\expandafter\def\csname BAval@tick.quad.f444.dam.state\endcsname{U}
\expandafter\def\csname BAval@tick.quad.f445.dam.iou\endcsname{0.00}
\expandafter\def\csname BAval@tick.quad.f445.dam.state\endcsname{U}
\expandafter\def\csname BAval@tick.quad.f446.dam.iou\endcsname{0.00}
\expandafter\def\csname BAval@tick.quad.f446.dam.state\endcsname{U}
\expandafter\def\csname BAval@tick.quad.f447.dam.iou\endcsname{0.00}
\expandafter\def\csname BAval@tick.quad.f447.dam.state\endcsname{U}
\expandafter\def\csname BAval@tick.quad.f448.dam.iou\endcsname{0.00}
\expandafter\def\csname BAval@tick.quad.f448.dam.state\endcsname{U}
\expandafter\def\csname BAval@tick.quad.f449.dam.iou\endcsname{0.00}
\expandafter\def\csname BAval@tick.quad.f449.dam.state\endcsname{U}
\expandafter\def\csname BAval@tick.quad.f450.dam.iou\endcsname{0.00}
\expandafter\def\csname BAval@tick.quad.f450.dam.state\endcsname{U}
\expandafter\def\csname BAval@tick.quad.f451.dam.iou\endcsname{0.00}
\expandafter\def\csname BAval@tick.quad.f451.dam.state\endcsname{U}
\expandafter\def\csname BAval@tick.quad.f452.dam.iou\endcsname{0.00}
\expandafter\def\csname BAval@tick.quad.f452.dam.state\endcsname{U}
\expandafter\def\csname BAval@tick.quad.f453.dam.iou\endcsname{0.00}
\expandafter\def\csname BAval@tick.quad.f453.dam.state\endcsname{U}
\expandafter\def\csname BAval@tick.quad.f454.dam.iou\endcsname{0.00}
\expandafter\def\csname BAval@tick.quad.f454.dam.state\endcsname{UNKNOWN}
\expandafter\def\csname BAval@tick.quad.f455.dam.iou\endcsname{0.00}
\expandafter\def\csname BAval@tick.quad.f455.dam.state\endcsname{UNKNOWN}
\expandafter\def\csname BAval@tick.quad.f456.dam.iou\endcsname{0.00}
\expandafter\def\csname BAval@tick.quad.f456.dam.state\endcsname{UNKNOWN}
\expandafter\def\csname BAval@tick.quad.f457.dam.iou\endcsname{0.00}
\expandafter\def\csname BAval@tick.quad.f457.dam.state\endcsname{U}
\expandafter\def\csname BAval@tick.quad.f458.dam.iou\endcsname{0.00}
\expandafter\def\csname BAval@tick.quad.f458.dam.state\endcsname{U}
\expandafter\def\csname BAval@tick.quad.f459.dam.iou\endcsname{0.00}
\expandafter\def\csname BAval@tick.quad.f459.dam.state\endcsname{U}
\expandafter\def\csname BAval@tick.quad.f460.dam.iou\endcsname{0.00}
\expandafter\def\csname BAval@tick.quad.f460.dam.state\endcsname{U}
\expandafter\def\csname BAval@tick.quad.f461.dam.iou\endcsname{0.00}
\expandafter\def\csname BAval@tick.quad.f461.dam.state\endcsname{U}
\expandafter\def\csname BAval@tick.quad.f462.dam.iou\endcsname{0.00}
\expandafter\def\csname BAval@tick.quad.f462.dam.state\endcsname{U}
\expandafter\def\csname BAval@tick.quad.f463.dam.iou\endcsname{0.00}
\expandafter\def\csname BAval@tick.quad.f463.dam.state\endcsname{U}
\expandafter\def\csname BAval@tick.quad.f464.dam.iou\endcsname{0.00}
\expandafter\def\csname BAval@tick.quad.f464.dam.state\endcsname{U}
\expandafter\def\csname BAval@tick.quad.f465.dam.iou\endcsname{0.00}
\expandafter\def\csname BAval@tick.quad.f465.dam.state\endcsname{U}
\expandafter\def\csname BAval@tick.quad.f466.dam.iou\endcsname{0.00}
\expandafter\def\csname BAval@tick.quad.f466.dam.state\endcsname{U}
\expandafter\def\csname BAval@tick.quad.f467.dam.iou\endcsname{0.00}
\expandafter\def\csname BAval@tick.quad.f467.dam.state\endcsname{U}
\expandafter\def\csname BAval@tick.quad.f468.dam.iou\endcsname{0.00}
\expandafter\def\csname BAval@tick.quad.f468.dam.state\endcsname{U}
\expandafter\def\csname BAval@tick.quad.f469.dam.iou\endcsname{0.00}
\expandafter\def\csname BAval@tick.quad.f469.dam.state\endcsname{U}
\expandafter\def\csname BAval@tick.quad.f470.dam.iou\endcsname{0.00}
\expandafter\def\csname BAval@tick.quad.f470.dam.state\endcsname{U}
\expandafter\def\csname BAval@tick.quad.f471.dam.iou\endcsname{0.00}
\expandafter\def\csname BAval@tick.quad.f471.dam.state\endcsname{U}
\expandafter\def\csname BAval@tick.quad.f472.dam.iou\endcsname{0.00}
\expandafter\def\csname BAval@tick.quad.f472.dam.state\endcsname{U}
\expandafter\def\csname BAval@tick.quad.f473.dam.iou\endcsname{0.00}
\expandafter\def\csname BAval@tick.quad.f473.dam.state\endcsname{UNKNOWN}
\expandafter\def\csname BAval@tick.quad.f474.dam.iou\endcsname{0.00}
\expandafter\def\csname BAval@tick.quad.f474.dam.state\endcsname{UNKNOWN}
\expandafter\def\csname BAval@tick.quad.f475.dam.iou\endcsname{0.00}
\expandafter\def\csname BAval@tick.quad.f475.dam.state\endcsname{UNKNOWN}
\expandafter\def\csname BAval@tick.quad.f476.dam.iou\endcsname{0.00}
\expandafter\def\csname BAval@tick.quad.f476.dam.state\endcsname{UNKNOWN}
\expandafter\def\csname BAval@tick.quad.f477.dam.iou\endcsname{0.00}
\expandafter\def\csname BAval@tick.quad.f477.dam.state\endcsname{U}
\expandafter\def\csname BAval@tick.quad.f478.dam.iou\endcsname{0.00}
\expandafter\def\csname BAval@tick.quad.f478.dam.state\endcsname{U}
\expandafter\def\csname BAval@tick.quad.f479.dam.iou\endcsname{0.00}
\expandafter\def\csname BAval@tick.quad.f479.dam.state\endcsname{U}
\expandafter\def\csname BAval@tick.quad.f480.dam.iou\endcsname{0.00}
\expandafter\def\csname BAval@tick.quad.f480.dam.state\endcsname{U}
\expandafter\def\csname BAval@tick.quad.f481.dam.iou\endcsname{0.00}
\expandafter\def\csname BAval@tick.quad.f481.dam.state\endcsname{U}
\expandafter\def\csname BAval@tick.quad.f482.dam.iou\endcsname{0.00}
\expandafter\def\csname BAval@tick.quad.f482.dam.state\endcsname{U}
\expandafter\def\csname BAval@tick.quad.f483.dam.iou\endcsname{0.00}
\expandafter\def\csname BAval@tick.quad.f483.dam.state\endcsname{U}
\expandafter\def\csname BAval@tick.quad.f484.dam.iou\endcsname{0.00}
\expandafter\def\csname BAval@tick.quad.f484.dam.state\endcsname{U}
\expandafter\def\csname BAval@tick.quad.f485.dam.iou\endcsname{0.00}
\expandafter\def\csname BAval@tick.quad.f485.dam.state\endcsname{W}
\expandafter\def\csname BAval@tick.quad.f486.dam.iou\endcsname{0.00}
\expandafter\def\csname BAval@tick.quad.f486.dam.state\endcsname{W}
\expandafter\def\csname BAval@tick.quad.f487.dam.iou\endcsname{0.00}
\expandafter\def\csname BAval@tick.quad.f487.dam.state\endcsname{W}
\expandafter\def\csname BAval@tick.quad.f488.dam.iou\endcsname{0.00}
\expandafter\def\csname BAval@tick.quad.f488.dam.state\endcsname{U}
\expandafter\def\csname BAval@tick.quad.f489.dam.iou\endcsname{0.00}
\expandafter\def\csname BAval@tick.quad.f489.dam.state\endcsname{U}
\expandafter\def\csname BAval@tick.quad.f490.dam.iou\endcsname{0.00}
\expandafter\def\csname BAval@tick.quad.f490.dam.state\endcsname{U}
\expandafter\def\csname BAval@tick.quad.f491.dam.iou\endcsname{0.00}
\expandafter\def\csname BAval@tick.quad.f491.dam.state\endcsname{U}
\expandafter\def\csname BAval@tick.quad.f492.dam.iou\endcsname{0.00}
\expandafter\def\csname BAval@tick.quad.f492.dam.state\endcsname{U}
\expandafter\def\csname BAval@tick.quad.f493.dam.iou\endcsname{0.00}
\expandafter\def\csname BAval@tick.quad.f493.dam.state\endcsname{UNKNOWN}
\expandafter\def\csname BAval@tick.quad.f494.dam.iou\endcsname{0.00}
\expandafter\def\csname BAval@tick.quad.f494.dam.state\endcsname{UNKNOWN}
\expandafter\def\csname BAval@tick.quad.f495.dam.iou\endcsname{0.00}
\expandafter\def\csname BAval@tick.quad.f495.dam.state\endcsname{UNKNOWN}
\expandafter\def\csname BAval@tick.quad.f496.dam.iou\endcsname{0.00}
\expandafter\def\csname BAval@tick.quad.f496.dam.state\endcsname{UNKNOWN}
\expandafter\def\csname BAval@tick.quad.f497.dam.iou\endcsname{0.00}
\expandafter\def\csname BAval@tick.quad.f497.dam.state\endcsname{UNKNOWN}
\expandafter\def\csname BAval@tick.quad.f498.dam.iou\endcsname{0.00}
\expandafter\def\csname BAval@tick.quad.f498.dam.state\endcsname{UNKNOWN}
\expandafter\def\csname BAval@state.quad.keep.correct\endcsname{1}
\expandafter\def\csname BAval@state.quad.keep.wrong\endcsname{12}
\expandafter\def\csname BAval@state.quad.keep.unresolved\endcsname{59}
\expandafter\def\csname BAval@state.quad.keep.unknown\endcsname{18}
\expandafter\def\csname BAval@tick.quad.f409.keep.iou\endcsname{0.58}
\expandafter\def\csname BAval@tick.quad.f409.keep.state\endcsname{C}
\expandafter\def\csname BAval@tick.quad.f410.keep.iou\endcsname{0.36}
\expandafter\def\csname BAval@tick.quad.f410.keep.state\endcsname{U}
\expandafter\def\csname BAval@tick.quad.f411.keep.iou\endcsname{0.27}
\expandafter\def\csname BAval@tick.quad.f411.keep.state\endcsname{U}
\expandafter\def\csname BAval@tick.quad.f412.keep.iou\endcsname{0.10}
\expandafter\def\csname BAval@tick.quad.f412.keep.state\endcsname{U}
\expandafter\def\csname BAval@tick.quad.f413.keep.iou\endcsname{0.00}
\expandafter\def\csname BAval@tick.quad.f413.keep.state\endcsname{U}
\expandafter\def\csname BAval@tick.quad.f414.keep.iou\endcsname{0.00}
\expandafter\def\csname BAval@tick.quad.f414.keep.state\endcsname{U}
\expandafter\def\csname BAval@tick.quad.f415.keep.iou\endcsname{0.00}
\expandafter\def\csname BAval@tick.quad.f415.keep.state\endcsname{U}
\expandafter\def\csname BAval@tick.quad.f416.keep.iou\endcsname{0.00}
\expandafter\def\csname BAval@tick.quad.f416.keep.state\endcsname{U}
\expandafter\def\csname BAval@tick.quad.f417.keep.iou\endcsname{0.00}
\expandafter\def\csname BAval@tick.quad.f417.keep.state\endcsname{U}
\expandafter\def\csname BAval@tick.quad.f418.keep.iou\endcsname{0.00}
\expandafter\def\csname BAval@tick.quad.f418.keep.state\endcsname{U}
\expandafter\def\csname BAval@tick.quad.f419.keep.iou\endcsname{0.00}
\expandafter\def\csname BAval@tick.quad.f419.keep.state\endcsname{U}
\expandafter\def\csname BAval@tick.quad.f420.keep.iou\endcsname{0.00}
\expandafter\def\csname BAval@tick.quad.f420.keep.state\endcsname{W}
\expandafter\def\csname BAval@tick.quad.f421.keep.iou\endcsname{0.00}
\expandafter\def\csname BAval@tick.quad.f421.keep.state\endcsname{U}
\expandafter\def\csname BAval@tick.quad.f422.keep.iou\endcsname{0.00}
\expandafter\def\csname BAval@tick.quad.f422.keep.state\endcsname{U}
\expandafter\def\csname BAval@tick.quad.f423.keep.iou\endcsname{0.00}
\expandafter\def\csname BAval@tick.quad.f423.keep.state\endcsname{U}
\expandafter\def\csname BAval@tick.quad.f424.keep.iou\endcsname{0.00}
\expandafter\def\csname BAval@tick.quad.f424.keep.state\endcsname{U}
\expandafter\def\csname BAval@tick.quad.f425.keep.iou\endcsname{0.00}
\expandafter\def\csname BAval@tick.quad.f425.keep.state\endcsname{W}
\expandafter\def\csname BAval@tick.quad.f426.keep.iou\endcsname{0.00}
\expandafter\def\csname BAval@tick.quad.f426.keep.state\endcsname{W}
\expandafter\def\csname BAval@tick.quad.f427.keep.iou\endcsname{0.00}
\expandafter\def\csname BAval@tick.quad.f427.keep.state\endcsname{W}
\expandafter\def\csname BAval@tick.quad.f428.keep.iou\endcsname{0.00}
\expandafter\def\csname BAval@tick.quad.f428.keep.state\endcsname{U}
\expandafter\def\csname BAval@tick.quad.f429.keep.iou\endcsname{0.00}
\expandafter\def\csname BAval@tick.quad.f429.keep.state\endcsname{U}
\expandafter\def\csname BAval@tick.quad.f430.keep.iou\endcsname{0.00}
\expandafter\def\csname BAval@tick.quad.f430.keep.state\endcsname{U}
\expandafter\def\csname BAval@tick.quad.f431.keep.iou\endcsname{0.00}
\expandafter\def\csname BAval@tick.quad.f431.keep.state\endcsname{U}
\expandafter\def\csname BAval@tick.quad.f432.keep.iou\endcsname{0.00}
\expandafter\def\csname BAval@tick.quad.f432.keep.state\endcsname{U}
\expandafter\def\csname BAval@tick.quad.f433.keep.iou\endcsname{0.00}
\expandafter\def\csname BAval@tick.quad.f433.keep.state\endcsname{U}
\expandafter\def\csname BAval@tick.quad.f434.keep.iou\endcsname{0.00}
\expandafter\def\csname BAval@tick.quad.f434.keep.state\endcsname{U}
\expandafter\def\csname BAval@tick.quad.f435.keep.iou\endcsname{0.00}
\expandafter\def\csname BAval@tick.quad.f435.keep.state\endcsname{U}
\expandafter\def\csname BAval@tick.quad.f436.keep.iou\endcsname{0.00}
\expandafter\def\csname BAval@tick.quad.f436.keep.state\endcsname{UNKNOWN}
\expandafter\def\csname BAval@tick.quad.f437.keep.iou\endcsname{0.00}
\expandafter\def\csname BAval@tick.quad.f437.keep.state\endcsname{UNKNOWN}
\expandafter\def\csname BAval@tick.quad.f438.keep.iou\endcsname{0.00}
\expandafter\def\csname BAval@tick.quad.f438.keep.state\endcsname{UNKNOWN}
\expandafter\def\csname BAval@tick.quad.f439.keep.iou\endcsname{0.00}
\expandafter\def\csname BAval@tick.quad.f439.keep.state\endcsname{UNKNOWN}
\expandafter\def\csname BAval@tick.quad.f440.keep.iou\endcsname{0.00}
\expandafter\def\csname BAval@tick.quad.f440.keep.state\endcsname{UNKNOWN}
\expandafter\def\csname BAval@tick.quad.f441.keep.iou\endcsname{0.00}
\expandafter\def\csname BAval@tick.quad.f441.keep.state\endcsname{W}
\expandafter\def\csname BAval@tick.quad.f442.keep.iou\endcsname{0.00}
\expandafter\def\csname BAval@tick.quad.f442.keep.state\endcsname{W}
\expandafter\def\csname BAval@tick.quad.f443.keep.iou\endcsname{0.00}
\expandafter\def\csname BAval@tick.quad.f443.keep.state\endcsname{W}
\expandafter\def\csname BAval@tick.quad.f444.keep.iou\endcsname{0.00}
\expandafter\def\csname BAval@tick.quad.f444.keep.state\endcsname{U}
\expandafter\def\csname BAval@tick.quad.f445.keep.iou\endcsname{0.00}
\expandafter\def\csname BAval@tick.quad.f445.keep.state\endcsname{U}
\expandafter\def\csname BAval@tick.quad.f446.keep.iou\endcsname{0.00}
\expandafter\def\csname BAval@tick.quad.f446.keep.state\endcsname{U}
\expandafter\def\csname BAval@tick.quad.f447.keep.iou\endcsname{0.00}
\expandafter\def\csname BAval@tick.quad.f447.keep.state\endcsname{U}
\expandafter\def\csname BAval@tick.quad.f448.keep.iou\endcsname{0.00}
\expandafter\def\csname BAval@tick.quad.f448.keep.state\endcsname{U}
\expandafter\def\csname BAval@tick.quad.f449.keep.iou\endcsname{0.00}
\expandafter\def\csname BAval@tick.quad.f449.keep.state\endcsname{U}
\expandafter\def\csname BAval@tick.quad.f450.keep.iou\endcsname{0.00}
\expandafter\def\csname BAval@tick.quad.f450.keep.state\endcsname{U}
\expandafter\def\csname BAval@tick.quad.f451.keep.iou\endcsname{0.00}
\expandafter\def\csname BAval@tick.quad.f451.keep.state\endcsname{U}
\expandafter\def\csname BAval@tick.quad.f452.keep.iou\endcsname{0.00}
\expandafter\def\csname BAval@tick.quad.f452.keep.state\endcsname{U}
\expandafter\def\csname BAval@tick.quad.f453.keep.iou\endcsname{0.00}
\expandafter\def\csname BAval@tick.quad.f453.keep.state\endcsname{U}
\expandafter\def\csname BAval@tick.quad.f454.keep.iou\endcsname{0.00}
\expandafter\def\csname BAval@tick.quad.f454.keep.state\endcsname{UNKNOWN}
\expandafter\def\csname BAval@tick.quad.f455.keep.iou\endcsname{0.00}
\expandafter\def\csname BAval@tick.quad.f455.keep.state\endcsname{UNKNOWN}
\expandafter\def\csname BAval@tick.quad.f456.keep.iou\endcsname{0.00}
\expandafter\def\csname BAval@tick.quad.f456.keep.state\endcsname{UNKNOWN}
\expandafter\def\csname BAval@tick.quad.f457.keep.iou\endcsname{0.00}
\expandafter\def\csname BAval@tick.quad.f457.keep.state\endcsname{W}
\expandafter\def\csname BAval@tick.quad.f458.keep.iou\endcsname{0.00}
\expandafter\def\csname BAval@tick.quad.f458.keep.state\endcsname{U}
\expandafter\def\csname BAval@tick.quad.f459.keep.iou\endcsname{0.00}
\expandafter\def\csname BAval@tick.quad.f459.keep.state\endcsname{U}
\expandafter\def\csname BAval@tick.quad.f460.keep.iou\endcsname{0.00}
\expandafter\def\csname BAval@tick.quad.f460.keep.state\endcsname{U}
\expandafter\def\csname BAval@tick.quad.f461.keep.iou\endcsname{0.00}
\expandafter\def\csname BAval@tick.quad.f461.keep.state\endcsname{U}
\expandafter\def\csname BAval@tick.quad.f462.keep.iou\endcsname{0.00}
\expandafter\def\csname BAval@tick.quad.f462.keep.state\endcsname{U}
\expandafter\def\csname BAval@tick.quad.f463.keep.iou\endcsname{0.00}
\expandafter\def\csname BAval@tick.quad.f463.keep.state\endcsname{U}
\expandafter\def\csname BAval@tick.quad.f464.keep.iou\endcsname{0.00}
\expandafter\def\csname BAval@tick.quad.f464.keep.state\endcsname{U}
\expandafter\def\csname BAval@tick.quad.f465.keep.iou\endcsname{0.00}
\expandafter\def\csname BAval@tick.quad.f465.keep.state\endcsname{U}
\expandafter\def\csname BAval@tick.quad.f466.keep.iou\endcsname{0.00}
\expandafter\def\csname BAval@tick.quad.f466.keep.state\endcsname{U}
\expandafter\def\csname BAval@tick.quad.f467.keep.iou\endcsname{0.00}
\expandafter\def\csname BAval@tick.quad.f467.keep.state\endcsname{U}
\expandafter\def\csname BAval@tick.quad.f468.keep.iou\endcsname{0.00}
\expandafter\def\csname BAval@tick.quad.f468.keep.state\endcsname{U}
\expandafter\def\csname BAval@tick.quad.f469.keep.iou\endcsname{0.00}
\expandafter\def\csname BAval@tick.quad.f469.keep.state\endcsname{U}
\expandafter\def\csname BAval@tick.quad.f470.keep.iou\endcsname{0.00}
\expandafter\def\csname BAval@tick.quad.f470.keep.state\endcsname{U}
\expandafter\def\csname BAval@tick.quad.f471.keep.iou\endcsname{0.00}
\expandafter\def\csname BAval@tick.quad.f471.keep.state\endcsname{U}
\expandafter\def\csname BAval@tick.quad.f472.keep.iou\endcsname{0.00}
\expandafter\def\csname BAval@tick.quad.f472.keep.state\endcsname{U}
\expandafter\def\csname BAval@tick.quad.f473.keep.iou\endcsname{0.00}
\expandafter\def\csname BAval@tick.quad.f473.keep.state\endcsname{UNKNOWN}
\expandafter\def\csname BAval@tick.quad.f474.keep.iou\endcsname{0.00}
\expandafter\def\csname BAval@tick.quad.f474.keep.state\endcsname{UNKNOWN}
\expandafter\def\csname BAval@tick.quad.f475.keep.iou\endcsname{0.00}
\expandafter\def\csname BAval@tick.quad.f475.keep.state\endcsname{UNKNOWN}
\expandafter\def\csname BAval@tick.quad.f476.keep.iou\endcsname{0.00}
\expandafter\def\csname BAval@tick.quad.f476.keep.state\endcsname{UNKNOWN}
\expandafter\def\csname BAval@tick.quad.f477.keep.iou\endcsname{0.00}
\expandafter\def\csname BAval@tick.quad.f477.keep.state\endcsname{U}
\expandafter\def\csname BAval@tick.quad.f478.keep.iou\endcsname{0.00}
\expandafter\def\csname BAval@tick.quad.f478.keep.state\endcsname{U}
\expandafter\def\csname BAval@tick.quad.f479.keep.iou\endcsname{0.00}
\expandafter\def\csname BAval@tick.quad.f479.keep.state\endcsname{U}
\expandafter\def\csname BAval@tick.quad.f480.keep.iou\endcsname{0.00}
\expandafter\def\csname BAval@tick.quad.f480.keep.state\endcsname{U}
\expandafter\def\csname BAval@tick.quad.f481.keep.iou\endcsname{0.00}
\expandafter\def\csname BAval@tick.quad.f481.keep.state\endcsname{U}
\expandafter\def\csname BAval@tick.quad.f482.keep.iou\endcsname{0.00}
\expandafter\def\csname BAval@tick.quad.f482.keep.state\endcsname{U}
\expandafter\def\csname BAval@tick.quad.f483.keep.iou\endcsname{0.00}
\expandafter\def\csname BAval@tick.quad.f483.keep.state\endcsname{U}
\expandafter\def\csname BAval@tick.quad.f484.keep.iou\endcsname{0.00}
\expandafter\def\csname BAval@tick.quad.f484.keep.state\endcsname{U}
\expandafter\def\csname BAval@tick.quad.f485.keep.iou\endcsname{0.00}
\expandafter\def\csname BAval@tick.quad.f485.keep.state\endcsname{U}
\expandafter\def\csname BAval@tick.quad.f486.keep.iou\endcsname{0.00}
\expandafter\def\csname BAval@tick.quad.f486.keep.state\endcsname{W}
\expandafter\def\csname BAval@tick.quad.f487.keep.iou\endcsname{0.00}
\expandafter\def\csname BAval@tick.quad.f487.keep.state\endcsname{W}
\expandafter\def\csname BAval@tick.quad.f488.keep.iou\endcsname{0.00}
\expandafter\def\csname BAval@tick.quad.f488.keep.state\endcsname{W}
\expandafter\def\csname BAval@tick.quad.f489.keep.iou\endcsname{0.00}
\expandafter\def\csname BAval@tick.quad.f489.keep.state\endcsname{W}
\expandafter\def\csname BAval@tick.quad.f490.keep.iou\endcsname{0.00}
\expandafter\def\csname BAval@tick.quad.f490.keep.state\endcsname{U}
\expandafter\def\csname BAval@tick.quad.f491.keep.iou\endcsname{0.00}
\expandafter\def\csname BAval@tick.quad.f491.keep.state\endcsname{U}
\expandafter\def\csname BAval@tick.quad.f492.keep.iou\endcsname{0.00}
\expandafter\def\csname BAval@tick.quad.f492.keep.state\endcsname{U}
\expandafter\def\csname BAval@tick.quad.f493.keep.iou\endcsname{0.00}
\expandafter\def\csname BAval@tick.quad.f493.keep.state\endcsname{UNKNOWN}
\expandafter\def\csname BAval@tick.quad.f494.keep.iou\endcsname{0.00}
\expandafter\def\csname BAval@tick.quad.f494.keep.state\endcsname{UNKNOWN}
\expandafter\def\csname BAval@tick.quad.f495.keep.iou\endcsname{0.00}
\expandafter\def\csname BAval@tick.quad.f495.keep.state\endcsname{UNKNOWN}
\expandafter\def\csname BAval@tick.quad.f496.keep.iou\endcsname{0.00}
\expandafter\def\csname BAval@tick.quad.f496.keep.state\endcsname{UNKNOWN}
\expandafter\def\csname BAval@tick.quad.f497.keep.iou\endcsname{0.00}
\expandafter\def\csname BAval@tick.quad.f497.keep.state\endcsname{UNKNOWN}
\expandafter\def\csname BAval@tick.quad.f498.keep.iou\endcsname{0.00}
\expandafter\def\csname BAval@tick.quad.f498.keep.state\endcsname{UNKNOWN}
\expandafter\def\csname BAval@state.quad.bind.correct\endcsname{1}
\expandafter\def\csname BAval@state.quad.bind.wrong\endcsname{12}
\expandafter\def\csname BAval@state.quad.bind.unresolved\endcsname{59}
\expandafter\def\csname BAval@state.quad.bind.unknown\endcsname{18}
\expandafter\def\csname BAval@tick.quad.f409.bind.iou\endcsname{0.58}
\expandafter\def\csname BAval@tick.quad.f409.bind.state\endcsname{C}
\expandafter\def\csname BAval@tick.quad.f410.bind.iou\endcsname{0.36}
\expandafter\def\csname BAval@tick.quad.f410.bind.state\endcsname{U}
\expandafter\def\csname BAval@tick.quad.f411.bind.iou\endcsname{0.27}
\expandafter\def\csname BAval@tick.quad.f411.bind.state\endcsname{U}
\expandafter\def\csname BAval@tick.quad.f412.bind.iou\endcsname{0.10}
\expandafter\def\csname BAval@tick.quad.f412.bind.state\endcsname{U}
\expandafter\def\csname BAval@tick.quad.f413.bind.iou\endcsname{0.00}
\expandafter\def\csname BAval@tick.quad.f413.bind.state\endcsname{U}
\expandafter\def\csname BAval@tick.quad.f414.bind.iou\endcsname{0.00}
\expandafter\def\csname BAval@tick.quad.f414.bind.state\endcsname{U}
\expandafter\def\csname BAval@tick.quad.f415.bind.iou\endcsname{0.00}
\expandafter\def\csname BAval@tick.quad.f415.bind.state\endcsname{U}
\expandafter\def\csname BAval@tick.quad.f416.bind.iou\endcsname{0.00}
\expandafter\def\csname BAval@tick.quad.f416.bind.state\endcsname{U}
\expandafter\def\csname BAval@tick.quad.f417.bind.iou\endcsname{0.00}
\expandafter\def\csname BAval@tick.quad.f417.bind.state\endcsname{U}
\expandafter\def\csname BAval@tick.quad.f418.bind.iou\endcsname{0.00}
\expandafter\def\csname BAval@tick.quad.f418.bind.state\endcsname{U}
\expandafter\def\csname BAval@tick.quad.f419.bind.iou\endcsname{0.00}
\expandafter\def\csname BAval@tick.quad.f419.bind.state\endcsname{U}
\expandafter\def\csname BAval@tick.quad.f420.bind.iou\endcsname{0.00}
\expandafter\def\csname BAval@tick.quad.f420.bind.state\endcsname{W}
\expandafter\def\csname BAval@tick.quad.f421.bind.iou\endcsname{0.00}
\expandafter\def\csname BAval@tick.quad.f421.bind.state\endcsname{U}
\expandafter\def\csname BAval@tick.quad.f422.bind.iou\endcsname{0.00}
\expandafter\def\csname BAval@tick.quad.f422.bind.state\endcsname{U}
\expandafter\def\csname BAval@tick.quad.f423.bind.iou\endcsname{0.00}
\expandafter\def\csname BAval@tick.quad.f423.bind.state\endcsname{U}
\expandafter\def\csname BAval@tick.quad.f424.bind.iou\endcsname{0.00}
\expandafter\def\csname BAval@tick.quad.f424.bind.state\endcsname{U}
\expandafter\def\csname BAval@tick.quad.f425.bind.iou\endcsname{0.00}
\expandafter\def\csname BAval@tick.quad.f425.bind.state\endcsname{W}
\expandafter\def\csname BAval@tick.quad.f426.bind.iou\endcsname{0.00}
\expandafter\def\csname BAval@tick.quad.f426.bind.state\endcsname{W}
\expandafter\def\csname BAval@tick.quad.f427.bind.iou\endcsname{0.00}
\expandafter\def\csname BAval@tick.quad.f427.bind.state\endcsname{W}
\expandafter\def\csname BAval@tick.quad.f428.bind.iou\endcsname{0.00}
\expandafter\def\csname BAval@tick.quad.f428.bind.state\endcsname{U}
\expandafter\def\csname BAval@tick.quad.f429.bind.iou\endcsname{0.00}
\expandafter\def\csname BAval@tick.quad.f429.bind.state\endcsname{U}
\expandafter\def\csname BAval@tick.quad.f430.bind.iou\endcsname{0.00}
\expandafter\def\csname BAval@tick.quad.f430.bind.state\endcsname{U}
\expandafter\def\csname BAval@tick.quad.f431.bind.iou\endcsname{0.00}
\expandafter\def\csname BAval@tick.quad.f431.bind.state\endcsname{U}
\expandafter\def\csname BAval@tick.quad.f432.bind.iou\endcsname{0.00}
\expandafter\def\csname BAval@tick.quad.f432.bind.state\endcsname{U}
\expandafter\def\csname BAval@tick.quad.f433.bind.iou\endcsname{0.00}
\expandafter\def\csname BAval@tick.quad.f433.bind.state\endcsname{U}
\expandafter\def\csname BAval@tick.quad.f434.bind.iou\endcsname{0.00}
\expandafter\def\csname BAval@tick.quad.f434.bind.state\endcsname{U}
\expandafter\def\csname BAval@tick.quad.f435.bind.iou\endcsname{0.00}
\expandafter\def\csname BAval@tick.quad.f435.bind.state\endcsname{U}
\expandafter\def\csname BAval@tick.quad.f436.bind.iou\endcsname{0.00}
\expandafter\def\csname BAval@tick.quad.f436.bind.state\endcsname{UNKNOWN}
\expandafter\def\csname BAval@tick.quad.f437.bind.iou\endcsname{0.00}
\expandafter\def\csname BAval@tick.quad.f437.bind.state\endcsname{UNKNOWN}
\expandafter\def\csname BAval@tick.quad.f438.bind.iou\endcsname{0.00}
\expandafter\def\csname BAval@tick.quad.f438.bind.state\endcsname{UNKNOWN}
\expandafter\def\csname BAval@tick.quad.f439.bind.iou\endcsname{0.00}
\expandafter\def\csname BAval@tick.quad.f439.bind.state\endcsname{UNKNOWN}
\expandafter\def\csname BAval@tick.quad.f440.bind.iou\endcsname{0.00}
\expandafter\def\csname BAval@tick.quad.f440.bind.state\endcsname{UNKNOWN}
\expandafter\def\csname BAval@tick.quad.f441.bind.iou\endcsname{0.00}
\expandafter\def\csname BAval@tick.quad.f441.bind.state\endcsname{W}
\expandafter\def\csname BAval@tick.quad.f442.bind.iou\endcsname{0.00}
\expandafter\def\csname BAval@tick.quad.f442.bind.state\endcsname{W}
\expandafter\def\csname BAval@tick.quad.f443.bind.iou\endcsname{0.00}
\expandafter\def\csname BAval@tick.quad.f443.bind.state\endcsname{W}
\expandafter\def\csname BAval@tick.quad.f444.bind.iou\endcsname{0.00}
\expandafter\def\csname BAval@tick.quad.f444.bind.state\endcsname{U}
\expandafter\def\csname BAval@tick.quad.f445.bind.iou\endcsname{0.00}
\expandafter\def\csname BAval@tick.quad.f445.bind.state\endcsname{U}
\expandafter\def\csname BAval@tick.quad.f446.bind.iou\endcsname{0.00}
\expandafter\def\csname BAval@tick.quad.f446.bind.state\endcsname{U}
\expandafter\def\csname BAval@tick.quad.f447.bind.iou\endcsname{0.00}
\expandafter\def\csname BAval@tick.quad.f447.bind.state\endcsname{U}
\expandafter\def\csname BAval@tick.quad.f448.bind.iou\endcsname{0.00}
\expandafter\def\csname BAval@tick.quad.f448.bind.state\endcsname{U}
\expandafter\def\csname BAval@tick.quad.f449.bind.iou\endcsname{0.00}
\expandafter\def\csname BAval@tick.quad.f449.bind.state\endcsname{U}
\expandafter\def\csname BAval@tick.quad.f450.bind.iou\endcsname{0.00}
\expandafter\def\csname BAval@tick.quad.f450.bind.state\endcsname{U}
\expandafter\def\csname BAval@tick.quad.f451.bind.iou\endcsname{0.00}
\expandafter\def\csname BAval@tick.quad.f451.bind.state\endcsname{U}
\expandafter\def\csname BAval@tick.quad.f452.bind.iou\endcsname{0.00}
\expandafter\def\csname BAval@tick.quad.f452.bind.state\endcsname{U}
\expandafter\def\csname BAval@tick.quad.f453.bind.iou\endcsname{0.00}
\expandafter\def\csname BAval@tick.quad.f453.bind.state\endcsname{U}
\expandafter\def\csname BAval@tick.quad.f454.bind.iou\endcsname{0.00}
\expandafter\def\csname BAval@tick.quad.f454.bind.state\endcsname{UNKNOWN}
\expandafter\def\csname BAval@tick.quad.f455.bind.iou\endcsname{0.00}
\expandafter\def\csname BAval@tick.quad.f455.bind.state\endcsname{UNKNOWN}
\expandafter\def\csname BAval@tick.quad.f456.bind.iou\endcsname{0.00}
\expandafter\def\csname BAval@tick.quad.f456.bind.state\endcsname{UNKNOWN}
\expandafter\def\csname BAval@tick.quad.f457.bind.iou\endcsname{0.00}
\expandafter\def\csname BAval@tick.quad.f457.bind.state\endcsname{W}
\expandafter\def\csname BAval@tick.quad.f458.bind.iou\endcsname{0.00}
\expandafter\def\csname BAval@tick.quad.f458.bind.state\endcsname{U}
\expandafter\def\csname BAval@tick.quad.f459.bind.iou\endcsname{0.00}
\expandafter\def\csname BAval@tick.quad.f459.bind.state\endcsname{U}
\expandafter\def\csname BAval@tick.quad.f460.bind.iou\endcsname{0.00}
\expandafter\def\csname BAval@tick.quad.f460.bind.state\endcsname{U}
\expandafter\def\csname BAval@tick.quad.f461.bind.iou\endcsname{0.00}
\expandafter\def\csname BAval@tick.quad.f461.bind.state\endcsname{U}
\expandafter\def\csname BAval@tick.quad.f462.bind.iou\endcsname{0.00}
\expandafter\def\csname BAval@tick.quad.f462.bind.state\endcsname{U}
\expandafter\def\csname BAval@tick.quad.f463.bind.iou\endcsname{0.00}
\expandafter\def\csname BAval@tick.quad.f463.bind.state\endcsname{U}
\expandafter\def\csname BAval@tick.quad.f464.bind.iou\endcsname{0.00}
\expandafter\def\csname BAval@tick.quad.f464.bind.state\endcsname{U}
\expandafter\def\csname BAval@tick.quad.f465.bind.iou\endcsname{0.00}
\expandafter\def\csname BAval@tick.quad.f465.bind.state\endcsname{U}
\expandafter\def\csname BAval@tick.quad.f466.bind.iou\endcsname{0.00}
\expandafter\def\csname BAval@tick.quad.f466.bind.state\endcsname{U}
\expandafter\def\csname BAval@tick.quad.f467.bind.iou\endcsname{0.00}
\expandafter\def\csname BAval@tick.quad.f467.bind.state\endcsname{U}
\expandafter\def\csname BAval@tick.quad.f468.bind.iou\endcsname{0.00}
\expandafter\def\csname BAval@tick.quad.f468.bind.state\endcsname{U}
\expandafter\def\csname BAval@tick.quad.f469.bind.iou\endcsname{0.00}
\expandafter\def\csname BAval@tick.quad.f469.bind.state\endcsname{U}
\expandafter\def\csname BAval@tick.quad.f470.bind.iou\endcsname{0.00}
\expandafter\def\csname BAval@tick.quad.f470.bind.state\endcsname{U}
\expandafter\def\csname BAval@tick.quad.f471.bind.iou\endcsname{0.00}
\expandafter\def\csname BAval@tick.quad.f471.bind.state\endcsname{U}
\expandafter\def\csname BAval@tick.quad.f472.bind.iou\endcsname{0.00}
\expandafter\def\csname BAval@tick.quad.f472.bind.state\endcsname{U}
\expandafter\def\csname BAval@tick.quad.f473.bind.iou\endcsname{0.00}
\expandafter\def\csname BAval@tick.quad.f473.bind.state\endcsname{UNKNOWN}
\expandafter\def\csname BAval@tick.quad.f474.bind.iou\endcsname{0.00}
\expandafter\def\csname BAval@tick.quad.f474.bind.state\endcsname{UNKNOWN}
\expandafter\def\csname BAval@tick.quad.f475.bind.iou\endcsname{0.00}
\expandafter\def\csname BAval@tick.quad.f475.bind.state\endcsname{UNKNOWN}
\expandafter\def\csname BAval@tick.quad.f476.bind.iou\endcsname{0.00}
\expandafter\def\csname BAval@tick.quad.f476.bind.state\endcsname{UNKNOWN}
\expandafter\def\csname BAval@tick.quad.f477.bind.iou\endcsname{0.00}
\expandafter\def\csname BAval@tick.quad.f477.bind.state\endcsname{U}
\expandafter\def\csname BAval@tick.quad.f478.bind.iou\endcsname{0.00}
\expandafter\def\csname BAval@tick.quad.f478.bind.state\endcsname{U}
\expandafter\def\csname BAval@tick.quad.f479.bind.iou\endcsname{0.00}
\expandafter\def\csname BAval@tick.quad.f479.bind.state\endcsname{U}
\expandafter\def\csname BAval@tick.quad.f480.bind.iou\endcsname{0.00}
\expandafter\def\csname BAval@tick.quad.f480.bind.state\endcsname{U}
\expandafter\def\csname BAval@tick.quad.f481.bind.iou\endcsname{0.00}
\expandafter\def\csname BAval@tick.quad.f481.bind.state\endcsname{U}
\expandafter\def\csname BAval@tick.quad.f482.bind.iou\endcsname{0.00}
\expandafter\def\csname BAval@tick.quad.f482.bind.state\endcsname{U}
\expandafter\def\csname BAval@tick.quad.f483.bind.iou\endcsname{0.00}
\expandafter\def\csname BAval@tick.quad.f483.bind.state\endcsname{U}
\expandafter\def\csname BAval@tick.quad.f484.bind.iou\endcsname{0.00}
\expandafter\def\csname BAval@tick.quad.f484.bind.state\endcsname{U}
\expandafter\def\csname BAval@tick.quad.f485.bind.iou\endcsname{0.00}
\expandafter\def\csname BAval@tick.quad.f485.bind.state\endcsname{U}
\expandafter\def\csname BAval@tick.quad.f486.bind.iou\endcsname{0.00}
\expandafter\def\csname BAval@tick.quad.f486.bind.state\endcsname{W}
\expandafter\def\csname BAval@tick.quad.f487.bind.iou\endcsname{0.00}
\expandafter\def\csname BAval@tick.quad.f487.bind.state\endcsname{W}
\expandafter\def\csname BAval@tick.quad.f488.bind.iou\endcsname{0.00}
\expandafter\def\csname BAval@tick.quad.f488.bind.state\endcsname{W}
\expandafter\def\csname BAval@tick.quad.f489.bind.iou\endcsname{0.00}
\expandafter\def\csname BAval@tick.quad.f489.bind.state\endcsname{W}
\expandafter\def\csname BAval@tick.quad.f490.bind.iou\endcsname{0.00}
\expandafter\def\csname BAval@tick.quad.f490.bind.state\endcsname{U}
\expandafter\def\csname BAval@tick.quad.f491.bind.iou\endcsname{0.00}
\expandafter\def\csname BAval@tick.quad.f491.bind.state\endcsname{U}
\expandafter\def\csname BAval@tick.quad.f492.bind.iou\endcsname{0.00}
\expandafter\def\csname BAval@tick.quad.f492.bind.state\endcsname{U}
\expandafter\def\csname BAval@tick.quad.f493.bind.iou\endcsname{0.00}
\expandafter\def\csname BAval@tick.quad.f493.bind.state\endcsname{UNKNOWN}
\expandafter\def\csname BAval@tick.quad.f494.bind.iou\endcsname{0.00}
\expandafter\def\csname BAval@tick.quad.f494.bind.state\endcsname{UNKNOWN}
\expandafter\def\csname BAval@tick.quad.f495.bind.iou\endcsname{0.00}
\expandafter\def\csname BAval@tick.quad.f495.bind.state\endcsname{UNKNOWN}
\expandafter\def\csname BAval@tick.quad.f496.bind.iou\endcsname{0.00}
\expandafter\def\csname BAval@tick.quad.f496.bind.state\endcsname{UNKNOWN}
\expandafter\def\csname BAval@tick.quad.f497.bind.iou\endcsname{0.00}
\expandafter\def\csname BAval@tick.quad.f497.bind.state\endcsname{UNKNOWN}
\expandafter\def\csname BAval@tick.quad.f498.bind.iou\endcsname{0.00}
\expandafter\def\csname BAval@tick.quad.f498.bind.state\endcsname{UNKNOWN}
\expandafter\def\csname BAval@rec.dance.attempts\endcsname{12}
\expandafter\def\csname BAval@rec.dance.complete\endcsname{0}
\expandafter\def\csname BAval@rec.dance.fallwait\endcsname{10}
\expandafter\def\csname BAval@rec.dance.expired\endcsname{2}
\expandafter\def\csname BAval@rec.dance.requests\endcsname{0}
\expandafter\def\csname BAval@rec.dance.returns\endcsname{0}
\expandafter\def\csname BAval@rec.dance.initialpub\endcsname{1}
\expandafter\def\csname BAval@rec.dance.laterpub\endcsname{0}
\expandafter\def\csname BAval@rec.dance.nativeupdates\endcsname{0}
\expandafter\def\csname BAval@rec.dance.ticks\endcsname{180}
\expandafter\def\csname BAval@state.dance.dam.correct\endcsname{78}
\expandafter\def\csname BAval@state.dance.dam.wrong\endcsname{20}
\expandafter\def\csname BAval@state.dance.dam.unresolved\endcsname{82}
\expandafter\def\csname BAval@state.dance.dam.unknown\endcsname{0}
\expandafter\def\csname BAval@tick.dance.f159.dam.iou\endcsname{0.99}
\expandafter\def\csname BAval@tick.dance.f159.dam.state\endcsname{C}
\expandafter\def\csname BAval@tick.dance.f160.dam.iou\endcsname{0.98}
\expandafter\def\csname BAval@tick.dance.f160.dam.state\endcsname{C}
\expandafter\def\csname BAval@tick.dance.f161.dam.iou\endcsname{0.97}
\expandafter\def\csname BAval@tick.dance.f161.dam.state\endcsname{C}
\expandafter\def\csname BAval@tick.dance.f162.dam.iou\endcsname{0.97}
\expandafter\def\csname BAval@tick.dance.f162.dam.state\endcsname{C}
\expandafter\def\csname BAval@tick.dance.f163.dam.iou\endcsname{0.95}
\expandafter\def\csname BAval@tick.dance.f163.dam.state\endcsname{C}
\expandafter\def\csname BAval@tick.dance.f164.dam.iou\endcsname{0.94}
\expandafter\def\csname BAval@tick.dance.f164.dam.state\endcsname{C}
\expandafter\def\csname BAval@tick.dance.f165.dam.iou\endcsname{0.92}
\expandafter\def\csname BAval@tick.dance.f165.dam.state\endcsname{C}
\expandafter\def\csname BAval@tick.dance.f166.dam.iou\endcsname{0.89}
\expandafter\def\csname BAval@tick.dance.f166.dam.state\endcsname{C}
\expandafter\def\csname BAval@tick.dance.f167.dam.iou\endcsname{0.87}
\expandafter\def\csname BAval@tick.dance.f167.dam.state\endcsname{C}
\expandafter\def\csname BAval@tick.dance.f168.dam.iou\endcsname{0.86}
\expandafter\def\csname BAval@tick.dance.f168.dam.state\endcsname{C}
\expandafter\def\csname BAval@tick.dance.f169.dam.iou\endcsname{0.83}
\expandafter\def\csname BAval@tick.dance.f169.dam.state\endcsname{C}
\expandafter\def\csname BAval@tick.dance.f170.dam.iou\endcsname{0.60}
\expandafter\def\csname BAval@tick.dance.f170.dam.state\endcsname{C}
\expandafter\def\csname BAval@tick.dance.f171.dam.iou\endcsname{0.52}
\expandafter\def\csname BAval@tick.dance.f171.dam.state\endcsname{C}
\expandafter\def\csname BAval@tick.dance.f172.dam.iou\endcsname{0.46}
\expandafter\def\csname BAval@tick.dance.f172.dam.state\endcsname{U}
\expandafter\def\csname BAval@tick.dance.f173.dam.iou\endcsname{0.51}
\expandafter\def\csname BAval@tick.dance.f173.dam.state\endcsname{C}
\expandafter\def\csname BAval@tick.dance.f174.dam.iou\endcsname{0.45}
\expandafter\def\csname BAval@tick.dance.f174.dam.state\endcsname{U}
\expandafter\def\csname BAval@tick.dance.f175.dam.iou\endcsname{0.44}
\expandafter\def\csname BAval@tick.dance.f175.dam.state\endcsname{U}
\expandafter\def\csname BAval@tick.dance.f176.dam.iou\endcsname{0.42}
\expandafter\def\csname BAval@tick.dance.f176.dam.state\endcsname{U}
\expandafter\def\csname BAval@tick.dance.f177.dam.iou\endcsname{0.42}
\expandafter\def\csname BAval@tick.dance.f177.dam.state\endcsname{U}
\expandafter\def\csname BAval@tick.dance.f178.dam.iou\endcsname{0.41}
\expandafter\def\csname BAval@tick.dance.f178.dam.state\endcsname{U}
\expandafter\def\csname BAval@tick.dance.f179.dam.iou\endcsname{0.39}
\expandafter\def\csname BAval@tick.dance.f179.dam.state\endcsname{U}
\expandafter\def\csname BAval@tick.dance.f180.dam.iou\endcsname{0.30}
\expandafter\def\csname BAval@tick.dance.f180.dam.state\endcsname{U}
\expandafter\def\csname BAval@tick.dance.f181.dam.iou\endcsname{0.27}
\expandafter\def\csname BAval@tick.dance.f181.dam.state\endcsname{U}
\expandafter\def\csname BAval@tick.dance.f182.dam.iou\endcsname{0.19}
\expandafter\def\csname BAval@tick.dance.f182.dam.state\endcsname{U}
\expandafter\def\csname BAval@tick.dance.f183.dam.iou\endcsname{0.14}
\expandafter\def\csname BAval@tick.dance.f183.dam.state\endcsname{U}
\expandafter\def\csname BAval@tick.dance.f184.dam.iou\endcsname{0.10}
\expandafter\def\csname BAval@tick.dance.f184.dam.state\endcsname{U}
\expandafter\def\csname BAval@tick.dance.f185.dam.iou\endcsname{0.08}
\expandafter\def\csname BAval@tick.dance.f185.dam.state\endcsname{W}
\expandafter\def\csname BAval@tick.dance.f186.dam.iou\endcsname{0.06}
\expandafter\def\csname BAval@tick.dance.f186.dam.state\endcsname{W}
\expandafter\def\csname BAval@tick.dance.f187.dam.iou\endcsname{0.04}
\expandafter\def\csname BAval@tick.dance.f187.dam.state\endcsname{W}
\expandafter\def\csname BAval@tick.dance.f188.dam.iou\endcsname{0.03}
\expandafter\def\csname BAval@tick.dance.f188.dam.state\endcsname{W}
\expandafter\def\csname BAval@tick.dance.f189.dam.iou\endcsname{0.02}
\expandafter\def\csname BAval@tick.dance.f189.dam.state\endcsname{W}
\expandafter\def\csname BAval@tick.dance.f190.dam.iou\endcsname{0.02}
\expandafter\def\csname BAval@tick.dance.f190.dam.state\endcsname{W}
\expandafter\def\csname BAval@tick.dance.f191.dam.iou\endcsname{0.02}
\expandafter\def\csname BAval@tick.dance.f191.dam.state\endcsname{W}
\expandafter\def\csname BAval@tick.dance.f192.dam.iou\endcsname{0.00}
\expandafter\def\csname BAval@tick.dance.f192.dam.state\endcsname{W}
\expandafter\def\csname BAval@tick.dance.f193.dam.iou\endcsname{0.00}
\expandafter\def\csname BAval@tick.dance.f193.dam.state\endcsname{W}
\expandafter\def\csname BAval@tick.dance.f194.dam.iou\endcsname{0.00}
\expandafter\def\csname BAval@tick.dance.f194.dam.state\endcsname{W}
\expandafter\def\csname BAval@tick.dance.f195.dam.iou\endcsname{0.00}
\expandafter\def\csname BAval@tick.dance.f195.dam.state\endcsname{W}
\expandafter\def\csname BAval@tick.dance.f196.dam.iou\endcsname{0.00}
\expandafter\def\csname BAval@tick.dance.f196.dam.state\endcsname{W}
\expandafter\def\csname BAval@tick.dance.f197.dam.iou\endcsname{0.00}
\expandafter\def\csname BAval@tick.dance.f197.dam.state\endcsname{W}
\expandafter\def\csname BAval@tick.dance.f198.dam.iou\endcsname{0.00}
\expandafter\def\csname BAval@tick.dance.f198.dam.state\endcsname{U}
\expandafter\def\csname BAval@tick.dance.f199.dam.iou\endcsname{0.00}
\expandafter\def\csname BAval@tick.dance.f199.dam.state\endcsname{U}
\expandafter\def\csname BAval@tick.dance.f200.dam.iou\endcsname{0.00}
\expandafter\def\csname BAval@tick.dance.f200.dam.state\endcsname{U}
\expandafter\def\csname BAval@tick.dance.f201.dam.iou\endcsname{0.00}
\expandafter\def\csname BAval@tick.dance.f201.dam.state\endcsname{U}
\expandafter\def\csname BAval@tick.dance.f202.dam.iou\endcsname{0.00}
\expandafter\def\csname BAval@tick.dance.f202.dam.state\endcsname{U}
\expandafter\def\csname BAval@tick.dance.f203.dam.iou\endcsname{0.00}
\expandafter\def\csname BAval@tick.dance.f203.dam.state\endcsname{U}
\expandafter\def\csname BAval@tick.dance.f204.dam.iou\endcsname{0.00}
\expandafter\def\csname BAval@tick.dance.f204.dam.state\endcsname{U}
\expandafter\def\csname BAval@tick.dance.f205.dam.iou\endcsname{0.00}
\expandafter\def\csname BAval@tick.dance.f205.dam.state\endcsname{U}
\expandafter\def\csname BAval@tick.dance.f206.dam.iou\endcsname{0.03}
\expandafter\def\csname BAval@tick.dance.f206.dam.state\endcsname{U}
\expandafter\def\csname BAval@tick.dance.f207.dam.iou\endcsname{0.04}
\expandafter\def\csname BAval@tick.dance.f207.dam.state\endcsname{U}
\expandafter\def\csname BAval@tick.dance.f208.dam.iou\endcsname{0.04}
\expandafter\def\csname BAval@tick.dance.f208.dam.state\endcsname{U}
\expandafter\def\csname BAval@tick.dance.f209.dam.iou\endcsname{0.04}
\expandafter\def\csname BAval@tick.dance.f209.dam.state\endcsname{U}
\expandafter\def\csname BAval@tick.dance.f210.dam.iou\endcsname{0.07}
\expandafter\def\csname BAval@tick.dance.f210.dam.state\endcsname{U}
\expandafter\def\csname BAval@tick.dance.f211.dam.iou\endcsname{0.09}
\expandafter\def\csname BAval@tick.dance.f211.dam.state\endcsname{U}
\expandafter\def\csname BAval@tick.dance.f212.dam.iou\endcsname{0.12}
\expandafter\def\csname BAval@tick.dance.f212.dam.state\endcsname{U}
\expandafter\def\csname BAval@tick.dance.f213.dam.iou\endcsname{0.13}
\expandafter\def\csname BAval@tick.dance.f213.dam.state\endcsname{U}
\expandafter\def\csname BAval@tick.dance.f214.dam.iou\endcsname{0.14}
\expandafter\def\csname BAval@tick.dance.f214.dam.state\endcsname{U}
\expandafter\def\csname BAval@tick.dance.f215.dam.iou\endcsname{0.15}
\expandafter\def\csname BAval@tick.dance.f215.dam.state\endcsname{U}
\expandafter\def\csname BAval@tick.dance.f216.dam.iou\endcsname{0.21}
\expandafter\def\csname BAval@tick.dance.f216.dam.state\endcsname{U}
\expandafter\def\csname BAval@tick.dance.f217.dam.iou\endcsname{0.24}
\expandafter\def\csname BAval@tick.dance.f217.dam.state\endcsname{U}
\expandafter\def\csname BAval@tick.dance.f218.dam.iou\endcsname{0.37}
\expandafter\def\csname BAval@tick.dance.f218.dam.state\endcsname{U}
\expandafter\def\csname BAval@tick.dance.f219.dam.iou\endcsname{0.41}
\expandafter\def\csname BAval@tick.dance.f219.dam.state\endcsname{U}
\expandafter\def\csname BAval@tick.dance.f220.dam.iou\endcsname{0.46}
\expandafter\def\csname BAval@tick.dance.f220.dam.state\endcsname{U}
\expandafter\def\csname BAval@tick.dance.f221.dam.iou\endcsname{0.51}
\expandafter\def\csname BAval@tick.dance.f221.dam.state\endcsname{C}
\expandafter\def\csname BAval@tick.dance.f222.dam.iou\endcsname{0.53}
\expandafter\def\csname BAval@tick.dance.f222.dam.state\endcsname{C}
\expandafter\def\csname BAval@tick.dance.f223.dam.iou\endcsname{0.53}
\expandafter\def\csname BAval@tick.dance.f223.dam.state\endcsname{C}
\expandafter\def\csname BAval@tick.dance.f224.dam.iou\endcsname{0.53}
\expandafter\def\csname BAval@tick.dance.f224.dam.state\endcsname{C}
\expandafter\def\csname BAval@tick.dance.f225.dam.iou\endcsname{0.54}
\expandafter\def\csname BAval@tick.dance.f225.dam.state\endcsname{C}
\expandafter\def\csname BAval@tick.dance.f226.dam.iou\endcsname{0.56}
\expandafter\def\csname BAval@tick.dance.f226.dam.state\endcsname{C}
\expandafter\def\csname BAval@tick.dance.f227.dam.iou\endcsname{0.55}
\expandafter\def\csname BAval@tick.dance.f227.dam.state\endcsname{C}
\expandafter\def\csname BAval@tick.dance.f228.dam.iou\endcsname{0.55}
\expandafter\def\csname BAval@tick.dance.f228.dam.state\endcsname{C}
\expandafter\def\csname BAval@tick.dance.f229.dam.iou\endcsname{0.54}
\expandafter\def\csname BAval@tick.dance.f229.dam.state\endcsname{C}
\expandafter\def\csname BAval@tick.dance.f230.dam.iou\endcsname{0.55}
\expandafter\def\csname BAval@tick.dance.f230.dam.state\endcsname{C}
\expandafter\def\csname BAval@tick.dance.f231.dam.iou\endcsname{0.55}
\expandafter\def\csname BAval@tick.dance.f231.dam.state\endcsname{C}
\expandafter\def\csname BAval@tick.dance.f232.dam.iou\endcsname{0.55}
\expandafter\def\csname BAval@tick.dance.f232.dam.state\endcsname{C}
\expandafter\def\csname BAval@tick.dance.f233.dam.iou\endcsname{0.56}
\expandafter\def\csname BAval@tick.dance.f233.dam.state\endcsname{C}
\expandafter\def\csname BAval@tick.dance.f234.dam.iou\endcsname{0.56}
\expandafter\def\csname BAval@tick.dance.f234.dam.state\endcsname{C}
\expandafter\def\csname BAval@tick.dance.f235.dam.iou\endcsname{0.56}
\expandafter\def\csname BAval@tick.dance.f235.dam.state\endcsname{C}
\expandafter\def\csname BAval@tick.dance.f236.dam.iou\endcsname{0.57}
\expandafter\def\csname BAval@tick.dance.f236.dam.state\endcsname{C}
\expandafter\def\csname BAval@tick.dance.f237.dam.iou\endcsname{0.58}
\expandafter\def\csname BAval@tick.dance.f237.dam.state\endcsname{C}
\expandafter\def\csname BAval@tick.dance.f238.dam.iou\endcsname{0.57}
\expandafter\def\csname BAval@tick.dance.f238.dam.state\endcsname{C}
\expandafter\def\csname BAval@tick.dance.f239.dam.iou\endcsname{0.58}
\expandafter\def\csname BAval@tick.dance.f239.dam.state\endcsname{C}
\expandafter\def\csname BAval@tick.dance.f240.dam.iou\endcsname{0.57}
\expandafter\def\csname BAval@tick.dance.f240.dam.state\endcsname{C}
\expandafter\def\csname BAval@tick.dance.f241.dam.iou\endcsname{0.58}
\expandafter\def\csname BAval@tick.dance.f241.dam.state\endcsname{C}
\expandafter\def\csname BAval@tick.dance.f242.dam.iou\endcsname{0.58}
\expandafter\def\csname BAval@tick.dance.f242.dam.state\endcsname{C}
\expandafter\def\csname BAval@tick.dance.f243.dam.iou\endcsname{0.59}
\expandafter\def\csname BAval@tick.dance.f243.dam.state\endcsname{C}
\expandafter\def\csname BAval@tick.dance.f244.dam.iou\endcsname{0.59}
\expandafter\def\csname BAval@tick.dance.f244.dam.state\endcsname{C}
\expandafter\def\csname BAval@tick.dance.f245.dam.iou\endcsname{0.60}
\expandafter\def\csname BAval@tick.dance.f245.dam.state\endcsname{C}
\expandafter\def\csname BAval@tick.dance.f246.dam.iou\endcsname{0.58}
\expandafter\def\csname BAval@tick.dance.f246.dam.state\endcsname{C}
\expandafter\def\csname BAval@tick.dance.f247.dam.iou\endcsname{0.60}
\expandafter\def\csname BAval@tick.dance.f247.dam.state\endcsname{C}
\expandafter\def\csname BAval@tick.dance.f248.dam.iou\endcsname{0.58}
\expandafter\def\csname BAval@tick.dance.f248.dam.state\endcsname{C}
\expandafter\def\csname BAval@tick.dance.f249.dam.iou\endcsname{0.57}
\expandafter\def\csname BAval@tick.dance.f249.dam.state\endcsname{C}
\expandafter\def\csname BAval@tick.dance.f250.dam.iou\endcsname{0.57}
\expandafter\def\csname BAval@tick.dance.f250.dam.state\endcsname{C}
\expandafter\def\csname BAval@tick.dance.f251.dam.iou\endcsname{0.57}
\expandafter\def\csname BAval@tick.dance.f251.dam.state\endcsname{C}
\expandafter\def\csname BAval@tick.dance.f252.dam.iou\endcsname{0.57}
\expandafter\def\csname BAval@tick.dance.f252.dam.state\endcsname{C}
\expandafter\def\csname BAval@tick.dance.f253.dam.iou\endcsname{0.56}
\expandafter\def\csname BAval@tick.dance.f253.dam.state\endcsname{C}
\expandafter\def\csname BAval@tick.dance.f254.dam.iou\endcsname{0.56}
\expandafter\def\csname BAval@tick.dance.f254.dam.state\endcsname{C}
\expandafter\def\csname BAval@tick.dance.f255.dam.iou\endcsname{0.56}
\expandafter\def\csname BAval@tick.dance.f255.dam.state\endcsname{C}
\expandafter\def\csname BAval@tick.dance.f256.dam.iou\endcsname{0.56}
\expandafter\def\csname BAval@tick.dance.f256.dam.state\endcsname{C}
\expandafter\def\csname BAval@tick.dance.f257.dam.iou\endcsname{0.56}
\expandafter\def\csname BAval@tick.dance.f257.dam.state\endcsname{C}
\expandafter\def\csname BAval@tick.dance.f258.dam.iou\endcsname{0.56}
\expandafter\def\csname BAval@tick.dance.f258.dam.state\endcsname{C}
\expandafter\def\csname BAval@tick.dance.f259.dam.iou\endcsname{0.56}
\expandafter\def\csname BAval@tick.dance.f259.dam.state\endcsname{C}
\expandafter\def\csname BAval@tick.dance.f260.dam.iou\endcsname{0.56}
\expandafter\def\csname BAval@tick.dance.f260.dam.state\endcsname{C}
\expandafter\def\csname BAval@tick.dance.f261.dam.iou\endcsname{0.56}
\expandafter\def\csname BAval@tick.dance.f261.dam.state\endcsname{C}
\expandafter\def\csname BAval@tick.dance.f262.dam.iou\endcsname{0.56}
\expandafter\def\csname BAval@tick.dance.f262.dam.state\endcsname{C}
\expandafter\def\csname BAval@tick.dance.f263.dam.iou\endcsname{0.56}
\expandafter\def\csname BAval@tick.dance.f263.dam.state\endcsname{C}
\expandafter\def\csname BAval@tick.dance.f264.dam.iou\endcsname{0.56}
\expandafter\def\csname BAval@tick.dance.f264.dam.state\endcsname{C}
\expandafter\def\csname BAval@tick.dance.f265.dam.iou\endcsname{0.56}
\expandafter\def\csname BAval@tick.dance.f265.dam.state\endcsname{C}
\expandafter\def\csname BAval@tick.dance.f266.dam.iou\endcsname{0.56}
\expandafter\def\csname BAval@tick.dance.f266.dam.state\endcsname{C}
\expandafter\def\csname BAval@tick.dance.f267.dam.iou\endcsname{0.56}
\expandafter\def\csname BAval@tick.dance.f267.dam.state\endcsname{C}
\expandafter\def\csname BAval@tick.dance.f268.dam.iou\endcsname{0.57}
\expandafter\def\csname BAval@tick.dance.f268.dam.state\endcsname{C}
\expandafter\def\csname BAval@tick.dance.f269.dam.iou\endcsname{0.58}
\expandafter\def\csname BAval@tick.dance.f269.dam.state\endcsname{C}
\expandafter\def\csname BAval@tick.dance.f270.dam.iou\endcsname{0.58}
\expandafter\def\csname BAval@tick.dance.f270.dam.state\endcsname{C}
\expandafter\def\csname BAval@tick.dance.f271.dam.iou\endcsname{0.59}
\expandafter\def\csname BAval@tick.dance.f271.dam.state\endcsname{C}
\expandafter\def\csname BAval@tick.dance.f272.dam.iou\endcsname{0.59}
\expandafter\def\csname BAval@tick.dance.f272.dam.state\endcsname{C}
\expandafter\def\csname BAval@tick.dance.f273.dam.iou\endcsname{0.60}
\expandafter\def\csname BAval@tick.dance.f273.dam.state\endcsname{C}
\expandafter\def\csname BAval@tick.dance.f274.dam.iou\endcsname{0.61}
\expandafter\def\csname BAval@tick.dance.f274.dam.state\endcsname{C}
\expandafter\def\csname BAval@tick.dance.f275.dam.iou\endcsname{0.62}
\expandafter\def\csname BAval@tick.dance.f275.dam.state\endcsname{C}
\expandafter\def\csname BAval@tick.dance.f276.dam.iou\endcsname{0.61}
\expandafter\def\csname BAval@tick.dance.f276.dam.state\endcsname{C}
\expandafter\def\csname BAval@tick.dance.f277.dam.iou\endcsname{0.61}
\expandafter\def\csname BAval@tick.dance.f277.dam.state\endcsname{C}
\expandafter\def\csname BAval@tick.dance.f278.dam.iou\endcsname{0.61}
\expandafter\def\csname BAval@tick.dance.f278.dam.state\endcsname{C}
\expandafter\def\csname BAval@tick.dance.f279.dam.iou\endcsname{0.54}
\expandafter\def\csname BAval@tick.dance.f279.dam.state\endcsname{C}
\expandafter\def\csname BAval@tick.dance.f280.dam.iou\endcsname{0.55}
\expandafter\def\csname BAval@tick.dance.f280.dam.state\endcsname{C}
\expandafter\def\csname BAval@tick.dance.f281.dam.iou\endcsname{0.57}
\expandafter\def\csname BAval@tick.dance.f281.dam.state\endcsname{C}
\expandafter\def\csname BAval@tick.dance.f282.dam.iou\endcsname{0.59}
\expandafter\def\csname BAval@tick.dance.f282.dam.state\endcsname{C}
\expandafter\def\csname BAval@tick.dance.f283.dam.iou\endcsname{0.64}
\expandafter\def\csname BAval@tick.dance.f283.dam.state\endcsname{C}
\expandafter\def\csname BAval@tick.dance.f284.dam.iou\endcsname{0.61}
\expandafter\def\csname BAval@tick.dance.f284.dam.state\endcsname{C}
\expandafter\def\csname BAval@tick.dance.f285.dam.iou\endcsname{0.47}
\expandafter\def\csname BAval@tick.dance.f285.dam.state\endcsname{U}
\expandafter\def\csname BAval@tick.dance.f286.dam.iou\endcsname{0.40}
\expandafter\def\csname BAval@tick.dance.f286.dam.state\endcsname{U}
\expandafter\def\csname BAval@tick.dance.f287.dam.iou\endcsname{0.30}
\expandafter\def\csname BAval@tick.dance.f287.dam.state\endcsname{U}
\expandafter\def\csname BAval@tick.dance.f288.dam.iou\endcsname{0.21}
\expandafter\def\csname BAval@tick.dance.f288.dam.state\endcsname{U}
\expandafter\def\csname BAval@tick.dance.f289.dam.iou\endcsname{0.15}
\expandafter\def\csname BAval@tick.dance.f289.dam.state\endcsname{U}
\expandafter\def\csname BAval@tick.dance.f290.dam.iou\endcsname{0.13}
\expandafter\def\csname BAval@tick.dance.f290.dam.state\endcsname{U}
\expandafter\def\csname BAval@tick.dance.f291.dam.iou\endcsname{0.05}
\expandafter\def\csname BAval@tick.dance.f291.dam.state\endcsname{U}
\expandafter\def\csname BAval@tick.dance.f292.dam.iou\endcsname{0.03}
\expandafter\def\csname BAval@tick.dance.f292.dam.state\endcsname{U}
\expandafter\def\csname BAval@tick.dance.f293.dam.iou\endcsname{0.00}
\expandafter\def\csname BAval@tick.dance.f293.dam.state\endcsname{W}
\expandafter\def\csname BAval@tick.dance.f294.dam.iou\endcsname{0.00}
\expandafter\def\csname BAval@tick.dance.f294.dam.state\endcsname{W}
\expandafter\def\csname BAval@tick.dance.f295.dam.iou\endcsname{0.00}
\expandafter\def\csname BAval@tick.dance.f295.dam.state\endcsname{W}
\expandafter\def\csname BAval@tick.dance.f296.dam.iou\endcsname{0.00}
\expandafter\def\csname BAval@tick.dance.f296.dam.state\endcsname{W}
\expandafter\def\csname BAval@tick.dance.f297.dam.iou\endcsname{0.00}
\expandafter\def\csname BAval@tick.dance.f297.dam.state\endcsname{W}
\expandafter\def\csname BAval@tick.dance.f298.dam.iou\endcsname{0.00}
\expandafter\def\csname BAval@tick.dance.f298.dam.state\endcsname{W}
\expandafter\def\csname BAval@tick.dance.f299.dam.iou\endcsname{0.00}
\expandafter\def\csname BAval@tick.dance.f299.dam.state\endcsname{U}
\expandafter\def\csname BAval@tick.dance.f300.dam.iou\endcsname{0.00}
\expandafter\def\csname BAval@tick.dance.f300.dam.state\endcsname{U}
\expandafter\def\csname BAval@tick.dance.f301.dam.iou\endcsname{0.00}
\expandafter\def\csname BAval@tick.dance.f301.dam.state\endcsname{U}
\expandafter\def\csname BAval@tick.dance.f302.dam.iou\endcsname{0.00}
\expandafter\def\csname BAval@tick.dance.f302.dam.state\endcsname{U}
\expandafter\def\csname BAval@tick.dance.f303.dam.iou\endcsname{0.00}
\expandafter\def\csname BAval@tick.dance.f303.dam.state\endcsname{U}
\expandafter\def\csname BAval@tick.dance.f304.dam.iou\endcsname{0.00}
\expandafter\def\csname BAval@tick.dance.f304.dam.state\endcsname{U}
\expandafter\def\csname BAval@tick.dance.f305.dam.iou\endcsname{0.00}
\expandafter\def\csname BAval@tick.dance.f305.dam.state\endcsname{U}
\expandafter\def\csname BAval@tick.dance.f306.dam.iou\endcsname{0.00}
\expandafter\def\csname BAval@tick.dance.f306.dam.state\endcsname{U}
\expandafter\def\csname BAval@tick.dance.f307.dam.iou\endcsname{0.00}
\expandafter\def\csname BAval@tick.dance.f307.dam.state\endcsname{U}
\expandafter\def\csname BAval@tick.dance.f308.dam.iou\endcsname{0.00}
\expandafter\def\csname BAval@tick.dance.f308.dam.state\endcsname{U}
\expandafter\def\csname BAval@tick.dance.f309.dam.iou\endcsname{0.00}
\expandafter\def\csname BAval@tick.dance.f309.dam.state\endcsname{U}
\expandafter\def\csname BAval@tick.dance.f310.dam.iou\endcsname{0.00}
\expandafter\def\csname BAval@tick.dance.f310.dam.state\endcsname{U}
\expandafter\def\csname BAval@tick.dance.f311.dam.iou\endcsname{0.00}
\expandafter\def\csname BAval@tick.dance.f311.dam.state\endcsname{U}
\expandafter\def\csname BAval@tick.dance.f312.dam.iou\endcsname{0.00}
\expandafter\def\csname BAval@tick.dance.f312.dam.state\endcsname{U}
\expandafter\def\csname BAval@tick.dance.f313.dam.iou\endcsname{0.00}
\expandafter\def\csname BAval@tick.dance.f313.dam.state\endcsname{U}
\expandafter\def\csname BAval@tick.dance.f314.dam.iou\endcsname{0.00}
\expandafter\def\csname BAval@tick.dance.f314.dam.state\endcsname{U}
\expandafter\def\csname BAval@tick.dance.f315.dam.iou\endcsname{0.00}
\expandafter\def\csname BAval@tick.dance.f315.dam.state\endcsname{U}
\expandafter\def\csname BAval@tick.dance.f316.dam.iou\endcsname{0.00}
\expandafter\def\csname BAval@tick.dance.f316.dam.state\endcsname{U}
\expandafter\def\csname BAval@tick.dance.f317.dam.iou\endcsname{0.00}
\expandafter\def\csname BAval@tick.dance.f317.dam.state\endcsname{W}
\expandafter\def\csname BAval@tick.dance.f318.dam.iou\endcsname{0.00}
\expandafter\def\csname BAval@tick.dance.f318.dam.state\endcsname{U}
\expandafter\def\csname BAval@tick.dance.f319.dam.iou\endcsname{0.00}
\expandafter\def\csname BAval@tick.dance.f319.dam.state\endcsname{U}
\expandafter\def\csname BAval@tick.dance.f320.dam.iou\endcsname{0.00}
\expandafter\def\csname BAval@tick.dance.f320.dam.state\endcsname{U}
\expandafter\def\csname BAval@tick.dance.f321.dam.iou\endcsname{0.00}
\expandafter\def\csname BAval@tick.dance.f321.dam.state\endcsname{U}
\expandafter\def\csname BAval@tick.dance.f322.dam.iou\endcsname{0.00}
\expandafter\def\csname BAval@tick.dance.f322.dam.state\endcsname{U}
\expandafter\def\csname BAval@tick.dance.f323.dam.iou\endcsname{0.00}
\expandafter\def\csname BAval@tick.dance.f323.dam.state\endcsname{U}
\expandafter\def\csname BAval@tick.dance.f324.dam.iou\endcsname{0.00}
\expandafter\def\csname BAval@tick.dance.f324.dam.state\endcsname{U}
\expandafter\def\csname BAval@tick.dance.f325.dam.iou\endcsname{0.00}
\expandafter\def\csname BAval@tick.dance.f325.dam.state\endcsname{U}
\expandafter\def\csname BAval@tick.dance.f326.dam.iou\endcsname{0.00}
\expandafter\def\csname BAval@tick.dance.f326.dam.state\endcsname{U}
\expandafter\def\csname BAval@tick.dance.f327.dam.iou\endcsname{0.00}
\expandafter\def\csname BAval@tick.dance.f327.dam.state\endcsname{U}
\expandafter\def\csname BAval@tick.dance.f328.dam.iou\endcsname{0.00}
\expandafter\def\csname BAval@tick.dance.f328.dam.state\endcsname{U}
\expandafter\def\csname BAval@tick.dance.f329.dam.iou\endcsname{0.00}
\expandafter\def\csname BAval@tick.dance.f329.dam.state\endcsname{U}
\expandafter\def\csname BAval@tick.dance.f330.dam.iou\endcsname{0.00}
\expandafter\def\csname BAval@tick.dance.f330.dam.state\endcsname{U}
\expandafter\def\csname BAval@tick.dance.f331.dam.iou\endcsname{0.00}
\expandafter\def\csname BAval@tick.dance.f331.dam.state\endcsname{U}
\expandafter\def\csname BAval@tick.dance.f332.dam.iou\endcsname{0.00}
\expandafter\def\csname BAval@tick.dance.f332.dam.state\endcsname{U}
\expandafter\def\csname BAval@tick.dance.f333.dam.iou\endcsname{0.00}
\expandafter\def\csname BAval@tick.dance.f333.dam.state\endcsname{U}
\expandafter\def\csname BAval@tick.dance.f334.dam.iou\endcsname{0.00}
\expandafter\def\csname BAval@tick.dance.f334.dam.state\endcsname{U}
\expandafter\def\csname BAval@tick.dance.f335.dam.iou\endcsname{0.00}
\expandafter\def\csname BAval@tick.dance.f335.dam.state\endcsname{U}
\expandafter\def\csname BAval@tick.dance.f336.dam.iou\endcsname{0.00}
\expandafter\def\csname BAval@tick.dance.f336.dam.state\endcsname{U}
\expandafter\def\csname BAval@tick.dance.f337.dam.iou\endcsname{0.00}
\expandafter\def\csname BAval@tick.dance.f337.dam.state\endcsname{U}
\expandafter\def\csname BAval@tick.dance.f338.dam.iou\endcsname{0.00}
\expandafter\def\csname BAval@tick.dance.f338.dam.state\endcsname{U}
\expandafter\def\csname BAval@state.dance.keep.correct\endcsname{72}
\expandafter\def\csname BAval@state.dance.keep.wrong\endcsname{23}
\expandafter\def\csname BAval@state.dance.keep.unresolved\endcsname{84}
\expandafter\def\csname BAval@state.dance.keep.unknown\endcsname{1}
\expandafter\def\csname BAval@tick.dance.f159.keep.iou\endcsname{0.99}
\expandafter\def\csname BAval@tick.dance.f159.keep.state\endcsname{C}
\expandafter\def\csname BAval@tick.dance.f160.keep.iou\endcsname{0.98}
\expandafter\def\csname BAval@tick.dance.f160.keep.state\endcsname{C}
\expandafter\def\csname BAval@tick.dance.f161.keep.iou\endcsname{0.97}
\expandafter\def\csname BAval@tick.dance.f161.keep.state\endcsname{C}
\expandafter\def\csname BAval@tick.dance.f162.keep.iou\endcsname{0.97}
\expandafter\def\csname BAval@tick.dance.f162.keep.state\endcsname{C}
\expandafter\def\csname BAval@tick.dance.f163.keep.iou\endcsname{0.95}
\expandafter\def\csname BAval@tick.dance.f163.keep.state\endcsname{C}
\expandafter\def\csname BAval@tick.dance.f164.keep.iou\endcsname{0.94}
\expandafter\def\csname BAval@tick.dance.f164.keep.state\endcsname{C}
\expandafter\def\csname BAval@tick.dance.f165.keep.iou\endcsname{0.92}
\expandafter\def\csname BAval@tick.dance.f165.keep.state\endcsname{C}
\expandafter\def\csname BAval@tick.dance.f166.keep.iou\endcsname{0.89}
\expandafter\def\csname BAval@tick.dance.f166.keep.state\endcsname{C}
\expandafter\def\csname BAval@tick.dance.f167.keep.iou\endcsname{0.87}
\expandafter\def\csname BAval@tick.dance.f167.keep.state\endcsname{C}
\expandafter\def\csname BAval@tick.dance.f168.keep.iou\endcsname{0.86}
\expandafter\def\csname BAval@tick.dance.f168.keep.state\endcsname{C}
\expandafter\def\csname BAval@tick.dance.f169.keep.iou\endcsname{0.83}
\expandafter\def\csname BAval@tick.dance.f169.keep.state\endcsname{C}
\expandafter\def\csname BAval@tick.dance.f170.keep.iou\endcsname{0.60}
\expandafter\def\csname BAval@tick.dance.f170.keep.state\endcsname{C}
\expandafter\def\csname BAval@tick.dance.f171.keep.iou\endcsname{0.52}
\expandafter\def\csname BAval@tick.dance.f171.keep.state\endcsname{C}
\expandafter\def\csname BAval@tick.dance.f172.keep.iou\endcsname{0.46}
\expandafter\def\csname BAval@tick.dance.f172.keep.state\endcsname{U}
\expandafter\def\csname BAval@tick.dance.f173.keep.iou\endcsname{0.51}
\expandafter\def\csname BAval@tick.dance.f173.keep.state\endcsname{C}
\expandafter\def\csname BAval@tick.dance.f174.keep.iou\endcsname{0.45}
\expandafter\def\csname BAval@tick.dance.f174.keep.state\endcsname{U}
\expandafter\def\csname BAval@tick.dance.f175.keep.iou\endcsname{0.44}
\expandafter\def\csname BAval@tick.dance.f175.keep.state\endcsname{U}
\expandafter\def\csname BAval@tick.dance.f176.keep.iou\endcsname{0.42}
\expandafter\def\csname BAval@tick.dance.f176.keep.state\endcsname{U}
\expandafter\def\csname BAval@tick.dance.f177.keep.iou\endcsname{0.42}
\expandafter\def\csname BAval@tick.dance.f177.keep.state\endcsname{U}
\expandafter\def\csname BAval@tick.dance.f178.keep.iou\endcsname{0.41}
\expandafter\def\csname BAval@tick.dance.f178.keep.state\endcsname{U}
\expandafter\def\csname BAval@tick.dance.f179.keep.iou\endcsname{0.39}
\expandafter\def\csname BAval@tick.dance.f179.keep.state\endcsname{U}
\expandafter\def\csname BAval@tick.dance.f180.keep.iou\endcsname{0.30}
\expandafter\def\csname BAval@tick.dance.f180.keep.state\endcsname{U}
\expandafter\def\csname BAval@tick.dance.f181.keep.iou\endcsname{0.27}
\expandafter\def\csname BAval@tick.dance.f181.keep.state\endcsname{U}
\expandafter\def\csname BAval@tick.dance.f182.keep.iou\endcsname{0.19}
\expandafter\def\csname BAval@tick.dance.f182.keep.state\endcsname{U}
\expandafter\def\csname BAval@tick.dance.f183.keep.iou\endcsname{0.14}
\expandafter\def\csname BAval@tick.dance.f183.keep.state\endcsname{U}
\expandafter\def\csname BAval@tick.dance.f184.keep.iou\endcsname{0.10}
\expandafter\def\csname BAval@tick.dance.f184.keep.state\endcsname{U}
\expandafter\def\csname BAval@tick.dance.f185.keep.iou\endcsname{0.08}
\expandafter\def\csname BAval@tick.dance.f185.keep.state\endcsname{W}
\expandafter\def\csname BAval@tick.dance.f186.keep.iou\endcsname{0.06}
\expandafter\def\csname BAval@tick.dance.f186.keep.state\endcsname{W}
\expandafter\def\csname BAval@tick.dance.f187.keep.iou\endcsname{0.04}
\expandafter\def\csname BAval@tick.dance.f187.keep.state\endcsname{W}
\expandafter\def\csname BAval@tick.dance.f188.keep.iou\endcsname{0.03}
\expandafter\def\csname BAval@tick.dance.f188.keep.state\endcsname{W}
\expandafter\def\csname BAval@tick.dance.f189.keep.iou\endcsname{0.02}
\expandafter\def\csname BAval@tick.dance.f189.keep.state\endcsname{W}
\expandafter\def\csname BAval@tick.dance.f190.keep.iou\endcsname{0.02}
\expandafter\def\csname BAval@tick.dance.f190.keep.state\endcsname{W}
\expandafter\def\csname BAval@tick.dance.f191.keep.iou\endcsname{0.02}
\expandafter\def\csname BAval@tick.dance.f191.keep.state\endcsname{W}
\expandafter\def\csname BAval@tick.dance.f192.keep.iou\endcsname{0.00}
\expandafter\def\csname BAval@tick.dance.f192.keep.state\endcsname{W}
\expandafter\def\csname BAval@tick.dance.f193.keep.iou\endcsname{0.00}
\expandafter\def\csname BAval@tick.dance.f193.keep.state\endcsname{W}
\expandafter\def\csname BAval@tick.dance.f194.keep.iou\endcsname{0.00}
\expandafter\def\csname BAval@tick.dance.f194.keep.state\endcsname{W}
\expandafter\def\csname BAval@tick.dance.f195.keep.iou\endcsname{0.00}
\expandafter\def\csname BAval@tick.dance.f195.keep.state\endcsname{W}
\expandafter\def\csname BAval@tick.dance.f196.keep.iou\endcsname{0.00}
\expandafter\def\csname BAval@tick.dance.f196.keep.state\endcsname{W}
\expandafter\def\csname BAval@tick.dance.f197.keep.iou\endcsname{0.00}
\expandafter\def\csname BAval@tick.dance.f197.keep.state\endcsname{W}
\expandafter\def\csname BAval@tick.dance.f198.keep.iou\endcsname{0.00}
\expandafter\def\csname BAval@tick.dance.f198.keep.state\endcsname{U}
\expandafter\def\csname BAval@tick.dance.f199.keep.iou\endcsname{0.00}
\expandafter\def\csname BAval@tick.dance.f199.keep.state\endcsname{U}
\expandafter\def\csname BAval@tick.dance.f200.keep.iou\endcsname{0.00}
\expandafter\def\csname BAval@tick.dance.f200.keep.state\endcsname{U}
\expandafter\def\csname BAval@tick.dance.f201.keep.iou\endcsname{0.00}
\expandafter\def\csname BAval@tick.dance.f201.keep.state\endcsname{U}
\expandafter\def\csname BAval@tick.dance.f202.keep.iou\endcsname{0.00}
\expandafter\def\csname BAval@tick.dance.f202.keep.state\endcsname{U}
\expandafter\def\csname BAval@tick.dance.f203.keep.iou\endcsname{0.00}
\expandafter\def\csname BAval@tick.dance.f203.keep.state\endcsname{U}
\expandafter\def\csname BAval@tick.dance.f204.keep.iou\endcsname{0.00}
\expandafter\def\csname BAval@tick.dance.f204.keep.state\endcsname{U}
\expandafter\def\csname BAval@tick.dance.f205.keep.iou\endcsname{0.00}
\expandafter\def\csname BAval@tick.dance.f205.keep.state\endcsname{U}
\expandafter\def\csname BAval@tick.dance.f206.keep.iou\endcsname{0.03}
\expandafter\def\csname BAval@tick.dance.f206.keep.state\endcsname{U}
\expandafter\def\csname BAval@tick.dance.f207.keep.iou\endcsname{0.04}
\expandafter\def\csname BAval@tick.dance.f207.keep.state\endcsname{U}
\expandafter\def\csname BAval@tick.dance.f208.keep.iou\endcsname{0.04}
\expandafter\def\csname BAval@tick.dance.f208.keep.state\endcsname{U}
\expandafter\def\csname BAval@tick.dance.f209.keep.iou\endcsname{0.04}
\expandafter\def\csname BAval@tick.dance.f209.keep.state\endcsname{U}
\expandafter\def\csname BAval@tick.dance.f210.keep.iou\endcsname{0.07}
\expandafter\def\csname BAval@tick.dance.f210.keep.state\endcsname{U}
\expandafter\def\csname BAval@tick.dance.f211.keep.iou\endcsname{0.09}
\expandafter\def\csname BAval@tick.dance.f211.keep.state\endcsname{U}
\expandafter\def\csname BAval@tick.dance.f212.keep.iou\endcsname{0.12}
\expandafter\def\csname BAval@tick.dance.f212.keep.state\endcsname{U}
\expandafter\def\csname BAval@tick.dance.f213.keep.iou\endcsname{0.13}
\expandafter\def\csname BAval@tick.dance.f213.keep.state\endcsname{U}
\expandafter\def\csname BAval@tick.dance.f214.keep.iou\endcsname{0.14}
\expandafter\def\csname BAval@tick.dance.f214.keep.state\endcsname{U}
\expandafter\def\csname BAval@tick.dance.f215.keep.iou\endcsname{0.15}
\expandafter\def\csname BAval@tick.dance.f215.keep.state\endcsname{U}
\expandafter\def\csname BAval@tick.dance.f216.keep.iou\endcsname{0.21}
\expandafter\def\csname BAval@tick.dance.f216.keep.state\endcsname{U}
\expandafter\def\csname BAval@tick.dance.f217.keep.iou\endcsname{0.24}
\expandafter\def\csname BAval@tick.dance.f217.keep.state\endcsname{U}
\expandafter\def\csname BAval@tick.dance.f218.keep.iou\endcsname{0.37}
\expandafter\def\csname BAval@tick.dance.f218.keep.state\endcsname{U}
\expandafter\def\csname BAval@tick.dance.f219.keep.iou\endcsname{0.41}
\expandafter\def\csname BAval@tick.dance.f219.keep.state\endcsname{U}
\expandafter\def\csname BAval@tick.dance.f220.keep.iou\endcsname{0.46}
\expandafter\def\csname BAval@tick.dance.f220.keep.state\endcsname{U}
\expandafter\def\csname BAval@tick.dance.f221.keep.iou\endcsname{0.51}
\expandafter\def\csname BAval@tick.dance.f221.keep.state\endcsname{C}
\expandafter\def\csname BAval@tick.dance.f222.keep.iou\endcsname{0.53}
\expandafter\def\csname BAval@tick.dance.f222.keep.state\endcsname{C}
\expandafter\def\csname BAval@tick.dance.f223.keep.iou\endcsname{0.53}
\expandafter\def\csname BAval@tick.dance.f223.keep.state\endcsname{C}
\expandafter\def\csname BAval@tick.dance.f224.keep.iou\endcsname{0.53}
\expandafter\def\csname BAval@tick.dance.f224.keep.state\endcsname{C}
\expandafter\def\csname BAval@tick.dance.f225.keep.iou\endcsname{0.54}
\expandafter\def\csname BAval@tick.dance.f225.keep.state\endcsname{C}
\expandafter\def\csname BAval@tick.dance.f226.keep.iou\endcsname{0.56}
\expandafter\def\csname BAval@tick.dance.f226.keep.state\endcsname{C}
\expandafter\def\csname BAval@tick.dance.f227.keep.iou\endcsname{0.55}
\expandafter\def\csname BAval@tick.dance.f227.keep.state\endcsname{C}
\expandafter\def\csname BAval@tick.dance.f228.keep.iou\endcsname{0.55}
\expandafter\def\csname BAval@tick.dance.f228.keep.state\endcsname{C}
\expandafter\def\csname BAval@tick.dance.f229.keep.iou\endcsname{0.54}
\expandafter\def\csname BAval@tick.dance.f229.keep.state\endcsname{C}
\expandafter\def\csname BAval@tick.dance.f230.keep.iou\endcsname{0.55}
\expandafter\def\csname BAval@tick.dance.f230.keep.state\endcsname{C}
\expandafter\def\csname BAval@tick.dance.f231.keep.iou\endcsname{0.55}
\expandafter\def\csname BAval@tick.dance.f231.keep.state\endcsname{C}
\expandafter\def\csname BAval@tick.dance.f232.keep.iou\endcsname{0.55}
\expandafter\def\csname BAval@tick.dance.f232.keep.state\endcsname{C}
\expandafter\def\csname BAval@tick.dance.f233.keep.iou\endcsname{0.56}
\expandafter\def\csname BAval@tick.dance.f233.keep.state\endcsname{C}
\expandafter\def\csname BAval@tick.dance.f234.keep.iou\endcsname{0.56}
\expandafter\def\csname BAval@tick.dance.f234.keep.state\endcsname{C}
\expandafter\def\csname BAval@tick.dance.f235.keep.iou\endcsname{0.56}
\expandafter\def\csname BAval@tick.dance.f235.keep.state\endcsname{C}
\expandafter\def\csname BAval@tick.dance.f236.keep.iou\endcsname{0.57}
\expandafter\def\csname BAval@tick.dance.f236.keep.state\endcsname{C}
\expandafter\def\csname BAval@tick.dance.f237.keep.iou\endcsname{0.58}
\expandafter\def\csname BAval@tick.dance.f237.keep.state\endcsname{C}
\expandafter\def\csname BAval@tick.dance.f238.keep.iou\endcsname{0.57}
\expandafter\def\csname BAval@tick.dance.f238.keep.state\endcsname{C}
\expandafter\def\csname BAval@tick.dance.f239.keep.iou\endcsname{0.58}
\expandafter\def\csname BAval@tick.dance.f239.keep.state\endcsname{C}
\expandafter\def\csname BAval@tick.dance.f240.keep.iou\endcsname{0.57}
\expandafter\def\csname BAval@tick.dance.f240.keep.state\endcsname{C}
\expandafter\def\csname BAval@tick.dance.f241.keep.iou\endcsname{0.58}
\expandafter\def\csname BAval@tick.dance.f241.keep.state\endcsname{C}
\expandafter\def\csname BAval@tick.dance.f242.keep.iou\endcsname{0.58}
\expandafter\def\csname BAval@tick.dance.f242.keep.state\endcsname{C}
\expandafter\def\csname BAval@tick.dance.f243.keep.iou\endcsname{0.59}
\expandafter\def\csname BAval@tick.dance.f243.keep.state\endcsname{C}
\expandafter\def\csname BAval@tick.dance.f244.keep.iou\endcsname{0.59}
\expandafter\def\csname BAval@tick.dance.f244.keep.state\endcsname{C}
\expandafter\def\csname BAval@tick.dance.f245.keep.iou\endcsname{0.60}
\expandafter\def\csname BAval@tick.dance.f245.keep.state\endcsname{C}
\expandafter\def\csname BAval@tick.dance.f246.keep.iou\endcsname{0.58}
\expandafter\def\csname BAval@tick.dance.f246.keep.state\endcsname{C}
\expandafter\def\csname BAval@tick.dance.f247.keep.iou\endcsname{0.60}
\expandafter\def\csname BAval@tick.dance.f247.keep.state\endcsname{C}
\expandafter\def\csname BAval@tick.dance.f248.keep.iou\endcsname{0.58}
\expandafter\def\csname BAval@tick.dance.f248.keep.state\endcsname{C}
\expandafter\def\csname BAval@tick.dance.f249.keep.iou\endcsname{0.57}
\expandafter\def\csname BAval@tick.dance.f249.keep.state\endcsname{C}
\expandafter\def\csname BAval@tick.dance.f250.keep.iou\endcsname{0.57}
\expandafter\def\csname BAval@tick.dance.f250.keep.state\endcsname{C}
\expandafter\def\csname BAval@tick.dance.f251.keep.iou\endcsname{0.57}
\expandafter\def\csname BAval@tick.dance.f251.keep.state\endcsname{C}
\expandafter\def\csname BAval@tick.dance.f252.keep.iou\endcsname{0.57}
\expandafter\def\csname BAval@tick.dance.f252.keep.state\endcsname{C}
\expandafter\def\csname BAval@tick.dance.f253.keep.iou\endcsname{0.56}
\expandafter\def\csname BAval@tick.dance.f253.keep.state\endcsname{C}
\expandafter\def\csname BAval@tick.dance.f254.keep.iou\endcsname{0.56}
\expandafter\def\csname BAval@tick.dance.f254.keep.state\endcsname{C}
\expandafter\def\csname BAval@tick.dance.f255.keep.iou\endcsname{0.56}
\expandafter\def\csname BAval@tick.dance.f255.keep.state\endcsname{C}
\expandafter\def\csname BAval@tick.dance.f256.keep.iou\endcsname{0.56}
\expandafter\def\csname BAval@tick.dance.f256.keep.state\endcsname{C}
\expandafter\def\csname BAval@tick.dance.f257.keep.iou\endcsname{0.56}
\expandafter\def\csname BAval@tick.dance.f257.keep.state\endcsname{C}
\expandafter\def\csname BAval@tick.dance.f258.keep.iou\endcsname{0.56}
\expandafter\def\csname BAval@tick.dance.f258.keep.state\endcsname{C}
\expandafter\def\csname BAval@tick.dance.f259.keep.iou\endcsname{0.56}
\expandafter\def\csname BAval@tick.dance.f259.keep.state\endcsname{C}
\expandafter\def\csname BAval@tick.dance.f260.keep.iou\endcsname{0.56}
\expandafter\def\csname BAval@tick.dance.f260.keep.state\endcsname{C}
\expandafter\def\csname BAval@tick.dance.f261.keep.iou\endcsname{0.56}
\expandafter\def\csname BAval@tick.dance.f261.keep.state\endcsname{C}
\expandafter\def\csname BAval@tick.dance.f262.keep.iou\endcsname{0.56}
\expandafter\def\csname BAval@tick.dance.f262.keep.state\endcsname{C}
\expandafter\def\csname BAval@tick.dance.f263.keep.iou\endcsname{0.56}
\expandafter\def\csname BAval@tick.dance.f263.keep.state\endcsname{C}
\expandafter\def\csname BAval@tick.dance.f264.keep.iou\endcsname{0.56}
\expandafter\def\csname BAval@tick.dance.f264.keep.state\endcsname{C}
\expandafter\def\csname BAval@tick.dance.f265.keep.iou\endcsname{0.56}
\expandafter\def\csname BAval@tick.dance.f265.keep.state\endcsname{C}
\expandafter\def\csname BAval@tick.dance.f266.keep.iou\endcsname{0.56}
\expandafter\def\csname BAval@tick.dance.f266.keep.state\endcsname{C}
\expandafter\def\csname BAval@tick.dance.f267.keep.iou\endcsname{0.56}
\expandafter\def\csname BAval@tick.dance.f267.keep.state\endcsname{C}
\expandafter\def\csname BAval@tick.dance.f268.keep.iou\endcsname{0.56}
\expandafter\def\csname BAval@tick.dance.f268.keep.state\endcsname{C}
\expandafter\def\csname BAval@tick.dance.f269.keep.iou\endcsname{0.57}
\expandafter\def\csname BAval@tick.dance.f269.keep.state\endcsname{C}
\expandafter\def\csname BAval@tick.dance.f270.keep.iou\endcsname{0.57}
\expandafter\def\csname BAval@tick.dance.f270.keep.state\endcsname{C}
\expandafter\def\csname BAval@tick.dance.f271.keep.iou\endcsname{0.57}
\expandafter\def\csname BAval@tick.dance.f271.keep.state\endcsname{C}
\expandafter\def\csname BAval@tick.dance.f272.keep.iou\endcsname{0.58}
\expandafter\def\csname BAval@tick.dance.f272.keep.state\endcsname{C}
\expandafter\def\csname BAval@tick.dance.f273.keep.iou\endcsname{0.58}
\expandafter\def\csname BAval@tick.dance.f273.keep.state\endcsname{C}
\expandafter\def\csname BAval@tick.dance.f274.keep.iou\endcsname{0.59}
\expandafter\def\csname BAval@tick.dance.f274.keep.state\endcsname{C}
\expandafter\def\csname BAval@tick.dance.f275.keep.iou\endcsname{0.60}
\expandafter\def\csname BAval@tick.dance.f275.keep.state\endcsname{C}
\expandafter\def\csname BAval@tick.dance.f276.keep.iou\endcsname{0.59}
\expandafter\def\csname BAval@tick.dance.f276.keep.state\endcsname{C}
\expandafter\def\csname BAval@tick.dance.f277.keep.iou\endcsname{0.59}
\expandafter\def\csname BAval@tick.dance.f277.keep.state\endcsname{C}
\expandafter\def\csname BAval@tick.dance.f278.keep.iou\endcsname{0.59}
\expandafter\def\csname BAval@tick.dance.f278.keep.state\endcsname{C}
\expandafter\def\csname BAval@tick.dance.f279.keep.iou\endcsname{0.41}
\expandafter\def\csname BAval@tick.dance.f279.keep.state\endcsname{U}
\expandafter\def\csname BAval@tick.dance.f280.keep.iou\endcsname{0.41}
\expandafter\def\csname BAval@tick.dance.f280.keep.state\endcsname{U}
\expandafter\def\csname BAval@tick.dance.f281.keep.iou\endcsname{0.43}
\expandafter\def\csname BAval@tick.dance.f281.keep.state\endcsname{U}
\expandafter\def\csname BAval@tick.dance.f282.keep.iou\endcsname{0.45}
\expandafter\def\csname BAval@tick.dance.f282.keep.state\endcsname{U}
\expandafter\def\csname BAval@tick.dance.f283.keep.iou\endcsname{0.50}
\expandafter\def\csname BAval@tick.dance.f283.keep.state\endcsname{U}
\expandafter\def\csname BAval@tick.dance.f284.keep.iou\endcsname{0.48}
\expandafter\def\csname BAval@tick.dance.f284.keep.state\endcsname{U}
\expandafter\def\csname BAval@tick.dance.f285.keep.iou\endcsname{0.37}
\expandafter\def\csname BAval@tick.dance.f285.keep.state\endcsname{U}
\expandafter\def\csname BAval@tick.dance.f286.keep.iou\endcsname{0.31}
\expandafter\def\csname BAval@tick.dance.f286.keep.state\endcsname{U}
\expandafter\def\csname BAval@tick.dance.f287.keep.iou\endcsname{0.22}
\expandafter\def\csname BAval@tick.dance.f287.keep.state\endcsname{U}
\expandafter\def\csname BAval@tick.dance.f288.keep.iou\endcsname{0.14}
\expandafter\def\csname BAval@tick.dance.f288.keep.state\endcsname{U}
\expandafter\def\csname BAval@tick.dance.f289.keep.iou\endcsname{0.09}
\expandafter\def\csname BAval@tick.dance.f289.keep.state\endcsname{U}
\expandafter\def\csname BAval@tick.dance.f290.keep.iou\endcsname{0.08}
\expandafter\def\csname BAval@tick.dance.f290.keep.state\endcsname{U}
\expandafter\def\csname BAval@tick.dance.f291.keep.iou\endcsname{0.01}
\expandafter\def\csname BAval@tick.dance.f291.keep.state\endcsname{U}
\expandafter\def\csname BAval@tick.dance.f292.keep.iou\endcsname{0.00}
\expandafter\def\csname BAval@tick.dance.f292.keep.state\endcsname{U}
\expandafter\def\csname BAval@tick.dance.f293.keep.iou\endcsname{0.00}
\expandafter\def\csname BAval@tick.dance.f293.keep.state\endcsname{W}
\expandafter\def\csname BAval@tick.dance.f294.keep.iou\endcsname{0.00}
\expandafter\def\csname BAval@tick.dance.f294.keep.state\endcsname{UNKNOWN}
\expandafter\def\csname BAval@tick.dance.f295.keep.iou\endcsname{0.00}
\expandafter\def\csname BAval@tick.dance.f295.keep.state\endcsname{W}
\expandafter\def\csname BAval@tick.dance.f296.keep.iou\endcsname{0.00}
\expandafter\def\csname BAval@tick.dance.f296.keep.state\endcsname{W}
\expandafter\def\csname BAval@tick.dance.f297.keep.iou\endcsname{0.00}
\expandafter\def\csname BAval@tick.dance.f297.keep.state\endcsname{W}
\expandafter\def\csname BAval@tick.dance.f298.keep.iou\endcsname{0.00}
\expandafter\def\csname BAval@tick.dance.f298.keep.state\endcsname{U}
\expandafter\def\csname BAval@tick.dance.f299.keep.iou\endcsname{0.00}
\expandafter\def\csname BAval@tick.dance.f299.keep.state\endcsname{U}
\expandafter\def\csname BAval@tick.dance.f300.keep.iou\endcsname{0.00}
\expandafter\def\csname BAval@tick.dance.f300.keep.state\endcsname{U}
\expandafter\def\csname BAval@tick.dance.f301.keep.iou\endcsname{0.00}
\expandafter\def\csname BAval@tick.dance.f301.keep.state\endcsname{U}
\expandafter\def\csname BAval@tick.dance.f302.keep.iou\endcsname{0.00}
\expandafter\def\csname BAval@tick.dance.f302.keep.state\endcsname{U}
\expandafter\def\csname BAval@tick.dance.f303.keep.iou\endcsname{0.00}
\expandafter\def\csname BAval@tick.dance.f303.keep.state\endcsname{U}
\expandafter\def\csname BAval@tick.dance.f304.keep.iou\endcsname{0.00}
\expandafter\def\csname BAval@tick.dance.f304.keep.state\endcsname{U}
\expandafter\def\csname BAval@tick.dance.f305.keep.iou\endcsname{0.00}
\expandafter\def\csname BAval@tick.dance.f305.keep.state\endcsname{U}
\expandafter\def\csname BAval@tick.dance.f306.keep.iou\endcsname{0.00}
\expandafter\def\csname BAval@tick.dance.f306.keep.state\endcsname{U}
\expandafter\def\csname BAval@tick.dance.f307.keep.iou\endcsname{0.00}
\expandafter\def\csname BAval@tick.dance.f307.keep.state\endcsname{U}
\expandafter\def\csname BAval@tick.dance.f308.keep.iou\endcsname{0.00}
\expandafter\def\csname BAval@tick.dance.f308.keep.state\endcsname{U}
\expandafter\def\csname BAval@tick.dance.f309.keep.iou\endcsname{0.00}
\expandafter\def\csname BAval@tick.dance.f309.keep.state\endcsname{U}
\expandafter\def\csname BAval@tick.dance.f310.keep.iou\endcsname{0.00}
\expandafter\def\csname BAval@tick.dance.f310.keep.state\endcsname{U}
\expandafter\def\csname BAval@tick.dance.f311.keep.iou\endcsname{0.00}
\expandafter\def\csname BAval@tick.dance.f311.keep.state\endcsname{U}
\expandafter\def\csname BAval@tick.dance.f312.keep.iou\endcsname{0.00}
\expandafter\def\csname BAval@tick.dance.f312.keep.state\endcsname{U}
\expandafter\def\csname BAval@tick.dance.f313.keep.iou\endcsname{0.00}
\expandafter\def\csname BAval@tick.dance.f313.keep.state\endcsname{U}
\expandafter\def\csname BAval@tick.dance.f314.keep.iou\endcsname{0.00}
\expandafter\def\csname BAval@tick.dance.f314.keep.state\endcsname{U}
\expandafter\def\csname BAval@tick.dance.f315.keep.iou\endcsname{0.00}
\expandafter\def\csname BAval@tick.dance.f315.keep.state\endcsname{U}
\expandafter\def\csname BAval@tick.dance.f316.keep.iou\endcsname{0.00}
\expandafter\def\csname BAval@tick.dance.f316.keep.state\endcsname{U}
\expandafter\def\csname BAval@tick.dance.f317.keep.iou\endcsname{0.00}
\expandafter\def\csname BAval@tick.dance.f317.keep.state\endcsname{W}
\expandafter\def\csname BAval@tick.dance.f318.keep.iou\endcsname{0.00}
\expandafter\def\csname BAval@tick.dance.f318.keep.state\endcsname{W}
\expandafter\def\csname BAval@tick.dance.f319.keep.iou\endcsname{0.00}
\expandafter\def\csname BAval@tick.dance.f319.keep.state\endcsname{U}
\expandafter\def\csname BAval@tick.dance.f320.keep.iou\endcsname{0.00}
\expandafter\def\csname BAval@tick.dance.f320.keep.state\endcsname{W}
\expandafter\def\csname BAval@tick.dance.f321.keep.iou\endcsname{0.00}
\expandafter\def\csname BAval@tick.dance.f321.keep.state\endcsname{W}
\expandafter\def\csname BAval@tick.dance.f322.keep.iou\endcsname{0.00}
\expandafter\def\csname BAval@tick.dance.f322.keep.state\endcsname{W}
\expandafter\def\csname BAval@tick.dance.f323.keep.iou\endcsname{0.00}
\expandafter\def\csname BAval@tick.dance.f323.keep.state\endcsname{W}
\expandafter\def\csname BAval@tick.dance.f324.keep.iou\endcsname{0.00}
\expandafter\def\csname BAval@tick.dance.f324.keep.state\endcsname{U}
\expandafter\def\csname BAval@tick.dance.f325.keep.iou\endcsname{0.00}
\expandafter\def\csname BAval@tick.dance.f325.keep.state\endcsname{U}
\expandafter\def\csname BAval@tick.dance.f326.keep.iou\endcsname{0.00}
\expandafter\def\csname BAval@tick.dance.f326.keep.state\endcsname{U}
\expandafter\def\csname BAval@tick.dance.f327.keep.iou\endcsname{0.00}
\expandafter\def\csname BAval@tick.dance.f327.keep.state\endcsname{U}
\expandafter\def\csname BAval@tick.dance.f328.keep.iou\endcsname{0.00}
\expandafter\def\csname BAval@tick.dance.f328.keep.state\endcsname{U}
\expandafter\def\csname BAval@tick.dance.f329.keep.iou\endcsname{0.00}
\expandafter\def\csname BAval@tick.dance.f329.keep.state\endcsname{U}
\expandafter\def\csname BAval@tick.dance.f330.keep.iou\endcsname{0.00}
\expandafter\def\csname BAval@tick.dance.f330.keep.state\endcsname{U}
\expandafter\def\csname BAval@tick.dance.f331.keep.iou\endcsname{0.00}
\expandafter\def\csname BAval@tick.dance.f331.keep.state\endcsname{U}
\expandafter\def\csname BAval@tick.dance.f332.keep.iou\endcsname{0.00}
\expandafter\def\csname BAval@tick.dance.f332.keep.state\endcsname{U}
\expandafter\def\csname BAval@tick.dance.f333.keep.iou\endcsname{0.00}
\expandafter\def\csname BAval@tick.dance.f333.keep.state\endcsname{U}
\expandafter\def\csname BAval@tick.dance.f334.keep.iou\endcsname{0.00}
\expandafter\def\csname BAval@tick.dance.f334.keep.state\endcsname{U}
\expandafter\def\csname BAval@tick.dance.f335.keep.iou\endcsname{0.00}
\expandafter\def\csname BAval@tick.dance.f335.keep.state\endcsname{U}
\expandafter\def\csname BAval@tick.dance.f336.keep.iou\endcsname{0.00}
\expandafter\def\csname BAval@tick.dance.f336.keep.state\endcsname{U}
\expandafter\def\csname BAval@tick.dance.f337.keep.iou\endcsname{0.00}
\expandafter\def\csname BAval@tick.dance.f337.keep.state\endcsname{U}
\expandafter\def\csname BAval@tick.dance.f338.keep.iou\endcsname{0.00}
\expandafter\def\csname BAval@tick.dance.f338.keep.state\endcsname{U}
\expandafter\def\csname BAval@state.dance.bind.correct\endcsname{78}
\expandafter\def\csname BAval@state.dance.bind.wrong\endcsname{23}
\expandafter\def\csname BAval@state.dance.bind.unresolved\endcsname{79}
\expandafter\def\csname BAval@state.dance.bind.unknown\endcsname{0}
\expandafter\def\csname BAval@tick.dance.f159.bind.iou\endcsname{0.99}
\expandafter\def\csname BAval@tick.dance.f159.bind.state\endcsname{C}
\expandafter\def\csname BAval@tick.dance.f160.bind.iou\endcsname{0.98}
\expandafter\def\csname BAval@tick.dance.f160.bind.state\endcsname{C}
\expandafter\def\csname BAval@tick.dance.f161.bind.iou\endcsname{0.97}
\expandafter\def\csname BAval@tick.dance.f161.bind.state\endcsname{C}
\expandafter\def\csname BAval@tick.dance.f162.bind.iou\endcsname{0.97}
\expandafter\def\csname BAval@tick.dance.f162.bind.state\endcsname{C}
\expandafter\def\csname BAval@tick.dance.f163.bind.iou\endcsname{0.95}
\expandafter\def\csname BAval@tick.dance.f163.bind.state\endcsname{C}
\expandafter\def\csname BAval@tick.dance.f164.bind.iou\endcsname{0.94}
\expandafter\def\csname BAval@tick.dance.f164.bind.state\endcsname{C}
\expandafter\def\csname BAval@tick.dance.f165.bind.iou\endcsname{0.92}
\expandafter\def\csname BAval@tick.dance.f165.bind.state\endcsname{C}
\expandafter\def\csname BAval@tick.dance.f166.bind.iou\endcsname{0.89}
\expandafter\def\csname BAval@tick.dance.f166.bind.state\endcsname{C}
\expandafter\def\csname BAval@tick.dance.f167.bind.iou\endcsname{0.87}
\expandafter\def\csname BAval@tick.dance.f167.bind.state\endcsname{C}
\expandafter\def\csname BAval@tick.dance.f168.bind.iou\endcsname{0.86}
\expandafter\def\csname BAval@tick.dance.f168.bind.state\endcsname{C}
\expandafter\def\csname BAval@tick.dance.f169.bind.iou\endcsname{0.83}
\expandafter\def\csname BAval@tick.dance.f169.bind.state\endcsname{C}
\expandafter\def\csname BAval@tick.dance.f170.bind.iou\endcsname{0.61}
\expandafter\def\csname BAval@tick.dance.f170.bind.state\endcsname{C}
\expandafter\def\csname BAval@tick.dance.f171.bind.iou\endcsname{0.52}
\expandafter\def\csname BAval@tick.dance.f171.bind.state\endcsname{C}
\expandafter\def\csname BAval@tick.dance.f172.bind.iou\endcsname{0.46}
\expandafter\def\csname BAval@tick.dance.f172.bind.state\endcsname{U}
\expandafter\def\csname BAval@tick.dance.f173.bind.iou\endcsname{0.51}
\expandafter\def\csname BAval@tick.dance.f173.bind.state\endcsname{C}
\expandafter\def\csname BAval@tick.dance.f174.bind.iou\endcsname{0.45}
\expandafter\def\csname BAval@tick.dance.f174.bind.state\endcsname{U}
\expandafter\def\csname BAval@tick.dance.f175.bind.iou\endcsname{0.44}
\expandafter\def\csname BAval@tick.dance.f175.bind.state\endcsname{U}
\expandafter\def\csname BAval@tick.dance.f176.bind.iou\endcsname{0.42}
\expandafter\def\csname BAval@tick.dance.f176.bind.state\endcsname{U}
\expandafter\def\csname BAval@tick.dance.f177.bind.iou\endcsname{0.42}
\expandafter\def\csname BAval@tick.dance.f177.bind.state\endcsname{U}
\expandafter\def\csname BAval@tick.dance.f178.bind.iou\endcsname{0.41}
\expandafter\def\csname BAval@tick.dance.f178.bind.state\endcsname{U}
\expandafter\def\csname BAval@tick.dance.f179.bind.iou\endcsname{0.39}
\expandafter\def\csname BAval@tick.dance.f179.bind.state\endcsname{U}
\expandafter\def\csname BAval@tick.dance.f180.bind.iou\endcsname{0.29}
\expandafter\def\csname BAval@tick.dance.f180.bind.state\endcsname{U}
\expandafter\def\csname BAval@tick.dance.f181.bind.iou\endcsname{0.27}
\expandafter\def\csname BAval@tick.dance.f181.bind.state\endcsname{U}
\expandafter\def\csname BAval@tick.dance.f182.bind.iou\endcsname{0.19}
\expandafter\def\csname BAval@tick.dance.f182.bind.state\endcsname{U}
\expandafter\def\csname BAval@tick.dance.f183.bind.iou\endcsname{0.14}
\expandafter\def\csname BAval@tick.dance.f183.bind.state\endcsname{U}
\expandafter\def\csname BAval@tick.dance.f184.bind.iou\endcsname{0.10}
\expandafter\def\csname BAval@tick.dance.f184.bind.state\endcsname{U}
\expandafter\def\csname BAval@tick.dance.f185.bind.iou\endcsname{0.08}
\expandafter\def\csname BAval@tick.dance.f185.bind.state\endcsname{W}
\expandafter\def\csname BAval@tick.dance.f186.bind.iou\endcsname{0.06}
\expandafter\def\csname BAval@tick.dance.f186.bind.state\endcsname{W}
\expandafter\def\csname BAval@tick.dance.f187.bind.iou\endcsname{0.04}
\expandafter\def\csname BAval@tick.dance.f187.bind.state\endcsname{W}
\expandafter\def\csname BAval@tick.dance.f188.bind.iou\endcsname{0.03}
\expandafter\def\csname BAval@tick.dance.f188.bind.state\endcsname{W}
\expandafter\def\csname BAval@tick.dance.f189.bind.iou\endcsname{0.02}
\expandafter\def\csname BAval@tick.dance.f189.bind.state\endcsname{W}
\expandafter\def\csname BAval@tick.dance.f190.bind.iou\endcsname{0.02}
\expandafter\def\csname BAval@tick.dance.f190.bind.state\endcsname{W}
\expandafter\def\csname BAval@tick.dance.f191.bind.iou\endcsname{0.02}
\expandafter\def\csname BAval@tick.dance.f191.bind.state\endcsname{W}
\expandafter\def\csname BAval@tick.dance.f192.bind.iou\endcsname{0.00}
\expandafter\def\csname BAval@tick.dance.f192.bind.state\endcsname{W}
\expandafter\def\csname BAval@tick.dance.f193.bind.iou\endcsname{0.00}
\expandafter\def\csname BAval@tick.dance.f193.bind.state\endcsname{W}
\expandafter\def\csname BAval@tick.dance.f194.bind.iou\endcsname{0.00}
\expandafter\def\csname BAval@tick.dance.f194.bind.state\endcsname{W}
\expandafter\def\csname BAval@tick.dance.f195.bind.iou\endcsname{0.00}
\expandafter\def\csname BAval@tick.dance.f195.bind.state\endcsname{W}
\expandafter\def\csname BAval@tick.dance.f196.bind.iou\endcsname{0.00}
\expandafter\def\csname BAval@tick.dance.f196.bind.state\endcsname{W}
\expandafter\def\csname BAval@tick.dance.f197.bind.iou\endcsname{0.00}
\expandafter\def\csname BAval@tick.dance.f197.bind.state\endcsname{W}
\expandafter\def\csname BAval@tick.dance.f198.bind.iou\endcsname{0.00}
\expandafter\def\csname BAval@tick.dance.f198.bind.state\endcsname{U}
\expandafter\def\csname BAval@tick.dance.f199.bind.iou\endcsname{0.00}
\expandafter\def\csname BAval@tick.dance.f199.bind.state\endcsname{U}
\expandafter\def\csname BAval@tick.dance.f200.bind.iou\endcsname{0.00}
\expandafter\def\csname BAval@tick.dance.f200.bind.state\endcsname{U}
\expandafter\def\csname BAval@tick.dance.f201.bind.iou\endcsname{0.00}
\expandafter\def\csname BAval@tick.dance.f201.bind.state\endcsname{U}
\expandafter\def\csname BAval@tick.dance.f202.bind.iou\endcsname{0.00}
\expandafter\def\csname BAval@tick.dance.f202.bind.state\endcsname{U}
\expandafter\def\csname BAval@tick.dance.f203.bind.iou\endcsname{0.00}
\expandafter\def\csname BAval@tick.dance.f203.bind.state\endcsname{U}
\expandafter\def\csname BAval@tick.dance.f204.bind.iou\endcsname{0.00}
\expandafter\def\csname BAval@tick.dance.f204.bind.state\endcsname{U}
\expandafter\def\csname BAval@tick.dance.f205.bind.iou\endcsname{0.00}
\expandafter\def\csname BAval@tick.dance.f205.bind.state\endcsname{U}
\expandafter\def\csname BAval@tick.dance.f206.bind.iou\endcsname{0.03}
\expandafter\def\csname BAval@tick.dance.f206.bind.state\endcsname{U}
\expandafter\def\csname BAval@tick.dance.f207.bind.iou\endcsname{0.04}
\expandafter\def\csname BAval@tick.dance.f207.bind.state\endcsname{U}
\expandafter\def\csname BAval@tick.dance.f208.bind.iou\endcsname{0.04}
\expandafter\def\csname BAval@tick.dance.f208.bind.state\endcsname{U}
\expandafter\def\csname BAval@tick.dance.f209.bind.iou\endcsname{0.05}
\expandafter\def\csname BAval@tick.dance.f209.bind.state\endcsname{U}
\expandafter\def\csname BAval@tick.dance.f210.bind.iou\endcsname{0.07}
\expandafter\def\csname BAval@tick.dance.f210.bind.state\endcsname{U}
\expandafter\def\csname BAval@tick.dance.f211.bind.iou\endcsname{0.09}
\expandafter\def\csname BAval@tick.dance.f211.bind.state\endcsname{U}
\expandafter\def\csname BAval@tick.dance.f212.bind.iou\endcsname{0.12}
\expandafter\def\csname BAval@tick.dance.f212.bind.state\endcsname{U}
\expandafter\def\csname BAval@tick.dance.f213.bind.iou\endcsname{0.13}
\expandafter\def\csname BAval@tick.dance.f213.bind.state\endcsname{U}
\expandafter\def\csname BAval@tick.dance.f214.bind.iou\endcsname{0.14}
\expandafter\def\csname BAval@tick.dance.f214.bind.state\endcsname{U}
\expandafter\def\csname BAval@tick.dance.f215.bind.iou\endcsname{0.15}
\expandafter\def\csname BAval@tick.dance.f215.bind.state\endcsname{U}
\expandafter\def\csname BAval@tick.dance.f216.bind.iou\endcsname{0.21}
\expandafter\def\csname BAval@tick.dance.f216.bind.state\endcsname{U}
\expandafter\def\csname BAval@tick.dance.f217.bind.iou\endcsname{0.24}
\expandafter\def\csname BAval@tick.dance.f217.bind.state\endcsname{U}
\expandafter\def\csname BAval@tick.dance.f218.bind.iou\endcsname{0.36}
\expandafter\def\csname BAval@tick.dance.f218.bind.state\endcsname{U}
\expandafter\def\csname BAval@tick.dance.f219.bind.iou\endcsname{0.41}
\expandafter\def\csname BAval@tick.dance.f219.bind.state\endcsname{U}
\expandafter\def\csname BAval@tick.dance.f220.bind.iou\endcsname{0.46}
\expandafter\def\csname BAval@tick.dance.f220.bind.state\endcsname{U}
\expandafter\def\csname BAval@tick.dance.f221.bind.iou\endcsname{0.51}
\expandafter\def\csname BAval@tick.dance.f221.bind.state\endcsname{C}
\expandafter\def\csname BAval@tick.dance.f222.bind.iou\endcsname{0.53}
\expandafter\def\csname BAval@tick.dance.f222.bind.state\endcsname{C}
\expandafter\def\csname BAval@tick.dance.f223.bind.iou\endcsname{0.53}
\expandafter\def\csname BAval@tick.dance.f223.bind.state\endcsname{C}
\expandafter\def\csname BAval@tick.dance.f224.bind.iou\endcsname{0.53}
\expandafter\def\csname BAval@tick.dance.f224.bind.state\endcsname{C}
\expandafter\def\csname BAval@tick.dance.f225.bind.iou\endcsname{0.54}
\expandafter\def\csname BAval@tick.dance.f225.bind.state\endcsname{C}
\expandafter\def\csname BAval@tick.dance.f226.bind.iou\endcsname{0.56}
\expandafter\def\csname BAval@tick.dance.f226.bind.state\endcsname{C}
\expandafter\def\csname BAval@tick.dance.f227.bind.iou\endcsname{0.55}
\expandafter\def\csname BAval@tick.dance.f227.bind.state\endcsname{C}
\expandafter\def\csname BAval@tick.dance.f228.bind.iou\endcsname{0.54}
\expandafter\def\csname BAval@tick.dance.f228.bind.state\endcsname{C}
\expandafter\def\csname BAval@tick.dance.f229.bind.iou\endcsname{0.53}
\expandafter\def\csname BAval@tick.dance.f229.bind.state\endcsname{C}
\expandafter\def\csname BAval@tick.dance.f230.bind.iou\endcsname{0.55}
\expandafter\def\csname BAval@tick.dance.f230.bind.state\endcsname{C}
\expandafter\def\csname BAval@tick.dance.f231.bind.iou\endcsname{0.55}
\expandafter\def\csname BAval@tick.dance.f231.bind.state\endcsname{C}
\expandafter\def\csname BAval@tick.dance.f232.bind.iou\endcsname{0.55}
\expandafter\def\csname BAval@tick.dance.f232.bind.state\endcsname{C}
\expandafter\def\csname BAval@tick.dance.f233.bind.iou\endcsname{0.56}
\expandafter\def\csname BAval@tick.dance.f233.bind.state\endcsname{C}
\expandafter\def\csname BAval@tick.dance.f234.bind.iou\endcsname{0.56}
\expandafter\def\csname BAval@tick.dance.f234.bind.state\endcsname{C}
\expandafter\def\csname BAval@tick.dance.f235.bind.iou\endcsname{0.56}
\expandafter\def\csname BAval@tick.dance.f235.bind.state\endcsname{C}
\expandafter\def\csname BAval@tick.dance.f236.bind.iou\endcsname{0.57}
\expandafter\def\csname BAval@tick.dance.f236.bind.state\endcsname{C}
\expandafter\def\csname BAval@tick.dance.f237.bind.iou\endcsname{0.58}
\expandafter\def\csname BAval@tick.dance.f237.bind.state\endcsname{C}
\expandafter\def\csname BAval@tick.dance.f238.bind.iou\endcsname{0.57}
\expandafter\def\csname BAval@tick.dance.f238.bind.state\endcsname{C}
\expandafter\def\csname BAval@tick.dance.f239.bind.iou\endcsname{0.58}
\expandafter\def\csname BAval@tick.dance.f239.bind.state\endcsname{C}
\expandafter\def\csname BAval@tick.dance.f240.bind.iou\endcsname{0.57}
\expandafter\def\csname BAval@tick.dance.f240.bind.state\endcsname{C}
\expandafter\def\csname BAval@tick.dance.f241.bind.iou\endcsname{0.58}
\expandafter\def\csname BAval@tick.dance.f241.bind.state\endcsname{C}
\expandafter\def\csname BAval@tick.dance.f242.bind.iou\endcsname{0.58}
\expandafter\def\csname BAval@tick.dance.f242.bind.state\endcsname{C}
\expandafter\def\csname BAval@tick.dance.f243.bind.iou\endcsname{0.58}
\expandafter\def\csname BAval@tick.dance.f243.bind.state\endcsname{C}
\expandafter\def\csname BAval@tick.dance.f244.bind.iou\endcsname{0.59}
\expandafter\def\csname BAval@tick.dance.f244.bind.state\endcsname{C}
\expandafter\def\csname BAval@tick.dance.f245.bind.iou\endcsname{0.59}
\expandafter\def\csname BAval@tick.dance.f245.bind.state\endcsname{C}
\expandafter\def\csname BAval@tick.dance.f246.bind.iou\endcsname{0.58}
\expandafter\def\csname BAval@tick.dance.f246.bind.state\endcsname{C}
\expandafter\def\csname BAval@tick.dance.f247.bind.iou\endcsname{0.60}
\expandafter\def\csname BAval@tick.dance.f247.bind.state\endcsname{C}
\expandafter\def\csname BAval@tick.dance.f248.bind.iou\endcsname{0.58}
\expandafter\def\csname BAval@tick.dance.f248.bind.state\endcsname{C}
\expandafter\def\csname BAval@tick.dance.f249.bind.iou\endcsname{0.57}
\expandafter\def\csname BAval@tick.dance.f249.bind.state\endcsname{C}
\expandafter\def\csname BAval@tick.dance.f250.bind.iou\endcsname{0.57}
\expandafter\def\csname BAval@tick.dance.f250.bind.state\endcsname{C}
\expandafter\def\csname BAval@tick.dance.f251.bind.iou\endcsname{0.57}
\expandafter\def\csname BAval@tick.dance.f251.bind.state\endcsname{C}
\expandafter\def\csname BAval@tick.dance.f252.bind.iou\endcsname{0.57}
\expandafter\def\csname BAval@tick.dance.f252.bind.state\endcsname{C}
\expandafter\def\csname BAval@tick.dance.f253.bind.iou\endcsname{0.56}
\expandafter\def\csname BAval@tick.dance.f253.bind.state\endcsname{C}
\expandafter\def\csname BAval@tick.dance.f254.bind.iou\endcsname{0.56}
\expandafter\def\csname BAval@tick.dance.f254.bind.state\endcsname{C}
\expandafter\def\csname BAval@tick.dance.f255.bind.iou\endcsname{0.56}
\expandafter\def\csname BAval@tick.dance.f255.bind.state\endcsname{C}
\expandafter\def\csname BAval@tick.dance.f256.bind.iou\endcsname{0.56}
\expandafter\def\csname BAval@tick.dance.f256.bind.state\endcsname{C}
\expandafter\def\csname BAval@tick.dance.f257.bind.iou\endcsname{0.56}
\expandafter\def\csname BAval@tick.dance.f257.bind.state\endcsname{C}
\expandafter\def\csname BAval@tick.dance.f258.bind.iou\endcsname{0.56}
\expandafter\def\csname BAval@tick.dance.f258.bind.state\endcsname{C}
\expandafter\def\csname BAval@tick.dance.f259.bind.iou\endcsname{0.56}
\expandafter\def\csname BAval@tick.dance.f259.bind.state\endcsname{C}
\expandafter\def\csname BAval@tick.dance.f260.bind.iou\endcsname{0.56}
\expandafter\def\csname BAval@tick.dance.f260.bind.state\endcsname{C}
\expandafter\def\csname BAval@tick.dance.f261.bind.iou\endcsname{0.56}
\expandafter\def\csname BAval@tick.dance.f261.bind.state\endcsname{C}
\expandafter\def\csname BAval@tick.dance.f262.bind.iou\endcsname{0.55}
\expandafter\def\csname BAval@tick.dance.f262.bind.state\endcsname{C}
\expandafter\def\csname BAval@tick.dance.f263.bind.iou\endcsname{0.55}
\expandafter\def\csname BAval@tick.dance.f263.bind.state\endcsname{C}
\expandafter\def\csname BAval@tick.dance.f264.bind.iou\endcsname{0.55}
\expandafter\def\csname BAval@tick.dance.f264.bind.state\endcsname{C}
\expandafter\def\csname BAval@tick.dance.f265.bind.iou\endcsname{0.56}
\expandafter\def\csname BAval@tick.dance.f265.bind.state\endcsname{C}
\expandafter\def\csname BAval@tick.dance.f266.bind.iou\endcsname{0.56}
\expandafter\def\csname BAval@tick.dance.f266.bind.state\endcsname{C}
\expandafter\def\csname BAval@tick.dance.f267.bind.iou\endcsname{0.55}
\expandafter\def\csname BAval@tick.dance.f267.bind.state\endcsname{C}
\expandafter\def\csname BAval@tick.dance.f268.bind.iou\endcsname{0.55}
\expandafter\def\csname BAval@tick.dance.f268.bind.state\endcsname{C}
\expandafter\def\csname BAval@tick.dance.f269.bind.iou\endcsname{0.56}
\expandafter\def\csname BAval@tick.dance.f269.bind.state\endcsname{C}
\expandafter\def\csname BAval@tick.dance.f270.bind.iou\endcsname{0.56}
\expandafter\def\csname BAval@tick.dance.f270.bind.state\endcsname{C}
\expandafter\def\csname BAval@tick.dance.f271.bind.iou\endcsname{0.57}
\expandafter\def\csname BAval@tick.dance.f271.bind.state\endcsname{C}
\expandafter\def\csname BAval@tick.dance.f272.bind.iou\endcsname{0.57}
\expandafter\def\csname BAval@tick.dance.f272.bind.state\endcsname{C}
\expandafter\def\csname BAval@tick.dance.f273.bind.iou\endcsname{0.58}
\expandafter\def\csname BAval@tick.dance.f273.bind.state\endcsname{C}
\expandafter\def\csname BAval@tick.dance.f274.bind.iou\endcsname{0.59}
\expandafter\def\csname BAval@tick.dance.f274.bind.state\endcsname{C}
\expandafter\def\csname BAval@tick.dance.f275.bind.iou\endcsname{0.60}
\expandafter\def\csname BAval@tick.dance.f275.bind.state\endcsname{C}
\expandafter\def\csname BAval@tick.dance.f276.bind.iou\endcsname{0.59}
\expandafter\def\csname BAval@tick.dance.f276.bind.state\endcsname{C}
\expandafter\def\csname BAval@tick.dance.f277.bind.iou\endcsname{0.59}
\expandafter\def\csname BAval@tick.dance.f277.bind.state\endcsname{C}
\expandafter\def\csname BAval@tick.dance.f278.bind.iou\endcsname{0.59}
\expandafter\def\csname BAval@tick.dance.f278.bind.state\endcsname{C}
\expandafter\def\csname BAval@tick.dance.f279.bind.iou\endcsname{0.52}
\expandafter\def\csname BAval@tick.dance.f279.bind.state\endcsname{C}
\expandafter\def\csname BAval@tick.dance.f280.bind.iou\endcsname{0.52}
\expandafter\def\csname BAval@tick.dance.f280.bind.state\endcsname{C}
\expandafter\def\csname BAval@tick.dance.f281.bind.iou\endcsname{0.54}
\expandafter\def\csname BAval@tick.dance.f281.bind.state\endcsname{C}
\expandafter\def\csname BAval@tick.dance.f282.bind.iou\endcsname{0.57}
\expandafter\def\csname BAval@tick.dance.f282.bind.state\endcsname{C}
\expandafter\def\csname BAval@tick.dance.f283.bind.iou\endcsname{0.62}
\expandafter\def\csname BAval@tick.dance.f283.bind.state\endcsname{C}
\expandafter\def\csname BAval@tick.dance.f284.bind.iou\endcsname{0.59}
\expandafter\def\csname BAval@tick.dance.f284.bind.state\endcsname{C}
\expandafter\def\csname BAval@tick.dance.f285.bind.iou\endcsname{0.46}
\expandafter\def\csname BAval@tick.dance.f285.bind.state\endcsname{U}
\expandafter\def\csname BAval@tick.dance.f286.bind.iou\endcsname{0.40}
\expandafter\def\csname BAval@tick.dance.f286.bind.state\endcsname{U}
\expandafter\def\csname BAval@tick.dance.f287.bind.iou\endcsname{0.29}
\expandafter\def\csname BAval@tick.dance.f287.bind.state\endcsname{U}
\expandafter\def\csname BAval@tick.dance.f288.bind.iou\endcsname{0.20}
\expandafter\def\csname BAval@tick.dance.f288.bind.state\endcsname{U}
\expandafter\def\csname BAval@tick.dance.f289.bind.iou\endcsname{0.15}
\expandafter\def\csname BAval@tick.dance.f289.bind.state\endcsname{U}
\expandafter\def\csname BAval@tick.dance.f290.bind.iou\endcsname{0.13}
\expandafter\def\csname BAval@tick.dance.f290.bind.state\endcsname{U}
\expandafter\def\csname BAval@tick.dance.f291.bind.iou\endcsname{0.05}
\expandafter\def\csname BAval@tick.dance.f291.bind.state\endcsname{U}
\expandafter\def\csname BAval@tick.dance.f292.bind.iou\endcsname{0.03}
\expandafter\def\csname BAval@tick.dance.f292.bind.state\endcsname{U}
\expandafter\def\csname BAval@tick.dance.f293.bind.iou\endcsname{0.00}
\expandafter\def\csname BAval@tick.dance.f293.bind.state\endcsname{W}
\expandafter\def\csname BAval@tick.dance.f294.bind.iou\endcsname{0.00}
\expandafter\def\csname BAval@tick.dance.f294.bind.state\endcsname{W}
\expandafter\def\csname BAval@tick.dance.f295.bind.iou\endcsname{0.00}
\expandafter\def\csname BAval@tick.dance.f295.bind.state\endcsname{W}
\expandafter\def\csname BAval@tick.dance.f296.bind.iou\endcsname{0.00}
\expandafter\def\csname BAval@tick.dance.f296.bind.state\endcsname{W}
\expandafter\def\csname BAval@tick.dance.f297.bind.iou\endcsname{0.00}
\expandafter\def\csname BAval@tick.dance.f297.bind.state\endcsname{W}
\expandafter\def\csname BAval@tick.dance.f298.bind.iou\endcsname{0.00}
\expandafter\def\csname BAval@tick.dance.f298.bind.state\endcsname{W}
\expandafter\def\csname BAval@tick.dance.f299.bind.iou\endcsname{0.00}
\expandafter\def\csname BAval@tick.dance.f299.bind.state\endcsname{U}
\expandafter\def\csname BAval@tick.dance.f300.bind.iou\endcsname{0.00}
\expandafter\def\csname BAval@tick.dance.f300.bind.state\endcsname{U}
\expandafter\def\csname BAval@tick.dance.f301.bind.iou\endcsname{0.00}
\expandafter\def\csname BAval@tick.dance.f301.bind.state\endcsname{U}
\expandafter\def\csname BAval@tick.dance.f302.bind.iou\endcsname{0.00}
\expandafter\def\csname BAval@tick.dance.f302.bind.state\endcsname{U}
\expandafter\def\csname BAval@tick.dance.f303.bind.iou\endcsname{0.00}
\expandafter\def\csname BAval@tick.dance.f303.bind.state\endcsname{U}
\expandafter\def\csname BAval@tick.dance.f304.bind.iou\endcsname{0.00}
\expandafter\def\csname BAval@tick.dance.f304.bind.state\endcsname{U}
\expandafter\def\csname BAval@tick.dance.f305.bind.iou\endcsname{0.00}
\expandafter\def\csname BAval@tick.dance.f305.bind.state\endcsname{U}
\expandafter\def\csname BAval@tick.dance.f306.bind.iou\endcsname{0.00}
\expandafter\def\csname BAval@tick.dance.f306.bind.state\endcsname{U}
\expandafter\def\csname BAval@tick.dance.f307.bind.iou\endcsname{0.00}
\expandafter\def\csname BAval@tick.dance.f307.bind.state\endcsname{U}
\expandafter\def\csname BAval@tick.dance.f308.bind.iou\endcsname{0.00}
\expandafter\def\csname BAval@tick.dance.f308.bind.state\endcsname{U}
\expandafter\def\csname BAval@tick.dance.f309.bind.iou\endcsname{0.00}
\expandafter\def\csname BAval@tick.dance.f309.bind.state\endcsname{U}
\expandafter\def\csname BAval@tick.dance.f310.bind.iou\endcsname{0.00}
\expandafter\def\csname BAval@tick.dance.f310.bind.state\endcsname{U}
\expandafter\def\csname BAval@tick.dance.f311.bind.iou\endcsname{0.00}
\expandafter\def\csname BAval@tick.dance.f311.bind.state\endcsname{U}
\expandafter\def\csname BAval@tick.dance.f312.bind.iou\endcsname{0.00}
\expandafter\def\csname BAval@tick.dance.f312.bind.state\endcsname{U}
\expandafter\def\csname BAval@tick.dance.f313.bind.iou\endcsname{0.00}
\expandafter\def\csname BAval@tick.dance.f313.bind.state\endcsname{U}
\expandafter\def\csname BAval@tick.dance.f314.bind.iou\endcsname{0.00}
\expandafter\def\csname BAval@tick.dance.f314.bind.state\endcsname{U}
\expandafter\def\csname BAval@tick.dance.f315.bind.iou\endcsname{0.00}
\expandafter\def\csname BAval@tick.dance.f315.bind.state\endcsname{U}
\expandafter\def\csname BAval@tick.dance.f316.bind.iou\endcsname{0.00}
\expandafter\def\csname BAval@tick.dance.f316.bind.state\endcsname{U}
\expandafter\def\csname BAval@tick.dance.f317.bind.iou\endcsname{0.00}
\expandafter\def\csname BAval@tick.dance.f317.bind.state\endcsname{W}
\expandafter\def\csname BAval@tick.dance.f318.bind.iou\endcsname{0.00}
\expandafter\def\csname BAval@tick.dance.f318.bind.state\endcsname{W}
\expandafter\def\csname BAval@tick.dance.f319.bind.iou\endcsname{0.00}
\expandafter\def\csname BAval@tick.dance.f319.bind.state\endcsname{U}
\expandafter\def\csname BAval@tick.dance.f320.bind.iou\endcsname{0.00}
\expandafter\def\csname BAval@tick.dance.f320.bind.state\endcsname{U}
\expandafter\def\csname BAval@tick.dance.f321.bind.iou\endcsname{0.00}
\expandafter\def\csname BAval@tick.dance.f321.bind.state\endcsname{W}
\expandafter\def\csname BAval@tick.dance.f322.bind.iou\endcsname{0.00}
\expandafter\def\csname BAval@tick.dance.f322.bind.state\endcsname{W}
\expandafter\def\csname BAval@tick.dance.f323.bind.iou\endcsname{0.00}
\expandafter\def\csname BAval@tick.dance.f323.bind.state\endcsname{U}
\expandafter\def\csname BAval@tick.dance.f324.bind.iou\endcsname{0.00}
\expandafter\def\csname BAval@tick.dance.f324.bind.state\endcsname{U}
\expandafter\def\csname BAval@tick.dance.f325.bind.iou\endcsname{0.00}
\expandafter\def\csname BAval@tick.dance.f325.bind.state\endcsname{U}
\expandafter\def\csname BAval@tick.dance.f326.bind.iou\endcsname{0.00}
\expandafter\def\csname BAval@tick.dance.f326.bind.state\endcsname{U}
\expandafter\def\csname BAval@tick.dance.f327.bind.iou\endcsname{0.00}
\expandafter\def\csname BAval@tick.dance.f327.bind.state\endcsname{U}
\expandafter\def\csname BAval@tick.dance.f328.bind.iou\endcsname{0.00}
\expandafter\def\csname BAval@tick.dance.f328.bind.state\endcsname{U}
\expandafter\def\csname BAval@tick.dance.f329.bind.iou\endcsname{0.00}
\expandafter\def\csname BAval@tick.dance.f329.bind.state\endcsname{U}
\expandafter\def\csname BAval@tick.dance.f330.bind.iou\endcsname{0.00}
\expandafter\def\csname BAval@tick.dance.f330.bind.state\endcsname{U}
\expandafter\def\csname BAval@tick.dance.f331.bind.iou\endcsname{0.00}
\expandafter\def\csname BAval@tick.dance.f331.bind.state\endcsname{U}
\expandafter\def\csname BAval@tick.dance.f332.bind.iou\endcsname{0.00}
\expandafter\def\csname BAval@tick.dance.f332.bind.state\endcsname{U}
\expandafter\def\csname BAval@tick.dance.f333.bind.iou\endcsname{0.00}
\expandafter\def\csname BAval@tick.dance.f333.bind.state\endcsname{U}
\expandafter\def\csname BAval@tick.dance.f334.bind.iou\endcsname{0.00}
\expandafter\def\csname BAval@tick.dance.f334.bind.state\endcsname{U}
\expandafter\def\csname BAval@tick.dance.f335.bind.iou\endcsname{0.00}
\expandafter\def\csname BAval@tick.dance.f335.bind.state\endcsname{U}
\expandafter\def\csname BAval@tick.dance.f336.bind.iou\endcsname{0.00}
\expandafter\def\csname BAval@tick.dance.f336.bind.state\endcsname{U}
\expandafter\def\csname BAval@tick.dance.f337.bind.iou\endcsname{0.00}
\expandafter\def\csname BAval@tick.dance.f337.bind.state\endcsname{U}
\expandafter\def\csname BAval@tick.dance.f338.bind.iou\endcsname{0.00}
\expandafter\def\csname BAval@tick.dance.f338.bind.state\endcsname{U}
\expandafter\def\csname BAval@pair.quad.dam.joint\endcsname{72}
\expandafter\def\csname BAval@pair.quad.dam.agree\endcsname{68}
\expandafter\def\csname BAval@pair.quad.dam.other\endcsname{4}
\expandafter\def\csname BAval@pair.quad.dam.unknown\endcsname{18}
\expandafter\def\csname BAval@pair.quad.dam.C.C\endcsname{1}
\expandafter\def\csname BAval@pair.quad.dam.C.W\endcsname{0}
\expandafter\def\csname BAval@pair.quad.dam.C.U\endcsname{0}
\expandafter\def\csname BAval@pair.quad.dam.C.UNKNOWN\endcsname{0}
\expandafter\def\csname BAval@pair.quad.dam.W.C\endcsname{0}
\expandafter\def\csname BAval@pair.quad.dam.W.W\endcsname{9}
\expandafter\def\csname BAval@pair.quad.dam.W.U\endcsname{3}
\expandafter\def\csname BAval@pair.quad.dam.W.UNKNOWN\endcsname{0}
\expandafter\def\csname BAval@pair.quad.dam.U.C\endcsname{0}
\expandafter\def\csname BAval@pair.quad.dam.U.W\endcsname{1}
\expandafter\def\csname BAval@pair.quad.dam.U.U\endcsname{58}
\expandafter\def\csname BAval@pair.quad.dam.U.UNKNOWN\endcsname{0}
\expandafter\def\csname BAval@pair.quad.dam.UNKNOWN.C\endcsname{0}
\expandafter\def\csname BAval@pair.quad.dam.UNKNOWN.W\endcsname{0}
\expandafter\def\csname BAval@pair.quad.dam.UNKNOWN.U\endcsname{0}
\expandafter\def\csname BAval@pair.quad.dam.UNKNOWN.UNKNOWN\endcsname{18}
\expandafter\def\csname BAval@pair.quad.keep.joint\endcsname{72}
\expandafter\def\csname BAval@pair.quad.keep.agree\endcsname{72}
\expandafter\def\csname BAval@pair.quad.keep.other\endcsname{0}
\expandafter\def\csname BAval@pair.quad.keep.unknown\endcsname{18}
\expandafter\def\csname BAval@pair.quad.keep.C.C\endcsname{1}
\expandafter\def\csname BAval@pair.quad.keep.C.W\endcsname{0}
\expandafter\def\csname BAval@pair.quad.keep.C.U\endcsname{0}
\expandafter\def\csname BAval@pair.quad.keep.C.UNKNOWN\endcsname{0}
\expandafter\def\csname BAval@pair.quad.keep.W.C\endcsname{0}
\expandafter\def\csname BAval@pair.quad.keep.W.W\endcsname{12}
\expandafter\def\csname BAval@pair.quad.keep.W.U\endcsname{0}
\expandafter\def\csname BAval@pair.quad.keep.W.UNKNOWN\endcsname{0}
\expandafter\def\csname BAval@pair.quad.keep.U.C\endcsname{0}
\expandafter\def\csname BAval@pair.quad.keep.U.W\endcsname{0}
\expandafter\def\csname BAval@pair.quad.keep.U.U\endcsname{59}
\expandafter\def\csname BAval@pair.quad.keep.U.UNKNOWN\endcsname{0}
\expandafter\def\csname BAval@pair.quad.keep.UNKNOWN.C\endcsname{0}
\expandafter\def\csname BAval@pair.quad.keep.UNKNOWN.W\endcsname{0}
\expandafter\def\csname BAval@pair.quad.keep.UNKNOWN.U\endcsname{0}
\expandafter\def\csname BAval@pair.quad.keep.UNKNOWN.UNKNOWN\endcsname{18}
\expandafter\def\csname BAval@pair.dance.dam.joint\endcsname{180}
\expandafter\def\csname BAval@pair.dance.dam.agree\endcsname{177}
\expandafter\def\csname BAval@pair.dance.dam.other\endcsname{3}
\expandafter\def\csname BAval@pair.dance.dam.unknown\endcsname{0}
\expandafter\def\csname BAval@pair.dance.dam.C.C\endcsname{78}
\expandafter\def\csname BAval@pair.dance.dam.C.W\endcsname{0}
\expandafter\def\csname BAval@pair.dance.dam.C.U\endcsname{0}
\expandafter\def\csname BAval@pair.dance.dam.C.UNKNOWN\endcsname{0}
\expandafter\def\csname BAval@pair.dance.dam.W.C\endcsname{0}
\expandafter\def\csname BAval@pair.dance.dam.W.W\endcsname{20}
\expandafter\def\csname BAval@pair.dance.dam.W.U\endcsname{3}
\expandafter\def\csname BAval@pair.dance.dam.W.UNKNOWN\endcsname{0}
\expandafter\def\csname BAval@pair.dance.dam.U.C\endcsname{0}
\expandafter\def\csname BAval@pair.dance.dam.U.W\endcsname{0}
\expandafter\def\csname BAval@pair.dance.dam.U.U\endcsname{79}
\expandafter\def\csname BAval@pair.dance.dam.U.UNKNOWN\endcsname{0}
\expandafter\def\csname BAval@pair.dance.dam.UNKNOWN.C\endcsname{0}
\expandafter\def\csname BAval@pair.dance.dam.UNKNOWN.W\endcsname{0}
\expandafter\def\csname BAval@pair.dance.dam.UNKNOWN.U\endcsname{0}
\expandafter\def\csname BAval@pair.dance.dam.UNKNOWN.UNKNOWN\endcsname{0}
\expandafter\def\csname BAval@pair.dance.keep.joint\endcsname{179}
\expandafter\def\csname BAval@pair.dance.keep.agree\endcsname{170}
\expandafter\def\csname BAval@pair.dance.keep.other\endcsname{9}
\expandafter\def\csname BAval@pair.dance.keep.unknown\endcsname{1}
\expandafter\def\csname BAval@pair.dance.keep.C.C\endcsname{72}
\expandafter\def\csname BAval@pair.dance.keep.C.W\endcsname{0}
\expandafter\def\csname BAval@pair.dance.keep.C.U\endcsname{6}
\expandafter\def\csname BAval@pair.dance.keep.C.UNKNOWN\endcsname{0}
\expandafter\def\csname BAval@pair.dance.keep.W.C\endcsname{0}
\expandafter\def\csname BAval@pair.dance.keep.W.W\endcsname{21}
\expandafter\def\csname BAval@pair.dance.keep.W.U\endcsname{1}
\expandafter\def\csname BAval@pair.dance.keep.W.UNKNOWN\endcsname{1}
\expandafter\def\csname BAval@pair.dance.keep.U.C\endcsname{0}
\expandafter\def\csname BAval@pair.dance.keep.U.W\endcsname{2}
\expandafter\def\csname BAval@pair.dance.keep.U.U\endcsname{77}
\expandafter\def\csname BAval@pair.dance.keep.U.UNKNOWN\endcsname{0}
\expandafter\def\csname BAval@pair.dance.keep.UNKNOWN.C\endcsname{0}
\expandafter\def\csname BAval@pair.dance.keep.UNKNOWN.W\endcsname{0}
\expandafter\def\csname BAval@pair.dance.keep.UNKNOWN.U\endcsname{0}
\expandafter\def\csname BAval@pair.dance.keep.UNKNOWN.UNKNOWN\endcsname{0}
\expandafter\def\csname BAval@pair.dam.joint\endcsname{252}
\expandafter\def\csname BAval@pair.dam.agree\endcsname{245}
\expandafter\def\csname BAval@pair.dam.other\endcsname{7}
\expandafter\def\csname BAval@pair.dam.unknown\endcsname{18}
\expandafter\def\csname BAval@pair.dam.C.C\endcsname{79}
\expandafter\def\csname BAval@pair.dam.C.W\endcsname{0}
\expandafter\def\csname BAval@pair.dam.C.U\endcsname{0}
\expandafter\def\csname BAval@pair.dam.C.UNKNOWN\endcsname{0}
\expandafter\def\csname BAval@pair.dam.W.C\endcsname{0}
\expandafter\def\csname BAval@pair.dam.W.W\endcsname{29}
\expandafter\def\csname BAval@pair.dam.W.U\endcsname{6}
\expandafter\def\csname BAval@pair.dam.W.UNKNOWN\endcsname{0}
\expandafter\def\csname BAval@pair.dam.U.C\endcsname{0}
\expandafter\def\csname BAval@pair.dam.U.W\endcsname{1}
\expandafter\def\csname BAval@pair.dam.U.U\endcsname{137}
\expandafter\def\csname BAval@pair.dam.U.UNKNOWN\endcsname{0}
\expandafter\def\csname BAval@pair.dam.UNKNOWN.C\endcsname{0}
\expandafter\def\csname BAval@pair.dam.UNKNOWN.W\endcsname{0}
\expandafter\def\csname BAval@pair.dam.UNKNOWN.U\endcsname{0}
\expandafter\def\csname BAval@pair.dam.UNKNOWN.UNKNOWN\endcsname{18}
\expandafter\def\csname BAval@pair.keep.joint\endcsname{251}
\expandafter\def\csname BAval@pair.keep.agree\endcsname{242}
\expandafter\def\csname BAval@pair.keep.other\endcsname{9}
\expandafter\def\csname BAval@pair.keep.unknown\endcsname{19}
\expandafter\def\csname BAval@pair.keep.C.C\endcsname{73}
\expandafter\def\csname BAval@pair.keep.C.W\endcsname{0}
\expandafter\def\csname BAval@pair.keep.C.U\endcsname{6}
\expandafter\def\csname BAval@pair.keep.C.UNKNOWN\endcsname{0}
\expandafter\def\csname BAval@pair.keep.W.C\endcsname{0}
\expandafter\def\csname BAval@pair.keep.W.W\endcsname{33}
\expandafter\def\csname BAval@pair.keep.W.U\endcsname{1}
\expandafter\def\csname BAval@pair.keep.W.UNKNOWN\endcsname{1}
\expandafter\def\csname BAval@pair.keep.U.C\endcsname{0}
\expandafter\def\csname BAval@pair.keep.U.W\endcsname{2}
\expandafter\def\csname BAval@pair.keep.U.U\endcsname{136}
\expandafter\def\csname BAval@pair.keep.U.UNKNOWN\endcsname{0}
\expandafter\def\csname BAval@pair.keep.UNKNOWN.C\endcsname{0}
\expandafter\def\csname BAval@pair.keep.UNKNOWN.W\endcsname{0}
\expandafter\def\csname BAval@pair.keep.UNKNOWN.U\endcsname{0}
\expandafter\def\csname BAval@pair.keep.UNKNOWN.UNKNOWN\endcsname{18}
\expandafter\def\csname BAval@rec.six.frames\endcsname{6}
\expandafter\def\csname BAval@rec.six.seconds\endcsname{0.30}
\expandafter\def\csname BAval@rec.six.start\endcsname{279}
\expandafter\def\csname BAval@rec.six.end\endcsname{284}
\expandafter\def\csname BAval@rec.six.anchor\endcsname{282}
\expandafter\def\csname BAval@rec.frame282.dam.iou\endcsname{0.59}
\expandafter\def\csname BAval@rec.frame282.keep.iou\endcsname{0.45}
\expandafter\def\csname BAval@rec.frame282.bind.iou\endcsname{0.57}
\expandafter\def\csname BAval@case.B.f120.frame\endcsname{120}
\expandafter\def\csname BAval@case.B.f120.kcf\endcsname{0.86}
\expandafter\def\csname BAval@case.B.f120.csrt\endcsname{0.66}
\expandafter\def\csname BAval@case.B.f140.frame\endcsname{140}
\expandafter\def\csname BAval@case.B.f140.kcf\endcsname{0.73}
\expandafter\def\csname BAval@case.B.f140.csrt\endcsname{0.62}
\expandafter\def\csname BAval@case.B.kcf.returned\endcsname{40}
\expandafter\def\csname BAval@case.B.kcf.evaluated.steps\endcsname{40}
\expandafter\def\csname BAval@case.B.kcf.mean\endcsname{0.877}
\expandafter\def\csname BAval@case.B.kcf.successes.iou50\endcsname{40}
\expandafter\def\csname BAval@case.B.csrt.returned\endcsname{40}
\expandafter\def\csname BAval@case.B.csrt.evaluated.steps\endcsname{40}
\expandafter\def\csname BAval@case.B.csrt.mean\endcsname{0.716}
\expandafter\def\csname BAval@case.B.csrt.successes.iou50\endcsname{40}
\expandafter\def\csname BAval@case.D.f272.frame\endcsname{272}
\expandafter\def\csname BAval@case.D.f272.kcf\endcsname{0.38}
\expandafter\def\csname BAval@case.D.f272.csrt\endcsname{0.43}
\expandafter\def\csname BAval@case.D.f300.frame\endcsname{300}
\expandafter\def\csname BAval@case.D.f300.kcf\endcsname{Lost}
\expandafter\def\csname BAval@case.D.f300.csrt\endcsname{0.57}
\expandafter\def\csname BAval@case.D.kcf.returned\endcsname{39}
\expandafter\def\csname BAval@case.D.kcf.evaluated.steps\endcsname{60}
\expandafter\def\csname BAval@case.D.csrt.returned\endcsname{60}
\expandafter\def\csname BAval@case.D.csrt.evaluated.steps\endcsname{60}
\expandafter\def\csname BAval@tab.delta.sid\endcsname{7.1}
\expandafter\def\csname BAval@tab.delta.wrong\endcsname{2.9}
\expandafter\def\csname BAval@tab.delta.recovery\endcsname{7.2}
\expandafter\def\csname BAval@tab.delta.t5\endcsname{0.6}
\expandafter\def\csname BAval@tab.delta.short\endcsname{4.1}
\expandafter\def\csname BAval@tab.delta.similar\endcsname{6.2}
\expandafter\def\csname BAval@tab.delta.long\endcsname{10.0}
\expandafter\def\csname BAval@tab.delta.ms\endcsname{0.5}
\expandafter\def\csname BAval@tab.abl.src\endcsname{3.0}
\expandafter\def\csname BAval@tab.abl.mem\endcsname{4.3}
\expandafter\def\csname BAval@tab.abl.tgt\endcsname{6.2}
\expandafter\def\csname BAval@tab.abl.ret\endcsname{4.4}
\expandafter\def\csname BAval@tab.abl.sep\endcsname{4.7}
\expandafter\def\csname BAval@tab.abl.fix\endcsname{8.5}
\expandafter\def\csname BAval@plan.strata.short.episodes\endcsname{120}
\expandafter\def\csname BAval@plan.strata.short.weight\endcsname{0.250}
\expandafter\def\csname BAval@plan.strata.similar.episodes\endcsname{180}
\expandafter\def\csname BAval@plan.strata.similar.weight\endcsname{0.375}
\expandafter\def\csname BAval@plan.strata.long.episodes\endcsname{180}
\expandafter\def\csname BAval@plan.strata.long.weight\endcsname{0.375}
\expandafter\def\csname BAval@plan.train.groups\endcsname{60}
\expandafter\def\csname BAval@plan.train.episodes\endcsname{1200}
\expandafter\def\csname BAval@plan.validation.groups\endcsname{20}
\expandafter\def\csname BAval@plan.validation.episodes\endcsname{200}
\expandafter\def\csname BAval@plan.test.groups\endcsname{40}
\expandafter\def\csname BAval@plan.test.episodes\endcsname{480}
\expandafter\def\csname BAval@t17.r01.axis\endcsname{0}
\expandafter\def\csname BAval@t17.r02.axis\endcsname{0.1}
\expandafter\def\csname BAval@t17.r03.axis\endcsname{0.25}
\expandafter\def\csname BAval@t17.r04.axis\endcsname{0.5}
\expandafter\def\csname BAval@t17.r05.axis\endcsname{1}
\expandafter\def\csname BAval@t17.r06.axis\endcsname{2}
\expandafter\def\csname BAval@t17.r07.axis\endcsname{0.5}
\expandafter\def\csname BAval@t17.r08.axis\endcsname{0.75}
\expandafter\def\csname BAval@t17.r09.axis\endcsname{1}
\expandafter\def\csname BAval@t17.r10.axis\endcsname{1.25}
\expandafter\def\csname BAval@t17.r11.axis\endcsname{1.5}
\expandafter\def\csname BAval@t17.r12.axis\endcsname{2}
\expandafter\def\csname BAval@t05.r01.axis\endcsname{0}
\expandafter\def\csname BAval@t05.r02.axis\endcsname{0.1}
\expandafter\def\csname BAval@t05.r03.axis\endcsname{0.25}
\expandafter\def\csname BAval@t05.r04.axis\endcsname{0.5}
\expandafter\def\csname BAval@t05.r05.axis\endcsname{1.0}
\expandafter\def\csname BAval@t05.r06.axis\endcsname{2.0}
\expandafter\def\csname BAval@t06.r01.axis\endcsname{0.5}
\expandafter\def\csname BAval@t06.r02.axis\endcsname{0.75}
\expandafter\def\csname BAval@t06.r03.axis\endcsname{1.0}
\expandafter\def\csname BAval@t06.r04.axis\endcsname{1.25}
\expandafter\def\csname BAval@t06.r05.axis\endcsname{1.5}
\expandafter\def\csname BAval@t06.r06.axis\endcsname{2.0}
\expandafter\def\csname BAval@t20.r01.axis\endcsname{0.0}
\expandafter\def\csname BAval@t20.r02.axis\endcsname{0.02}
\expandafter\def\csname BAval@t20.r03.axis\endcsname{0.04}
\expandafter\def\csname BAval@t20.r04.axis\endcsname{0.06}
\expandafter\def\csname BAval@t20.r05.axis\endcsname{0.08}
\expandafter\def\csname BAval@t20.r06.axis\endcsname{0.1}
\expandafter\def\csname BAval@t20.r07.axis\endcsname{0.12}
\expandafter\def\csname BAval@t20.r08.axis\endcsname{0.14}
\expandafter\def\csname BAval@t20.r09.axis\endcsname{0.16}
\expandafter\def\csname BAval@t20.r10.axis\endcsname{0.18}
\expandafter\def\csname BAval@t20.r11.axis\endcsname{0.2}
\expandafter\def\csname BAval@t20.r12.axis\endcsname{0.22}
\expandafter\def\csname BAval@t20.r13.axis\endcsname{0.24}
\expandafter\def\csname BAval@t20.r14.axis\endcsname{0.26}
\expandafter\def\csname BAval@t20.r15.axis\endcsname{0.28}
\expandafter\def\csname BAval@t20.r16.axis\endcsname{0.3}
\expandafter\def\csname BAval@t20.r17.axis\endcsname{0.32}
\expandafter\def\csname BAval@t20.r18.axis\endcsname{0.34}
\expandafter\def\csname BAval@t20.r19.axis\endcsname{0.36}
\expandafter\def\csname BAval@t20.r20.axis\endcsname{0.38}
\expandafter\def\csname BAval@t20.r21.axis\endcsname{0.4}
\expandafter\def\csname BAval@t20.r22.axis\endcsname{0.42}
\expandafter\def\csname BAval@t20.r23.axis\endcsname{0.44}
\expandafter\def\csname BAval@t20.r24.axis\endcsname{0.46}
\expandafter\def\csname BAval@t20.r25.axis\endcsname{0.48}
\expandafter\def\csname BAval@t20.r26.axis\endcsname{0.5}
\expandafter\def\csname BAval@t20.r27.axis\endcsname{0.52}
\expandafter\def\csname BAval@t20.r28.axis\endcsname{0.54}
\expandafter\def\csname BAval@t20.r29.axis\endcsname{0.56}
\expandafter\def\csname BAval@t20.r30.axis\endcsname{0.58}
\expandafter\def\csname BAval@t20.r31.axis\endcsname{0.6}
\expandafter\def\csname BAval@t20.r32.axis\endcsname{0.62}
\expandafter\def\csname BAval@t20.r33.axis\endcsname{0.64}
\expandafter\def\csname BAval@t20.r34.axis\endcsname{0.66}
\expandafter\def\csname BAval@t20.r35.axis\endcsname{0.68}
\expandafter\def\csname BAval@t20.r36.axis\endcsname{0.7}
\expandafter\def\csname BAval@t20.r37.axis\endcsname{0.72}
\expandafter\def\csname BAval@t20.r38.axis\endcsname{0.74}
\expandafter\def\csname BAval@t20.r39.axis\endcsname{0.76}
\expandafter\def\csname BAval@t20.r40.axis\endcsname{0.78}
\expandafter\def\csname BAval@t20.r41.axis\endcsname{0.8}
\expandafter\def\csname BAval@t20.r42.axis\endcsname{0.82}
\expandafter\def\csname BAval@t20.r43.axis\endcsname{0.84}
\expandafter\def\csname BAval@t20.r44.axis\endcsname{0.86}
\expandafter\def\csname BAval@t20.r45.axis\endcsname{0.88}
\expandafter\def\csname BAval@t20.r46.axis\endcsname{0.9}
\expandafter\def\csname BAval@t20.r47.axis\endcsname{0.92}
\expandafter\def\csname BAval@t20.r48.axis\endcsname{0.94}
\expandafter\def\csname BAval@t20.r49.axis\endcsname{0.96}
\expandafter\def\csname BAval@t20.r50.axis\endcsname{0.98}
\expandafter\def\csname BAval@t20.r51.axis\endcsname{1.0}
\expandafter\def\csname BAval@t20.r52.axis\endcsname{0.0}
\expandafter\def\csname BAval@t20.r53.axis\endcsname{0.02}
\expandafter\def\csname BAval@t20.r54.axis\endcsname{0.04}
\expandafter\def\csname BAval@t20.r55.axis\endcsname{0.06}
\expandafter\def\csname BAval@t20.r56.axis\endcsname{0.08}
\expandafter\def\csname BAval@t20.r57.axis\endcsname{0.1}
\expandafter\def\csname BAval@t20.r58.axis\endcsname{0.12}
\expandafter\def\csname BAval@t20.r59.axis\endcsname{0.14}
\expandafter\def\csname BAval@t20.r60.axis\endcsname{0.16}
\expandafter\def\csname BAval@t20.r61.axis\endcsname{0.18}
\expandafter\def\csname BAval@t20.r62.axis\endcsname{0.2}
\expandafter\def\csname BAval@t20.r63.axis\endcsname{0.22}
\expandafter\def\csname BAval@t20.r64.axis\endcsname{0.24}
\expandafter\def\csname BAval@t20.r65.axis\endcsname{0.26}
\expandafter\def\csname BAval@t20.r66.axis\endcsname{0.28}
\expandafter\def\csname BAval@t20.r67.axis\endcsname{0.3}
\expandafter\def\csname BAval@t20.r68.axis\endcsname{0.32}
\expandafter\def\csname BAval@t20.r69.axis\endcsname{0.34}
\expandafter\def\csname BAval@t20.r70.axis\endcsname{0.36}
\expandafter\def\csname BAval@t20.r71.axis\endcsname{0.38}
\expandafter\def\csname BAval@t20.r72.axis\endcsname{0.4}
\expandafter\def\csname BAval@t20.r73.axis\endcsname{0.42}
\expandafter\def\csname BAval@t20.r74.axis\endcsname{0.44}
\expandafter\def\csname BAval@t20.r75.axis\endcsname{0.46}
\expandafter\def\csname BAval@t20.r76.axis\endcsname{0.48}
\expandafter\def\csname BAval@t20.r77.axis\endcsname{0.5}
\expandafter\def\csname BAval@t20.r78.axis\endcsname{0.52}
\expandafter\def\csname BAval@t20.r79.axis\endcsname{0.54}
\expandafter\def\csname BAval@t20.r80.axis\endcsname{0.56}
\expandafter\def\csname BAval@t20.r81.axis\endcsname{0.58}
\expandafter\def\csname BAval@t20.r82.axis\endcsname{0.6}
\expandafter\def\csname BAval@t20.r83.axis\endcsname{0.62}
\expandafter\def\csname BAval@t20.r84.axis\endcsname{0.64}
\expandafter\def\csname BAval@t20.r85.axis\endcsname{0.66}
\expandafter\def\csname BAval@t20.r86.axis\endcsname{0.68}
\expandafter\def\csname BAval@t20.r87.axis\endcsname{0.7}
\expandafter\def\csname BAval@t20.r88.axis\endcsname{0.72}
\expandafter\def\csname BAval@t20.r89.axis\endcsname{0.74}
\expandafter\def\csname BAval@t20.r90.axis\endcsname{0.76}
\expandafter\def\csname BAval@t20.r91.axis\endcsname{0.78}
\expandafter\def\csname BAval@t20.r92.axis\endcsname{0.8}
\expandafter\def\csname BAval@t20.r93.axis\endcsname{0.82}
\expandafter\def\csname BAval@t20.r94.axis\endcsname{0.84}
\expandafter\def\csname BAval@t20.r95.axis\endcsname{0.86}
\expandafter\def\csname BAval@t20.r96.axis\endcsname{0.88}
\expandafter\def\csname BAval@t20.r97.axis\endcsname{0.9}
\expandafter\def\csname BAval@t20.r98.axis\endcsname{0.92}
\expandafter\def\csname BAval@t20.r99.axis\endcsname{0.94}
\expandafter\def\csname BAval@t20.r100.axis\endcsname{0.96}
\expandafter\def\csname BAval@t20.r101.axis\endcsname{0.98}
\expandafter\def\csname BAval@t20.r102.axis\endcsname{1.0}
\expandafter\def\csname BAval@t20.r103.axis\endcsname{0.0}
\expandafter\def\csname BAval@t20.r104.axis\endcsname{0.02}
\expandafter\def\csname BAval@t20.r105.axis\endcsname{0.04}
\expandafter\def\csname BAval@t20.r106.axis\endcsname{0.06}
\expandafter\def\csname BAval@t20.r107.axis\endcsname{0.08}
\expandafter\def\csname BAval@t20.r108.axis\endcsname{0.1}
\expandafter\def\csname BAval@t20.r109.axis\endcsname{0.12}
\expandafter\def\csname BAval@t20.r110.axis\endcsname{0.14}
\expandafter\def\csname BAval@t20.r111.axis\endcsname{0.16}
\expandafter\def\csname BAval@t20.r112.axis\endcsname{0.18}
\expandafter\def\csname BAval@t20.r113.axis\endcsname{0.2}
\expandafter\def\csname BAval@t20.r114.axis\endcsname{0.22}
\expandafter\def\csname BAval@t20.r115.axis\endcsname{0.24}
\expandafter\def\csname BAval@t20.r116.axis\endcsname{0.26}
\expandafter\def\csname BAval@t20.r117.axis\endcsname{0.28}
\expandafter\def\csname BAval@t20.r118.axis\endcsname{0.3}
\expandafter\def\csname BAval@t20.r119.axis\endcsname{0.32}
\expandafter\def\csname BAval@t20.r120.axis\endcsname{0.34}
\expandafter\def\csname BAval@t20.r121.axis\endcsname{0.36}
\expandafter\def\csname BAval@t20.r122.axis\endcsname{0.38}
\expandafter\def\csname BAval@t20.r123.axis\endcsname{0.4}
\expandafter\def\csname BAval@t20.r124.axis\endcsname{0.42}
\expandafter\def\csname BAval@t20.r125.axis\endcsname{0.44}
\expandafter\def\csname BAval@t20.r126.axis\endcsname{0.46}
\expandafter\def\csname BAval@t20.r127.axis\endcsname{0.48}
\expandafter\def\csname BAval@t20.r128.axis\endcsname{0.5}
\expandafter\def\csname BAval@t20.r129.axis\endcsname{0.52}
\expandafter\def\csname BAval@t20.r130.axis\endcsname{0.54}
\expandafter\def\csname BAval@t20.r131.axis\endcsname{0.56}
\expandafter\def\csname BAval@t20.r132.axis\endcsname{0.58}
\expandafter\def\csname BAval@t20.r133.axis\endcsname{0.6}
\expandafter\def\csname BAval@t20.r134.axis\endcsname{0.62}
\expandafter\def\csname BAval@t20.r135.axis\endcsname{0.64}
\expandafter\def\csname BAval@t20.r136.axis\endcsname{0.66}
\expandafter\def\csname BAval@t20.r137.axis\endcsname{0.68}
\expandafter\def\csname BAval@t20.r138.axis\endcsname{0.7}
\expandafter\def\csname BAval@t20.r139.axis\endcsname{0.72}
\expandafter\def\csname BAval@t20.r140.axis\endcsname{0.74}
\expandafter\def\csname BAval@t20.r141.axis\endcsname{0.76}
\expandafter\def\csname BAval@t20.r142.axis\endcsname{0.78}
\expandafter\def\csname BAval@t20.r143.axis\endcsname{0.8}
\expandafter\def\csname BAval@t20.r144.axis\endcsname{0.82}
\expandafter\def\csname BAval@t20.r145.axis\endcsname{0.84}
\expandafter\def\csname BAval@t20.r146.axis\endcsname{0.86}
\expandafter\def\csname BAval@t20.r147.axis\endcsname{0.88}
\expandafter\def\csname BAval@t20.r148.axis\endcsname{0.9}
\expandafter\def\csname BAval@t20.r149.axis\endcsname{0.92}
\expandafter\def\csname BAval@t20.r150.axis\endcsname{0.94}
\expandafter\def\csname BAval@t20.r151.axis\endcsname{0.96}
\expandafter\def\csname BAval@t20.r152.axis\endcsname{0.98}
\expandafter\def\csname BAval@t20.r153.axis\endcsname{1.0}
\expandafter\def\csname BAval@t20.r154.axis\endcsname{0.0}
\expandafter\def\csname BAval@t20.r155.axis\endcsname{0.02}
\expandafter\def\csname BAval@t20.r156.axis\endcsname{0.04}
\expandafter\def\csname BAval@t20.r157.axis\endcsname{0.06}
\expandafter\def\csname BAval@t20.r158.axis\endcsname{0.08}
\expandafter\def\csname BAval@t20.r159.axis\endcsname{0.1}
\expandafter\def\csname BAval@t20.r160.axis\endcsname{0.12}
\expandafter\def\csname BAval@t20.r161.axis\endcsname{0.14}
\expandafter\def\csname BAval@t20.r162.axis\endcsname{0.16}
\expandafter\def\csname BAval@t20.r163.axis\endcsname{0.18}
\expandafter\def\csname BAval@t20.r164.axis\endcsname{0.2}
\expandafter\def\csname BAval@t20.r165.axis\endcsname{0.22}
\expandafter\def\csname BAval@t20.r166.axis\endcsname{0.24}
\expandafter\def\csname BAval@t20.r167.axis\endcsname{0.26}
\expandafter\def\csname BAval@t20.r168.axis\endcsname{0.28}
\expandafter\def\csname BAval@t20.r169.axis\endcsname{0.3}
\expandafter\def\csname BAval@t20.r170.axis\endcsname{0.32}
\expandafter\def\csname BAval@t20.r171.axis\endcsname{0.34}
\expandafter\def\csname BAval@t20.r172.axis\endcsname{0.36}
\expandafter\def\csname BAval@t20.r173.axis\endcsname{0.38}
\expandafter\def\csname BAval@t20.r174.axis\endcsname{0.4}
\expandafter\def\csname BAval@t20.r175.axis\endcsname{0.42}
\expandafter\def\csname BAval@t20.r176.axis\endcsname{0.44}
\expandafter\def\csname BAval@t20.r177.axis\endcsname{0.46}
\expandafter\def\csname BAval@t20.r178.axis\endcsname{0.48}
\expandafter\def\csname BAval@t20.r179.axis\endcsname{0.5}
\expandafter\def\csname BAval@t20.r180.axis\endcsname{0.52}
\expandafter\def\csname BAval@t20.r181.axis\endcsname{0.54}
\expandafter\def\csname BAval@t20.r182.axis\endcsname{0.56}
\expandafter\def\csname BAval@t20.r183.axis\endcsname{0.58}
\expandafter\def\csname BAval@t20.r184.axis\endcsname{0.6}
\expandafter\def\csname BAval@t20.r185.axis\endcsname{0.62}
\expandafter\def\csname BAval@t20.r186.axis\endcsname{0.64}
\expandafter\def\csname BAval@t20.r187.axis\endcsname{0.66}
\expandafter\def\csname BAval@t20.r188.axis\endcsname{0.68}
\expandafter\def\csname BAval@t20.r189.axis\endcsname{0.7}
\expandafter\def\csname BAval@t20.r190.axis\endcsname{0.72}
\expandafter\def\csname BAval@t20.r191.axis\endcsname{0.74}
\expandafter\def\csname BAval@t20.r192.axis\endcsname{0.76}
\expandafter\def\csname BAval@t20.r193.axis\endcsname{0.78}
\expandafter\def\csname BAval@t20.r194.axis\endcsname{0.8}
\expandafter\def\csname BAval@t20.r195.axis\endcsname{0.82}
\expandafter\def\csname BAval@t20.r196.axis\endcsname{0.84}
\expandafter\def\csname BAval@t20.r197.axis\endcsname{0.86}
\expandafter\def\csname BAval@t20.r198.axis\endcsname{0.88}
\expandafter\def\csname BAval@t20.r199.axis\endcsname{0.9}
\expandafter\def\csname BAval@t20.r200.axis\endcsname{0.92}
\expandafter\def\csname BAval@t20.r201.axis\endcsname{0.94}
\expandafter\def\csname BAval@t20.r202.axis\endcsname{0.96}
\expandafter\def\csname BAval@t20.r203.axis\endcsname{0.98}
\expandafter\def\csname BAval@t20.r204.axis\endcsname{1.0}
\expandafter\def\csname BAval@t20.r205.axis\endcsname{0.0}
\expandafter\def\csname BAval@t20.r206.axis\endcsname{0.02}
\expandafter\def\csname BAval@t20.r207.axis\endcsname{0.04}
\expandafter\def\csname BAval@t20.r208.axis\endcsname{0.06}
\expandafter\def\csname BAval@t20.r209.axis\endcsname{0.08}
\expandafter\def\csname BAval@t20.r210.axis\endcsname{0.1}
\expandafter\def\csname BAval@t20.r211.axis\endcsname{0.12}
\expandafter\def\csname BAval@t20.r212.axis\endcsname{0.14}
\expandafter\def\csname BAval@t20.r213.axis\endcsname{0.16}
\expandafter\def\csname BAval@t20.r214.axis\endcsname{0.18}
\expandafter\def\csname BAval@t20.r215.axis\endcsname{0.2}
\expandafter\def\csname BAval@t20.r216.axis\endcsname{0.22}
\expandafter\def\csname BAval@t20.r217.axis\endcsname{0.24}
\expandafter\def\csname BAval@t20.r218.axis\endcsname{0.26}
\expandafter\def\csname BAval@t20.r219.axis\endcsname{0.28}
\expandafter\def\csname BAval@t20.r220.axis\endcsname{0.3}
\expandafter\def\csname BAval@t20.r221.axis\endcsname{0.32}
\expandafter\def\csname BAval@t20.r222.axis\endcsname{0.34}
\expandafter\def\csname BAval@t20.r223.axis\endcsname{0.36}
\expandafter\def\csname BAval@t20.r224.axis\endcsname{0.38}
\expandafter\def\csname BAval@t20.r225.axis\endcsname{0.4}
\expandafter\def\csname BAval@t20.r226.axis\endcsname{0.42}
\expandafter\def\csname BAval@t20.r227.axis\endcsname{0.44}
\expandafter\def\csname BAval@t20.r228.axis\endcsname{0.46}
\expandafter\def\csname BAval@t20.r229.axis\endcsname{0.48}
\expandafter\def\csname BAval@t20.r230.axis\endcsname{0.5}
\expandafter\def\csname BAval@t20.r231.axis\endcsname{0.52}
\expandafter\def\csname BAval@t20.r232.axis\endcsname{0.54}
\expandafter\def\csname BAval@t20.r233.axis\endcsname{0.56}
\expandafter\def\csname BAval@t20.r234.axis\endcsname{0.58}
\expandafter\def\csname BAval@t20.r235.axis\endcsname{0.6}
\expandafter\def\csname BAval@t20.r236.axis\endcsname{0.62}
\expandafter\def\csname BAval@t20.r237.axis\endcsname{0.64}
\expandafter\def\csname BAval@t20.r238.axis\endcsname{0.66}
\expandafter\def\csname BAval@t20.r239.axis\endcsname{0.68}
\expandafter\def\csname BAval@t20.r240.axis\endcsname{0.7}
\expandafter\def\csname BAval@t20.r241.axis\endcsname{0.72}
\expandafter\def\csname BAval@t20.r242.axis\endcsname{0.74}
\expandafter\def\csname BAval@t20.r243.axis\endcsname{0.76}
\expandafter\def\csname BAval@t20.r244.axis\endcsname{0.78}
\expandafter\def\csname BAval@t20.r245.axis\endcsname{0.8}
\expandafter\def\csname BAval@t20.r246.axis\endcsname{0.82}
\expandafter\def\csname BAval@t20.r247.axis\endcsname{0.84}
\expandafter\def\csname BAval@t20.r248.axis\endcsname{0.86}
\expandafter\def\csname BAval@t20.r249.axis\endcsname{0.88}
\expandafter\def\csname BAval@t20.r250.axis\endcsname{0.9}
\expandafter\def\csname BAval@t20.r251.axis\endcsname{0.92}
\expandafter\def\csname BAval@t20.r252.axis\endcsname{0.94}
\expandafter\def\csname BAval@t20.r253.axis\endcsname{0.96}
\expandafter\def\csname BAval@t20.r254.axis\endcsname{0.98}
\expandafter\def\csname BAval@t20.r255.axis\endcsname{1.0}
\expandafter\def\csname BAval@t20.r256.axis\endcsname{0.0}
\expandafter\def\csname BAval@t20.r257.axis\endcsname{0.02}
\expandafter\def\csname BAval@t20.r258.axis\endcsname{0.04}
\expandafter\def\csname BAval@t20.r259.axis\endcsname{0.06}
\expandafter\def\csname BAval@t20.r260.axis\endcsname{0.08}
\expandafter\def\csname BAval@t20.r261.axis\endcsname{0.1}
\expandafter\def\csname BAval@t20.r262.axis\endcsname{0.12}
\expandafter\def\csname BAval@t20.r263.axis\endcsname{0.14}
\expandafter\def\csname BAval@t20.r264.axis\endcsname{0.16}
\expandafter\def\csname BAval@t20.r265.axis\endcsname{0.18}
\expandafter\def\csname BAval@t20.r266.axis\endcsname{0.2}
\expandafter\def\csname BAval@t20.r267.axis\endcsname{0.22}
\expandafter\def\csname BAval@t20.r268.axis\endcsname{0.24}
\expandafter\def\csname BAval@t20.r269.axis\endcsname{0.26}
\expandafter\def\csname BAval@t20.r270.axis\endcsname{0.28}
\expandafter\def\csname BAval@t20.r271.axis\endcsname{0.3}
\expandafter\def\csname BAval@t20.r272.axis\endcsname{0.32}
\expandafter\def\csname BAval@t20.r273.axis\endcsname{0.34}
\expandafter\def\csname BAval@t20.r274.axis\endcsname{0.36}
\expandafter\def\csname BAval@t20.r275.axis\endcsname{0.38}
\expandafter\def\csname BAval@t20.r276.axis\endcsname{0.4}
\expandafter\def\csname BAval@t20.r277.axis\endcsname{0.42}
\expandafter\def\csname BAval@t20.r278.axis\endcsname{0.44}
\expandafter\def\csname BAval@t20.r279.axis\endcsname{0.46}
\expandafter\def\csname BAval@t20.r280.axis\endcsname{0.48}
\expandafter\def\csname BAval@t20.r281.axis\endcsname{0.5}
\expandafter\def\csname BAval@t20.r282.axis\endcsname{0.52}
\expandafter\def\csname BAval@t20.r283.axis\endcsname{0.54}
\expandafter\def\csname BAval@t20.r284.axis\endcsname{0.56}
\expandafter\def\csname BAval@t20.r285.axis\endcsname{0.58}
\expandafter\def\csname BAval@t20.r286.axis\endcsname{0.6}
\expandafter\def\csname BAval@t20.r287.axis\endcsname{0.62}
\expandafter\def\csname BAval@t20.r288.axis\endcsname{0.64}
\expandafter\def\csname BAval@t20.r289.axis\endcsname{0.66}
\expandafter\def\csname BAval@t20.r290.axis\endcsname{0.68}
\expandafter\def\csname BAval@t20.r291.axis\endcsname{0.7}
\expandafter\def\csname BAval@t20.r292.axis\endcsname{0.72}
\expandafter\def\csname BAval@t20.r293.axis\endcsname{0.74}
\expandafter\def\csname BAval@t20.r294.axis\endcsname{0.76}
\expandafter\def\csname BAval@t20.r295.axis\endcsname{0.78}
\expandafter\def\csname BAval@t20.r296.axis\endcsname{0.8}
\expandafter\def\csname BAval@t20.r297.axis\endcsname{0.82}
\expandafter\def\csname BAval@t20.r298.axis\endcsname{0.84}
\expandafter\def\csname BAval@t20.r299.axis\endcsname{0.86}
\expandafter\def\csname BAval@t20.r300.axis\endcsname{0.88}
\expandafter\def\csname BAval@t20.r301.axis\endcsname{0.9}
\expandafter\def\csname BAval@t20.r302.axis\endcsname{0.92}
\expandafter\def\csname BAval@t20.r303.axis\endcsname{0.94}
\expandafter\def\csname BAval@t20.r304.axis\endcsname{0.96}
\expandafter\def\csname BAval@t20.r305.axis\endcsname{0.98}
\expandafter\def\csname BAval@t20.r306.axis\endcsname{1.0}
\expandafter\def\csname BAval@ladder.step.law\endcsname{0.5}
\expandafter\def\csname BAval@ladder.step.id\endcsname{3.3}
\expandafter\def\csname BAval@ladder.step.future\endcsname{1.8}
\expandafter\def\csname BAval@ladder.step.zerodelay\endcsname{6.2}
\expandafter\def\csname BAval@ladder.step.fullview\endcsname{3.9}
\expandafter\def\csname BAval@gap.external.abs\endcsname{12.7}
\expandafter\def\csname BAval@diag.late.full\endcsname{75.8}
\expandafter\def\csname BAval@diag.late.ablated\endcsname{67.7}
\expandafter\def\csname BAval@diag.late.drop\endcsname{8.2}
\expandafter\def\csname BAval@diag.early.full\endcsname{87.0}
\expandafter\def\csname BAval@diag.early.ablated\endcsname{85.8}
\expandafter\def\csname BAval@diag.early.drop\endcsname{1.1}
\expandafter\def\csname BAval@diag.corr.full\endcsname{80.1}
\expandafter\def\csname BAval@diag.corr.ablated\endcsname{73.8}
\expandafter\def\csname BAval@diag.corr.drop\endcsname{6.3}
\expandafter\def\csname BAval@diag.indep.full\endcsname{84.6}
\expandafter\def\csname BAval@diag.indep.ablated\endcsname{84.2}
\expandafter\def\csname BAval@diag.indep.drop\endcsname{0.4}
\expandafter\def\csname BAval@svc.direct.sid\endcsname{78.1}
\expandafter\def\csname BAval@svc.recal.sid\endcsname{81.0}
\expandafter\def\csname BAval@svc.retrain.sid\endcsname{81.7}
\expandafter\def\csname BAval@delta.svc.recal\endcsname{2.9}
\expandafter\def\csname BAval@delta.svc.retrain\endcsname{0.6}
\expandafter\def\csname BAval@lambda.sens.range\endcsname{0.7}
\expandafter\def\csname BAval@t24.r01.lambda.w\endcsname{1}
\expandafter\def\csname BAval@t24.r01.lambda.u\endcsname{1}
\expandafter\def\csname BAval@t24.r02.lambda.w\endcsname{2}
\expandafter\def\csname BAval@t24.r02.lambda.u\endcsname{1}
\expandafter\def\csname BAval@t24.r03.lambda.w\endcsname{4}
\expandafter\def\csname BAval@t24.r03.lambda.u\endcsname{1}
\expandafter\def\csname BAval@t24.r04.lambda.w\endcsname{8}
\expandafter\def\csname BAval@t24.r04.lambda.u\endcsname{1}
\expandafter\def\csname BAval@lambda.w\endcsname{4}
\expandafter\def\csname BAval@lambda.u\endcsname{1}
\expandafter\def\csname BAval@weight.gamma.state\endcsname{1.00}
\expandafter\def\csname BAval@weight.gamma.return\endcsname{0.50}
\expandafter\def\csname BAval@weight.gamma.cost\endcsname{0.10}
\expandafter\def\csname BAval@weight.gamma.rank\endcsname{1.00}
\expandafter\def\csname BAval@weight.gamma.calibration\endcsname{0.20}
\expandafter\def\csname BAval@weight.soft.compute.penalty\endcsname{0.01}
\expandafter\def\csname BAval@weight.soft.motion.penalty\endcsname{0.01}
\expandafter\def\csname BAval@weight.uncertainty.weight\endcsname{0.05}
\expandafter\def\csname BAval@clip.quad.dam.sid\endcsname{1.1}
\expandafter\def\csname BAval@clip.quad.dam.wrong\endcsname{11.1}
\expandafter\def\csname BAval@clip.quad.keep.sid\endcsname{1.1}
\expandafter\def\csname BAval@clip.quad.keep.wrong\endcsname{13.3}
\expandafter\def\csname BAval@clip.dance.dam.sid\endcsname{43.3}
\expandafter\def\csname BAval@clip.dance.dam.wrong\endcsname{11.1}
\expandafter\def\csname BAval@clip.dance.keep.sid\endcsname{40.0}
\expandafter\def\csname BAval@clip.dance.keep.wrong\endcsname{12.8}
\expandafter\def\csname BAval@tick.dance.f272.dam.revision\endcsname{1}
\expandafter\def\csname BAval@tick.dance.f272.dam.sourceage\endcsname{5.70}
\expandafter\def\csname BAval@tick.dance.f272.dam.pubage\endcsname{0.25}
\expandafter\def\csname BAval@tick.dance.f272.dam.id\endcsname{5}
\expandafter\def\csname BAval@tick.dance.f272.dam.maxcompiou\endcsname{0.320}
\expandafter\def\csname BAval@tick.dance.f272.keep.revision\endcsname{0}
\expandafter\def\csname BAval@tick.dance.f272.keep.sourceage\endcsname{5.70}
\expandafter\def\csname BAval@tick.dance.f272.keep.pubage\endcsname{5.70}
\expandafter\def\csname BAval@tick.dance.f272.keep.id\endcsname{5}
\expandafter\def\csname BAval@tick.dance.f272.keep.maxcompiou\endcsname{0.342}
\expandafter\def\csname BAval@tick.dance.f272.bind.revision\endcsname{0}
\expandafter\def\csname BAval@tick.dance.f272.bind.sourceage\endcsname{5.70}
\expandafter\def\csname BAval@tick.dance.f272.bind.pubage\endcsname{5.70}
\expandafter\def\csname BAval@tick.dance.f272.bind.id\endcsname{5}
\expandafter\def\csname BAval@tick.dance.f272.bind.maxcompiou\endcsname{0.342}
\expandafter\def\csname BAval@tick.dance.f279.dam.revision\endcsname{1}
\expandafter\def\csname BAval@tick.dance.f279.dam.sourceage\endcsname{6.05}
\expandafter\def\csname BAval@tick.dance.f279.dam.pubage\endcsname{0.60}
\expandafter\def\csname BAval@tick.dance.f279.dam.id\endcsname{5}
\expandafter\def\csname BAval@tick.dance.f279.dam.maxcompiou\endcsname{0.314}
\expandafter\def\csname BAval@tick.dance.f279.keep.revision\endcsname{1}
\expandafter\def\csname BAval@tick.dance.f279.keep.sourceage\endcsname{6.05}
\expandafter\def\csname BAval@tick.dance.f279.keep.pubage\endcsname{0.02}
\expandafter\def\csname BAval@tick.dance.f279.keep.id\endcsname{5}
\expandafter\def\csname BAval@tick.dance.f279.keep.maxcompiou\endcsname{0.278}
\expandafter\def\csname BAval@tick.dance.f279.bind.revision\endcsname{0}
\expandafter\def\csname BAval@tick.dance.f279.bind.sourceage\endcsname{6.05}
\expandafter\def\csname BAval@tick.dance.f279.bind.pubage\endcsname{6.05}
\expandafter\def\csname BAval@tick.dance.f279.bind.id\endcsname{5}
\expandafter\def\csname BAval@tick.dance.f279.bind.maxcompiou\endcsname{0.337}
\expandafter\def\csname BAval@tick.dance.f280.dam.revision\endcsname{1}
\expandafter\def\csname BAval@tick.dance.f280.dam.sourceage\endcsname{6.10}
\expandafter\def\csname BAval@tick.dance.f280.dam.pubage\endcsname{0.65}
\expandafter\def\csname BAval@tick.dance.f280.dam.id\endcsname{5}
\expandafter\def\csname BAval@tick.dance.f280.dam.maxcompiou\endcsname{0.332}
\expandafter\def\csname BAval@tick.dance.f280.keep.revision\endcsname{1}
\expandafter\def\csname BAval@tick.dance.f280.keep.sourceage\endcsname{6.10}
\expandafter\def\csname BAval@tick.dance.f280.keep.pubage\endcsname{0.07}
\expandafter\def\csname BAval@tick.dance.f280.keep.id\endcsname{5}
\expandafter\def\csname BAval@tick.dance.f280.keep.maxcompiou\endcsname{0.293}
\expandafter\def\csname BAval@tick.dance.f280.bind.revision\endcsname{0}
\expandafter\def\csname BAval@tick.dance.f280.bind.sourceage\endcsname{6.10}
\expandafter\def\csname BAval@tick.dance.f280.bind.pubage\endcsname{6.10}
\expandafter\def\csname BAval@tick.dance.f280.bind.id\endcsname{5}
\expandafter\def\csname BAval@tick.dance.f280.bind.maxcompiou\endcsname{0.355}
\expandafter\def\csname BAval@tick.dance.f281.dam.revision\endcsname{1}
\expandafter\def\csname BAval@tick.dance.f281.dam.sourceage\endcsname{6.15}
\expandafter\def\csname BAval@tick.dance.f281.dam.pubage\endcsname{0.70}
\expandafter\def\csname BAval@tick.dance.f281.dam.id\endcsname{5}
\expandafter\def\csname BAval@tick.dance.f281.dam.maxcompiou\endcsname{0.343}
\expandafter\def\csname BAval@tick.dance.f281.keep.revision\endcsname{1}
\expandafter\def\csname BAval@tick.dance.f281.keep.sourceage\endcsname{6.15}
\expandafter\def\csname BAval@tick.dance.f281.keep.pubage\endcsname{0.12}
\expandafter\def\csname BAval@tick.dance.f281.keep.id\endcsname{5}
\expandafter\def\csname BAval@tick.dance.f281.keep.maxcompiou\endcsname{0.301}
\expandafter\def\csname BAval@tick.dance.f281.bind.revision\endcsname{0}
\expandafter\def\csname BAval@tick.dance.f281.bind.sourceage\endcsname{6.15}
\expandafter\def\csname BAval@tick.dance.f281.bind.pubage\endcsname{6.15}
\expandafter\def\csname BAval@tick.dance.f281.bind.id\endcsname{5}
\expandafter\def\csname BAval@tick.dance.f281.bind.maxcompiou\endcsname{0.366}
\expandafter\def\csname BAval@tick.dance.f282.dam.revision\endcsname{1}
\expandafter\def\csname BAval@tick.dance.f282.dam.sourceage\endcsname{6.20}
\expandafter\def\csname BAval@tick.dance.f282.dam.pubage\endcsname{0.75}
\expandafter\def\csname BAval@tick.dance.f282.dam.id\endcsname{5}
\expandafter\def\csname BAval@tick.dance.f282.dam.maxcompiou\endcsname{0.336}
\expandafter\def\csname BAval@tick.dance.f282.keep.revision\endcsname{1}
\expandafter\def\csname BAval@tick.dance.f282.keep.sourceage\endcsname{6.20}
\expandafter\def\csname BAval@tick.dance.f282.keep.pubage\endcsname{0.17}
\expandafter\def\csname BAval@tick.dance.f282.keep.id\endcsname{5}
\expandafter\def\csname BAval@tick.dance.f282.keep.maxcompiou\endcsname{0.292}
\expandafter\def\csname BAval@tick.dance.f282.bind.revision\endcsname{0}
\expandafter\def\csname BAval@tick.dance.f282.bind.sourceage\endcsname{6.20}
\expandafter\def\csname BAval@tick.dance.f282.bind.pubage\endcsname{6.20}
\expandafter\def\csname BAval@tick.dance.f282.bind.id\endcsname{5}
\expandafter\def\csname BAval@tick.dance.f282.bind.maxcompiou\endcsname{0.358}
\expandafter\def\csname BAval@tick.dance.f283.dam.revision\endcsname{1}
\expandafter\def\csname BAval@tick.dance.f283.dam.sourceage\endcsname{6.25}
\expandafter\def\csname BAval@tick.dance.f283.dam.pubage\endcsname{0.80}
\expandafter\def\csname BAval@tick.dance.f283.dam.id\endcsname{5}
\expandafter\def\csname BAval@tick.dance.f283.dam.maxcompiou\endcsname{0.281}
\expandafter\def\csname BAval@tick.dance.f283.keep.revision\endcsname{1}
\expandafter\def\csname BAval@tick.dance.f283.keep.sourceage\endcsname{6.25}
\expandafter\def\csname BAval@tick.dance.f283.keep.pubage\endcsname{0.22}
\expandafter\def\csname BAval@tick.dance.f283.keep.id\endcsname{5}
\expandafter\def\csname BAval@tick.dance.f283.keep.maxcompiou\endcsname{0.226}
\expandafter\def\csname BAval@tick.dance.f283.bind.revision\endcsname{0}
\expandafter\def\csname BAval@tick.dance.f283.bind.sourceage\endcsname{6.25}
\expandafter\def\csname BAval@tick.dance.f283.bind.pubage\endcsname{6.25}
\expandafter\def\csname BAval@tick.dance.f283.bind.id\endcsname{5}
\expandafter\def\csname BAval@tick.dance.f283.bind.maxcompiou\endcsname{0.297}
\expandafter\def\csname BAval@tick.dance.f284.dam.revision\endcsname{1}
\expandafter\def\csname BAval@tick.dance.f284.dam.sourceage\endcsname{6.30}
\expandafter\def\csname BAval@tick.dance.f284.dam.pubage\endcsname{0.85}
\expandafter\def\csname BAval@tick.dance.f284.dam.id\endcsname{5}
\expandafter\def\csname BAval@tick.dance.f284.dam.maxcompiou\endcsname{0.298}
\expandafter\def\csname BAval@tick.dance.f284.keep.revision\endcsname{1}
\expandafter\def\csname BAval@tick.dance.f284.keep.sourceage\endcsname{6.30}
\expandafter\def\csname BAval@tick.dance.f284.keep.pubage\endcsname{0.27}
\expandafter\def\csname BAval@tick.dance.f284.keep.id\endcsname{5}
\expandafter\def\csname BAval@tick.dance.f284.keep.maxcompiou\endcsname{0.242}
\expandafter\def\csname BAval@tick.dance.f284.bind.revision\endcsname{0}
\expandafter\def\csname BAval@tick.dance.f284.bind.sourceage\endcsname{6.30}
\expandafter\def\csname BAval@tick.dance.f284.bind.pubage\endcsname{6.30}
\expandafter\def\csname BAval@tick.dance.f284.bind.id\endcsname{5}
\expandafter\def\csname BAval@tick.dance.f284.bind.maxcompiou\endcsname{0.314}
\expandafter\def\csname BAval@tick.dance.f296.dam.revision\endcsname{1}
\expandafter\def\csname BAval@tick.dance.f296.dam.sourceage\endcsname{6.90}
\expandafter\def\csname BAval@tick.dance.f296.dam.pubage\endcsname{1.45}
\expandafter\def\csname BAval@tick.dance.f296.dam.id\endcsname{5}
\expandafter\def\csname BAval@tick.dance.f296.dam.competitor\endcsname{6}
\expandafter\def\csname BAval@tick.dance.f296.dam.compiou\endcsname{0.822}
\expandafter\def\csname BAval@tick.dance.f296.dam.maxcompiou\endcsname{0.822}
\expandafter\def\csname BAval@tick.dance.f296.keep.revision\endcsname{1}
\expandafter\def\csname BAval@tick.dance.f296.keep.sourceage\endcsname{6.90}
\expandafter\def\csname BAval@tick.dance.f296.keep.pubage\endcsname{0.87}
\expandafter\def\csname BAval@tick.dance.f296.keep.id\endcsname{5}
\expandafter\def\csname BAval@tick.dance.f296.keep.competitor\endcsname{6}
\expandafter\def\csname BAval@tick.dance.f296.keep.compiou\endcsname{0.792}
\expandafter\def\csname BAval@tick.dance.f296.keep.maxcompiou\endcsname{0.792}
\expandafter\def\csname BAval@tick.dance.f296.bind.revision\endcsname{0}
\expandafter\def\csname BAval@tick.dance.f296.bind.sourceage\endcsname{6.90}
\expandafter\def\csname BAval@tick.dance.f296.bind.pubage\endcsname{6.90}
\expandafter\def\csname BAval@tick.dance.f296.bind.id\endcsname{5}
\expandafter\def\csname BAval@tick.dance.f296.bind.competitor\endcsname{6}
\expandafter\def\csname BAval@tick.dance.f296.bind.compiou\endcsname{0.786}
\expandafter\def\csname BAval@tick.dance.f296.bind.maxcompiou\endcsname{0.786}
\expandafter\def\csname BAval@tick.dance.f320.dam.revision\endcsname{1}
\expandafter\def\csname BAval@tick.dance.f320.dam.sourceage\endcsname{8.10}
\expandafter\def\csname BAval@tick.dance.f320.dam.pubage\endcsname{2.65}
\expandafter\def\csname BAval@tick.dance.f320.dam.id\endcsname{5}
\expandafter\def\csname BAval@tick.dance.f320.dam.maxcompiou\endcsname{0.474}
\expandafter\def\csname BAval@tick.dance.f320.keep.revision\endcsname{1}
\expandafter\def\csname BAval@tick.dance.f320.keep.sourceage\endcsname{8.10}
\expandafter\def\csname BAval@tick.dance.f320.keep.pubage\endcsname{2.07}
\expandafter\def\csname BAval@tick.dance.f320.keep.id\endcsname{5}
\expandafter\def\csname BAval@tick.dance.f320.keep.competitor\endcsname{4}
\expandafter\def\csname BAval@tick.dance.f320.keep.compiou\endcsname{0.502}
\expandafter\def\csname BAval@tick.dance.f320.keep.maxcompiou\endcsname{0.502}
\expandafter\def\csname BAval@tick.dance.f320.bind.revision\endcsname{0}
\expandafter\def\csname BAval@tick.dance.f320.bind.sourceage\endcsname{8.10}
\expandafter\def\csname BAval@tick.dance.f320.bind.pubage\endcsname{8.10}
\expandafter\def\csname BAval@tick.dance.f320.bind.id\endcsname{5}
\expandafter\def\csname BAval@tick.dance.f320.bind.maxcompiou\endcsname{0.493}
\expandafter\def\csname BAval@tick.quad.f465.dam.revision\endcsname{1}
\expandafter\def\csname BAval@tick.quad.f465.dam.sourceage\endcsname{5.70}
\expandafter\def\csname BAval@tick.quad.f465.dam.pubage\endcsname{2.25}
\expandafter\def\csname BAval@tick.quad.f465.dam.id\endcsname{59}
\expandafter\def\csname BAval@tick.quad.f465.dam.maxcompiou\endcsname{0.192}
\expandafter\def\csname BAval@tick.quad.f465.keep.revision\endcsname{1}
\expandafter\def\csname BAval@tick.quad.f465.keep.sourceage\endcsname{5.70}
\expandafter\def\csname BAval@tick.quad.f465.keep.pubage\endcsname{0.94}
\expandafter\def\csname BAval@tick.quad.f465.keep.id\endcsname{59}
\expandafter\def\csname BAval@tick.quad.f465.keep.maxcompiou\endcsname{0.232}
\expandafter\def\csname BAval@tick.quad.f465.bind.revision\endcsname{0}
\expandafter\def\csname BAval@tick.quad.f465.bind.sourceage\endcsname{5.70}
\expandafter\def\csname BAval@tick.quad.f465.bind.pubage\endcsname{5.70}
\expandafter\def\csname BAval@tick.quad.f465.bind.id\endcsname{59}
\expandafter\def\csname BAval@tick.quad.f465.bind.maxcompiou\endcsname{0.225}
\expandafter\def\csname BAval@tick.quad.f477.dam.revision\endcsname{1}
\expandafter\def\csname BAval@tick.quad.f477.dam.sourceage\endcsname{6.90}
\expandafter\def\csname BAval@tick.quad.f477.dam.pubage\endcsname{3.45}
\expandafter\def\csname BAval@tick.quad.f477.dam.id\endcsname{59}
\expandafter\def\csname BAval@tick.quad.f477.dam.maxcompiou\endcsname{0.007}
\expandafter\def\csname BAval@tick.quad.f477.keep.revision\endcsname{1}
\expandafter\def\csname BAval@tick.quad.f477.keep.sourceage\endcsname{6.90}
\expandafter\def\csname BAval@tick.quad.f477.keep.pubage\endcsname{2.14}
\expandafter\def\csname BAval@tick.quad.f477.keep.id\endcsname{59}
\expandafter\def\csname BAval@tick.quad.f477.keep.maxcompiou\endcsname{0.030}
\expandafter\def\csname BAval@tick.quad.f477.bind.revision\endcsname{0}
\expandafter\def\csname BAval@tick.quad.f477.bind.sourceage\endcsname{6.90}
\expandafter\def\csname BAval@tick.quad.f477.bind.pubage\endcsname{6.90}
\expandafter\def\csname BAval@tick.quad.f477.bind.id\endcsname{59}
\expandafter\def\csname BAval@tick.quad.f477.bind.maxcompiou\endcsname{0.027}
\expandafter\def\csname BAval@tick.quad.f489.dam.revision\endcsname{2}
\expandafter\def\csname BAval@tick.quad.f489.dam.sourceage\endcsname{4.69}
\expandafter\def\csname BAval@tick.quad.f489.dam.pubage\endcsname{0.44}
\expandafter\def\csname BAval@tick.quad.f489.dam.id\endcsname{59}
\expandafter\def\csname BAval@tick.quad.f489.dam.maxcompiou\endcsname{0.363}
\expandafter\def\csname BAval@tick.quad.f489.keep.revision\endcsname{1}
\expandafter\def\csname BAval@tick.quad.f489.keep.sourceage\endcsname{8.10}
\expandafter\def\csname BAval@tick.quad.f489.keep.pubage\endcsname{3.34}
\expandafter\def\csname BAval@tick.quad.f489.keep.id\endcsname{59}
\expandafter\def\csname BAval@tick.quad.f489.keep.competitor\endcsname{101}
\expandafter\def\csname BAval@tick.quad.f489.keep.compiou\endcsname{0.547}
\expandafter\def\csname BAval@tick.quad.f489.keep.maxcompiou\endcsname{0.547}
\expandafter\def\csname BAval@tick.quad.f489.bind.revision\endcsname{0}
\expandafter\def\csname BAval@tick.quad.f489.bind.sourceage\endcsname{8.10}
\expandafter\def\csname BAval@tick.quad.f489.bind.pubage\endcsname{8.10}
\expandafter\def\csname BAval@tick.quad.f489.bind.id\endcsname{59}
\expandafter\def\csname BAval@tick.quad.f489.bind.competitor\endcsname{101}
\expandafter\def\csname BAval@tick.quad.f489.bind.compiou\endcsname{0.531}
\expandafter\def\csname BAval@tick.quad.f489.bind.maxcompiou\endcsname{0.531}
\expandafter\def\csname BAval@goal.B.f100.x\endcsname{647}
\expandafter\def\csname BAval@goal.B.f100.y\endcsname{438}
\expandafter\def\csname BAval@goal.B.f100.width\endcsname{100}
\expandafter\def\csname BAval@goal.B.f100.height\endcsname{305}
\expandafter\def\csname BAval@goal.B.f120.x\endcsname{676}
\expandafter\def\csname BAval@goal.B.f120.y\endcsname{438}
\expandafter\def\csname BAval@goal.B.f120.width\endcsname{97}
\expandafter\def\csname BAval@goal.B.f120.height\endcsname{318}
\expandafter\def\csname BAval@goal.B.f140.x\endcsname{693}
\expandafter\def\csname BAval@goal.B.f140.y\endcsname{434}
\expandafter\def\csname BAval@goal.B.f140.width\endcsname{109}
\expandafter\def\csname BAval@goal.B.f140.height\endcsname{328}
\expandafter\def\csname BAval@goal.D.f240.x\endcsname{348}
\expandafter\def\csname BAval@goal.D.f240.y\endcsname{202}
\expandafter\def\csname BAval@goal.D.f240.width\endcsname{50}
\expandafter\def\csname BAval@goal.D.f240.height\endcsname{165}
\expandafter\def\csname BAval@goal.D.f272.x\endcsname{324}
\expandafter\def\csname BAval@goal.D.f272.y\endcsname{295}
\expandafter\def\csname BAval@goal.D.f272.width\endcsname{89}
\expandafter\def\csname BAval@goal.D.f272.height\endcsname{242}
\expandafter\def\csname BAval@goal.D.f300.x\endcsname{634}
\expandafter\def\csname BAval@goal.D.f300.y\endcsname{228}
\expandafter\def\csname BAval@goal.D.f300.width\endcsname{69}
\expandafter\def\csname BAval@goal.D.f300.height\endcsname{212}

%% file: numbers_v38.tex
\expandafter\def\csname BAval@sd.ext.advat\endcsname{0.3}
\expandafter\def\csname BAval@sd.ext.oavat\endcsname{0.2}
\expandafter\def\csname BAval@sd.ext.pvt\endcsname{0.4}
\expandafter\def\csname BAval@sd.ext.artrack\endcsname{0.3}
\expandafter\def\csname BAval@sd.ext.chanakya\endcsname{0.6}
\expandafter\def\csname BAval@sd.ext.joint\endcsname{0.8}
\expandafter\def\csname BAval@sd.ext.best\endcsname{0.2}
\expandafter\def\csname BAval@sd.ext.bind\endcsname{0.3}
\expandafter\def\csname BAval@sd.t03.r01\endcsname{0.6}
\expandafter\def\csname BAval@sd.t03.r02\endcsname{0.9}
\expandafter\def\csname BAval@sd.t03.r03\endcsname{0.3}
\expandafter\def\csname BAval@sd.t03.r04\endcsname{0.3}
\expandafter\def\csname BAval@t03.r01.sidx\endcsname{65.9}
\expandafter\def\csname BAval@t03.r02.sidx\endcsname{70.3}
\expandafter\def\csname BAval@t03.r03.sidx\endcsname{77.6}
\expandafter\def\csname BAval@t03.r04.sidx\endcsname{82.3}
\expandafter\def\csname BAval@sd.t04.r01\endcsname{0.3}
\expandafter\def\csname BAval@sd.t04.r02\endcsname{0.2}
\expandafter\def\csname BAval@sd.t04.r03\endcsname{0.2}
\expandafter\def\csname BAval@sd.t04.r04\endcsname{0.9}
\expandafter\def\csname BAval@sd.t04.r05\endcsname{1.1}
\expandafter\def\csname BAval@sd.t04.r06\endcsname{0.3}
\expandafter\def\csname BAval@sd.t04.r07\endcsname{0.3}
\expandafter\def\csname BAval@abl.unc.sid\endcsname{80.5}
\expandafter\def\csname BAval@abl.unc.wrong\endcsname{5.8}
\expandafter\def\csname BAval@abl.unc.t5\endcsname{1.5}
\expandafter\def\csname BAval@abl.unc.drop\endcsname{1.8}
\expandafter\def\csname BAval@sd.abl.unc\endcsname{0.1}
\expandafter\def\csname BAval@abl.imm.sid\endcsname{76.2}
\expandafter\def\csname BAval@abl.imm.wrong\endcsname{7.1}
\expandafter\def\csname BAval@abl.imm.t5\endcsname{1.7}
\expandafter\def\csname BAval@abl.imm.drop\endcsname{6.1}
\expandafter\def\csname BAval@rule.prox.sid\endcsname{73.6}
\expandafter\def\csname BAval@rule.prox.wrong\endcsname{8.0}
\expandafter\def\csname BAval@rule.prox.t5\endcsname{1.8}
\expandafter\def\csname BAval@rule.prox.drop\endcsname{8.7}
\expandafter\def\csname BAval@rule.prox.k\endcsname{1.5}
\expandafter\def\csname BAval@rule.rate.sid\endcsname{71.4}
\expandafter\def\csname BAval@rule.rate.wrong\endcsname{8.8}
\expandafter\def\csname BAval@rule.rate.t5\endcsname{1.6}
\expandafter\def\csname BAval@rule.rate.drop\endcsname{10.9}
\expandafter\def\csname BAval@rule.rate.period\endcsname{0.6}
\expandafter\def\csname BAval@rule.prox.long\endcsname{61.4}
\expandafter\def\csname BAval@rule.rate.long\endcsname{58.9}
\expandafter\def\csname BAval@ci.lo\endcsname{5.1}
\expandafter\def\csname BAval@ci.hi\endcsname{9.0}
\expandafter\def\csname BAval@ci.groups\endcsname{40}
\expandafter\def\csname BAval@ci.resamples\endcsname{10{,}000}
\expandafter\def\csname BAval@idf1.best\endcsname{77.5}
\expandafter\def\csname BAval@idf1.bind\endcsname{83.5}
\expandafter\def\csname BAval@hota.best\endcsname{61.8}
\expandafter\def\csname BAval@hota.bind\endcsname{67.3}
\expandafter\def\csname BAval@tune.configs\endcsname{24}
\expandafter\def\csname BAval@tune.seeds\endcsname{3}
\expandafter\def\csname BAval@tune.groups\endcsname{20}
\expandafter\def\csname BAval@retsweep.episodes\endcsname{243}
\expandafter\def\csname BAval@retsweep.early\endcsname{94.2}
\expandafter\def\csname BAval@retsweep.zero\endcsname{80.7}
\expandafter\def\csname BAval@retsweep.late\endcsname{56.8}
\expandafter\def\csname BAval@attr.dam.total\endcsname{137}
\expandafter\def\csname BAval@attr.dam.after\endcsname{89}
\expandafter\def\csname BAval@attr.dam.pct\endcsname{65.0}
\expandafter\def\csname BAval@attr.keep.total\endcsname{151}
\expandafter\def\csname BAval@attr.keep.after\endcsname{87}
\expandafter\def\csname BAval@attr.keep.pct\endcsname{57.6}
\expandafter\def\csname BAval@attr.best.total\endcsname{97}
\expandafter\def\csname BAval@attr.best.after\endcsname{61}
\expandafter\def\csname BAval@attr.best.pct\endcsname{62.9}
\expandafter\def\csname BAval@attr.bind.total\endcsname{53}
\expandafter\def\csname BAval@attr.bind.after\endcsname{18}
\expandafter\def\csname BAval@attr.bind.pct\endcsname{34.0}
\expandafter\def\csname BAval@plan.mean\endcsname{2.1}
\expandafter\def\csname BAval@plan.p95\endcsname{3.8}
\expandafter\def\csname BAval@plan.decisions\endcsname{10{,}000}
\expandafter\def\csname BAval@plan.limit\endcsname{10}
\expandafter\def\csname BAval@cost.mixformer\endcsname{19.8}
\expandafter\def\csname BAval@cost.hiptrack\endcsname{27.5}
\expandafter\def\csname BAval@cost.odtrack\endcsname{22.2}
\expandafter\def\csname BAval@cost.ltmu\endcsname{31.8}
\expandafter\def\csname BAval@cost.keeptrack\endcsname{29.1}
\expandafter\def\csname BAval@cost.advat\endcsname{18.6}
\expandafter\def\csname BAval@cost.joint\endcsname{38.1}
\expandafter\def\csname BAval@sens.h1\endcsname{78.6}
\expandafter\def\csname BAval@sens.h15\endcsname{80.9}
\expandafter\def\csname BAval@sens.h2\endcsname{82.3}
\expandafter\def\csname BAval@sens.h3\endcsname{81.9}
\expandafter\def\csname BAval@sens.h1.ms\endcsname{1.38}
\expandafter\def\csname BAval@sens.h15.ms\endcsname{1.72}
\expandafter\def\csname BAval@sens.h2.ms\endcsname{2.09}
\expandafter\def\csname BAval@sens.h3.ms\endcsname{2.73}
\expandafter\def\csname BAval@sens.b4\endcsname{79.8}
\expandafter\def\csname BAval@sens.b8\endcsname{82.3}
\expandafter\def\csname BAval@sens.b16\endcsname{82.4}
\expandafter\def\csname BAval@sens.b4.ms\endcsname{1.42}
\expandafter\def\csname BAval@sens.b8.ms\endcsname{2.09}
\expandafter\def\csname BAval@sens.b16.ms\endcsname{3.61}
\expandafter\def\csname BAval@sens.d2\endcsname{79.9}
\expandafter\def\csname BAval@sens.d3\endcsname{82.3}
\expandafter\def\csname BAval@sens.d4\endcsname{82.4}
\expandafter\def\csname BAval@sens.d2.ms\endcsname{1.26}
\expandafter\def\csname BAval@sens.d3.ms\endcsname{2.09}
\expandafter\def\csname BAval@sens.d4.ms\endcsname{3.84}
\expandafter\def\csname BAval@t27.r01.correct.s\endcsname{7.40}
\expandafter\def\csname BAval@t27.r01.wrong.s\endcsname{0.21}
\expandafter\def\csname BAval@t27.r01.sid\endcsname{82.2}
\expandafter\def\csname BAval@t27.r01.wrong\endcsname{2.4}
\expandafter\def\csname BAval@t27.r02.correct.s\endcsname{7.07}
\expandafter\def\csname BAval@t27.r02.wrong.s\endcsname{0.35}
\expandafter\def\csname BAval@t27.r02.sid\endcsname{78.5}
\expandafter\def\csname BAval@t27.r02.wrong\endcsname{3.9}
\expandafter\def\csname BAval@sens.h.drop\endcsname{3.6}
\expandafter\def\csname BAval@sens.search.gain\endcsname{0.1}

%% file: numbers_v41.tex
\expandafter\def\csname BAval@svc.retrain.sid\endcsname{81.6}
\expandafter\def\csname BAval@delta.svc.gap\endcsname{3.5}

%% file: numbers_update_20261117.tex
\expandafter\def\csname BAval@t17.r01.dam4sam.chanakya.j\endcsname{80.12}
\expandafter\def\csname BAval@t17.r01.bindadapt\endcsname{84.22}
\expandafter\def\csname BAval@t17.r02.dam4sam.chanakya.j\endcsname{79.36}
\expandafter\def\csname BAval@t17.r02.bindadapt\endcsname{84.09}
\expandafter\def\csname BAval@t17.r03.dam4sam.chanakya.j\endcsname{77.84}
\expandafter\def\csname BAval@t17.r03.bindadapt\endcsname{83.46}
\expandafter\def\csname BAval@t17.r04.dam4sam.chanakya.j\endcsname{75.20}
\expandafter\def\csname BAval@t17.r04.bindadapt\endcsname{82.30}
\expandafter\def\csname BAval@t17.r05.dam4sam.chanakya.j\endcsname{68.46}
\expandafter\def\csname BAval@t17.r05.bindadapt\endcsname{78.13}
\expandafter\def\csname BAval@t17.r06.dam4sam.chanakya.j\endcsname{54.82}
\expandafter\def\csname BAval@t17.r06.bindadapt\endcsname{67.02}
\expandafter\def\csname BAval@delta.sweep.zero\endcsname{4.10}
\expandafter\def\csname BAval@delta.sweep.ref\endcsname{7.10}
\expandafter\def\csname BAval@delta.sweep.two\endcsname{12.20}
\expandafter\def\csname BAval@svc.direct.sid\endcsname{81.34}
\expandafter\def\csname BAval@svc.recal.sid\endcsname{83.62}
\expandafter\def\csname BAval@svc.retrain.sid\endcsname{84.03}
\expandafter\def\csname BAval@delta.svc.recal\endcsname{2.28}
\expandafter\def\csname BAval@delta.svc.retrain\endcsname{0.41}
\expandafter\def\csname BAval@delta.svc.gap\endcsname{2.69}

%% file: numbers_v44.tex
\expandafter\def\csname BAval@delta.sweep.zero\endcsname{4.1}
\expandafter\def\csname BAval@delta.sweep.ref\endcsname{7.1}
\expandafter\def\csname BAval@delta.sweep.two\endcsname{12.2}
\expandafter\def\csname BAval@svc.direct.sid\endcsname{81.3}
\expandafter\def\csname BAval@svc.recal.sid\endcsname{83.6}
\expandafter\def\csname BAval@svc.retrain.sid\endcsname{84.0}
\expandafter\def\csname BAval@delta.svc.recal\endcsname{2.3}
\expandafter\def\csname BAval@delta.svc.retrain\endcsname{0.4}
\expandafter\def\csname BAval@delta.svc.gap\endcsname{2.7}

\expandafter\def\csname BAval@tab.delta.ms\endcsname{0.4}

%% file: unsure_flags.tex
\expandafter\def\csname BAflag@int.joint.recovery\endcsname{}
\expandafter\def\csname BAflag@int.joint.t5\endcsname{}
\expandafter\def\csname BAflag@int.oracle.recovery\endcsname{}
\expandafter\def\csname BAflag@int.oracle.t5\endcsname{}
\expandafter\def\csname BAflag@int.oracle.wrong\endcsname{}
\expandafter\def\csname BAflag@t01.r01.recovery\endcsname{}
\expandafter\def\csname BAflag@t01.r01.sid\endcsname{}
\expandafter\def\csname BAflag@t01.r01.t5\endcsname{}
\expandafter\def\csname BAflag@t01.r01.wrong\endcsname{}
\expandafter\def\csname BAflag@t01.r02.recovery\endcsname{}
\expandafter\def\csname BAflag@t01.r02.sid\endcsname{}
\expandafter\def\csname BAflag@t01.r02.t5\endcsname{}
\expandafter\def\csname BAflag@t01.r02.wrong\endcsname{}
\expandafter\def\csname BAflag@t01.r03.recovery\endcsname{}
\expandafter\def\csname BAflag@t01.r03.sid\endcsname{}
\expandafter\def\csname BAflag@t01.r03.t5\endcsname{}
\expandafter\def\csname BAflag@t01.r03.wrong\endcsname{}
\expandafter\def\csname BAflag@t01.r04.recovery\endcsname{}
\expandafter\def\csname BAflag@t01.r04.sid\endcsname{}
\expandafter\def\csname BAflag@t01.r04.t5\endcsname{}
\expandafter\def\csname BAflag@t01.r04.wrong\endcsname{}
\expandafter\def\csname BAflag@t01.r05.recovery\endcsname{}
\expandafter\def\csname BAflag@t01.r05.sid\endcsname{}
\expandafter\def\csname BAflag@t01.r05.t5\endcsname{}
\expandafter\def\csname BAflag@t01.r05.wrong\endcsname{}
\expandafter\def\csname BAflag@t01.r07.recovery\endcsname{}
\expandafter\def\csname BAflag@t01.r07.t5\endcsname{}
\expandafter\def\csname BAflag@t01.r09.recovery\endcsname{}
\expandafter\def\csname BAflag@t01.r09.t5\endcsname{}
\expandafter\def\csname BAflag@t01.r09.wrong\endcsname{}
\expandafter\def\csname BAflag@t02.r01.long\endcsname{}
\expandafter\def\csname BAflag@t02.r01.short\endcsname{}
\expandafter\def\csname BAflag@t02.r01.similar\endcsname{}
\expandafter\def\csname BAflag@t02.r02.long\endcsname{}
\expandafter\def\csname BAflag@t02.r02.short\endcsname{}
\expandafter\def\csname BAflag@t02.r02.similar\endcsname{}
\expandafter\def\csname BAflag@t02.r03.long\endcsname{}
\expandafter\def\csname BAflag@t02.r03.short\endcsname{}
\expandafter\def\csname BAflag@t02.r03.similar\endcsname{}
\expandafter\def\csname BAflag@t02.r04.long\endcsname{}
\expandafter\def\csname BAflag@t02.r04.short\endcsname{}
\expandafter\def\csname BAflag@t02.r04.similar\endcsname{}
\expandafter\def\csname BAflag@t02.r06.long\endcsname{}
\expandafter\def\csname BAflag@t02.r06.short\endcsname{}
\expandafter\def\csname BAflag@t02.r06.similar\endcsname{}

%% file: sec/0_abstract.tex
\begin{abstract}
\begingroup\parfillskip=0pt
Pan--tilt--zoom (PTZ) tracking can maintain a confident bounding box while following the wrong person. Recovering the designated identity depends on when verification results become available, because tracker memory continues to change while computation is pending. In controlled crossings, delaying the same identity check from before to after a crossing reduces two-second identity retention from 94.2\% to 56.8\%. We propose \textbf{BindAdapt}, which jointly schedules camera motion, visual operators, and operator start times to preserve target identity under computation and motion budgets. A provenance graph records the shared origins of observations and tracker-memory cues. An identity-response model, trained on paired rendered branches from the same tracker state, predicts result return times and subsequent identity-binding trajectories. A receding-horizon scheduler uses these predictions to select timed actions that minimize expected identity loss. Across 480 PTZ episodes rendered from panoramas, BindAdapt increases correct-target tracking time from 75.2\% to 82.3\% and reduces wrong binding from 7.3\% to 4.4\% over DAM4SAM with Chanakya-J at matched service time. Gains persist across three fixed tracking backbones and without artificial delay. On a real PTZ camera, identity success improves from 74.8\% with Chanakya-J to 80.3\%.\par\endgroup

\end{abstract}

%% file: sec/1_introduction.tex
\section{Introduction}
\begin{figure}[t]
\centering\includegraphics[width=\linewidth]{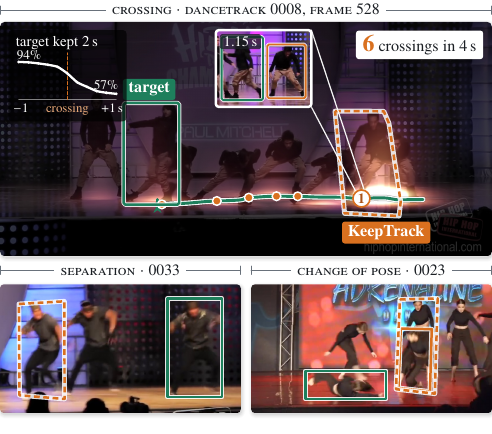}
\caption{\textbf{A confident box can follow the wrong person.} Top: DanceTrack 0008 at frame 528, 4\,s after frame 448. The target and its path are green, and KeepTrack, in orange, has moved to a look-alike. Dots mark the six crossings, the first at 1.15\,s, and the brightness under each person is its color-histogram similarity to the target at frame 448. Inset: share of \BAval{retsweep.episodes} crossing episodes in which one fixed identity check keeps the target for 2\,s, as its result returns from 1\,s before to 1\,s after the crossing. Bottom: the same failure after two dancers separate and after the target drops to the floor, with DAM4SAM solid and KeepTrack dashed. DanceTrack is fixed-camera footage, used here to show identity drift.}
\label{fig:motivation}
\end{figure}

A pan--tilt--zoom (PTZ) tracker receives a box around a designated person in its first view and outputs the box of that person in every later frame, while it steers the camera that produces those frames. It pairs a perception tracker, which localizes the person in each view, with control that decides where the camera points and which visual operator runs.

\begin{figure*}[t]
\centering\includegraphics[width=\textwidth]{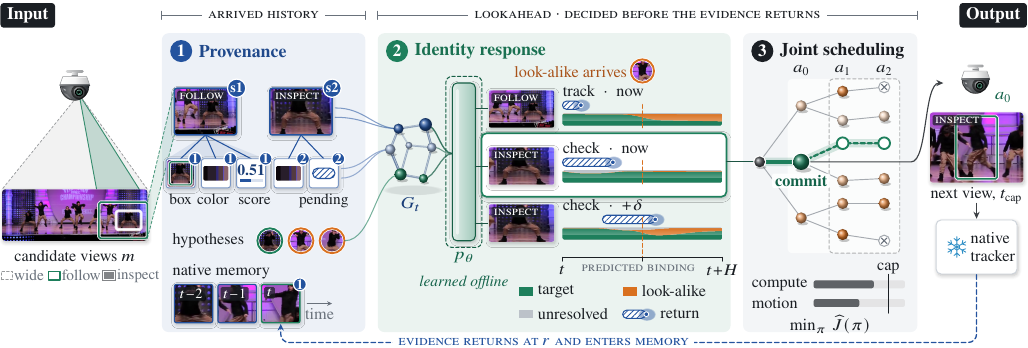}
\caption{\textbf{Overview of BindAdapt.} (1)~Each view and every quantity computed from it share a source identifier and, with native memory and identity hypotheses, form the provenance graph $G_t$. (2)~For each candidate view, operator, and start time, the identity response $p_\theta$, learned offline from common-prefix branches, predicts when the result returns and how the binding to the target, a look-alike, or neither evolves. (3)~The joint scheduler expands a beam of timed sequences under the caps and commits only $a_0$, captured at $t_{\mathrm{cap}}$ and entering native memory at its return time $r$. Photos: DanceTrack 0008, frames 448--528. Ribbons, return times, and bars are illustrative.}
\label{fig:method}
\end{figure*}

Work on PTZ tracking has advanced on both parts. Perception trackers have gained strong representations~\cite{dimp,stark,ostrack,mixformer,hiptrack,odtrack,loratv2}, and distractor-aware memory and long-term re-detection reduce drift and loss~\cite{keeptrack,dasiam,dam4sam,ltmu}. On the control side, active trackers learn camera behavior from observations~\cite{active,advat,oavat}, and runtime controllers decide how much computation each frame receives~\cite{chanakya,artrackac,avtrack}. Both directions model the designated target through templates, distractor memory, or instance prototypes, but mainly judge their output by localization, through overlap, tracking reward, or latency-aware accuracy, and rarely ask whose box it is over time. Yet a box can overlap a person well and still belong to the wrong person. Figure~\ref{fig:motivation} follows the target and KeepTrack~\cite{keeptrack} from frame 448 of a DanceTrack sequence~\cite{dancetrack}. The target crosses six dancers, and by frame 528 KeepTrack has moved to a look-alike while its box stays stable and confident. Across \BAval{t09.r03.episodes} test episodes under a common PTZ controller, the eight perception trackers of Table~\ref{tab:external} follow a wrong person for \BAval{t12.r08.wrong}--\BAval{t12.r01.wrong}\% of the time.

To understand why these failures persist, we ask when the evidence that could prevent them reaches the tracker. An identity check can fail because its result is wrong, or because a correct result arrives too late. Stronger representations and re-detection address the first case, so we examine the second. While the check runs, the target may leave the view, the tracker may write the look-alike into memory, or a faster request may return first. The slow result then describes an earlier view, yet it is inserted into the current tracker state. In a controlled sweep that fixes one identity check and its input frame, captured before the earliest return, and moves only the release of its result, the target is kept for two seconds in \BAval{retsweep.early}\% of \BAval{retsweep.episodes} crossing episodes when the result returns one second before the crossing, but in only \BAval{retsweep.late}\% when it returns one second after, as the inset of Figure~\ref{fig:motivation} shows. Likewise, \BAval{attr.dam.after} of the \BAval{attr.dam.total} wrong bindings of DAM4SAM on the test episodes follow a crossing whose latest evidence returned after it. A result thus acts on the tracker memory that exists at its return.
This finding explains why strengthening individual modules falls short. A more accurate tracker cannot choose the view it receives, and a faster one still spends its budget on frames whose results return after native memory flips. Active camera policies such as OA-VAT~\cite{oavat} plan trajectories for occlusion recovery while toggling detection by fixed confidence rules, whereas runtime controllers such as Chanakya~\cite{chanakya} allocate operators over an input stream they cannot steer. Keeping the target instead forces camera motion, visual operator, and start time to be chosen together, by how returned evidence will alter native memory.
To this end, we propose \textbf{BindAdapt}, which treats a camera trajectory, a visual operator, and its start time as one timed action and judges candidate plans by how long the target stays correctly bound on the physical clock, while the tracker's native memory and update rule stay intact. Figure~\ref{fig:method} traces its three components from arrived history to commitment. First, because a detector box, a feature crop, and a memory update from one frame are correlated, a provenance graph links every feature, hypothesis, and native-memory entry to its source frame, so the response can recognize repeated reads of one view instead of treating them as independent evidence. Second, because a recorded run never shows what an untaken view would have written into memory, an identity response learned from common-prefix branches, restarted from identical snapshots of camera, queue, and memory, predicts when each candidate action returns, what it costs, and how correct, wrong, and unresolved binding evolve. Third, because camera settling and operator execution compete for one return window and budget, a joint scheduler searches timed sequences on these predictions and commits the best first action.

Our contributions are as follows:
\begin{itemize}
\setlength{\itemsep}{0pt}\setlength{\parskip}{0pt}
\item We show by controlled interventions that the identity a PTZ tracker keeps depends on \textbf{when a result returns} relative to memory changes, not only on whether it is correct, and we formulate PTZ identity tracking as a joint choice of camera motion, operator, and start time.
\item We propose \textbf{BINDADAPT}, which predicts the identity response of each timed action from a provenance graph, learns it from common-prefix branches, and schedules camera and computation jointly.
\item Under shared budgets, BindAdapt raises the share of time on the original target by \textbf{\BAval{delta.ext.sid} points} over the strongest of 15 compared trackers, controllers, schedulers, and combinations, at matched service time. It gains \BAval{delta.backbone.dam}--\BAval{delta.backbone.ost} points over Chanakya-J with each of three backbones held fixed, \BAval{delta.sweep.zero} points without added delay, and about 5.5 points on a real PTZ camera, and every ablation lowers this share.
\end{itemize}

%% file: sec/2_related.tex
\section{Related Work}
\paragraph{Visual object tracking.} Discriminative trackers learn to separate the target from nearby background~\cite{atom,dimp,tomp}. Transformer trackers combine matching and localization through spatiotemporal interaction~\cite{stark,ostrack,mixformer}, and autoregressive models cast tracking as coordinate or sequence prediction~\cite{artrack,seqtrack}. Later work conditions the representation on preceding observations~\cite{hiptrack,odtrack}, reduces its cost~\cite{mixformerv2}, and adapts large pretrained backbones such as DINOv2~\cite{dinov2} to tracking~\cite{lorat,loratv2}. To keep the same person over time, distractor-aware models, candidate association, and template updating address different sources of drift~\cite{dasiam,keeptrack,updatenet}, and long-term trackers add verification and re-detection~\cite{ltmu,siamrcnn,ltmetrics}. PTAV runs a verifier in parallel with the tracker and feeds its corrections back asynchronously~\cite{ptav}, and we include such asynchronous verification as a control. Memory-based segmentation propagates a designated object through video~\cite{stm,rmem,sam2,xmem,cutie}, SAM2Long maintains alternative memory trajectories~\cite{sam2long}, and DAM4SAM protects its memory against distractors~\cite{dam4sam}. All of these trackers, however, process an exogenous frame stream under fixed execution rules. BindAdapt keeps their native memory and update rules intact while deciding which views to acquire, which operators feed that memory, and when results return.

\paragraph{Active and resource-aware perception.} Visual active tracking (VAT) learns camera or agent actions from observations~\cite{active,advat,distractionvat}. Streaming perception scores the latest available prediction on the physical clock~\cite{streaming}, PVT++ predicts ahead to offset tracking latency~\cite{pvtpp}, and Chanakya learns runtime configuration decisions~\cite{chanakya}. ARTrack-AC and AVTrack adapt computation within a single tracker~\cite{artrackac,avtrackicml,avtrack}. The closest to ours are OA-VAT~\cite{oavat} and Chanakya. OA-VAT couples instance prototypes with occlusion-aware recovery planning, and its asynchronous prototype updates recognize that perception can evolve alongside control. BindAdapt differs from both in two ways. It chooses camera motion, operator, and start time as one action, whereas OA-VAT chooses motion and Chanakya chooses computation. It also predicts the identity state that holds after a result returns, whereas OA-VAT imitates paths that restore the target's view and Chanakya optimizes latency-aware accuracy.

\paragraph{Identity over time.} Tracking benchmarks mostly judge where a box lies. OTB, TrackingNet, LaSOT, and GOT-10k established overlap and center-precision measures~\cite{otb,trackingnet,lasot,got10k}, and UAV123, NFS, TLP, and long-term protocols expose failures under viewpoint change, high frame rate, long duration, and absence~\cite{uav123,nfs,tlp,ltmetrics}. Panoramic, crowded-scene, and mask benchmarks add controllable views, persistent people, and region measures~\cite{omni,jrdb,panotrack,vot360,dancetrack,davis}. None of these single-target measures separates wrong binding from localization failure and unresolved output on the physical clock, and multi-object measures such as the identification F1 score (IDF1) and higher order tracking accuracy (HOTA)~\cite{idf1,hota} score association among all annotated people, not the time spent on the person designated in the first view. Decision-focused learning evaluates predictions by the decisions they support~\cite{spo}, and metareasoning weighs computation against policy gains~\cite{meta}. We therefore report standard measures alongside physical-time identity outcomes for the designated person.

%% file: sec/3_methodology.tex
\section{Method}
\label{sec:method}
Figure~\ref{fig:method} shows the three components of BindAdapt. A returned result enters the tracker as an observation that its own association and memory update may accept or reject. BindAdapt never edits memory entries or network weights, so it decides only what the tracker observes and computes.
\subsection{Problem formulation}
\label{sec:problem}
An episode begins with a designated target in the initial view. Let $\mathcal{H}_t$ denote the history available at time $t$, which contains acquired views, completed computations, candidate identity hypotheses, native tracker memory, camera state, and the queue of pending requests. Each entry carries a source identifier that records the acquisition or computation event it came from. Apart from the initial box, online decisions cannot access ground-truth boxes, true identities, future frames, or image regions the camera has not viewed.

BindAdapt decides at events, which are acquisitions, returned results, and timeouts. At each event it chooses an action $a_t=(m_t,o_t,\delta_t)$, where $m_t$ is a camera trajectory, $o_t$ is a visual operator, and $\delta_t$ is the start time of the operator relative to the event. Operators include local tracking, an identity check, wide-view search, and waiting. A positive $\delta_t$ keeps a non-preemptive worker free until the camera has slewed, or until separating motion makes a check informative. Because camera motion and settling precede acquisition, the result of the chosen operator $o_t$ returns at
\begin{equation}
r_t=\max\bigl(t+\delta_t,\,q_t,\,t_{\mathrm{cap}}\bigr)+\tau_{o_t}+\tau_{\mathrm{tr}},
\label{eq:ready}
\end{equation}
where $q_t$ is the time at which the compute device becomes free, $t_{\mathrm{cap}}$ is the capture time of the frame after the camera settles, $\tau_{o_t}$ is the service time of operator $o_t$ on that input, and $\tau_{\mathrm{tr}}$ covers post-processing and transfer. Operator execution thus cannot begin before its frame exists and its device is free, and the tracker can use the result only after $r_t$.

The return time matters because it decides which identity a result can still change. We distinguish two states. The memory state is the identity held in native tracker memory, and the binding state $z_j\in\{\mathrm{c},\mathrm{w},\mathrm{u}\}$ labels the box that the tracker outputs at evaluation tick $t_j$. A result changes the memory state at its return, and the memory state determines later outputs, so the binding state is its observable consequence. Correct binding, labeled $\mathrm{c}$, requires the original target to be in view and the box to reach an intersection-over-union (IoU) of at least $0.5$ with it. Otherwise, wrong binding, labeled $\mathrm{w}$, requires an IoU of at least $0.5$ with an annotated competing person, so a box that covers both counts as correct. Every other outcome, including a missing output or a localization failure, is unresolved and labeled $\mathrm{u}$. Unresolved is a state of the tracker, whereas ticks without a definite annotation are marked as insufficiently annotated and are excluded from all three labels. For an action sequence $\pi=(a_0,a_1,\dots)$ and a horizon $H$, the identity loss accumulated over evaluation ticks $t_j\in[t,t+H]$ is defined as
\begin{equation}
\ell(\mathcal{H}_t,\pi)=\sum_{j}d_j\bigl(\lambda_{\mathrm{w}}\mathbf{1}[z_j=\mathrm{w}]+\lambda_{\mathrm{u}}\mathbf{1}[z_j=\mathrm{u}]\bigr),
\label{eq:loss}
\end{equation}
where $d_j$ is the physical duration of tick $j$ and $\lambda_{\mathrm{w}}\ge\lambda_{\mathrm{u}}>0$. The default $\lambda_{\mathrm{w}}>\lambda_{\mathrm{u}}$ penalizes following a different person more than remaining unresolved. The scheduler minimizes the expected loss under hard caps on the computation and motion budgets. The computation cap covers feature extraction, response inference, planning, and replay, and the motion cap covers angular travel, speed, and acceleration. Estimating it first requires a summary of the history.

\subsection{Provenance graph and native memory}
\label{sec:graph}
This summary must recognize shared origins, because several computations on one frame can look like independent evidence. BindAdapt therefore represents $\mathcal{H}_t$ as a typed provenance graph $G_t$, drawn in the first panel of Figure~\ref{fig:method}. Source nodes describe captured frames and their geometry, computation nodes record the operator with its inputs, capture and return times, and returned features, hypothesis nodes describe candidate identities, memory nodes summarize native entries with their provenance, and queue nodes record pending requests without revealing their content. Edges connect each derived quantity to its source, so a detector box, a feature crop, and a confidence score from one frame share an ancestor, and that ancestry survives memory updates. Shared ancestry does not make a second computation worthless, because two operators on the same pixels can expose different information, so candidate-conditioned attention, anchored on the target box in the first view, learns which dependencies matter for each candidate action.

Each candidate action forms a query from its camera geometry, operator, source, and start time. A shared graph encoder produces history tokens, and cross-attention from the query yields the representation of the action.
\subsection{Identity response}
\label{sec:response}
The identity response predicts the binding states that an action produces. Because a result changes the memory state that exists at its return, it models return time and binding jointly through the following conditional distribution
\begin{equation}
p_\theta\bigl(z_{t:t+H},\,\mathbf{r},\,\mathbf{k}\mid\mathcal{H}_t,\pi\bigr),
\label{eq:response}
\end{equation}
where $\theta$ denotes the network parameters, $z_{t:t+H}$ is the binding trajectory over the horizon, $\mathbf{r}$ lists the return times of the requests in $\pi$, and $\mathbf{k}$ is the total computation and motion cost. A discrete hazard head predicts the return time over future event bins. Time-indexed state heads estimate the binding probabilities $p_{\mathrm{c}}$, $p_{\mathrm{w}}$, and $p_{\mathrm{u}}$, which form the future risk in the second panel of Figure~\ref{fig:method}. A cost head predicts $\mathbf{k}$, and a predictive uncertainty term flags states and actions with sparse training support. Compared with regressing a scalar value, predicting when evidence arrives and which binding state it produces gives training a denser, time-resolved signal that separates results landing before and after a memory update.

For multi-step planning, a transition head propagates candidate identities and geometry, a native-memory summary, camera pose, pending requests, and remaining budgets, and predicts the next acquisition, return, or timeout with the binding trajectory up to it. These rollouts never read future images and never modify the real tracker.

Training requires the outcomes of actions that were not taken, and a recorded run both hides untaken branches and confounds hard states with the corrective checks it triggers. We therefore generate paired outcomes from a common prefix: we snapshot camera, queue, hypotheses, native memory, and random state, then execute each candidate branch over the same physical horizon. A panorama renderer synthesizes each PTZ view from the panoramic source, so pixels outside the current view are available to the environment and the offline labeler but never to the policy. Each branch restores the same tracker tensors, queue, cached features, and random state before rendering its next view, and only its own returned evidence can update the restored memory. Because branches share everything except the action, their outcome difference isolates its effect on the original target, in the spirit of decision-focused learning~\cite{spo}.

The loss sums five weighted terms: a state term, the cross-entropy of binding states over the horizon, with insufficiently annotated ticks masked although their timing and cost still train the return and cost heads; a censored discrete hazard loss over return-time bins; a regression on computation and motion demand; a ranking term that prefers the paired branch with lower realized loss under Eq.~\eqref{eq:loss}; and a calibration term against overconfident wrong-binding predictions. Each sequence is supervised only by its own continuation, and ground truth never enters the input.
\begin{algorithm}[t]
\caption{One joint camera--computation update}\label{alg:online}
\begin{algorithmic}[1]
\Require History $\mathcal{H}_t$, pending queue, remaining budgets
\State Assimilate completed results in event order
\State Update identity hypotheses and native memory
\State Construct provenance graph $G_t$ and encode history
\State Generate timed actions that satisfy the remaining caps
\For{each candidate action $a$}
  \State Predict its next event and binding with $p_\theta$
\EndFor
\State Expand sequences by beam search and score them by $\widehat{J}$
\State Commit the first action $a_0$ of the best sequence
\State Recheck pending requests and replan at the next event
\end{algorithmic}
\end{algorithm}

\subsection{Receding-horizon joint scheduling}
\label{sec:schedule}
The number of timed sequences grows with every planning step, and the search itself is charged to the compute budget. At each real event, the scheduler therefore expands feasible sequences by beam search and scores each one by the expected identity loss under the response,
\begin{equation}
\widehat{J}(\pi)=\sum_j d_j\bigl(\lambda_{\mathrm{w}}\hat{p}_{\mathrm{w},j}+\lambda_{\mathrm{u}}\hat{p}_{\mathrm{u},j}\bigr),
\label{eq:plan}
\end{equation}
subject to the hard budgets, where hats denote predictions of Eq.~\eqref{eq:response}. This score is the expected value of Eq.~\eqref{eq:loss}. The response is carried across decisions by state particles, samples that hold candidate identities, request identifiers, and return times. Small penalties on predicted computation and motion cost and on the spread of cumulative loss across these particles separate otherwise similar feasible actions. Lookahead lets the committed action steer the view now so that a later operator can verify identity before a crossing, or hold the worker while an in-flight request returns. Algorithm~\ref{alg:online} summarizes the joint scheduling update at each decision event.

%% file: sec/4_experiments.tex
\section{Experiments}
\label{sec:experiments}

\begin{table*}[t]
\centering\small\setlength{\tabcolsep}{1.5pt}\renewcommand{\arraystretch}{1.0}\setlength{\heavyrulewidth}{1pt}
\begin{tabularx}{\textwidth}{@{}llc*{3}{>{\centering\arraybackslash}X}@{\hspace{9pt}}*{3}{>{\centering\arraybackslash}X}@{\hspace{9pt}}*{2}{>{\centering\arraybackslash}X}@{\hspace{9pt}}c@{}}\toprule
\multirow{2}{*}{\textbf{Method}} & \multirow{2}{*}{\textbf{Tracker}} & \multicolumn{4}{c}{\textbf{Identity on the physical clock}} & \multicolumn{3}{c}{\textbf{$\boldsymbol{\sid}$ by event type}} & \multicolumn{2}{c}{\textbf{Fixed view}} & \textbf{Service time}\\
\cmidrule(lr){3-6}\cmidrule(lr){7-9}\cmidrule(lr){10-11}\cmidrule(l){12-12}
 & & $\sid\!\uparrow$ & $W\!\downarrow$ & $R\!\uparrow$ & $T_5$\,(s)$\downarrow$ & short$\uparrow$ & similar$\uparrow$ & long$\uparrow$ & AO$\uparrow$ & sAO$\uparrow$ & (ms)$\downarrow$\\\midrule
\BAtable tables/external_full.tex \bottomrule\end{tabularx}
\caption{\textbf{BindAdapt keeps the original target longest under shared budgets.} Identity success $\sid$, wrong-binding rate $W$, and five-second recovery $R$ in \%, mean recovery time $T_5$ in s. Events are short occlusion, similar-person distractor, and long absence, and AO and sAO use fixed-view replay. $\Delta$ is against DAM4SAM + Chanakya-J in points, s, and ms. Tracker names the perception module (\emph{self}: the row's own tracker; \emph{built-in}: inside the policy). Learned methods report three-seed means with the $\sid$ deviation in gray. Service time per step includes the scheduling decision and excludes the $0.5$\,s added delay used for the other columns. Dashes mark settings not evaluated. Bold and underline mark the best and second best except for service time.}
\label{tab:external}
\end{table*}

\subsection{Experimental setup}
\paragraph{Datasets.} Episodes are rendered as PTZ views from panoramic recordings in QuadTrack and JRDB~\cite{omni,jrdb,panotrack}. We split by recording or scene group before generating branches, so crops, people, and branches from one recording never cross splits. Training receives \BAval{t09.r01.groups} groups with \BAval{t09.r01.episodes} episodes, validation \BAval{t09.r02.groups} groups with \BAval{t09.r02.episodes} episodes, and testing \BAval{t09.r03.groups} groups with \BAval{t09.r03.episodes} episodes. The test set contains \BAval{plan.strata.short.episodes} short occlusions, \BAval{plan.strata.similar.episodes} similar-person distractors, and \BAval{plan.strata.long.episodes} long absences, weighted by count as $0.25$, $0.375$, and $0.375$.

\paragraph{Baselines.} We compare BindAdapt with 15 methods and combinations, grouped in Table~\ref{tab:external} by how they receive camera actions. Perception trackers keep their native models and memory under a common causal PTZ controller, active controllers act in the common PTZ action space, PVT++ predicts delivery, ARTrack-AC selects capacity, and the joint policies choose from the same joint action catalog as BindAdapt. OA-VAT retrains its recovery planner in the PTZ action space, and Chanakya-J keeps its latency-aware reward with the same causal history, queue, and budget inputs as BindAdapt. Every learned method, BindAdapt included, is tuned with the same validation budget of \BAval{tune.configs} configurations, each trained with \BAval{tune.seeds} seeds and scored on the \BAval{tune.groups} validation groups. All methods share the initial box, the permitted views and actions, and the motion and compute caps, and no policy can see a view before acquiring it.

\paragraph{Metrics.} Identity success $\sid$ is the primary metric. For an episode of duration $T$, $\sid=D_{\mathrm{c}}/T$, where $D_{\mathrm{c}}$ is the total duration of ticks labeled $\mathrm{c}$. The wrong-binding rate $W=D_{\mathrm{w}}/T$ uses the confirmed wrong-person duration $D_{\mathrm{w}}$. Ticks with insufficient annotation stay in $T$ and count toward neither numerator, so $\sid$ and $W$ are lower bounds on confirmed durations. Episode ratios are averaged within each event type and then combined with the weights above. Five-second recovery $R$ is the fraction of reappearance events in which correct binding resumes within five seconds and persists for one second, and $T_5$ is the mean time to that recovery, with five seconds assigned to unrecovered events. Following standard benchmarks~\cite{otb,trackingnet,lasot,got10k}, we also report on fixed-view replay the average overlap (AO), which equals the area under the success curve (AUC), together with streaming AO (sAO), which scores the latest causal output on the physical clock~\cite{streaming}.

These metrics separate localization from identity: a box on a distractor can overlap that person accurately yet contributes to $W$ rather than $\sid$. Likewise, a quickly returned result is useful only if subsequent outputs remain on the designated target. Reporting duration and recovery together exposes both prolonged drift and delayed restoration.

\paragraph{Implementation details.} The graph encoder uses 96-dimensional tokens, three relation-aware attention layers, and four attention heads, and the state capacity keeps up to six hypotheses, sixteen source nodes, and sixteen memory entries. The action catalog combines six camera trajectories, four operators, and two start times, giving at most 48 actions. The scheduler uses a two-second horizon, beam width eight, and depth three. Default normalized caps are $1.0$ for motion and compute. As a stress condition, $0.5$\,s of waiting is added after acquisition, motion, and computation, whereas the device experiments use measured timing, 332.7\,ms on average from command to usable result. Differences are computed before rounding. The binding weights are $\lambda_{\mathrm{w}}=\BAval{lambda.w}$ and $\lambda_{\mathrm{u}}=\BAval{lambda.u}$, and the training and penalty weights are listed in the supplementary material.

\begin{table}[t]
\centering\small\setlength{\tabcolsep}{1.8pt}\setlength{\heavyrulewidth}{1pt}\renewcommand{\arraystretch}{1.08}
\begin{tabular*}{\columnwidth}{@{\extracolsep{\fill}}lccccc@{}}\toprule
\multirow{2}{*}{\textbf{Backbone}} & \multicolumn{4}{c}{\textbf{Scheduler, $\boldsymbol{\sid}\!\uparrow$}} & \multirow{2}{*}{$\Delta$}\\\cmidrule(lr){2-5}
 & {\footnotesize\shortstack{Common\\controller}} & {\footnotesize OA-VAT-PTZ} & {\footnotesize Chanakya-J} & \cellcolor{BAours}{\footnotesize\textbf{BindAdapt}} & \\\midrule
\BAtable tables/backbone_styled.tex \bottomrule\end{tabular*}
\caption{\textbf{The gain holds with each backbone fixed.} $\sid$ in \% under shared actions and costs. $\Delta$ is the $\sid$ gain of BindAdapt over Chanakya-J in percentage points.}\label{tab:backbone}
\end{table}

\subsection{Comparison with external methods}
Table~\ref{tab:external} shows a progression across families. Stronger perception lifts identity success from \BAval{t12.r01.sid}\% for OSTrack to \BAval{t12.r08.sid}\% for DAM4SAM, and latency extrapolation and adaptive capacity, at \BAval{t12.r11.sid}\% and \BAval{t12.r12.sid}\%, trim spatial lag without steering views. Even the joint policies plateau near three-quarters of episode time, at \BAval{t12.r14.sid}\% and \BAval{ext.best.sid}\%, because their checks often return after the crossing. On the same backbone and action catalog, BindAdapt reaches \BAval{ext.bind.sid}\% and lowers the wrong-binding rate from \BAval{ext.best.wrong}\% for DAM4SAM with Chanakya-J to \BAval{ext.bind.wrong}\%. Five-second recovery also rises from \BAval{ext.best.recovery}\% to \BAval{ext.bind.recovery}\%, and the mean recovery time falls from \BAval{ext.best.t5} to \BAval{ext.bind.t5}\,s. A paired bootstrap over the \BAval{ci.groups} test scene groups places the \BAval{delta.ext.sid}-point gain in a 95\% interval of [\BAval{ci.lo}, \BAval{ci.hi}], and IDF1~\cite{idf1} and HOTA~\cite{hota} agree, rising from \BAval{idf1.best}\% to \BAval{idf1.bind}\% and from \BAval{hota.best}\% to \BAval{hota.bind}\%.

The margin widens on the physical clock. On fixed-view replay, where every method receives the same frames, ARTrack-AC ranks second in streaming AO at \BAval{t13.r10.sao.pct}\% by cutting network latency, and BindAdapt reaches \BAval{t13.r11.sao.pct}\%. Over its own DAM4SAM backbone, BindAdapt gains \BAval{delta.fixed.dam.ao} points in AO, which ignores latency, but \BAval{delta.fixed.dam.sao} points in streaming AO, so much of the gain comes from boxes that arrive in time to affect tracking.

\subsection{Backbone control}
The gain transfers across perception backbones. Replacing Chanakya-J by BindAdapt with the weights of OSTrack, LoRATv2, and DAM4SAM frozen and their native memory updates unchanged raises identity success by \BAval{delta.backbone.ost}, \BAval{delta.backbone.lorat}, and \BAval{delta.backbone.dam} points (Table~\ref{tab:backbone}), so the gain does not rest on one backbone. Memoryless OSTrack gains most, from \BAval{t15.r01.common.controller}\% to \BAval{t15.r01.bindadapt}\%, and surpasses DAM4SAM with Chanakya-J at \BAval{t15.r03.chanakya.j}\%: a timely check compensates for missing memory. This comparison changes the scheduler while preserving the weights and native update rule of each tracker. It therefore tests the benefit of choosing views and computation times for each of the three evaluated tracking backbones.
\begin{table}[t]
\centering\small\setlength{\tabcolsep}{1.6pt}\setlength{\heavyrulewidth}{1pt}
\begin{tabularx}{\columnwidth}{@{}Xcccc@{}}\toprule
\textbf{Variant} & $\sid\!\uparrow$ & $\Delta\sid$\,(pp) & $W\!\downarrow$ & $T_5\!\downarrow$\\\midrule
\BAtable tables/ablation_styled.tex \bottomrule
\end{tabularx}
\caption{\textbf{Every ablation lowers identity success.} DAM4SAM backbone, default budgets, $\Delta\sid$ as variant minus full, $T_5$ in s, with three-seed standard deviations shown in gray.}\label{tab:ablation}
\vspace{7pt}
\setlength{\tabcolsep}{4pt}
\begin{tabular*}{\columnwidth}{@{\extracolsep{\fill}}lccc@{}}\toprule
\multirow{2}{*}{\textbf{Response}} & \multicolumn{2}{c}{\textbf{Scheduler}, $\boldsymbol{\sid}\!\uparrow$} & \multirow{2}{*}{\shortstack{Joint $-$\\Separable}}\\\cmidrule(lr){2-3}
 & Separable & Joint & \\\midrule
Reference response & \BAval{t03.r01.sid}\BApm{\BAval{sd.t03.r01}} & \BAval{t03.r02.sid}\BApm{\BAval{sd.t03.r02}} & \textcolor{BAgain}{$+$\BAval{delta.swap.refplanner}}\\
\rowcolor{BAours}BindAdapt response & \BAval{t03.r03.sid}\BApm{\BAval{sd.t03.r03}} & \textbf{\BAval{t03.r04.sid}}\BApm{\BAval{sd.t03.r04}} & \textcolor{BAgain}{$+$\BAval{delta.swap.bindplanner}}\\\midrule
BindAdapt $-$ Reference & \textcolor{BAgain}{$+$\BAval{delta.swap.sepresponse}} & \textcolor{BAgain}{$+$\BAval{delta.swap.jointresponse}} & \\\bottomrule
\end{tabular*}
\caption{\textbf{The response improves identity success under both schedulers, and joint scheduling adds a further gain.} Reference is a joint-belief response of equal size and runtime, and Separable selects camera motion before computation. Gains in $\sid$ points.}\label{tab:swap}
\end{table}

\subsection{Response and scheduler contributions}
To separate prediction from search, Table~\ref{tab:swap} crosses two responses with two schedulers on shared branches in the setting of Table~\ref{tab:ablation}. Its reference response is trained on the same branches with our parameter count and runtime, so it differs from the joint-belief composition of Table~\ref{tab:external}. Our response adds \BAval{delta.swap.sepresponse} and \BAval{delta.swap.jointresponse} points under the two schedulers, and joint scheduling adds \BAval{delta.swap.refplanner} and \BAval{delta.swap.bindplanner} points under the two responses. Both contribute almost additively, and the response contributes more. A matched temporal direct-$Q$ model on the same branches reaches only \BAval{int.temporal.sid}\%, ordering \BAval{t07.r02.pair.rank.pct}\% of action pairs against \BAval{t07.r04.pair.rank.pct}\% for our response, because its scalar target gives no direct supervision on whether evidence returns before a change in the tracker's native memory.

\begin{figure*}[t]
\centering\includegraphics[width=\textwidth]{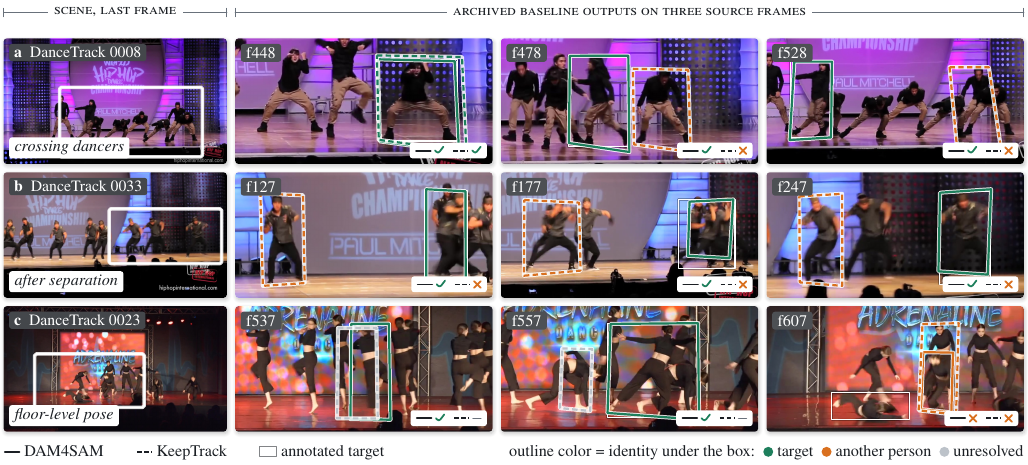}
\caption{\textbf{Baseline trackers switch to another dancer at crossings and pose changes.} Rows a to c show three DanceTrack sequences, with the last frame and its zoom window on the left and the archived DAM4SAM and KeepTrack outputs as solid and dashed outlines on three source frames. These fixed-camera frames illustrate identity drift qualitatively.}
\label{fig:recorded}
\end{figure*}

\begin{figure}[t]
\centering\includegraphics[width=\linewidth]{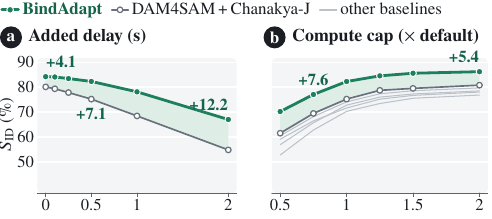}
\caption{\textbf{The gain grows with delay and holds under every compute cap.} Labels and the green fill give the difference from DAM4SAM with Chanakya-J in points under (a)~added delay and (b)~the normalized caps on total computation.}
\label{fig:robust}
\end{figure}

Every variant in Table~\ref{tab:ablation} lowers identity success. Fixing the computation type costs the most, from \BAval{t04.r01.sid}\% to \BAval{t04.r06.sid}\%, so camera control alone does not explain the gain. Among the response inputs, target conditioning matters most: without it identity success falls to \BAval{t04.r03.sid}\% and wrong binding rises from \BAval{t04.r01.wrong}\% to \BAval{t04.r03.wrong}\%, because the response can no longer separate the target from a look-alike. Removing the uncertainty term or the memory summary costs \BAval{abl.unc.drop} and \BAval{tab.abl.mem} points, and starting every operator at once costs \BAval{abl.imm.drop}, so the start time is a decision in its own right. Validated rules that check when another person approaches or every \BAval{rule.rate.period}\,s reach \BAval{rule.prox.sid}\% and \BAval{rule.rate.sid}\%. Both already exceed all eight perception trackers of Table~\ref{tab:external}, yet they trail BindAdapt by \BAval{rule.prox.drop} and \BAval{rule.rate.drop} points and fall to \BAval{rule.prox.long}\% and \BAval{rule.rate.long}\% on long absences, because a rule checks without predicting when its result returns.

Stratified interventions isolate the mechanisms directly. Removing return time costs \BAval{diag.late.drop} points when evidence returns after memory changes but \BAval{diag.early.drop} when it returns before, and removing source dependencies costs \BAval{diag.corr.drop} points on same-source cues but \BAval{diag.indep.drop} otherwise. Across the test episodes, wrong bindings that follow a post-crossing return fall from \BAval{attr.best.after} of \BAval{attr.best.total} for DAM4SAM with Chanakya-J to \BAval{attr.bind.after} of \BAval{attr.bind.total}, closing the loop opened in the introduction, and BindAdapt closes \BAval{gap.external}\% of the gap to an oracle over the same finite action catalog, whose identity success is \BAval{t08.r04.sid}\%.

\subsection{Robustness and computational cost}
As the return-time mechanism predicts, the lead widens with delay and event difficulty. Figure~\ref{fig:robust} shows that the difference from DAM4SAM with Chanakya-J rises from \BAval{delta.sweep.zero} points without added delay to \BAval{delta.sweep.two} points at two seconds of delay. A longer delay lets more results meet a changed memory. Across the event types of Table~\ref{tab:external}, the difference grows from \BAval{delta.strata.short} points for short occlusions to \BAval{delta.strata.long} points for long absences, where a late result is most likely to describe a person who has already left the view. The advantage persists at every compute cap, and at a cap of $0.75$ BindAdapt already exceeds the strongest baseline at the default cap, with identity success of \BAval{t17.r08.bindadapt}\% versus \BAval{t17.r09.dam4sam.chanakya.j}\%.\par The gain is stable across the tested settings. Identity success changes by at most \BAval{lambda.sens.range} points across binding weights from \BAval{t24.r01.lambda.w} to \BAval{t24.r04.lambda.w}, and wider beams or deeper search add at most \BAval{sens.search.gain} points. Under measured device latencies without added waiting, calibrating only the service model raises identity success from \BAval{svc.direct.sid}\% to \BAval{svc.recal.sid}\%, close to the \BAval{svc.retrain.sid}\% of full retraining and 5.2 points above a recalibrated Chanakya-J, as the return-time error falls from 64.8 to 19.3\,ms.

BindAdapt also matches the service time of the strongest baseline, \BAval{runtime.bind.mean} against \BAval{runtime.best.mean}\,ms per step under device-neutral profiles, including its \BAval{plan.mean}\,ms decision.

\subsection{Real PTZ camera}
The gain carries over to a physical camera. Table~\ref{tab:realptz} runs 20 two-person scripts three times per method on a real PTZ camera, with actual device timing and no added waiting, and scores one reappearance event per run. BindAdapt raises identity success from 74.8\% for Chanakya-J to 80.3\%, lowers wrong binding from 8.8\% to 5.6\%, and shortens $T_5$ from 2.12 to 1.55\,s. It also exceeds asynchronous verification and a variant without the return-time model, so predicting when evidence returns pays off on hardware as well. Its compute service time stays within 0.3\,ms of Chanakya-J. The physical run includes camera settling and processing delays throughout each of the shared two-person scripts.

\input{tables/real_ptz_updated}
\input{tables/cases_updated}

\subsection{Recorded clips}
In every baseline failure of Figure~\ref{fig:recorded}, at a crossing, a separation, or a pose change, the box moves to the wrong person by the deadline for an identity check. To score such failures under one protocol, Table~\ref{tab:casesupdated} reruns every scheduler on DanceTrack 0001 and QuadTrack 0025 from the same initialization. On fixed-camera DanceTrack, pan, tilt, and zoom move and resize a digital viewport inside the original frame, whereas QuadTrack views are rendered from its panorama. BindAdapt keeps the target for 73.9\% and 76.4\% of the two clips, 6.7 and 6.9 points more than Chanakya-J, and cuts wrong binding by 2.2 and 2.7 points.

%% file: tables/real_ptz_updated.tex
\begin{table}[t]
\centering\small\setlength{\tabcolsep}{1.6pt}\setlength{\heavyrulewidth}{1pt}
\begin{tabular*}{\columnwidth}{@{\extracolsep{\fill}}lcccccc@{}}\toprule
\textbf{Scheduler} & $\sid\!\uparrow$ & $W\!\downarrow$ & $U\!\downarrow$ & $R\!\uparrow$ & $T_5\!\downarrow$ & ms\\\midrule
Common controller & 70.9 & 10.7 & 18.4 & 70.0 & 2.44 & 35.9\\
Chanakya-J & 74.8 & 8.8 & 16.4 & 76.7 & 2.12 & 38.2\\
Async.\ verification & 76.1 & 7.9 & 16.0 & 78.3 & 2.01 & 37.9\\
No return-time model & 77.2 & 7.1 & 15.7 & 80.0 & 1.86 & 37.6\\
\rowcolor{BAours}\textbf{BindAdapt} & \textbf{80.3} & \textbf{5.6} & \textbf{14.1} & \textbf{85.0} & \textbf{1.55} & 38.5\\\bottomrule
\end{tabular*}
\caption{\textbf{BindAdapt keeps the target longest on a real PTZ camera.} 20 two-person scripts, three runs each per method, with actual device timing and no added waiting. $U$ is the unresolved share, $T_5$ in s, and ms is compute service time.}\label{tab:realptz}
\end{table}

%% file: tables/cases_updated.tex
\begin{table}[t]
\centering\small\setlength{\tabcolsep}{2pt}\setlength{\heavyrulewidth}{1pt}
\begin{tabular*}{\columnwidth}{@{\extracolsep{\fill}}lcccccc@{}}\toprule
\multirow{2}{*}{\textbf{Scheduler}} & \multicolumn{3}{c}{\textbf{DanceTrack 0001}} & \multicolumn{3}{c}{\textbf{QuadTrack 0025}}\\\cmidrule(lr){2-4}\cmidrule(l){5-7}
 & $\sid\!\uparrow$ & $W\!\downarrow$ & $U\!\downarrow$ & $\sid\!\uparrow$ & $W\!\downarrow$ & $U\!\downarrow$\\\midrule
Common controller & 59.8 & 9.7 & 30.5 & 61.7 & 10.4 & 27.9\\
Chanakya-J & 67.3 & 7.8 & 24.9 & 69.5 & 8.2 & 22.4\\
\rowcolor{BAours}\textbf{BindAdapt} & \textbf{73.9} & \textbf{5.6} & \textbf{20.5} & \textbf{76.4} & \textbf{5.5} & \textbf{18.2}\\\bottomrule
\end{tabular*}
\caption{\textbf{The gain holds when every method is rerun on the recorded clips.} Shares of clip duration in \%, fully annotated, with shared initialization, backbone, operators, and caps.}\label{tab:casesupdated}
\end{table}

%% file: sec/5_conclusion.tex
\section{Conclusion}
\begingroup\parfillskip=0pt
BindAdapt jointly schedules camera motion, visual operators, and their start times for PTZ identity tracking. Asynchronous evidence acts on the memory present when a result returns. A provenance graph tracks shared sources, an identity-response model predicts binding after each timed action, and a receding-horizon scheduler coordinates these actions under shared motion and compute budgets. With native tracker updates unchanged, BindAdapt keeps the designated target longer than the compared methods at matched service time. The gain persists across the three evaluated backbones and compute caps, grows with delay and event difficulty, and carries over to a real PTZ camera, where identity success rises from 74.8\% to 80.3\%. These results connect persistent target identity to the acquired evidence and its arrival time in tracker memory.\par\endgroup

%% file: supp/s01_implementation.tex
\section{Implementation and Training}\label{sec:simplementation}
BindAdapt enforces strict causality, deterministic event ordering, and exact state restoration, so the online policy never sees unviewed pixels, future frames, or ground-truth identities, which only the offline branch supervisor reads. Every decision records the camera pose at acquisition, the queue state at dispatch, and the native memory receiving each result, so online updates and offline branches are replayable.

\subsection{Causal observations and snapshot separation}
Online decisions read only acquired PTZ viewports, completed operator outputs, native tracker state, camera pose, and queue status. After the first view supplies the initial box or mask for the designated person, each acquisition or computation event receives a source identifier that propagates to every feature derived from it. A future acquisition token specifies its capture rule, proposed time, trajectory, pose, parent event, and readiness condition, while its pixels remain unreadable until capture completes. Once a camera action is chosen, the renderer projects only the requested viewport from the panorama into $\mathcal{H}_t$, whereas the full panorama and true identities remain restricted to the offline branch supervisor.

Two separate snapshots enforce this boundary during data generation and training. The policy reads only the executable snapshot, which holds camera and queue state, native tracker tensors, completed observations, and random-generator state, so every training input tensor traces back to an observed source event. The evaluator snapshot adds identity labels and future source frames solely to score executed branches offline.

\subsection{Event timing and deterministic return order}
\begin{figure*}[!t]
\centering\includegraphics[width=\textwidth]{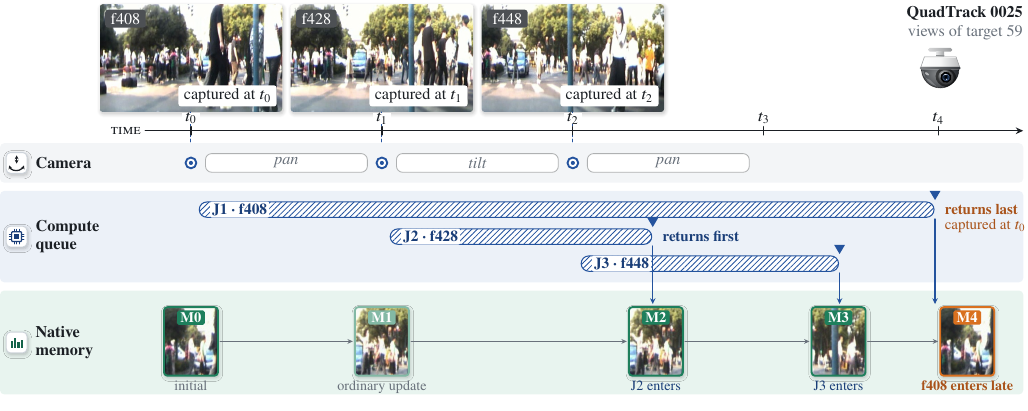}
\caption{\textbf{Capture time and return time differ, and memory changes in between.} Three jobs use QuadTrack 0025 views of frames 408, 428, and 448, and each bar spans dispatch to publication, including queueing, transfer, and added waiting. Job~2 returns first, so its result enters memory first, and Job~1 returns last and writes evidence from frame 408 into a memory that has changed three times since that capture. Each memory card shows the frame whose result wrote it, and M1 is an ordinary tracker transition. The timing is schematic.}
\label{fig:sclock}
\end{figure*}

An operator starts processing only after its source frame is captured, its start-time offset elapses, and the compute worker becomes free, matching Eq.~(1) of the main paper. Each request records source readiness $t_{\mathrm{cap}}$, requested dispatch time $t+\delta_t$, device availability $q_t$, and return time $r_t$. An operator scheduled on an unfinished acquisition stays a deferred intent that occupies no in-flight slot until its source arrives, at which point the runtime rechecks motion, source validity, and remaining budget before replanning and dispatch. An invalid or expired source cancels the intent before dispatch rather than delivering a stale result, whereas a timeout triggers replanning while keeping already dispatched work in flight. Figure~\ref{fig:sclock} illustrates how capture time and return time separate on the physical clock while the evaluator continuously scores the active tracker output across camera motion, inference, transfer, and waiting.

Simultaneous timestamps follow a fixed priority order so that queue and memory transitions remain deterministic. Acquisitions complete first and service returns next, after which the runtime updates the state, forms the observation, selects a feasible action, and dispatches newly chosen work. Each result is consumed at most once, timeouts never convert unfinished work into evidence, already spent compute remains charged upon cancellation, and pending requests stay in the queue across replanning without relaunching. At each evaluation tick, the evaluator scores the latest causal output against the current target position without backward interpolation, holding or extrapolating the prior box through one fixed adapter rule across all service profiles when no new result has arrived.

\subsection{Deterministic versus predicted decision state}
One joint update of Algorithm~1 in the main paper separates deterministic runtime bookkeeping from learned state predictions, as detailed in Table~\ref{tab:sstate}. Planning runs on the physical clock, so its decision latency is charged to the compute budget and included in service time. Restoring the source frames, queue, tracker tensors, and random-generator state listed in Table~\ref{tab:sstate} reproduces any branch from its prefix.

\begin{table*}[!t]\caption{\textbf{State, actions, and outputs of one joint update.} Deterministic entries come from runtime records and the scheduler transition $\mathcal{D}$, and predicted entries come from the identity response $p_\theta$.}\label{tab:sstate}\centering\small\setlength{\tabcolsep}{4pt}
\begin{tabular*}{\textwidth}{@{\extracolsep{\fill}}>{\raggedright\arraybackslash}p{0.17\textwidth}>{\raggedright\arraybackslash}p{0.47\textwidth}>{\raggedright\arraybackslash}p{0.28\textwidth}@{}}\toprule
Part & Content & Advanced by\\\midrule
Camera & Current pose, trajectory in execution, speed and acceleration limits & Deterministic camera model\\
Queue & Request identifiers, source pointers, capture, dispatch, and return times of completed requests, in-flight slots (at most three) & Deterministic runtime rules\\
Budgets & Remaining compute and motion allowance, reserved upper bounds of pending requests & Deterministic reservation and reconciliation\\
Provenance & Source identifiers and ancestry of features, hypotheses, and memory entries & Deterministic bookkeeping\\
Candidates & Identity hypotheses and geometry, including one slot for new or unidentifiable people & Predicted by $p_\theta$ between real events\\
Native memory & Summary of the tracker memory, whose real content changes only through the tracker's own update & Predicted summary, real update by the tracker\\
Pending returns & Return time of each in-flight request, sampled once per particle and kept across depths & Predicted by the hazard head\\\midrule
Action & Camera trajectory (6), operator (4: local tracking, identity check, wide-view search, wait), start-time offset (2) & Chosen by beam search\\
Outputs & Binding trajectory over $\mathrm{c}$, $\mathrm{w}$, $\mathrm{u}$, return times, compute and motion cost, the committed first action & Response and scheduler\\\bottomrule
\end{tabular*}\end{table*}

\begin{figure*}[!t]
\centering\includegraphics[width=\textwidth]{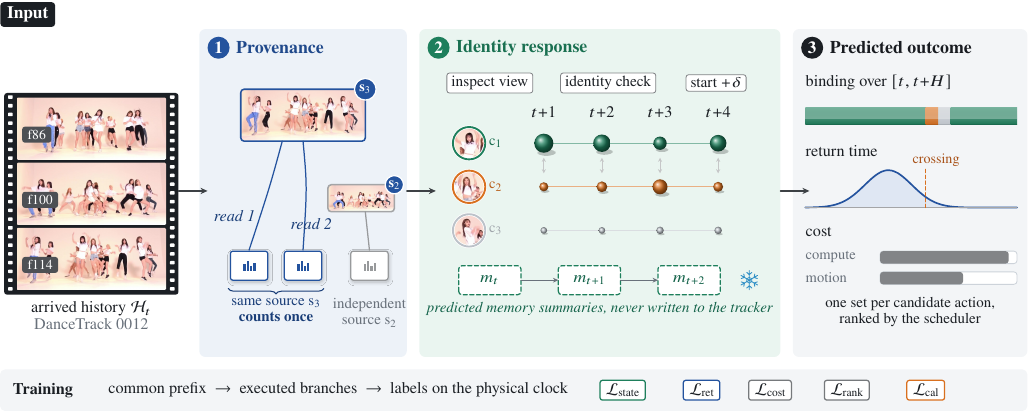}
\caption{\textbf{The identity response keeps evidence origin, return time, and native-memory dependence.} Two reads of one frame carry the same source and count once. An action query predicts how each candidate evolves over the horizon, together with memory summaries that never update the real tracker, and returns the binding trajectory, return time, and cost that the scheduler ranks. The frames are DanceTrack 0012 frames 86, 100, and 114, the candidates are crops of frame 114, and the graph, spheres, and curves are schematic.}
\label{fig:sresponse}
\end{figure*}

\subsection{Provenance graph capacity and state restoration}
The provenance graph compresses observation history for scheduling while native tracker memory retains the high-dimensional features that drive perception. Offline branch restoration recovers both representations, including hidden templates, optimizer states, and segmentation-memory entries read by later updates. The graph keeps at most sixteen source nodes, sixteen memory entries, six candidate hypotheses, and the active queue, retaining source records for all active jobs until completion. Admission prioritizes the most recent completed acquisitions alongside earlier identity-informative views with deterministic tie-breaking, and node eviction preserves ancestor identifiers in surviving descendants so repeated reads of one capture remain linked.

Relation-aware attention maps this bounded graph to action-conditioned predictions at the same parameter count and runtime allowance as the reference response. Graph edges distinguish shared pixels, successive state updates, candidate association, and queue dependence across three four-head layers with 96-dimensional tokens. As Figure~\ref{fig:sresponse} illustrates, each query encodes the proposed camera trajectory, operator, source frame, start-time offset, estimated resource demand, and remaining budget, then attends to the encoded history to predict binding states, return times, and the cost of the action.

\subsection{Eventwise latent transitions for joint search}
Multi-step lookahead rolls decision states forward from event to event through compact latent summaries rather than future images. At the search root, the encoder maps $\mathcal{H}_t$ to candidate identity and geometry distributions across observed tracks plus one slot for newly appearing or unidentifiable people, together with a native-memory summary, camera pose, pending work, and remaining budgets. Native tracker tensors stay outside the planning state, and a fixed extractor applied to offline branch tracker states provides target memory summaries during training. Each beam node carries particles of the decision state $x_e$ at event time $\tau_e$ under a timed continuation $\pi=(a_0,\ldots,a_{k-1})$ with $k\leq3$. Over each segment, the response kernel is
\begin{equation}
K_\theta\bigl(\widetilde{x}_{e+1},z_{I_e},\eta_e,\Delta c_e\mid x_e,b_e,d_e\bigr),
\label{eq:seventkernel}
\end{equation}
where $b_e$ is the active action or wait, $d_e=\min(\tau_{\mathrm{dec}},T_{\mathrm{end}})$ is the continuation cutoff, $\eta_e$ is the next event, $\widetilde{x}_{e+1}$ samples candidate, geometry, and memory summaries jointly with the event time, and $\Delta c_e$ is paid segment compute while motion cost $c_{\mathrm{mot}}$ sums angular travel and zoom. A deterministic transition $\mathcal{D}$ then updates request identifiers, source pointers, camera constraints, queue occupancy, and budget reservations via $x_{e+1}=\mathcal{D}(\widetilde{x}_{e+1},x_e,b_e,\eta_e)$. Only newly selected actions advance the search depth, pending requests retain their sampled return times within each particle across depths, and descendant features inherit the ancestor identifier of their predicted acquisition.

Partitioning the two-second horizon at automatic and planned events allows one kernel to compose arbitrary timed continuations. With global endpoint $T_{\mathrm{end}}=t+H$, each segment ends at the earliest acquisition, return, timeout, decision, or horizon limit,
\begin{equation}
\tau_{e+1}=\min\{\tau_{\mathrm{acq}},\tau_{\mathrm{ret}},\tau_{\mathrm{timeout}},\tau_{\mathrm{dec}},T_{\mathrm{end}}\}.
\label{eq:ssegment}
\end{equation}
The intervals $I_e=[\tau_e,\tau_{e+1})$ partition the horizon with the endpoint tick assigned to the final segment and zero-duration ties resolved by the priority order above. Conditioning on $d_e$ distinguishes an early verification cutoff from continued waiting even under the same active action, while an earlier return truncates the segment and updates the next state. Summing the segment identity loss $\ell_e$ and compute increments gives
\begin{equation}
\ell(\mathcal{H}_t,\pi)=\sum_e\ell_e(x_e,b_e;\pi),\qquad c_{\mathrm{comp}}(\pi)=\sum_e\Delta c_e.
\label{eq:ssequence}
\end{equation}
Beyond depth $k$, waiting fills the remainder of the horizon while pending returns and native updates proceed. Algorithm~\ref{alg:srollout} evaluates four particles per continuation, enforcing hard feasibility through reserved resource upper bounds and ranking sequences by mean particle loss, soft resource penalties, and cumulative loss spread $\hat{s}$ weighted by $\beta_s=\BAval{weight.uncertainty.weight}$, with both particle count and $\beta_s$ selected on validation scenes under the shared planning budget. Only $a_0$ executes before the next real event updates the native tracker and rebuilds the root.

\begin{algorithm}[t]
\caption{Eventwise evaluation of a timed continuation}\label{alg:srollout}
\begin{algorithmic}[1]
\Require Root particles, continuation $\pi$, endpoint $T_{\mathrm{end}}$
\For{each particle}
\While{$\tau_e<T_{\mathrm{end}}$}
\State Resolve active action $b_e$ and cutoff $d_e$ from $\pi$
\State Query $K_\theta(\cdot\mid x_e,b_e,d_e)$ for the next segment
\State Apply $\mathcal{D}$ to predicted summaries and scheduler state
\State Add this segment's loss and costs once
\State Advance to $\tau_{e+1}$ and retain pending request samples
\EndWhile
\EndFor
\State Return mean cumulative loss, costs, and loss spread $\hat{s}$
\end{algorithmic}
\end{algorithm}

\subsection{Counterfactual supervision from common-prefix branches}
Restarting all feasible actions from an identical snapshot isolates how a single timed choice alters subsequent target binding. At each training decision point, every branch restores the same source timestamp, acquisition history, camera pose, tracker tensors, candidate state, queue, random state, and remaining budgets. From the 48 combinations of six camera trajectories, four operators, and two start-time offsets, infeasible motions and immediate requests on unreadable sources are pruned, and future-source requests are admitted only when a feasible acquisition precedes them, their resource upper bound can be reserved, and source readability is verified before dispatch. Executing the queried sequence over the two-second horizon records the $\mathrm{c}$, $\mathrm{w}$, and $\mathrm{u}$ trajectory, return events, compute and motion costs, segment boundaries, active actions, acquired sources, candidate geometry, memory summaries, endpoint pending work, and whether each planned action was dispatched, deferred, canceled, or replaced by waiting.

Conditioning every label on its exact continuation prevents diverging future trajectories from contaminating supervision. Single-action branches execute $\pi=(a,\kappa_0)$, where $\kappa_0$ dispatches no further operator while pending work, camera motion, waiting, returns, and native updates run to completion, and whole-window labels from other continuations are never reassigned to $\pi$. When two branches share an initial wide-view acquisition before one verifies a competitor and the other waits for a tracker return, they share supervision only up to the divergence event. Uncertain ground-truth spans remain masked or interval-valued rather than filled by teacher predictions.

\subsection{Training losses and hyperparameter weights}
Five loss terms supervise binding states, return timing, resource demand, action ordering, and wrong-binding calibration without feeding offline future states into the input. The state loss $\mathcal{L}_{\mathrm{state}}$ combines masked cross-entropy on identity states, distribution matching on candidate probabilities, and regression on geometry and memory summaries, trained with teacher forcing followed by multi-step unrolling under the recorded continuation and accumulated once across the two-second window. Return timing uses a right-censored discrete hazard loss $\mathcal{L}_{\mathrm{ret}}$ while $\mathcal{L}_{\mathrm{cost}}$ regresses compute and motion demand. Over uncertain annotations, each branch retains its compatible loss interval so $\mathcal{L}_{\mathrm{rank}}$ orders a pair only when one branch's upper loss bound lies strictly below the other's lower bound, while timing, cost, and observable transitions stay supervised, and $\mathcal{L}_{\mathrm{cal}}$ penalizes overconfident wrong-binding predictions on held-out prefixes.

All loss and planner weights are chosen once on validation scene groups, and the same weights and stopping point serve every test episode. Validation selects loss weights $\gamma$ of $\BAval{weight.gamma.state}$, $\BAval{weight.gamma.return}$, $\BAval{weight.gamma.cost}$, $\BAval{weight.gamma.rank}$, and $\BAval{weight.gamma.calibration}$ for $\mathcal{L}_{\mathrm{state}}$, $\mathcal{L}_{\mathrm{ret}}$, $\mathcal{L}_{\mathrm{cost}}$, $\mathcal{L}_{\mathrm{rank}}$, and $\mathcal{L}_{\mathrm{cal}}$. During online beam search over a two-second horizon with depth three and beam width eight, soft compute and motion penalties both carry weight $\BAval{weight.soft.compute.penalty}$, particle spread carries weight $\beta_s=\BAval{weight.uncertainty.weight}$, and binding weights are $\lambda_{\mathrm{w}}=\BAval{lambda.w}$ and $\lambda_{\mathrm{u}}=\BAval{lambda.u}$. History-encoding cache maintenance is charged to decision overhead, and all learned comparators share the same batching.

%% file: supp/s02_data_adapters.tex
\section{Data and Baseline Adaptations}\label{sec:sdata}
All comparisons enforce scene-group independence across data splits and preserve each baseline's native perception, memory, and learning mechanism under shared PTZ action spaces and resource caps.

\subsection{Panoramic rendering and scene-group splits}
Rendering PTZ viewports on demand from panoramic recordings supplies controllable camera motion together with multi-person identity ground truth. Indoor and outdoor recordings come from QuadTrack and JRDB~\cite{omni,jrdb,panotrack}, where each selected camera action projects only the acquired viewport with its calibrated pose and resolution while withholding the unviewed panorama. Identity evaluation requires labels for both the designated person and competing people around crossings, which QuadTrack and JRDB provide whereas 360VOT~\cite{vot360} annotates a single target only.

Grouping every recording by physical scene prevents overlapping subjects, backgrounds, and camera setups from leaking across splits. Adjacent clips from one recording and separate recordings of the same physical location always share a scene group rather than splitting by timestamp. This partition yields \BAval{t09.r01.groups} training, \BAval{t09.r02.groups} validation, and \BAval{t09.r03.groups} test scene groups containing \BAval{t09.r01.episodes}, \BAval{t09.r02.episodes}, and \BAval{t09.r03.episodes} episodes, respectively. Each episode fixes the initial reference box, designated person, duration, reappearance events, and admissible camera region. The test split comprises \BAval{plan.strata.short.episodes} short-occlusion, \BAval{plan.strata.similar.episodes} similar-person distractor, and \BAval{plan.strata.long.episodes} long-absence episodes combined with pre-fixed weights \BAval{plan.strata.short.weight}, \BAval{plan.strata.similar.weight}, and \BAval{plan.strata.long.weight}.

\subsection{Identity labels and pre-defined reappearances}
Admitted episodes separate confirmed target binding, confirmed wrong binding, localization failure, target absence, and annotation uncertainty. Every admitted episode provides a continuous identity for the designated target, boxes or masks for overlap scoring, target-presence flags, and identities for competing people during ambiguous encounters. Clearly identified distractors are never marked unknown, and indistinguishable candidates are never assigned the target identity by proximity to the prior box. An unknown annotation marks evaluation uncertainty rather than the tracker's unresolved state $\mathrm{u}$, and keeping uncertain spans in the full episode duration defines both the conservative lower bound $\sid$ and an upper bound that adds only uncertain spans consistent with the target identity and IoU criterion, alongside the count and duration of uncertain intervals.

Reappearance events are fixed from the source annotation before any method executes so no policy can inflate recovery by detecting only easy returns. Each event records when the designated target re-enters the admissible scene, whether a legal camera view can reach it, and whether visual cues distinguish its identity, enabling separate analysis of recoverable, visually ambiguous, and unreachable absences.

\subsection{Common PTZ controller for perception trackers}
Every fixed-stream perception tracker in main Table~1 operates under the same causal common PTZ controller while keeping its native representation, candidate logic, and memory update intact. Because fixed-stream trackers do not output camera moves, the common PTZ controller centers the predicted target while tracker confidence stays above a threshold tuned on validation scenes and switches to a shared wide-area search schedule below that threshold.

All perception trackers also share identical viewport delivery, queue discipline, and compute accounting. Each request receives an already acquired viewport with its geometry, preserving native template, crop, and resolution conventions within the shared compute cap while charging all resampling and feature extraction. When a tracker falls behind the requested update rate, the shared runtime rule governs queueing, cancellation, and next-operator dispatch, charging spent compute on canceled or failed requests and selecting initialization settings on validation scenes alone.

\subsection{One-stream and temporal trackers}
Five one-stream and temporal trackers test whether stronger visual representations or frame-to-frame history suffice under the common PTZ controller. OSTrack provides the one-stream transformer baseline~\cite{ostrack}, MixFormerV2 supplies an efficient transformer operating point~\cite{mixformerv2}, HIPTrack and ODTrack incorporate historical prompts and dense temporal tokens~\cite{hiptrack,odtrack}, and LoRATv2 adapts large pretrained backbones with low-cost temporal modeling~\cite{lorat,loratv2}, all running with pretrained heads and native temporal updates intact.

Holding a perception backbone fixed across schedulers, as in main Table~2, isolates the effect of timed camera and operator decisions. Within each backbone row, all compared schedulers start from identical frozen checkpoints, action catalogs, operator implementations, native update rules, cost models, and initial tracker and queue states, so performance differences come from the views, operators, and start times each policy chooses.

\subsection{Long-term re-detection and distractor memory}
Long-term and distractor-aware trackers retain their native re-detection, candidate association, and memory protection strictly over acquired PTZ views. LTMU runs its meta-updated long-term verifier and global re-detector across the acquired viewport without accessing the hidden panorama~\cite{ltmu}, and KeepTrack maintains its learned target--candidate association over inferred rather than ground-truth hypotheses~\cite{keeptrack}. DAM4SAM preserves its distractor-aware segmentation memory~\cite{dam4sam}, initialized from the same first-frame reference as all baselines and converted from mask to box by one fixed rule that counts an empty mask as missing.

Pairing frozen DAM4SAM with alternative joint policies over the same 48-action catalog isolates the decision mechanism in main Tables~1 and~4. Combining DAM4SAM with an external joint-belief planner reaches \BAval{ext.joint.sid}\% identity success in main Table~1, whereas pairing the internal joint scheduler with the capacity-matched reference response reaches \BAval{int.joint.sid}\% in main Table~4.

\subsection{Active camera controllers in PTZ space}
Adapting embodied active trackers to a fixed-center PTZ camera replaces translational motion around obstacles with the shared pan, tilt, and zoom catalog while preserving each method's control architecture and training principle. Because a fixed-center camera rotates and zooms without translating, recovery trajectories are learned directly over legal PTZ moves and causal observations.

Both active baselines retain their core learning mechanisms under this PTZ action space. AD-VAT+-PTZ preserves asymmetric adversarial target--tracker training~\cite{advat} over the shared PTZ catalog using training-split scene groups and identities, supervised during privileged training by physical-clock rewards that penalize wrong and unresolved spans while deploying strictly on causal history. OA-VAT-PTZ preserves instance prototypes, online queue-based prototype enhancement from completed results, confidence-aware tracking, and its conditional-diffusion planner retrained for PTZ occlusion recovery~\cite{oavat}, and every initial reference or augmentation it receives is given to all other methods as well.

\subsection{Latency extrapolation and adaptive capacity}
PVT++ and ARTrack-AC test whether motion extrapolation or per-frame capacity switching can compensate for asynchronous return delays without steering views. PVT++ extrapolates boxes from completed causal states with its prediction overhead charged to compute~\cite{pvtpp}, refining spatial location without acquiring fresh identity cues and evaluated under both its released latency-aware evaluator and streaming overlap. ARTrack-AC switches between its native high-capacity and low-capacity modes from causal history~\cite{artrackac}, charging both mode selection and any probe pass to the shared compute cap so adaptive capacity never receives uncounted computation.

\subsection{Joint runtime schedulers and value baselines}
Chanakya-J and temporal direct-$Q$ share BindAdapt's causal inputs and 48-action catalog to isolate what return-time binding prediction adds beyond latency-aware accuracy and scalar value regression. Extending Chanakya~\cite{chanakya} to choose camera trajectory, visual operator, and start-time offset including waiting, Chanakya-J receives the same causal history summary, verification queue, queue ages, and budget state as BindAdapt while retaining Chanakya's contextual decision model and latency-aware accuracy reward. Main Table~1 reports both Chanakya-J with its built-in tracker and DAM4SAM + Chanakya-J, whereas comparisons that fix the backbone across rows, such as main Tables~2 and~6 and Table~\ref{tab:sschedulingupdated}, refer to the shared-backbone policy concisely as Chanakya-J. Temporal direct-$Q$ instead trains at matched model capacity on BindAdapt's common-prefix branches using the scalar identity loss, isolating the gain of time-resolved binding and return-time supervision and every comparator pays its own decision cost.

%% file: supp/s03_measures.tex
\section{Evaluation Measures}\label{sec:smetrics}
Physical-clock identity metrics measure how long a tracker stays bound to the designated person, complementing standard overlap, long-term absence, latency-aware, and camera-centering measures by reporting correct, wrong, and unresolved binding time separately.

\subsection{Overlap and center precision}
Fixed-view replay weights visible frames equally whereas streaming overlap weights each causal output by its physical duration. Given a predicted box $B_t$ and the designated-target box $G_t$, frame overlap is $o_t=|B_t\cap G_t|/|B_t\cup G_t|$ with missing required predictions scored as zero. Over $N$ target-visible samples, equal-weight average overlap is $\mathrm{AO}=N^{-1}\sum_t o_t$ and duration-weighted overlap is $\sum_t w_t o_t/\sum_t w_t$. Integrating the success curve $S(\alpha)=N^{-1}\sum_t\mathbf1[o_t\geq\alpha]$ across thresholds recovers average overlap,
\begin{equation}
\int_0^1S(\alpha)\,d\alpha=\frac1N\sum_t o_t=\mathrm{AO},
\end{equation}
so AUC and AO coincide under continuous thresholds and identical aggregation while discrete benchmark grids introduce minor differences, and we additionally report GOT-10k success rates at $0.5$ and $0.75$~\cite{got10k}.

Center-error metrics complement box overlap at fixed and target-normalized scales. From center offsets $\Delta x_t$ and $\Delta y_t$, $P_{20}$ measures the share of visible samples with $\sqrt{\Delta x_t^2+\Delta y_t^2}\leq20$ pixels at fixed resolution~\cite{otb}, while normalized error $e_t=\sqrt{(\Delta x_t/w_t^{\rm gt})^2+(\Delta y_t/h_t^{\rm gt})^2}$ defines normalized precision at the fixed threshold $0.20$, $P_N^{.20}=N^{-1}\sum_t\mathbf1[e_t\leq0.20]$, which we report as an endpoint score distinct from the integrated normalized precision of LaSOT~\cite{lasot} and TrackingNet~\cite{trackingnet}.

\subsection{Physical-clock identity and recovery}
\begin{figure}[t]
\centering\includegraphics[width=\columnwidth]{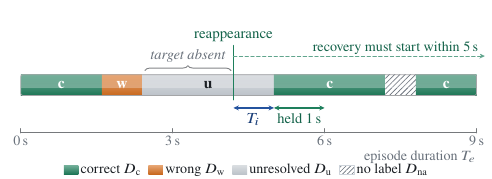}
\caption{\textbf{Scoring one episode on the physical clock.} Identity success and wrong binding are the shares of correct and wrong time in the full duration $T_e$, and time without a definite label stays in the denominator. Recovery must start within five seconds of reappearance and then hold for one second, and $T_i$ runs to its start. Durations are illustrative.}
\label{fig:sbands}
\end{figure}

Each episode duration $T_e$ is partitioned on the physical clock into mutually exclusive intervals of correct binding $D_{\mathrm{c},e}$, wrong binding $D_{\mathrm{w},e}$, unresolved tracker output $D_{\mathrm{u},e}$, and insufficient annotation $D_{\mathrm{na},e}$ so that $T_e=D_{\mathrm{c},e}+D_{\mathrm{w},e}+D_{\mathrm{u},e}+D_{\mathrm{na},e}$, as illustrated in Figure~\ref{fig:sbands}. Correct binding requires the designated target to be in view with box IoU at least $0.5$. Otherwise, wrong binding requires box IoU at least $0.5$ with an annotated competing person, so a box covering both counts as correct, and all remaining annotated ticks, including missing outputs, target absence, ambiguous candidates, and low IoU, count as unresolved.

Keeping insufficiently annotated time in the full denominator makes both episode rates conservative lower bounds on confirmed durations. Episode identity success $S_{\mathrm{ID},e}=D_{\mathrm{c},e}/T_e$ and wrong-binding rate $W_e=D_{\mathrm{w},e}/T_e$, reported as percentages, exclude $D_{\mathrm{na},e}$ from both numerators and convert from scorable-time rates $W^{\rm sc}_e=D_{\mathrm{w},e}/(T_e-D_{\mathrm{na},e})$ via $W_e=W^{\rm sc}_e(1-D_{\mathrm{na},e}/T_e)$ per episode prior to averaging. Episodes receive equal weight within each event stratum $s$ as $W_s=|E_s|^{-1}\sum_{e\in E_s}W_e$ over admitted episodes $E_s$, and both $\sid$ and $W$ combine strata as $\sum_s\omega_s W_s$ with weights $0.25$, $0.375$, and $0.375$ for short occlusions, similar-person distractors, and long absences while treating scene groups as the independent sampling units.

Recovery measures require sustained re-binding rather than a momentary overlap after reappearance. Once the designated target re-enters the admissible scene and can be evaluated, an event counts as recovered if correct binding resumes within five seconds and holds continuously for one second, with $T_i$ measuring elapsed time from reappearance to the start of that sustained interval. Over all $N_{\rm rec}$ admitted reappearance events including failures, five-second recovery $R$ is the recovered fraction and mean recovery time is $T_5=N_{\rm rec}^{-1}\sum_i\min(T_i,5)$, with five seconds assigned to each unrecovered event.

\subsection{Long-term precision, recall, and F-score}
Long-term tracking precision and recall jointly penalize false box reports during absence and missed localizations during presence~\cite{ltmetrics}. At confidence threshold $\xi$, let $N_p(\xi)$ count nonempty predictions and $N_g$ count target-present samples within a sequence. Dividing summed intersection overlap by $N_p(\xi)$ and $N_g$ yields sequence precision and recall, whose equal-weight sequence averages $\mathrm{Pr}(\xi)$ and $\mathrm{Re}(\xi)$ combine as
\begin{equation}
F(\xi)=\frac{2\mathrm{Pr}(\xi)\mathrm{Re}(\xi)}{\mathrm{Pr}(\xi)+\mathrm{Re}(\xi)}.
\end{equation}
Peak $F_{\max}$ selects a single best threshold across precision and recall, whereas operating $F_v$ applies the threshold fixed on validation scenes.

\begin{figure*}[!t]
\centering\includegraphics[width=\textwidth]{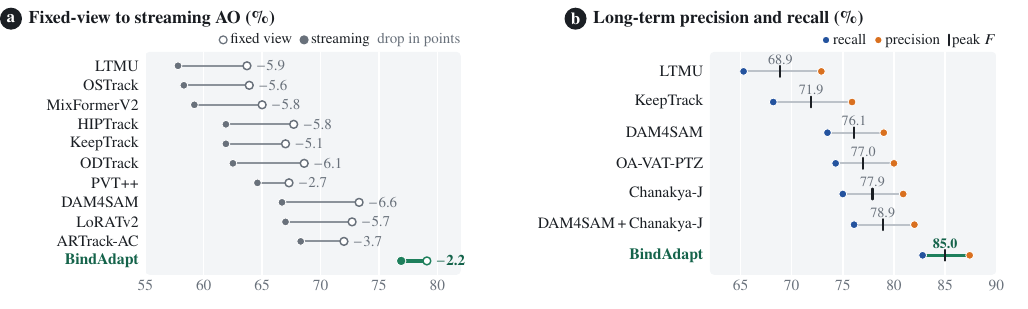}
\caption{\textbf{BindAdapt loses the least overlap under streaming evaluation and leads in long-term precision and recall.} The left panel connects fixed-view AO with streaming AO for every tracker in Table~\ref{tab:sstandard}, sorted by streaming AO, and BindAdapt drops by only 2.2 points. The right panel shows recall, precision, and the peak F-score at the threshold that maximizes each method's F-score from Table~\ref{tab:slt}.}
\label{fig:sstandard}
\end{figure*}

\begin{table*}[!t]\caption{\textbf{All standard and streaming measures on fixed-view replay (\%).} Every method receives the same frames. Main Table~1 reports AO and streaming AO.}\label{tab:sstandard}\centering\small\setlength{\tabcolsep}{4.8pt}\begin{tabularx}{\textwidth}{@{}l>{\centering\arraybackslash}X>{\centering\arraybackslash}X>{\centering\arraybackslash}X>{\centering\arraybackslash}X>{\centering\arraybackslash}X>{\centering\arraybackslash}X>{\centering\arraybackslash}X>{\centering\arraybackslash}X@{}}\toprule
Method & AUC/AO & $P_{20}$ & $P_N^{.20}$ & SR$_{.5}$ & SR$_{.75}$ & sAO & sSR$_{.5}$ & sSR$_{.75}$\\\midrule
\BAinputrows tables/sstandard.tex
\bottomrule\end{tabularx}\end{table*}

\subsection{Streaming versus native active metrics}
Scoring every baseline under both streaming overlap and its native clock separates timing sensitivity from protocol conventions. Streaming AO and streaming success evaluate the latest completed causal box at each physical tick~\cite{streaming}, whereas PVT++ applies frame-indexed latency deadlines~\cite{pvtpp} and active tracking baselines retain their source reward and success definitions. Unlike those localization protocols, $\sid$ requires the output box to stay on the designated person under a fixed physical-time denominator and absence rule.

\subsection{Latency-aware evaluation across grace deadlines}
We adopt the frame-indexed deadline and grace grid of the released PVT++ evaluator~\cite{pvtpp} to test whether accuracy depends on timing grace. Frame $i$ at frame rate $f$ receives deadline $t_i=(i+\sigma)/f$ for grace fraction $\sigma$, scoring the latest prediction completed by $t_i$ via AUC and 20-pixel distance precision and our summary integrates over $\sigma\in\{0,0.02,\ldots,1.0\}$ by the trapezoidal rule to obtain mAUC and mDP. At zero grace La$_0$, each method's service latency applies in full without extra allowance, so predictive motion extrapolation still takes effect only upon completion, and shared frames, temporal weights, and completion records make AUC@La$_0$ equal sAO for every shared method while larger $\sigma$ relaxes the deadline.

\subsection{Camera alignment and the identity trade-off}
We adapt the correct action rate CAR of OA-VAT~\cite{oavat} to the shared PTZ action space, where it scores single-step camera alignment on a shared view set rather than closed-loop target identity. Projecting the designated target's center under each legal PTZ move onto a common prerecorded view, an action succeeds outside a fixed centering deadzone only if it strictly reduces center distance relative to holding still, with ties counted as failures, while holding still succeeds inside the deadzone. CAR therefore tests whether an action steers the optical axis toward the target, whereas $\sid$ tests whether the tracker's box and memory stay bound to that person across crossings and absences.

%% file: supp/s04_robustness.tex
\section{Robustness of the Main Comparison}\label{sec:sresults}
BindAdapt keeps its identity lead across every event stratum, external localization and absence protocol, multi-object association measure, random seed, and scene-group bootstrap, and it nearly matches the best camera-centering controller. Here and in the next two sections, every reported difference comes from the unrounded scores.

\subsection{Event strata}
\begin{table}[t]\caption{\textbf{Identity success $\sid$ (\%) by event stratum.} The weighted column combines the strata with weights $0.25$, $0.375$, and $0.375$ for every method.}\label{tab:sstrata}\centering\footnotesize\setlength{\tabcolsep}{3pt}\begin{tabularx}{\columnwidth}{@{}l>{\centering\arraybackslash}X>{\centering\arraybackslash}X>{\centering\arraybackslash}X>{\centering\arraybackslash}X@{}}\toprule
Method & Short & Similar & Long & Weighted\\\midrule
\BAinputrows tables/sstrata.tex
\bottomrule\end{tabularx}\end{table}
BindAdapt improves identity success in every event stratum and achieves its widest margin on long absences, where delayed results are most likely to describe a person who has already left the view. Table~\ref{tab:sstrata} shows that across short occlusions, similar-person distractors, and long absences, BindAdapt raises $\sid$ from $\BAval{t14.r05.short}\%$, $\BAval{t14.r05.similar}\%$, and $\BAval{t14.r05.long}\%$ for DAM4SAM with Chanakya-J to $\BAval{t14.r06.short}\%$, $\BAval{t14.r06.similar}\%$, and $\BAval{t14.r06.long}\%$, widening the gain from $\BAval{delta.strata.short}$ and $\BAval{delta.strata.similar}$ points to $\BAval{delta.strata.long}$ points.

\subsection{Long-term, latency-aware, and fixed-view benchmarks}
\begin{table}[t]\caption{\textbf{Long-term tracking precision, recall, and F-score (\%).} Stars denote the common confidence threshold that achieves the peak F-score, and subscript $v$ denotes the threshold fixed on validation.}\label{tab:slt}\centering\footnotesize\setlength{\tabcolsep}{3pt}\begin{tabularx}{\columnwidth}{@{}l>{\centering\arraybackslash}X>{\centering\arraybackslash}X>{\centering\arraybackslash}X>{\centering\arraybackslash}X>{\centering\arraybackslash}X>{\centering\arraybackslash}X@{}}\toprule
Method & Pr$^*$ & Re$^*$ & $F_{\max}$ & Pr$_v$ & Re$_v$ & $F_v$\\\midrule
\BAinputrows tables/slt.tex
\bottomrule\end{tabularx}\end{table}
\begin{table}[t]\caption{\textbf{Latency-aware tracking with the released PVT++ evaluator (\%).} mAUC and mDP integrate the grace sweep $\sigma\in[0,1]$ by the trapezoidal rule, which changes the time allowed for each prediction while keeping the source frames fixed.}\label{tab:selae}\centering\footnotesize\setlength{\tabcolsep}{3pt}\begin{tabularx}{\columnwidth}{@{}l>{\centering\arraybackslash}X>{\centering\arraybackslash}X>{\centering\arraybackslash}X>{\centering\arraybackslash}X@{}}\toprule
Method & AUC@La$_0$ & DP@La$_0$ & mAUC & mDP\\\midrule
\BAinputrows tables/selae.tex
\bottomrule\end{tabularx}\end{table}
BindAdapt raises long-term precision and recall together rather than trading abstention during absence against rediscovery after return. Precision penalizes false box reports while the target is absent, and recall penalizes missed localizations while the target is visible. Table~\ref{tab:slt} shows that BindAdapt lifts peak $F_{\max}$ from $\BAval{t16.r06.fmax.pct}\%$ for DAM4SAM with Chanakya-J to $\BAval{t16.r07.fmax.pct}\%$ and lifts operating $F_v$ under the validation threshold from $\BAval{t16.r06.f.operating.pct}\%$ to $\BAval{t16.r07.f.operating.pct}\%$.

Under the released PVT++ evaluator, the advantage holds at zero grace and over the integrated grace range. Varying the post-acquisition deadline fraction $\sigma$ in Table~\ref{tab:selae}, BindAdapt reaches $\BAval{t19.r06.auc.la0.pct}\%$ in AUC@La$_0$ and $\BAval{t19.r06.mauc.pct}\%$ and $\BAval{t19.r06.mdp.pct}\%$ in mAUC and mDP, outperforming Chanakya-J at $\BAval{t19.r05.auc.la0.pct}\%$, $\BAval{t19.r05.mauc.pct}\%$, and $\BAval{t19.r05.mdp.pct}\%$. At zero grace, each shared fixed-view run uses the same completed outputs and source frames as streaming overlap.

On fixed-view replay, BindAdapt leads every overlap and center-precision metric while losing the least accuracy when scored on the physical streaming clock. Across Table~\ref{tab:sstandard} and Figure~\ref{fig:sstandard}, BindAdapt reaches $\BAval{t13.r11.ao}\%$ AO, $\BAval{t13.r11.p20.pct}\%$ $P_{20}$, and $\BAval{t13.r11.pn20}\%$ $P_N^{.20}$ and drops by only $2.2$ points to $\BAval{t13.r11.sao.pct}\%$ in streaming AO, whereas standalone DAM4SAM drops from $\BAval{t13.r08.ao}\%$ to $\BAval{t13.r08.sao.pct}\%$.

\subsection{Camera alignment}
BindAdapt nearly matches specialized active controllers on single-step geometric centering while keeping identity far better. Table~\ref{tab:scar} shows that BindAdapt reaches $\BAval{t21.r05.car.pct}\%$ CAR against $\BAval{t21.r03.car.pct}\%$ for OA-VAT-PTZ, exceeding both Chanakya-J at $\BAval{t21.r04.car.pct}\%$ and the common PTZ controller at $\BAval{t21.r01.car.pct}\%$. Whereas OA-VAT-PTZ optimizes view centering yet reaches $\BAval{t14.r02.weighted}\%$ $\sid$, BindAdapt reaches $\BAval{t14.r06.weighted}\%$ $\sid$, since a camera can move toward the designated person after the tracker has already switched to a look-alike.

\begin{table}[t]\caption{\textbf{Correct action rate (CAR, \%) adapted to projected PTZ centering.} CAR scores whether a legal action moves the original target toward the center of the projected view.}\label{tab:scar}\centering\small\setlength{\tabcolsep}{4pt}\begin{tabularx}{\columnwidth}{@{}l>{\centering\arraybackslash}X@{}}\toprule
Policy & PTZ CAR\\\midrule
\BAinputrows tables/scar.tex
\bottomrule\end{tabularx}\end{table}

\subsection{Seed variation, paired bootstrap, and multi-object metrics}\label{sec:sseeds}
Run-to-run variation across training seeds is far smaller than BindAdapt's gain under a matched validation budget. Every learned policy selects among $24$ configurations with three seeds per configuration on the $20$ validation scene groups. Across all eight learned methods in Table~\ref{tab:sseeds}, the largest three-seed standard deviation of $\sid$ is $0.8$ points and BindAdapt's is $0.3$ points, compared with its $7.1$-point lead over DAM4SAM with Chanakya-J.

\begin{table}[t]\caption{\textbf{Mean and seed standard deviation of the learned methods.} Scores in \%, $T_5$ in seconds, three seeds each.}\label{tab:sseeds}\centering\footnotesize\setlength{\tabcolsep}{2pt}
\begin{tabularx}{\columnwidth}{@{}Xcccc@{}}\toprule
Method & $\sid\uparrow$ & $W\downarrow$ & $R\uparrow$ & $T_5\downarrow$\\\midrule
AD-VAT+-PTZ & 66.5$\pm$0.3 & 10.9$\pm$0.20 & 68.0$\pm$0.8 & 2.57$\pm$0.03\\
OA-VAT-PTZ & 72.8$\pm$0.2 & 7.8$\pm$0.03 & 76.1$\pm$0.2 & 2.12$\pm$0.03\\
PVT++ & 68.0$\pm$0.4 & 10.1$\pm$0.11 & 69.0$\pm$0.4 & 2.50$\pm$0.01\\
ARTrack-AC & 71.6$\pm$0.3 & 8.6$\pm$0.09 & 73.9$\pm$0.4 & 2.21$\pm$0.01\\
Chanakya-J & 74.1$\pm$0.6 & 7.4$\pm$0.08 & 77.1$\pm$0.1 & 2.09$\pm$0.06\\
DAM4SAM + joint belief & 74.6$\pm$0.8 & 6.9$\pm$0.11 & 77.9$\pm$0.5 & 2.00$\pm$0.01\\
DAM4SAM + Chanakya-J & 75.2$\pm$0.2 & 7.3$\pm$0.16 & 79.0$\pm$0.6 & 2.03$\pm$0.01\\
\rowcolor{BAours}BindAdapt & \textbf{82.3}$\pm$0.3 & \textbf{4.4}$\pm$0.15 & \textbf{86.2}$\pm$0.2 & \textbf{1.39}$\pm$0.06\\\bottomrule
\end{tabularx}\end{table}

Paired scene-group resampling and standard multi-object association metrics independently confirm this gain. Resampling the $40$ test scene groups with replacement $10{,}000$ times while evaluating both methods on identical resampled groups yields a paired $\sid$ gain of $7.06$ points with a 95\% confidence interval of $[5.14, 9.03]$ excluding zero. On the test episodes, BindAdapt also raises identity true positives from $73{,}562$ for DAM4SAM with Chanakya-J to $80{,}391$ while cutting identity false positives from $18{,}129$ to $13{,}944$ and false negatives from $24{,}638$ to $17{,}809$, lifting IDF1 from $77.48\%$ to $83.51\%$ and HOTA from $61.84\%$ to $67.28\%$.

%% file: supp/s05_mechanism.tex
\section{Timing, Resources, and Mechanism}\label{sec:smechanism}
Timing audits, controlled sweeps, and ablations show that BindAdapt runs at matched service cost, leads with or without added waiting, and draws its gain from when evidence returns relative to memory updates.

\subsection{Device-neutral profiles and physical return latency}
\begin{table}[t]\caption{\textbf{Device-neutral service times.} Mean service includes decision overhead, P95 is the 95th percentile, and nominal FPS equals $1000$ divided by the unrounded mean in milliseconds. Added waiting is excluded from this rate.}\label{tab:sruntime}\centering\footnotesize\setlength{\tabcolsep}{3pt}\begin{tabularx}{\columnwidth}{@{}l>{\centering\arraybackslash}X>{\centering\arraybackslash}X>{\centering\arraybackslash}X>{\centering\arraybackslash}X@{}}\toprule
Method & Mean ms & P95 ms & Nominal FPS & Decision ms\\\midrule
\BAinputrows tables/sruntime.tex
\bottomrule\end{tabularx}\end{table}
\begin{table}[t]\caption{\textbf{Measured stages of one request on the physical PTZ device.} Times in ms over the same requests. Stage means add up to the mean total before rounding, and the total P95 is taken over per-request totals.}\label{tab:slatencyupdated}\centering\small\setlength{\tabcolsep}{4pt}\begin{tabularx}{\columnwidth}{@{}l>{\centering\arraybackslash}X>{\centering\arraybackslash}X@{}}\toprule
Stage & Mean & P95\\\midrule
Camera motion & 118.6 & 301.8\\
Settling after motion & 78.4 & 164.7\\
Capture and frame availability & 29.8 & 50.1\\
Queueing after frame availability & 47.8 & 141.4\\
Computation & 37.4 & 51.7\\
Postprocessing and transfer & 20.6 & 47.3\\\midrule
Command to usable result & 332.7 & 679.2\\\bottomrule
\end{tabularx}\end{table}
BindAdapt matches the per-step service time of the strongest baseline when its scheduling decision is charged in full. Under the hardware-independent profiles of Table~\ref{tab:sruntime}, which fix visual inference, decision computation, data movement, and declared queueing across processors and batch sizes, BindAdapt runs at $\BAval{t18.r09.mean.inference.ms}$\,ms mean and $\BAval{t18.r09.p95.inference.ms}$\,ms P95 service time including its $\BAval{t18.r09.decision.overhead.ms}$\,ms decision cost, matching DAM4SAM with Chanakya-J at $\BAval{t18.r08.mean.inference.ms}$\,ms mean and $\BAval{t18.r08.p95.inference.ms}$\,ms P95, or $\BAval{t18.r09.fps}$ versus $\BAval{t18.r08.fps}$ nominal FPS.

Queue occupancy and variable transfer times separate per-step service time from the physical moment a result updates tracker memory. One non-preemptive first-in, first-out worker executes at most three dispatched requests in flight, releasing the worker after processing while transfer and $0.5$\,s of added waiting at the main stress operating point precede publication. At the three-slot limit, new compute dispatches wait while the waiting action remains legal and camera motion overlaps processing and transfer. Because results apply in publication order under native tracker transitions, a slow earlier request can arrive after a fast later one, where a delayed identity cue can still resolve ambiguity whereas a stale local box can reinforce a wrong person. The identity response $p_\theta$ predicts this outcome from the memory state at arrival, and the compute charge stays in the ledger even when the tracker rejects a returned result.

Stage measurements on the physical PTZ camera confirm that camera motion, settling, and queueing dominate total return latency. Across the six stages of Table~\ref{tab:slatencyupdated}, one request takes $332.7$\,ms on average and $679.2$\,ms at P95 from command to usable result, with camera motion and settling taking $197.0$\,ms, capture taking $29.8$\,ms, and post-capture queueing, postprocessing, and transfer taking $68.5$\,ms against only $37.4$\,ms for computation. BindAdapt schedules against this full return latency, which the $0.5$\,s added waiting stresses further in simulation while hardware experiments use measured timing alone.

\subsection{Shared budget accounting and hard caps}
Every method operates under identical compute and motion caps with upfront upper-bound reservation and no refund for spent work. Paid compute sums feature extraction, wide-view search, verification, response inference, planning, and cache maintenance against a shared normalized cap of $1.0$, while motion cost sums angular travel and zoom within hard speed, acceleration, and view-reachability bounds. Soft compute and motion penalties act only among feasible plans, and every scheduler reserves a request's declared upper bound before dispatch, reconciles actual cost upon completion, releases only unspent work upon cancellation, and charges completed results even if native association rejects them.

\subsection{Delay and compute sweeps}
BindAdapt already leads with zero added waiting, and its advantage grows monotonically with the added delay. Across the post-computation added waiting sweep of $0$, $0.1$, $0.25$, $0.5$, $1.0$, and $2.0$\,s in Table~\ref{tab:sdelayupdated}, corresponding to main Figure~4a, BindAdapt leads DAM4SAM with Chanakya-J by $\BAval{delta.sweep.zero}$ points at $0$\,s, $\BAval{delta.sweep.ref}$ points at $0.5$\,s, and $\BAval{delta.sweep.two}$ points at $2.0$\,s, while keeping wrong binding between $4.0\%$ and $10.3\%$ against $5.8\%$ and $15.9\%$ for the baseline.

\begin{table}[t]\caption{\textbf{Added waiting with the DAM4SAM backbone.} Base is DAM4SAM with Chanakya-J, $\sid$ and $W$ are in \%, and $\Delta$ is the $\sid$ gain of BindAdapt in points, computed before rounding.}\label{tab:sdelayupdated}\centering\small\setlength{\tabcolsep}{3pt}\begin{tabularx}{\columnwidth}{@{}Xccccc@{}}\toprule
\multirow{2}{*}{Added (s)} & \multicolumn{2}{c}{$\sid\uparrow$} & \multirow{2}{*}{$\Delta$} & \multicolumn{2}{c}{$W\downarrow$}\\\cmidrule(lr){2-3}\cmidrule(l){5-6}
 & Base & BindAdapt & & Base & BindAdapt\\\midrule
$0$ & 80.1 & \textbf{84.2} & 4.1 & 5.8 & \textbf{4.0}\\
$0.1$ & 79.4 & \textbf{84.1} & 4.7 & 6.1 & \textbf{4.0}\\
$0.25$ & 77.8 & \textbf{83.5} & 5.6 & 6.6 & \textbf{4.2}\\
$0.5$ & 75.2 & \textbf{82.3} & 7.1 & 7.3 & \textbf{4.4}\\
$1.0$ & 68.5 & \textbf{78.1} & 9.7 & 10.2 & \textbf{6.3}\\
$2.0$ & 54.8 & \textbf{67.0} & 12.2 & 15.9 & \textbf{10.3}\\\bottomrule
\end{tabularx}\end{table}

BindAdapt also leads at every compute cap and beats the default-budget baseline while using a $0.75$ cap. Fixing motion limits across normalized compute caps of $0.5$, $0.75$, $1.0$, $1.25$, $1.5$, and $2.0$ as plotted in main Figure~4b, BindAdapt reaches $\BAval{t17.r07.bindadapt}\%$, $\BAval{t17.r08.bindadapt}\%$, $\BAval{t17.r09.bindadapt}\%$, $\BAval{t17.r10.bindadapt}\%$, $\BAval{t17.r11.bindadapt}\%$, and $\BAval{t17.r12.bindadapt}\%$ identity success against $\BAval{t17.r07.dam4sam.chanakya.j}\%$, $\BAval{t17.r08.dam4sam.chanakya.j}\%$, $\BAval{t17.r09.dam4sam.chanakya.j}\%$, $\BAval{t17.r10.dam4sam.chanakya.j}\%$, $\BAval{t17.r11.dam4sam.chanakya.j}\%$, and $\BAval{t17.r12.dam4sam.chanakya.j}\%$ for DAM4SAM with Chanakya-J. At the $0.75$ cap, BindAdapt already reaches $\BAval{budget.bind075.sid}\%$, surpassing the $\BAval{ext.best.sid}\%$ of DAM4SAM with Chanakya-J at the $1.0$ default cap.

\subsection{Return-time sweep}\label{sec:ssweep}
Holding the input frame and operator output fixed shows that shifting result arrival across a crossing sharply lowers target retention. To generate the inset of main Figure~1 across $243$ crossing episodes, one identity check is captured before the earliest $-1.0$\,s release so no condition reads a future frame, and only its release into the tracker shifts from $-1.0$\,s to $1.0$\,s relative to the crossing in $0.2$\,s steps while ordinary frame-by-frame updates continue in between. Table~\ref{tab:ssweep} shows that two-second retention stays high from $94.2\%$ across $229$ episodes at $-1.0$\,s to $88.5\%$ across $215$ episodes at $-0.2$\,s, then drops sharply across the crossing to $80.7\%$ at $0$\,s, $70.4\%$ at $0.2$\,s, and $56.8\%$ across $138$ episodes at $1.0$\,s because late arrivals enter a memory that look-alike updates have already altered.

\begin{table}[t]\caption{\textbf{Return-time sweep over $243$ crossing episodes.} Offset is the release time minus the crossing time, and kept counts the episodes in which the target is kept for two seconds.}\label{tab:ssweep}\centering\small\setlength{\tabcolsep}{4pt}
\begin{tabularx}{\columnwidth}{@{}>{\centering\arraybackslash}X>{\centering\arraybackslash}X>{\centering\arraybackslash}X@{\hspace{10pt}}>{\centering\arraybackslash}X>{\centering\arraybackslash}X>{\centering\arraybackslash}X@{}}\toprule
Offset (s) & Kept & Rate (\%) & Offset (s) & Kept & Rate (\%)\\\cmidrule(r{6pt}){1-3}\cmidrule{4-6}
$-1.0$ & 229 & 94.2 & $0.2$ & 171 & 70.4\\
$-0.8$ & 228 & 93.8 & $0.4$ & 158 & 65.0\\
$-0.6$ & 225 & 92.6 & $0.6$ & 146 & 60.1\\
$-0.4$ & 221 & 90.9 & $0.8$ & 141 & 58.0\\
$-0.2$ & 215 & 88.5 & $1.0$ & 138 & 56.8\\
$0$ & 196 & 80.7 & & & \\\bottomrule
\end{tabularx}\end{table}

\subsection{Rule-based checks and error attribution}\label{sec:srules}
Heuristic rules that trigger checks by proximity or fixed period fall well short of BindAdapt, especially on long absences. Replacing only the scheduler under the DAM4SAM backbone and default budgets of main Table~3, validation over $24$ configurations selects trigger threshold $k=1.5$ for the proximity rule and period $T=0.6$\,s for the fixed-rate rule. Table~\ref{tab:srules} shows that the two rules reach $73.6\%$ and $71.4\%$ $\sid$ against $82.3\%$ for BindAdapt and $61.4\%$ and $58.9\%$ on long absences against $73.9\%$.

\begin{table}[t]\caption{\textbf{Rule-based checks by event type.} $\sid$ by event type, its weighted average, $W$, and $R$ are in \%, and $T_5$ is in seconds.}\label{tab:srules}\centering\small\setlength{\tabcolsep}{2pt}
\begin{tabularx}{\columnwidth}{@{}Xccccccc@{}}\toprule
Method & Short & Similar & Long & $\sid$ & $W$ & $R$ & $T_5$\\\midrule
Proximity rule & 85.1 & 78.2 & 61.4 & 73.6 & 8.0 & 78.3 & 1.76\\
Fixed-rate rule & 83.7 & 75.6 & 58.9 & 71.4 & 8.8 & 76.1 & 1.58\\
\rowcolor{BAours}BindAdapt & \textbf{90.2} & \textbf{85.4} & \textbf{73.9} & \textbf{82.3} & \textbf{4.4} & \textbf{86.2} & \textbf{1.39}\\\bottomrule
\end{tabularx}\end{table}
\begin{table}[t]\caption{\textbf{When the latest evidence returned before a wrong binding.} Counts on the test episodes. The last column gives the share returned after the crossing.}\label{tab:sattr}\centering\small\setlength{\tabcolsep}{2pt}
\begin{tabularx}{\columnwidth}{@{}l>{\centering\arraybackslash}X>{\centering\arraybackslash}X>{\centering\arraybackslash}X>{\centering\arraybackslash}X>{\centering\arraybackslash}X@{}}\toprule
Method & Total & After & Before & None & After (\%)\\\midrule
DAM4SAM & 137 & 89 & 29 & 19 & 65.0\\
KeepTrack & 151 & 87 & 41 & 23 & 57.6\\
DAM4SAM + Chanakya-J & 97 & 61 & 24 & 12 & 62.9\\
\rowcolor{BAours}BindAdapt & \textbf{53} & 18 & 27 & 8 & \textbf{34.0}\\\bottomrule
\end{tabularx}\end{table}

Error attribution on the test episodes shows that BindAdapt removes mainly the wrong bindings that follow evidence returned after a crossing. Tracing the latest returned identity evidence before each wrong binding in Table~\ref{tab:sattr}, post-crossing returns precede $89$ of $137$ wrong bindings at $65.0\%$ for DAM4SAM, $87$ of $151$ at $57.6\%$ for KeepTrack, and $61$ of $97$ at $62.9\%$ for DAM4SAM with Chanakya-J. BindAdapt lowers total wrong bindings from $97$ to $53$, and those after post-crossing returns fall from $61$ to $18$, now $34.0\%$ of its errors, whereas errors after pre-crossing returns stay comparable at $27$ versus $24$.

\subsection{Crossing the response with the scheduler}
Crossing two responses with two schedulers under matched parameters and branches confirms that response modeling and joint search contribute nearly additively. Extending main Table~4 across shared scenes, branches, budgets, and training allowance in Table~\ref{tab:sswap}, switching from separable to joint scheduling raises $\sid$ by $\BAval{delta.swap.refplanner}$ points from $\BAval{t03.r01.sid}\%$ to $\BAval{t03.r02.sid}\%$ under the reference response and by $\BAval{delta.swap.bindplanner}$ points from $\BAval{t03.r03.sid}\%$ to $\BAval{t03.r04.sid}\%$ under the BindAdapt response. Holding the scheduler fixed instead yields response gains of $\BAval{delta.swap.sepresponse}$ and $\BAval{delta.swap.jointresponse}$ points while halving wrong binding from $\BAval{t03.r02.wrong}\%$ to $\BAval{t03.r04.wrong}\%$ under joint search, with an interaction below half a point.

\begin{table}[t]\caption{\textbf{Crossed response--scheduler control.} $\sid$ and $W$ are percentages. Each difference in the text holds the other factor fixed.}\label{tab:sswap}\centering\small\setlength{\tabcolsep}{4pt}\begin{tabularx}{\columnwidth}{@{}l>{\centering\arraybackslash}X>{\centering\arraybackslash}X@{}}\toprule
Response / scheduler & $\sid$ & $W$\\\midrule
\BAinputrows tables/sswap.tex
\bottomrule\end{tabularx}\end{table}

\subsection{Component interventions}
Removing any graph cue, response input, or scheduling choice lowers identity success. Every intervention in Table~\ref{tab:sablation} runs as a complete pipeline under the shared DAM4SAM backbone, scene groups, and resource caps, replacing removed inputs with neutral representations and releasing unused compute from the ledger.

\begin{table}[t]\caption{\textbf{Component interventions under the common backbone and resource caps.} $\sid$ and $W$ are percentages, and $T_5$ is the mean recovery time in seconds.}\label{tab:sablation}\centering\small\setlength{\tabcolsep}{4pt}\begin{tabularx}{\columnwidth}{@{}l>{\centering\arraybackslash}X>{\centering\arraybackslash}X>{\centering\arraybackslash}X@{}}\toprule
Variant & $\sid$ & $W$ & $T_5$\\\midrule
\BAinputrows tables/sablation.tex
\bottomrule\end{tabularx}\end{table}

Fixing the computation type lowers $\sid$ most, from $\BAval{t04.r01.sid}\%$ to $\BAval{t04.r06.sid}\%$, and slows mean recovery time $T_5$ from $\BAval{t04.r01.t5}$ to $\BAval{t04.r06.t5}$\,s, so camera control alone does not explain the gain. Among response inputs, removing target conditioning raises wrong binding $W$ highest from $\BAval{t04.r01.wrong}\%$ to $\BAval{t04.r03.wrong}\%$ and drops $\sid$ to $\BAval{t04.r03.sid}\%$, followed by removing return time at $\BAval{t04.r02.sid}\%$, the memory summary at $\BAval{t04.r04.sid}\%$, source dependencies at $\BAval{t04.r05.sid}\%$, and the uncertainty term at $\BAval{t04.r08.sid}\%$. Forcing immediate starts drops $\sid$ to $\BAval{abl.imm.sid}\%$ with $\BAval{abl.imm.wrong}\%$ wrong binding and separable scheduling drops it to $\BAval{t04.r07.sid}\%$, so start-time offsets and joint search both contribute.

Stratified interventions show that return-time modeling and provenance edges matter most where memory changes or cues share a frame. Removing return time lowers $\sid$ from $\BAval{diag.late.full}\%$ to $\BAval{diag.late.ablated}\%$ when evidence returns after tracker memory changes, versus $\BAval{diag.early.full}\%$ to $\BAval{diag.early.ablated}\%$ when it returns before. Removing source dependencies lowers $\sid$ from $\BAval{diag.corr.full}\%$ to $\BAval{diag.corr.ablated}\%$ on shared-frame cues, versus $\BAval{diag.indep.full}\%$ to $\BAval{diag.indep.ablated}\%$ on independent sources, while the memory intervention withholds the summary only from $p_\theta$ without altering native tracker updates.

\subsection{Action ranking and decision regret}
Predicting time-resolved binding trajectories improves pairwise action ranking and cuts physical-time decision regret relative to scalar value regression. On each common prefix with realized branch loss $L(a)$ and predicted loss $\widehat L(a)$, pairwise ranking accuracy counts properly ordered legal action pairs excluding exact ties by a fixed rule, risk MAE averages $|L(a)-\widehat L(a)|$, calibration error compares binned wrong-binding predictions against observed branch frequencies, and action regret measures
\begin{equation}
\mathrm{Reg}=L\bigl(\arg\min_a\widehat L(a)\bigr)-\min_aL(a),
\end{equation}
which is nonnegative because both terms evaluate executed branches from the same prefix.

\begin{table}[t]\caption{\textbf{Decision diagnostics on common-prefix action branches.} Ranking and regret use the same set of legal candidate actions at each prefix, with the continuation and cost accounting fixed.}\label{tab:sresponse}\centering\footnotesize\setlength{\tabcolsep}{3pt}\begin{tabularx}{\columnwidth}{@{}l>{\centering\arraybackslash}X>{\centering\arraybackslash}X>{\centering\arraybackslash}X>{\centering\arraybackslash}X@{}}\toprule
Estimator & Rank \% & Risk MAE & Calib.\ error & Regret\\\midrule
\BAinputrows tables/sresponse.tex
\bottomrule\end{tabularx}\end{table}

Across these shared prefixes, Table~\ref{tab:sresponse} shows that the BindAdapt response orders $\BAval{t07.r04.pair.rank.pct}\%$ of candidate pairs against $\BAval{t07.r02.pair.rank.pct}\%$ for temporal direct-$Q$ and cuts regret from $\BAval{t07.r02.action.regret}$ to $\BAval{t07.r04.action.regret}$. Whereas ranking accuracy treats every misordered pair equally, regret weights mistakes by lost physical-time binding, penalizing a missed recovery window during a brief reappearance far more than swapping two nearly tied moves.

\subsection{Sensitivity to weights and search settings}\label{sec:ssens}
Performance stays stable across an eightfold range of binding weights and plateaus at the default search horizon, beam width, and depth. Varying one parameter at a time around $\lambda_{\mathrm{w}}/\lambda_{\mathrm{u}}=4$, $H=2$\,s, beam width $8$, and depth $3$ in Table~\ref{tab:ssens}, $\sid$ shifts by at most \BAval{lambda.sens.range} points across weight ratios from $1$ to $8$. Shortening the horizon to $1.0$\,s loses $3.6$ points whereas extending it to $3.0$\,s yields no further gain, and increasing beam width to $16$ or depth to $4$ adds at most $0.1$ points while decision latency rises to $3.6$ and $3.8$\,ms.

\begin{table}[t]\caption{\textbf{Sensitivity of BindAdapt.} $\sid$ in \% and mean decision time in ms, with defaults in bold.}\label{tab:ssens}\centering\small\setlength{\tabcolsep}{3pt}
\begin{tabularx}{\columnwidth}{@{}l>{\centering\arraybackslash}X>{\centering\arraybackslash}X>{\centering\arraybackslash}X>{\centering\arraybackslash}X@{}}\toprule
$\lambda_{\mathrm{w}}/\lambda_{\mathrm{u}}$ & 1 & 2 & \textbf{4} & 8\\
$\sid$ & \BAval{t24.r01.sid} & \BAval{t24.r02.sid} & \textbf{\BAval{t24.r03.sid}} & \BAval{t24.r04.sid}\\\midrule
Horizon $H$ (s) & 1.0 & 1.5 & \textbf{2.0} & 3.0\\
$\sid$ & 78.6 & 80.9 & \textbf{82.3} & 81.9\\
Decision (ms) & 1.38 & 1.72 & \textbf{2.09} & 2.73\\\midrule
Beam width & 4 & \textbf{8} & 16 & \\
$\sid$ & 79.8 & \textbf{82.3} & 82.4 & \\
Decision (ms) & 1.42 & \textbf{2.09} & 3.61 & \\\midrule
Depth & 2 & \textbf{3} & 4 & \\
$\sid$ & 79.9 & \textbf{82.3} & 82.4 & \\
Decision (ms) & 1.26 & \textbf{2.09} & 3.84 & \\\bottomrule
\end{tabularx}\end{table}

Across $10{,}000$ online updates under the reference service profile, decision latency averages $2.09$\,ms with a 95th percentile of $3.76$\,ms, well below the $10$\,ms planning limit.

%% file: supp/s06_transfer.tex
\section{Transfer and Recorded Footage}\label{sec:stransfer}
BindAdapt preserves its identity advantage over parallel verifiers and learned schedulers across held-out distribution shifts, lightweight hardware timing calibration, closed-loop physical PTZ runs, and recorded video clips.

\subsection{Asynchronous verification}
Adding return-time prediction already improves a parallel verifier, and coupling it with joint view, operator, and start-time control delivers a much larger gain. Adapting PTAV-style parallel tracking and verification~\cite{ptav} to the common PTZ controller with DAM4SAM and $0.5$\,s of added waiting in Table~\ref{tab:sschedulingupdated}, triggering checks periodically or at low confidence lifts $\sid$ from $70.3\%$ to $73.8\%$, and predicting check return times raises it to $76.2\%$, edging past Chanakya-J on the same backbone at $75.2\%$. Choosing camera motion, visual operator, and start-time offset together lifts BindAdapt by a further $6.1$ points to $82.3\%$, whereas dropping the return-time model or delayed starts lowers it back to $77.9\%$ and $76.2\%$, matching main Table~3.

\begin{table}[t]\caption{\textbf{Asynchronous verification under the common PTZ controller.} All rows use DAM4SAM, the shared operator set, and $0.5$\,s of added waiting. $\sid$ and $W$ are in \%, $T_5$ is in s, and ms is the compute service time.}\label{tab:sschedulingupdated}\centering\small\setlength{\tabcolsep}{2pt}\begin{tabularx}{\columnwidth}{@{}l>{\centering\arraybackslash}X>{\centering\arraybackslash}X>{\centering\arraybackslash}X>{\centering\arraybackslash}X@{}}\toprule
Scheduler & $\sid\uparrow$ & $W\downarrow$ & $T_5\downarrow$ & ms\\\midrule
Common PTZ controller & 70.3 & 8.1 & 2.2 & 36.2\\
Async.\ verification, fixed trigger & 73.8 & 7.4 & 2.1 & 37.6\\
Async.\ verification, predicted return & 76.2 & 6.6 & 1.9 & 37.7\\
Chanakya-J & 75.2 & 7.3 & 2.0 & 37.8\\
BindAdapt, no return-time model & 77.9 & 7.1 & 1.9 & 37.2\\
BindAdapt, immediate starts only & 76.2 & 7.1 & 1.7 & 37.3\\
\rowcolor{BAours}BindAdapt & \textbf{82.3} & \textbf{4.4} & \textbf{1.4} & 37.4\\\bottomrule
\end{tabularx}\end{table}

\subsection{Held-out generalization axes}
Isolating five distribution shifts one at a time shows that BindAdapt leads on unseen scenes, people, latency profiles, motion statistics, and tracker backbones. Unseen target identities test appearance generalization within familiar scene types, unseen service profiles test return-time extrapolation beyond training delays, unseen motion statistics alter crossing timing, and a changed backbone alters the native memory dynamics that returned evidence enters. Across all five rows of Table~\ref{tab:sgen}, with source scene groups kept disjoint across training, calibration, and model selection, BindAdapt outperforms both the separable scheduler and temporal direct-$Q$, reaching from $\BAval{t10.r01.bindadapt}\%$ on unseen scene groups to $\BAval{t10.r04.bindadapt}\%$ on unseen motion statistics.

\begin{table}[t]\caption{\textbf{Identity success $\sid$ (\%) under held-out conditions.} Each row holds out one factor among source groups, target identities, device delays, motion statistics, and the tracker backbone.}\label{tab:sgen}\centering\small\setlength{\tabcolsep}{2.2pt}\begin{tabularx}{\columnwidth}{@{}l>{\centering\arraybackslash}X>{\centering\arraybackslash}X>{\centering\arraybackslash}X@{}}\toprule
Held-out condition & Separable & Temp.\ direct-$Q$ & BindAdapt\\\midrule
\BAinputrows tables/sgen.tex
\bottomrule\end{tabularx}\end{table}

\subsection{Service-time calibration to physical latency}
Refitting only the service-time model adapts BindAdapt to physical hardware latency nearly as well as full retraining and requires no new training branches. Evaluating under the measured device latency without added waiting in Table~\ref{tab:scalibrationupdated}, uncalibrated BindAdapt already reaches $\BAval{svc.direct.sid}\%$. Calibrating only the service-time model with the binding response and scheduler frozen cuts return-time MAE from $64.8$ to $19.3$\,ms and raises $\sid$ to $\BAval{svc.recal.sid}\%$, coming within $\BAval{delta.svc.retrain}$ points of full retraining at $\BAval{svc.retrain.sid}\%$ and maintaining a $5.2$-point lead over cost-recalibrated Chanakya-J at $78.5\%$.

\begin{table}[t]\caption{\textbf{Transfer to the measured latency of the physical device.} No added waiting. Service-model calibration refits the service-time model with the response and the scheduler fixed, and full retraining refits every model. $\sid$ and $W$ are in \%, $T_5$ is in s, and MAE is the error in ms of the predicted return time.}\label{tab:scalibrationupdated}\centering\small\setlength{\tabcolsep}{2pt}\begin{tabularx}{\columnwidth}{@{}l>{\centering\arraybackslash}X>{\centering\arraybackslash}X>{\centering\arraybackslash}X>{\centering\arraybackslash}X@{}}\toprule
Method and calibration & $\sid\uparrow$ & $W\downarrow$ & $T_5\downarrow$ & MAE\\\midrule
Chanakya-J, recalibrated costs & 78.5 & 6.6 & 1.84 & --\\
BindAdapt, no calibration & 81.3 & 5.3 & 1.63 & 64.8\\
BindAdapt, service model only & 83.6 & 4.5 & 1.42 & 19.3\\
BindAdapt, full retraining & \textbf{84.0} & \textbf{4.4} & \textbf{1.39} & \textbf{18.7}\\\bottomrule
\end{tabularx}\end{table}

\begin{figure*}[!t]
\centering\includegraphics[width=\textwidth]{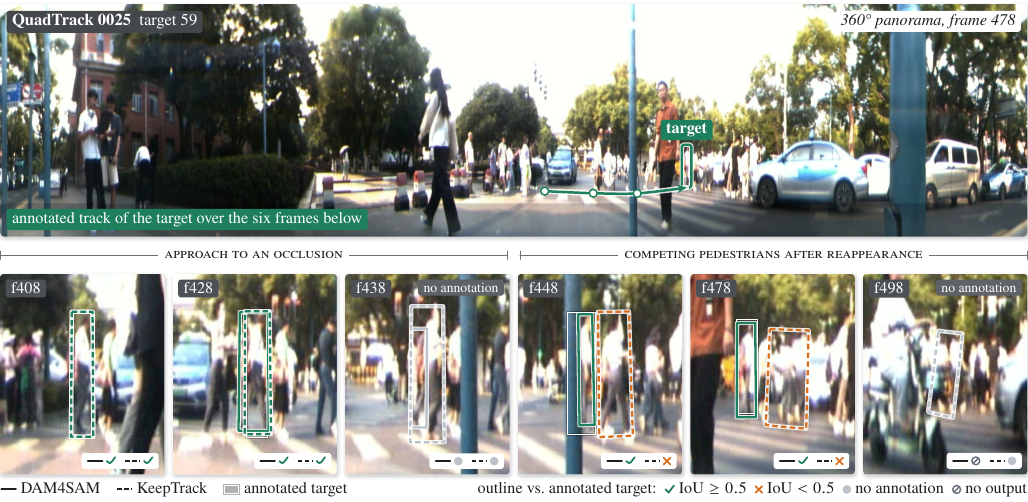}
\caption{\textbf{QuadTrack 0025, target 59. KeepTrack moves to a neighboring pedestrian after the target reappears.} The banner shows panorama frame 478 with the annotated track of the target over the six frames below. Each crop shows the archived DAM4SAM output as a solid outline and the KeepTrack output as a dashed one, with the annotated target as a thin white box. An outline is green when its IoU with the annotated target is at least 0.5, orange otherwise, and grey on frames without annotation.}
\label{fig:squadtrace}
\end{figure*}

\begin{figure*}[!t]
\centering\includegraphics[width=\textwidth]{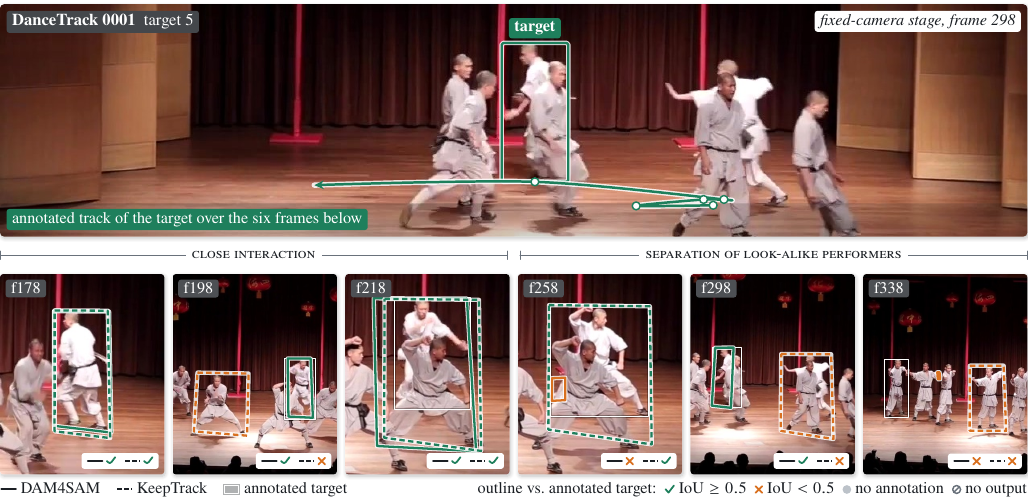}
\caption{\textbf{DanceTrack 0001, target 5. Both baselines leave the target as look-alike performers separate.} The banner shows frame 298 with the annotated track of the target over the six frames below. Outlines and colors follow Figure~\ref{fig:squadtrace}, and small predictions keep their original extent.}
\label{fig:sdancetrace}
\end{figure*}

\input{tables/supp_cases_updated}

\subsection{Physical PTZ camera}
Closed-loop hardware runs confirm that scheduling by predicted return time transfers to real optics, pan--tilt--zoom mechanics, and device queueing without synthetic waiting. In the protocol of main Table~5, each method executes $20$ two-person scripts three times for $60$ runs total under the measured latency of Table~\ref{tab:slatencyupdated} with no added waiting, scoring one reappearance event and full-duration identity per run under the same criteria as rendered episodes. Recovery succeeds in $42$, $46$, $47$, $48$, and $51$ of the $60$ runs for the common PTZ controller, Chanakya-J, asynchronous verification, BindAdapt without the return-time model, and BindAdapt, giving BindAdapt a $5.4$-point $\sid$ gain over Chanakya-J and a $4.2$-point gain over asynchronous verification while its compute service time of $38.5$\,ms stays within $0.3$\,ms of Chanakya-J.

\subsection{Unified reruns on two recorded clips}\label{sec:recordedwindows}
Rerunning the compared schedulers from identical initializations on recorded pedestrian and stage footage reproduces the gain under both panoramic and digital-viewport PTZ control. Extending main Table~6, Table~\ref{tab:srecstates} runs the common PTZ controller, Chanakya-J, and BindAdapt on DanceTrack 0001 and QuadTrack 0025 with the same backbone, operator catalog, resource caps, and scoring rule, where correct, wrong, and unresolved shares sum to the full clip. On fixed-camera DanceTrack~\cite{dancetrack}, pan and tilt translate a digital crop center and zoom resizes its extent inside the source frame so the policy observes only that crop, whereas QuadTrack renders viewports from its panorama. BindAdapt keeps the designated target for $73.9\%$ and $76.4\%$ of the two clips, beating Chanakya-J by $6.7$ and $6.9$ points while cutting wrong binding by $2.2$ and $2.7$ points.

\subsection{Frame-by-frame baseline drift on recorded clips}
Archived fixed-view outputs of DAM4SAM and KeepTrack on the same two clips show frame by frame how fixed-view trackers lose the designated person, complementing main Figure~3. Each panel maps the archived fixed-view boxes of both baselines through the recorded camera poses onto zoomed source frames and colors each outline by its IoU with the annotated target. In QuadTrack 0025, Figure~\ref{fig:squadtrace} tracks target 59 into an occlusion and subsequent pedestrian separation, where both baselines hold the target beforehand but KeepTrack switches to a neighboring pedestrian after reappearance and DAM4SAM outputs nothing at frame 498. In DanceTrack 0001, Figure~\ref{fig:sdancetrace} tracks target 5 among uniformly dressed performers, where KeepTrack jumps to another performer at frames 198 and 298, DAM4SAM shrinks onto a fragment at frame 258, and both settle on wrong positions by frame 338.

%% file: tables/supp_cases_updated.tex
\begin{table}[t]
\caption{\textbf{Unified reruns on two recorded clips.} Correct, wrong, and unresolved shares of the clip duration in \%, which add up to 100 before rounding.}\label{tab:srecstates}
\centering\small\setlength{\tabcolsep}{2pt}
\begin{tabularx}{\columnwidth}{@{}l>{\centering\arraybackslash}X>{\centering\arraybackslash}X>{\centering\arraybackslash}X>{\centering\arraybackslash}X>{\centering\arraybackslash}X>{\centering\arraybackslash}X@{}}\toprule
\multirow{2}{*}{Scheduler} & \multicolumn{3}{c}{DanceTrack 0001} & \multicolumn{3}{c}{QuadTrack 0025}\\\cmidrule(lr){2-4}\cmidrule(l){5-7}
 & $\sid\uparrow$ & $W\downarrow$ & $U\downarrow$ & $\sid\uparrow$ & $W\downarrow$ & $U\downarrow$\\\midrule
Common controller & 59.8 & 9.7 & 30.5 & 61.7 & 10.4 & 27.9\\
Chanakya-J & 67.3 & 7.8 & 24.9 & 69.5 & 8.2 & 22.4\\
\rowcolor{BAours}BindAdapt & \textbf{73.9} & \textbf{5.6} & \textbf{20.5} & \textbf{76.4} & \textbf{5.5} & \textbf{18.2}\\\bottomrule
\end{tabularx}
\end{table}

%% file: supp/s07_oracle.tex
\section{Finite-Action Reference and Replay}\label{sec:soracle}
Comparing online BindAdapt against a nested hierarchy of privileged policies shows that it closes over half of the baseline gap to the finite-action oracle, loses only $\BAval{ladder.step.law}$ points to response approximation error, and gains specifically when identity ambiguity and service delay coexist. Each successive class relaxes one information or physical constraint on the same designated target and admissible camera moves so every step carries a direct interpretation.

\begin{figure}[!b]
\centering\includegraphics[width=\columnwidth]{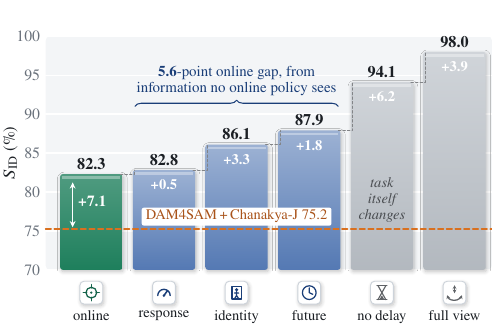}
\caption{\textbf{Online BindAdapt comes within 5.6 points of the finite-action reference, and unobserved identity and future outcomes account for most of that gap.} Each step adds one relaxation to online BindAdapt. The three blue steps add the exact response, perfect identity on acquired frames, and future outcomes of the finite action catalog, information that no online policy observes, and together form the 5.6-point gap to the finite-action reference. The two grey steps remove service delay and the field-of-view limit, which changes the task itself. The dashed line marks DAM4SAM with Chanakya-J.}
\label{fig:supper}
\end{figure}

\subsection{Nested feasible policy classes}
Six nested policy classes separate online modeling error, unobserved perceptual and motion outcomes, and physical task limits. Class $\Pi_0$ contains online policies that observe only causal history $\mathcal{H}_t$ under the shared motion, compute, and timing constraints. Class $\Pi_1$ adds the exact conditional identity response for every feasible action, $\Pi_2$ adds ground-truth identity labels on acquired frames, and $\Pi_3$ adds realized future outcomes across the finite action catalog while keeping physical acquisition and return timing intact, defining the finite-action reference reported as the oracle in the main paper. Beyond online decision limits, $\Pi_4$ removes service delay and $\Pi_5$ exposes the full panoramic scene. Table~\ref{tab:supper} reports $\sid$ across all six levels.

Because each richer class only adds information or relaxes a constraint, a policy in $\Pi_{i+1}$ can always ignore the relaxation and emulate any member of $\Pi_i$. Under identical episodes, objectives, and unrelaxed constraints, $\Pi_i\subseteq\Pi_{i+1}$ implies
\begin{equation}
\sup_{\pi\in\Pi_i}\mathbb E[\sid(\pi)]\leq
\sup_{\pi\in\Pi_{i+1}}\mathbb E[\sid(\pi)].
\end{equation}

\begin{table}[t]\caption{\textbf{Identity success $\sid$ (\%) by information level.} Each level relaxes one information or service constraint while keeping the same identity measure and the same designated target.}\label{tab:supper}\centering\small\setlength{\tabcolsep}{4pt}\begin{tabularx}{\columnwidth}{@{}l>{\centering\arraybackslash}X@{}}\toprule
Information and constraints & $\sid$\\\midrule
\BAinputrows tables/supper.tex
\bottomrule\end{tabularx}\end{table}

\subsection{Finite-action oracle construction}
Exhaustive branch enumeration with full native-state restoration makes $\Pi_3$ an exact upper bound over the shared action catalog. Over the fixed horizon, budgets, camera resolution, start-time grid, and trajectory set, the oracle evaluates every feasible sequence without beam pruning or greedy truncation. At each common prefix, every branch restores native tracker tensors, queues, random state, and cached features, whereas restoring only the output box would leak templates or segmentation memory written by an earlier branch into subsequent branches.

\subsection{Oracle gap decomposition}
Replacing the learned response $p_\theta$ with the exact conditional law adds only \BAval{ladder.step.law} points, showing that unobserved identity labels and future branch realizations account for nearly all of the remaining online gap. Across Table~\ref{tab:supper} and Figure~\ref{fig:supper}, $\sid$ moves from $\BAval{t08.r01.sid}\%$ at $\Pi_0$ to $\BAval{t08.r02.sid}\%$, $\BAval{t08.r03.sid}\%$, $\BAval{t08.r04.sid}\%$, $\BAval{t08.r05.sid}\%$, and $\BAval{t08.r06.sid}\%$, with step gains of \BAval{ladder.step.law}, \BAval{ladder.step.id}, \BAval{ladder.step.future}, \BAval{ladder.step.zerodelay}, and \BAval{ladder.step.fullview} points. The first three steps sum to the $\BAval{gap.remaining}$-point gap to the finite-action reference, dominated by \BAval{ladder.step.id} points from perfect acquired-frame identity and \BAval{ladder.step.future} points from realized future outcomes. Removing service delay at $\Pi_4$ then adds \BAval{ladder.step.zerodelay} points and full visibility at $\Pi_5$ adds \BAval{ladder.step.fullview} points, which alter the physical tracking task itself.

Measured against the $\BAval{t08.r04.sid}\%$ finite-action reference, BindAdapt closes more than half of the gap left by the strongest external baseline. DAM4SAM with Chanakya-J reaches $\BAval{ext.best.sid}\%$, trailing $\Pi_3$ by \BAval{gap.external.abs} points, whereas BindAdapt reaches $\BAval{ext.bind.sid}\%$ and closes $(82.27-75.21)/(87.89-75.21)=\BAval{gap.external}\%$ of that gap from unrounded scores, and closes $(82.27-70.29)/(87.89-70.29)=\BAval{gap.internal}\%$ of the gap from the internal joint-belief planner at $70.29\%$.

\begin{table}[t]\caption{\textbf{Ceiling and negative controls.} Ref.\ is the reference response with the separable scheduler and Ours is BindAdapt, entries are identity success in \%, and $\Delta$ is their difference in points. The gain is large when identity is ambiguous and computation is delayed.}\label{tab:sceil}\centering\small\setlength{\tabcolsep}{3pt}\begin{tabularx}{\columnwidth}{@{}>{\raggedright\arraybackslash}X ccc@{}}\toprule
Control & Ref. & Ours & $\Delta$\\\midrule
\BAinputrows tables/sceil.tex
\bottomrule\end{tabularx}\end{table}

\subsection{Ceiling and negative controls}
Ceiling and negative controls confirm that BindAdapt gains specifically when identity is ambiguous and computation is delayed, and gains almost nothing when timing is unneeded or identity is unobservable. In Table~\ref{tab:sceil}, removing look-alike ambiguity or freezing the scene under negligible delay leaves little for return-time scheduling to repair, limiting the gain over the reference response with the separable scheduler to $\BAval{t11.r01.gain.pp}$ and $\BAval{t11.r02.gain.pp}$ points. Withholding distinguishing identity cues or keeping the absent target outside every reachable view leaves no causal view that can resolve identity, shrinking the difference to $\BAval{t11.r04.gain.pp}$ and $\BAval{t11.r05.gain.pp}$ points. When ambiguity and delayed computation coincide, BindAdapt raises $\sid$ from $\BAval{t11.r03.reference}\%$ to $\BAval{t11.r03.bindadapt}\%$, a $\BAval{t11.r03.gain.pp}$-point gain from timely evidence about the target.